\documentclass[10pt,twocolumn,letterpaper]{article}

\usepackage[pagenumbers]{cvpr} %

\usepackage{tabularx}
\usepackage{multirow}
\usepackage[per-mode=symbol, group-digits=integer, group-minimum-digits = 4, detect-all=true, input-symbols={()}]{siunitx}

\usepackage[percent]{overpic} %
\usepackage{fp} %
\usepackage{graphicx}
\usepackage{tikz}
\usetikzlibrary{positioning}
\usetikzlibrary{shapes.geometric, shapes, arrows, arrows.meta, bending}
\usetikzlibrary{backgrounds}
\usetikzlibrary{spy}
\usepackage{pgfplots}
\usepackage{bbding}
\usepackage{dsfont}
\usepackage{fontawesome5}
\usepackage{calc}
\pgfplotsset{compat=1.17}
\usepgfplotslibrary{groupplots}
\usepackage[outline]{contour} 
\contourlength{0.02em}
\usepackage[table]{xcolor}
\usepackage{makecell}
\usepackage{pifont}%
\newcommand{\cmark}{\ding{51}}%
\newcommand{\xmark}{\ding{55}}%
\usepackage[scaled=0.97]{newtxtt}

\newcommand*{\inparagraph}[1]{\smallskip\noindent\textbf{#1}\hspace{0.2em}}
\newcommand*{\inparagraphnospace}[1]{\smallskip\noindent\textbf{#1}}

\newcolumntype{Z}{S[table-format=2.1]}
\newcolumntype{Y}{>{\centering\arraybackslash}X} %

\definecolor{cvprcolor}{RGB}{127,127,255}
\definecolor{tud0d}{RGB}{83,83,83}
\definecolor{tud0c}{RGB}{137,137,137}
\definecolor{tud0b}{RGB}{181,181,181}
\definecolor{tud0a}{RGB}{220,220,220}
\definecolor{tud1a}{RGB}{93,133,195}
\definecolor{tud2a}{RGB}{0,156,218}
\definecolor{tud3a}{RGB}{80,182,149}
\definecolor{tud4a}{RGB}{175,204,80}
\definecolor{tud5a}{RGB}{221,223,72}
\definecolor{tud6a}{RGB}{255,224,92}
\definecolor{tud7a}{RGB}{248,186,60}
\definecolor{tud8a}{RGB}{238,122,52}
\definecolor{tud9a}{RGB}{233,80,62}
\definecolor{tud10a}{RGB}{201,48,142}
\definecolor{tud11a}{RGB}{128,69,151}
\definecolor{tud1b}{RGB}{0,90,169}
\definecolor{tud2b}{RGB}{0,131,204}
\definecolor{tud3b}{RGB}{0,157,129}
\definecolor{tud4b}{RGB}{153,192,0}
\definecolor{tud5b}{RGB}{201,212,0}
\definecolor{tud6b}{RGB}{253,202,0}
\definecolor{tud7b}{RGB}{245,163,0}
\definecolor{tud8b}{RGB}{236,101,0}
\definecolor{tud9b}{RGB}{230,0,26}
\definecolor{tud10b}{RGB}{166,0,132}
\definecolor{tud11b}{RGB}{114,16,133}
\definecolor{tud1c}{RGB}{0,78,138}
\definecolor{tud2c}{RGB}{0,104,157}
\definecolor{tud3c}{RGB}{0,136,119}
\definecolor{tud4c}{RGB}{127,171,22}
\definecolor{tud5c}{RGB}{177,189,0}
\definecolor{tud6c}{RGB}{215,172,0}
\definecolor{tud7c}{RGB}{210,135,0}
\definecolor{tud8c}{RGB}{204,76,3}
\definecolor{tud9c}{RGB}{185,15,34}
\definecolor{tud10c}{RGB}{149,17,105}
\definecolor{tud11c}{RGB}{97,28,115}
\definecolor{tud1d}{RGB}{36,53,114}
\definecolor{tud2d}{RGB}{0,78,115}
\definecolor{tud3d}{RGB}{0,113,94}
\definecolor{tud4d}{RGB}{106,139,55}
\definecolor{tud5d}{RGB}{153,166,4}
\definecolor{tud6d}{RGB}{174,142,0}
\definecolor{tud7d}{RGB}{190,111,0}
\definecolor{tud8d}{RGB}{169,73,19}
\definecolor{tud9d}{RGB}{156,28,38}
\definecolor{tud10d}{RGB}{115,32,84}
\definecolor{tud11d}{RGB}{76,34,106}

\definecolor{our3}{RGB}{220, 38, 127}

\definecolor{our1}{RGB}{13, 146, 244}
\definecolor{our2}{RGB}{119, 205, 255}
\definecolor{our3}{RGB}{249, 84, 84}
\definecolor{our4}{RGB}{198, 46, 46}

\definecolor{unlabeled}{RGB}{0,0,0}
\definecolor{egovehicle}{RGB}{0,0,0}
\definecolor{rectification border}{RGB}{0,0,0}
\definecolor{outofroi}{RGB}{0,0,0}
\definecolor{static}{RGB}{0,0,0}
\definecolor{dynamic}{RGB}{111,74,0}
\definecolor{ground}{RGB}{81,0,81}
\definecolor{road}{RGB}{128,64,128}
\definecolor{sidewalk}{RGB}{244,35,232}
\definecolor{parking}{RGB}{250,170,160}
\definecolor{rail track}{RGB}{230,150,140}
\definecolor{building}{RGB}{70,70,70}
\definecolor{wall}{RGB}{102,102,156}
\definecolor{fence}{RGB}{190,153,153}
\definecolor{guard rail}{RGB}{180,165,180}
\definecolor{bridge}{RGB}{150,100,100}
\definecolor{tunnel}{RGB}{150,120,90}
\definecolor{pole}{RGB}{153,153,153}
\definecolor{polegroup}{RGB}{153,153,153}
\definecolor{trafficlight}{RGB}{250,170,30}
\definecolor{trafficsign}{RGB}{220,220,0}
\definecolor{vegetation}{RGB}{107,142,35}
\definecolor{terrain}{RGB}{152,251,152}
\definecolor{sky}{RGB}{70,130,180}
\definecolor{skylight}{RGB}{98,182,252}
\definecolor{person}{RGB}{220,20,60}
\definecolor{rider}{RGB}{255,0,0}
\definecolor{car}{RGB}{0,0,142}
\definecolor{truck}{RGB}{0,0,70}
\definecolor{bus}{RGB}{0,60,100}
\definecolor{caravan}{RGB}{0,0,90}
\definecolor{trailer}{RGB}{0,0,110}
\definecolor{train}{RGB}{0,80,100}
\definecolor{motorcycle}{RGB}{0,0,230}
\definecolor{bicycle}{RGB}{119,11,32}
\definecolor{licenseplate}{RGB}{0,0,142}

\newcommand{\dlt}[1]{{\textcolor{tud3c}{#1}}}

\newcommand{\MethodName}{VideoCUPS\@\xspace} %

\usepackage{pict2e,picture,graphicx}

\makeatletter
\DeclareRobustCommand{\Arrow}[1][]{%
\check@mathfonts
\if\relax\detokenize{#1}\relax
\settowidth{\dimen@}{$\m@th\rightarrow$}%
\else
\setlength{\dimen@}{#1}%
\fi
\sbox\z@{\usefont{U}{lasy}{m}{n}\symbol{41}}%
\begin{picture}(\dimen@,\ht\z@)
\roundcap
\put(\dimexpr\dimen@-.7\wd\z@,0){\usebox\z@}
\put(0,\fontdimen22\textfont2){\line(1,0){\dimen@}}
\end{picture}%
}
\makeatother
\newcommand{\veryshortrightarrow}{\hspace{.2mm}\scalebox{.8}{\Arrow[.1cm]}\hspace{.2mm}}

\tikzset{
    double_trapezium/.pic={
        \pgfkeys{/double trapezium/width/.initial=1.15cm}
        \pgfkeys{/double trapezium/height/.initial=0.5cm}
        \pgfkeys{/double trapezium/midheight/.initial=0.25cm}
        \pgfkeys{/double trapezium/color/.initial=pole} %
        \pgfkeysgetvalue{/double trapezium/width}{\width}
        \pgfkeysgetvalue{/double trapezium/height}{\height}
        \pgfkeysgetvalue{/double trapezium/midheight}{\midheight}
        \pgfkeysgetvalue{/double trapezium/color}{\colorvar} %

        \filldraw[left color=\colorvar!50, right color=\colorvar!10, draw=\colorvar!10, rounded corners=1] 
            ({-\width}, \height) -- (0, {\midheight}) -- (0, {-\midheight}) -- ({-\width}, -\height) -- cycle;

        \filldraw[left color=\colorvar!10, right color=\colorvar!50, draw=\colorvar!10, rounded corners=1] 
            ({\width}, \height) -- (0, {\midheight}) -- (0, {-\midheight}) -- ({\width}, -\height) -- cycle;

        \filldraw[draw=black, fill=none, thick, rounded corners=1] 
            ({-\width}, \height) -- (0, {\midheight}) -- ({\width}, \height) -- 
            ({\width}, -\height) -- (0, {-\midheight}) -- ({-\width}, -\height) -- cycle;

    }
}

\tikzset{
    double_trapezium_small/.pic={
        \pgfkeys{/double trapezium/width/.initial=0.9625cm}
        \pgfkeys{/double trapezium/height/.initial=0.4375cm}
        \pgfkeys{/double trapezium/midheight/.initial=0.21875cm}
        \pgfkeys{/double trapezium/color/.initial=pole} %
        \pgfkeysgetvalue{/double trapezium/width}{\width}
        \pgfkeysgetvalue{/double trapezium/height}{\height}
        \pgfkeysgetvalue{/double trapezium/midheight}{\midheight}
        \pgfkeysgetvalue{/double trapezium/color}{\colorvar} %

        \filldraw[left color=\colorvar!50, right color=\colorvar!10, draw=\colorvar!10, rounded corners=1] 
            ({-\width}, \height) -- (0, {\midheight}) -- (0, {-\midheight}) -- ({-\width}, -\height) -- cycle;

        \filldraw[left color=\colorvar!10, right color=\colorvar!50, draw=\colorvar!10, rounded corners=1] 
            ({\width}, \height) -- (0, {\midheight}) -- (0, {-\midheight}) -- ({\width}, -\height) -- cycle;

        \filldraw[draw=black, fill=none, rounded corners=1] 
            ({-\width}, \height) -- (0, {\midheight}) -- ({\width}, \height) -- 
            ({\width}, -\height) -- (0, {-\midheight}) -- ({-\width}, -\height) -- cycle;

    }
}

\definecolor{cvprblue}{rgb}{0.21,0.49,0.74}
\usepackage[pagebackref,breaklinks,colorlinks,allcolors=cvprblue]{hyperref}

\def\paperID{}
\def\confName{CVPR}
\def\confYear{2026}

\title{Scene-Centric Unsupervised Video Panoptic Segmentation}

\newcommand{\authorstep}{\hspace{0.8cm}}
\newcommand{\affiliationstep}{\hspace{0.155cm}}
\author{
Christoph Reich\textsuperscript{\normalfont{}* 1,2,5,6}
\authorstep Oliver Hahn\textsuperscript{\normalfont{}* 2,3}
\authorstep Nikita Araslanov\textsuperscript{\normalfont{}\,1,5}
\authorstep Laura Leal-Taixé\textsuperscript{\normalfont{}\,3}\\
Christian Rupprecht\textsuperscript{\normalfont{}\,4}
\authorstep Daniel Cremers\textsuperscript{\normalfont{}\textdagger{} 1,5,6}
\authorstep Stefan Roth\textsuperscript{\normalfont{}\textdagger{} 2,6,7}\\[0.0pt]
\small{\textsuperscript{1}TU Munich\affiliationstep \textsuperscript{2}TU Darmstadt \affiliationstep 
\textsuperscript{3}NVIDIA \affiliationstep 
\textsuperscript{4}University of Oxford \affiliationstep \textsuperscript{5}MCML\affiliationstep \textsuperscript{6}ELIZA\affiliationstep \textsuperscript{7}hessian.AI\affiliationstep
\textsuperscript{*}equal contribution\affiliationstep
\textsuperscript{\textdagger}equal advising}\\[-2pt]\small {\url{https://visinf.github.io/videocups}}}

\usepackage{fancyhdr}
\usepackage{setspace}

\fancypagestyle{firststyle}
{

	\fancyhf{}
	\lfoot{{\footnotesize\begin{spacing}{.5}\parbox{\linewidth}{\vspace{-1.39em}%
					To appear in \emph{Proceedings of the IEEE/CVF Conference on Computer Vision and Pattern Recognition}, Denver, CO, USA, 2026.%
					\\\hrule\vspace{\baselineskip}
					\copyright~2026 IEEE. Personal use of this material is permitted. Permission from IEEE must be obtained for all other uses, in any current or future media, including reprinting/republishing this material for advertising or promotional purposes, creating new collective works, for resale or redistribution to servers or lists, or reuse of any copyrighted component of this work in other works. 
	}\end{spacing}}}
}

\begin{document}
\twocolumn[{%
\renewcommand\twocolumn[1][]{#1}%
\maketitle
\vspace{-2.4em}
\tikzset{every picture/.style={font=\sffamily\scriptsize}}

\pgfplotsset{
  teaser axis style/.style={
    width=3.5cm,
    height=3.5cm,
    ybar,
    bar width=17pt,
    xtick=data,
    ymin=14.5,
    ytick=\empty,
    axis x line=bottom,
    axis y line=none,
    axis line style={-},
    xtick style={draw=none},
    enlarge x limits=1.7,
    nodes near coords,
    nodes near coords={\pgfmathprintnumber[assume math mode=true]{\pgfplotspointmeta}},
    every node near coord/.append style={
      yshift=-10pt,
      color=white,
      font=\sffamily\scriptsize,
      /pgf/number format/.cd,
      fixed,
      fixed zerofill,
      precision=1,
    },
    legend style={
      at={(-0.0, 0.85)},
      anchor=north,
      draw=none,
      legend columns=1,
    },
    reverse legend,
    legend cell align=left,
    legend image code/.code={
      \draw[fill=#1, draw=none] (-0.08cm,-0.08cm) rectangle (0.12cm,0.12cm);
    },
  }
}

\newcommand{\teaserimgscale}{1.0}
\FPeval{\csteaserimgscale}{round(\teaserimgscale*0.2,6)}
\FPeval{\kitticutoff}{round((((3.369565217391304/2.0)-1)/4),8)}
\newcommand{\imgoverlayopacity}{0.45}

\scriptsize
\sffamily
\setlength{\tabcolsep}{0pt}
\renewcommand{\arraystretch}{0.975}

\begin{tabular}{
>{\centering\arraybackslash} m{0.155\textwidth}
>{\centering\arraybackslash} m{0.01\textwidth}
>{\centering\arraybackslash} m{0.155\textwidth}
>{\centering\arraybackslash} m{0.01\textwidth}
>{\centering\arraybackslash} m{0.155\textwidth}
>{\centering\arraybackslash} m{0.029\textwidth}
>{\centering\arraybackslash} m{0.155\textwidth}
>{\centering\arraybackslash} m{0.01\textwidth}
>{\centering\arraybackslash} m{0.155\textwidth}
>{\centering\arraybackslash} m{0.01\textwidth}
>{\centering\arraybackslash} m{0.155\textwidth}
}

\multicolumn{5}{c}{\rule[0.5ex]{2.6cm}{0.4pt}\textbf{ Cityscapes-VPS }\rule[0.5ex]{2.6cm}{0.4pt}}
& 
& \multicolumn{5}{c}{\rule[0.5ex]{2.95cm}{0.4pt}\textbf{ KITTI-STEP\vphantom{g} }\rule[0.5ex]{2.95cm}{0.4pt}} \\[0.25pt]

\includegraphics[width=\linewidth]{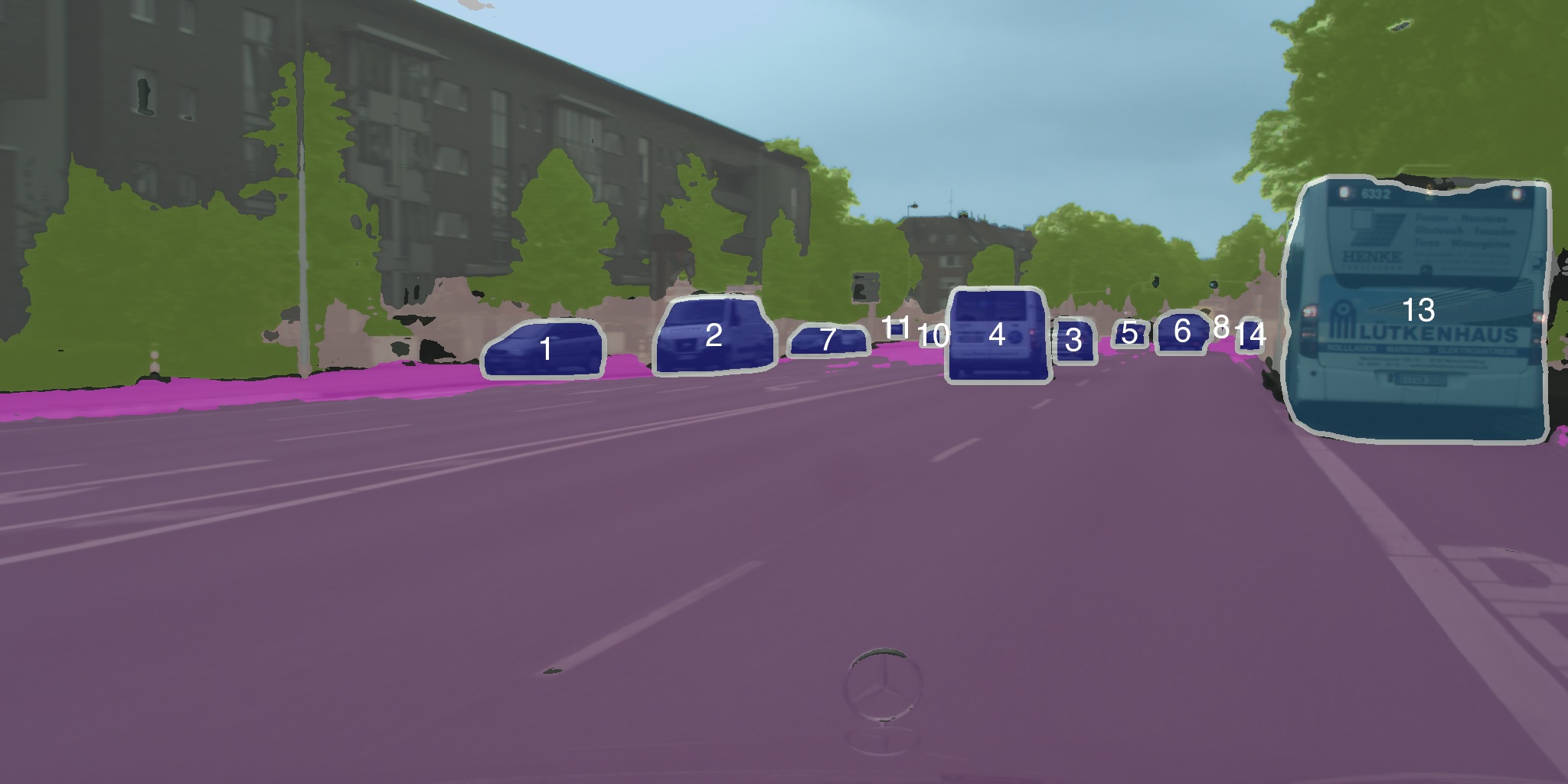} 
& %
& \includegraphics[width=\linewidth]{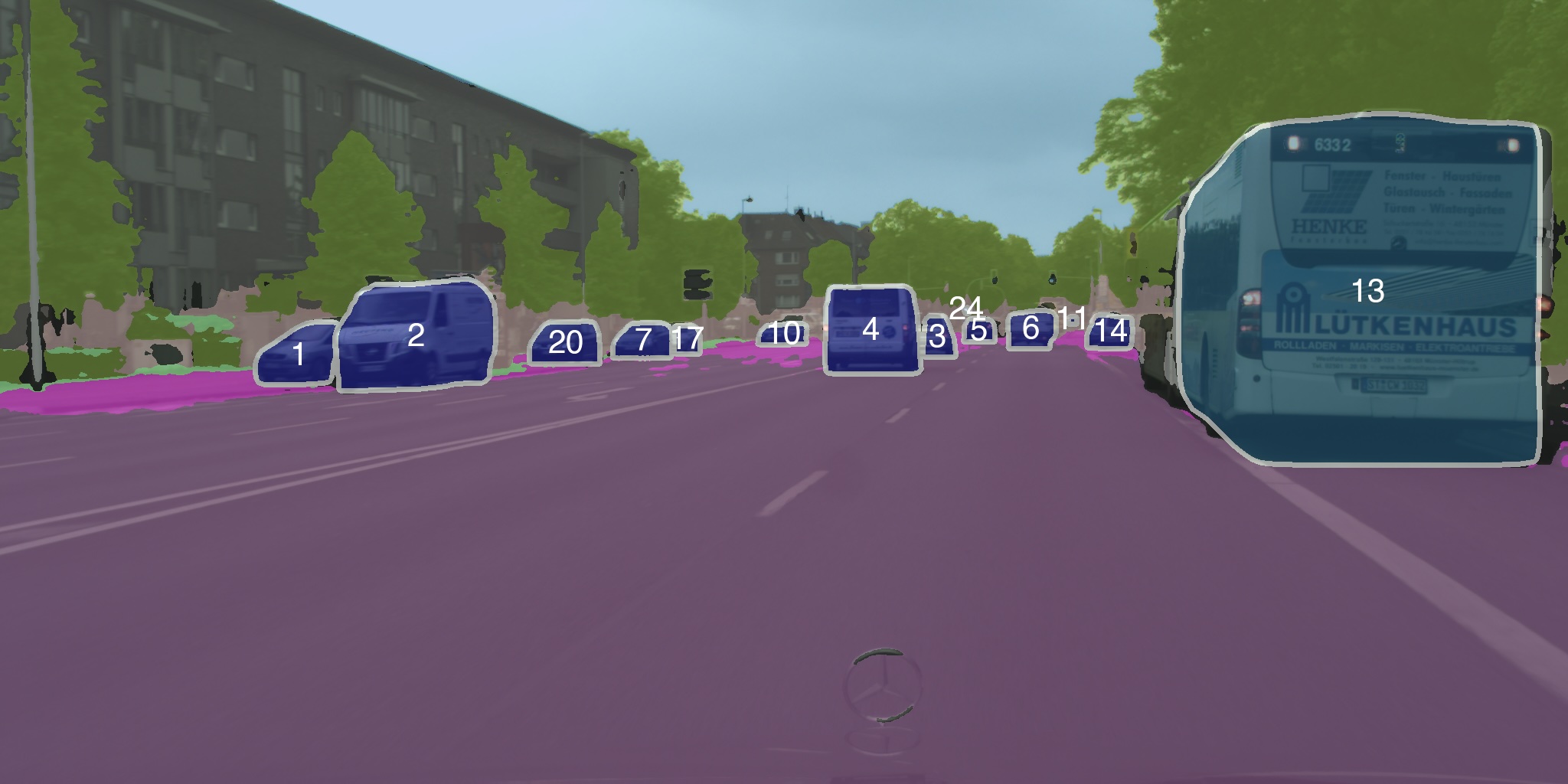}
& %
& \includegraphics[width=\linewidth]{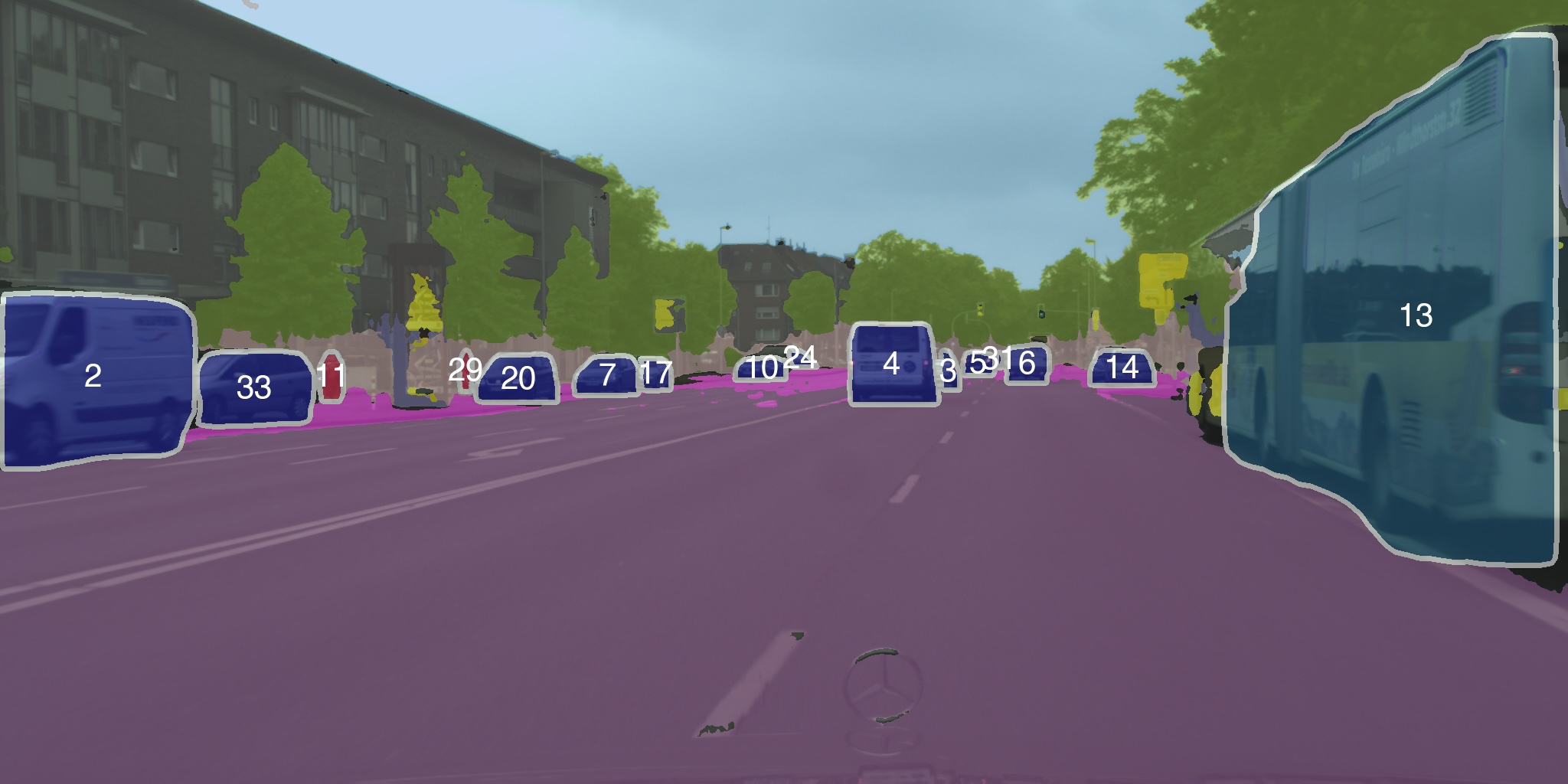}
&
& \includegraphics[width=\linewidth]{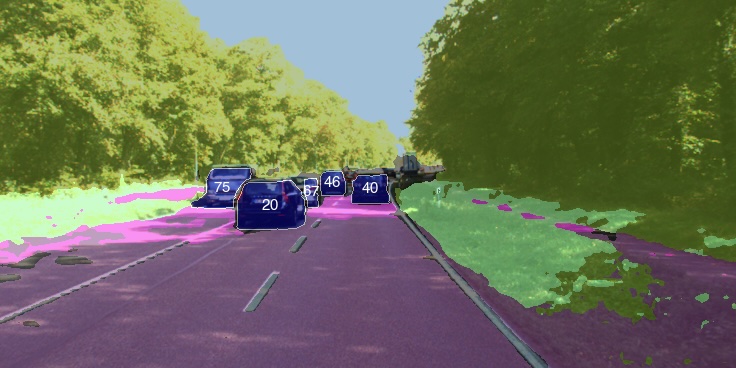}
& %
& \includegraphics[width=\linewidth]{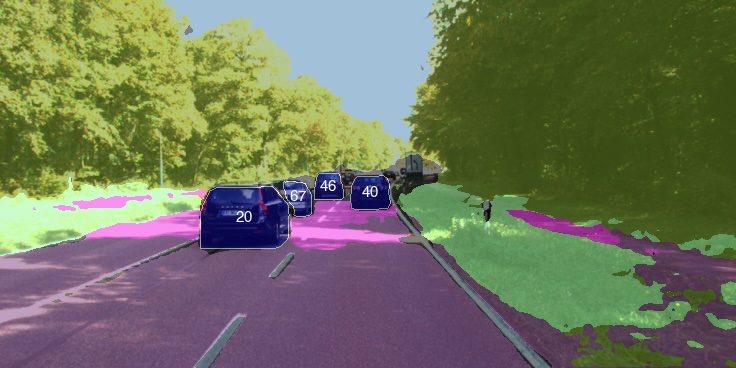}
& %
& \includegraphics[width=\linewidth]{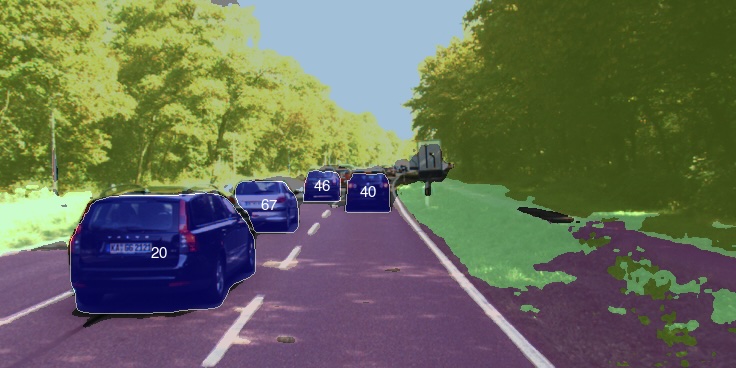} \\[-1.5pt]

\multicolumn{5}{c}{\rule[0.5ex]{3.1cm}{0.4pt}\textbf{ Waymo }\rule[0.5ex]{3.1cm}{0.4pt}}
& 
& \multicolumn{5}{c}{\rule[0.5ex]{3.3cm}{0.4pt}\textbf{ MOTS }\rule[0.5ex]{3.3cm}{0.4pt}} \\[0.0pt]

\includegraphics[width=\linewidth]{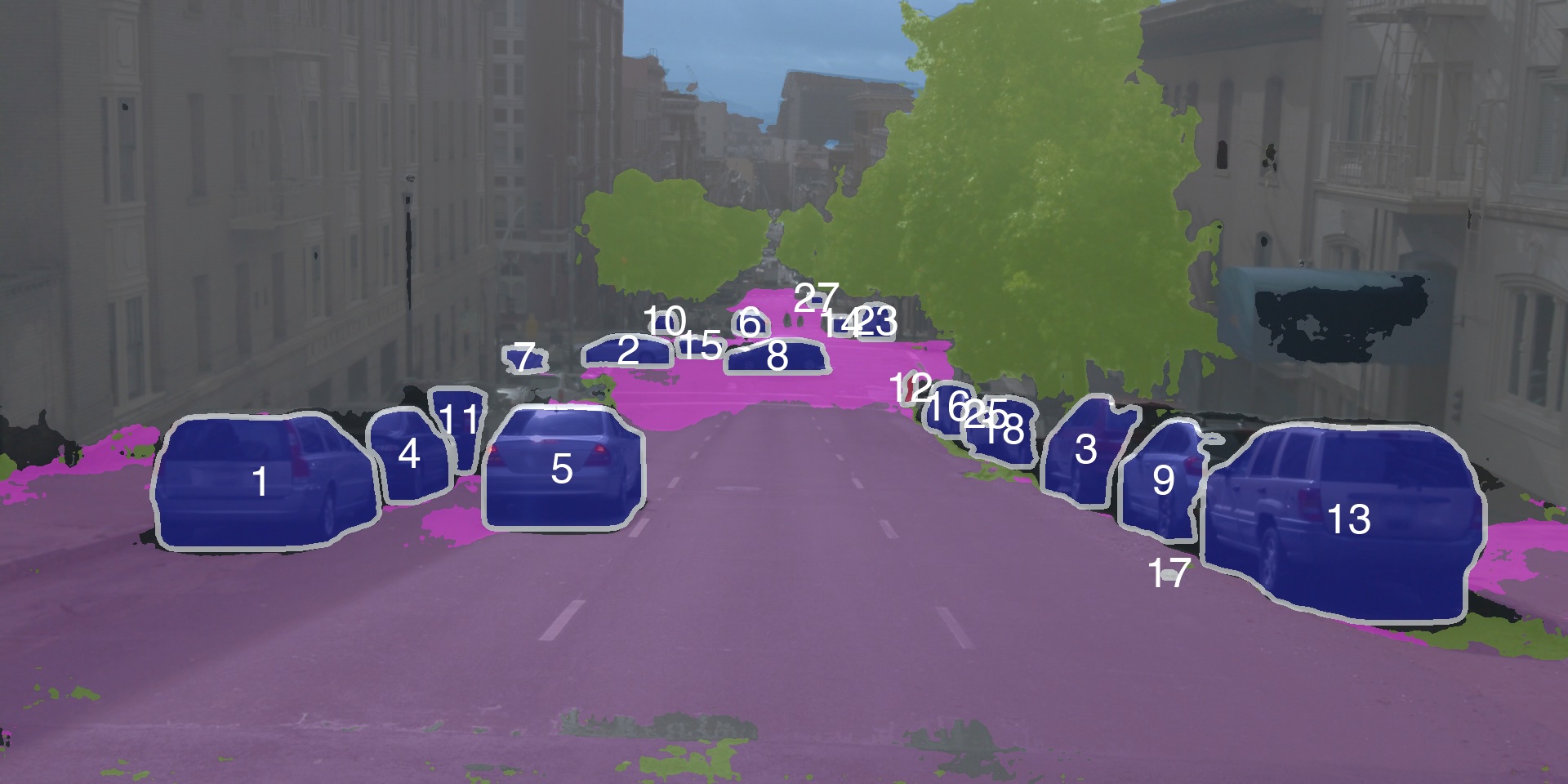} 
& %
& \includegraphics[width=\linewidth]{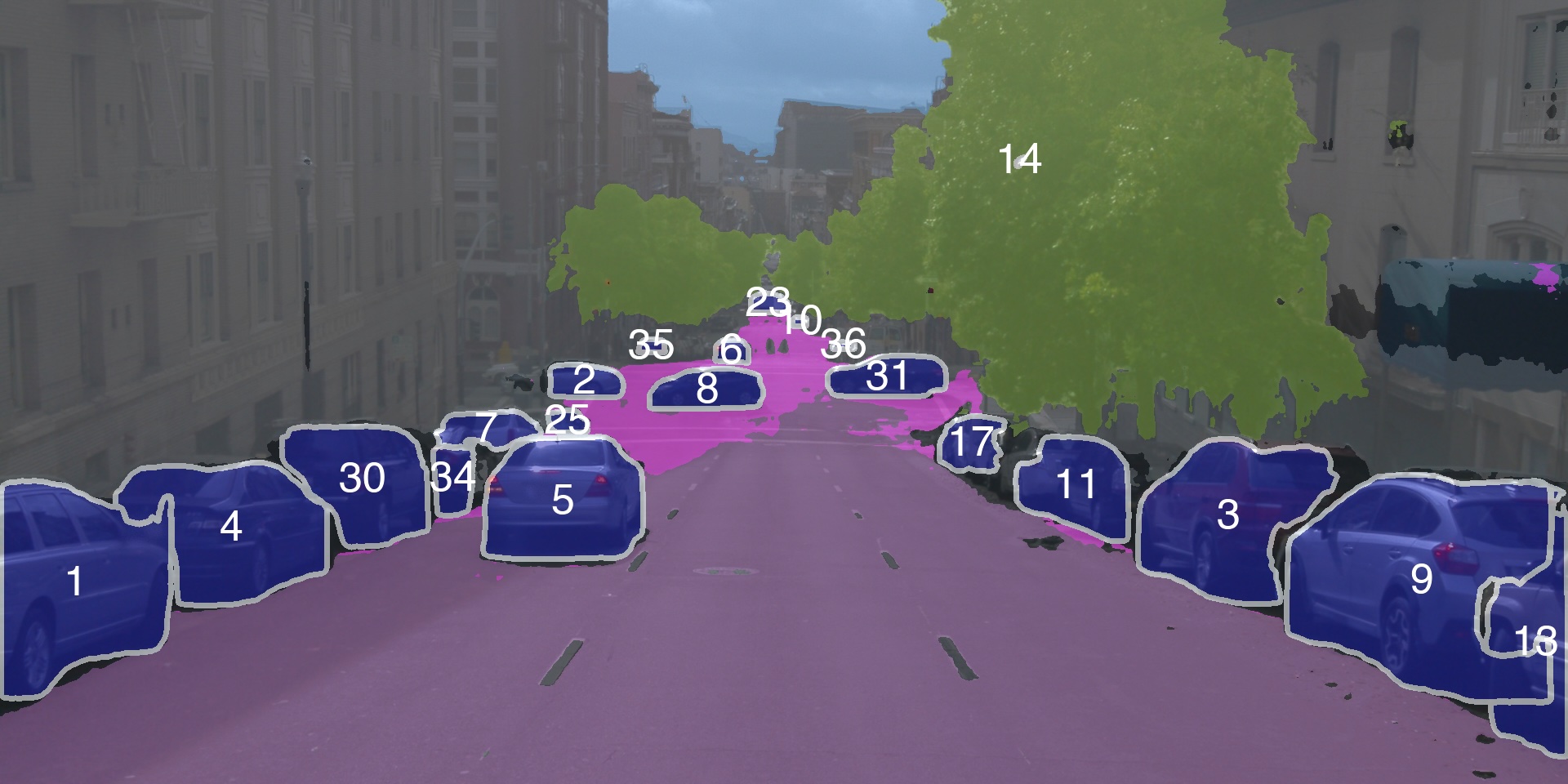} 
& %
& \includegraphics[width=\linewidth]{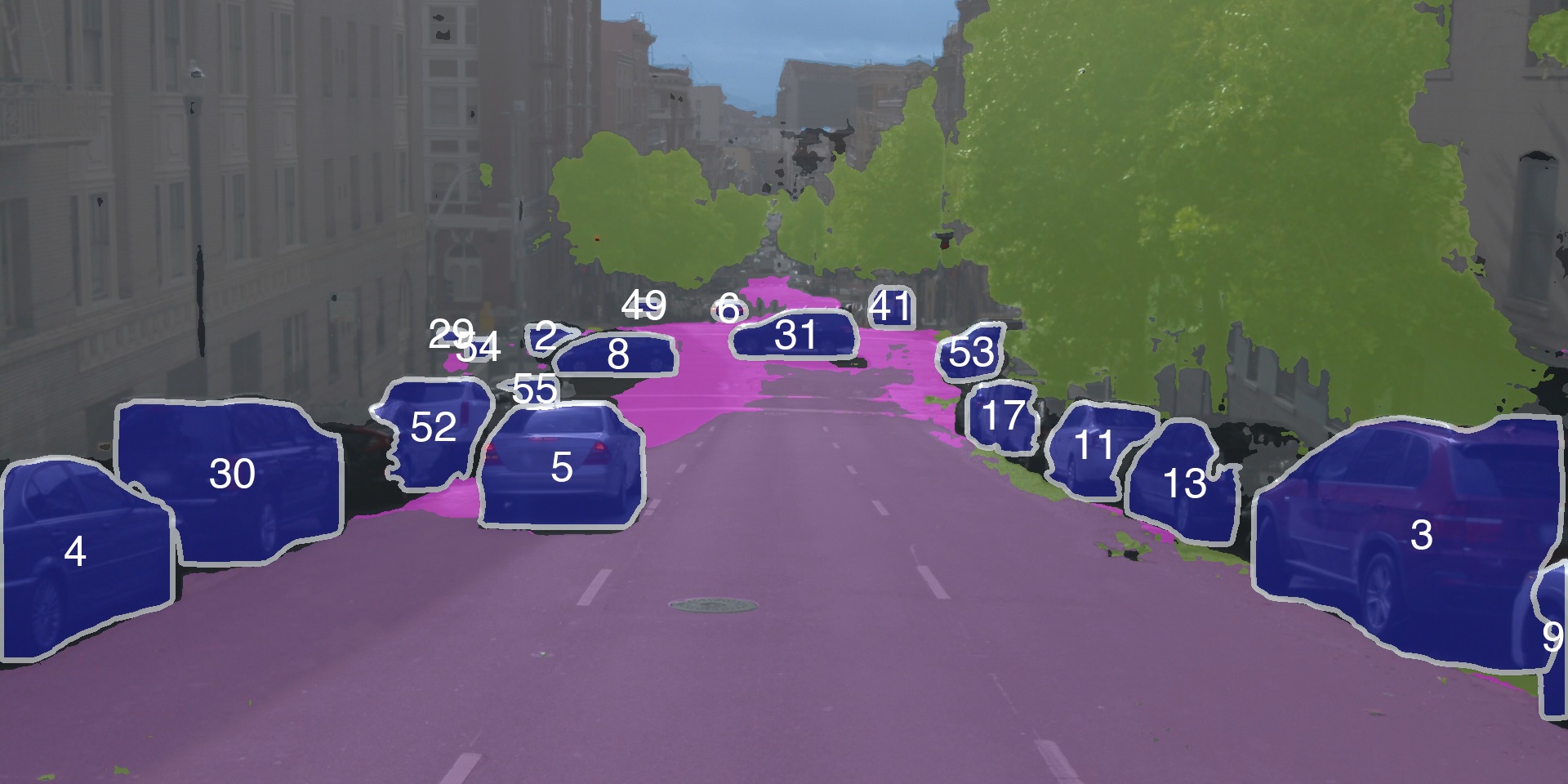} 
&
& \includegraphics[width=\linewidth]{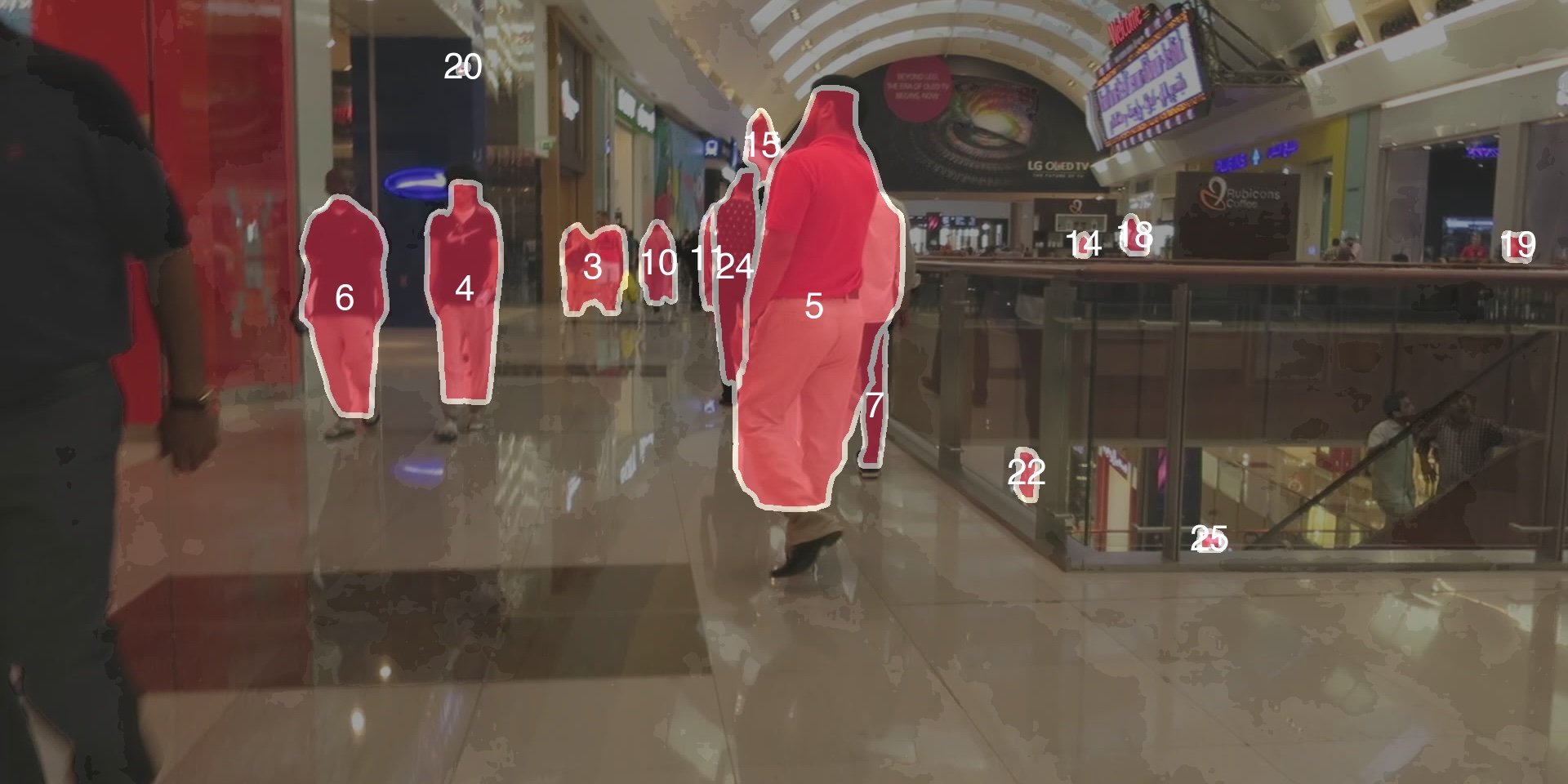}
& %
& \includegraphics[width=\linewidth]{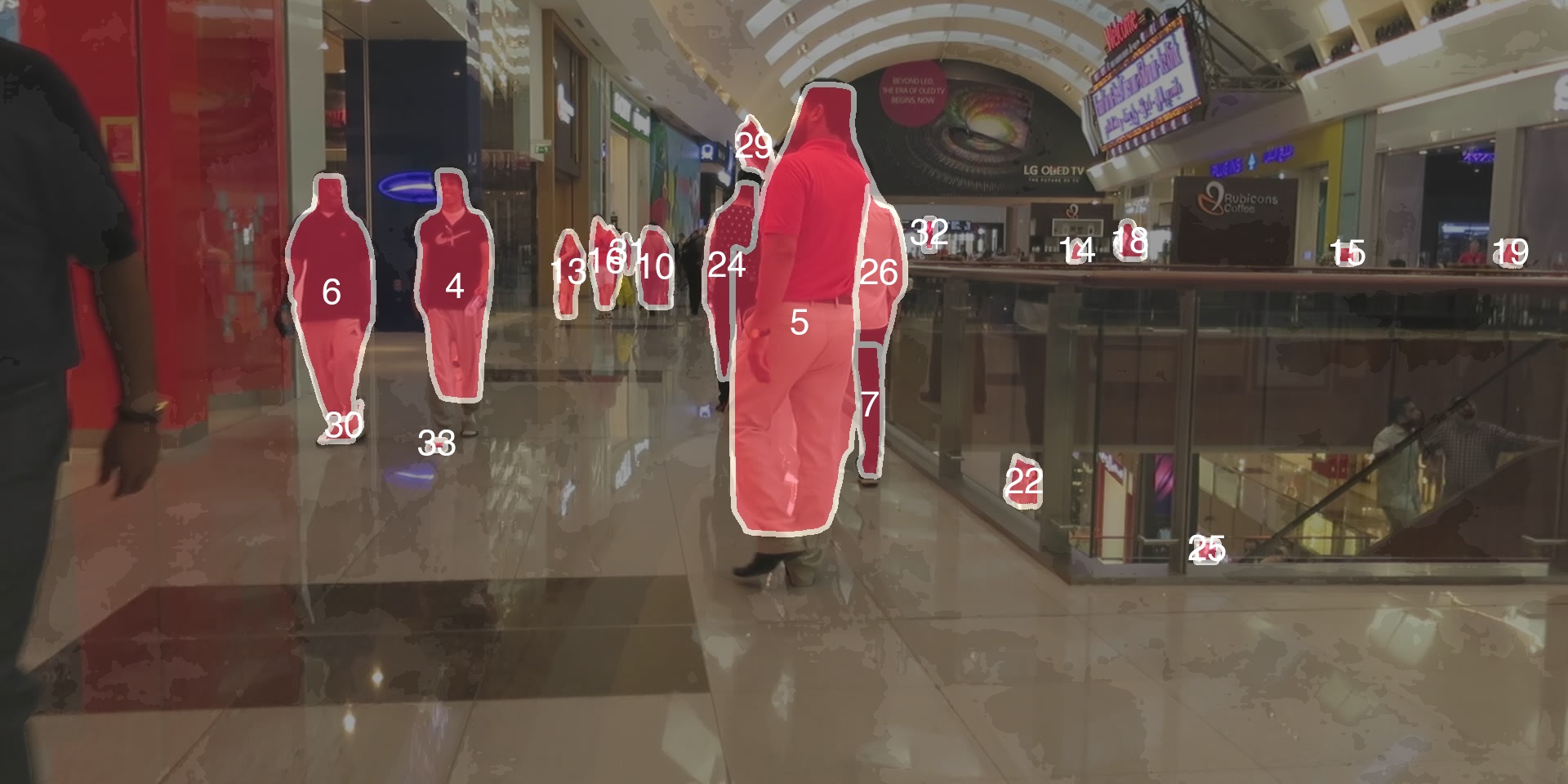}
& %
& \includegraphics[width=\linewidth]{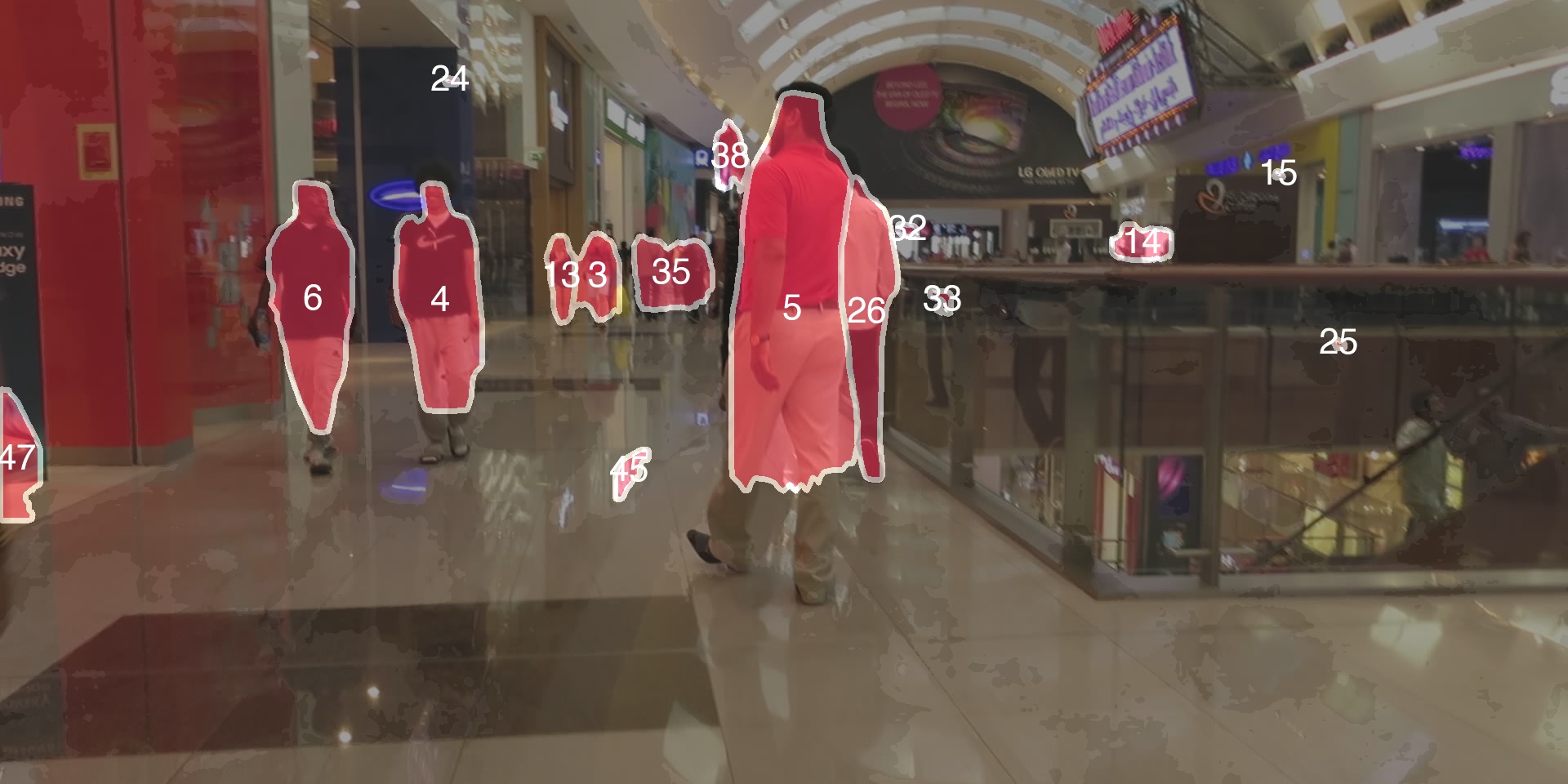} \\[-3.25pt]

\tiny{$\phantom{.}t_0$}
& 
&\tiny{$\phantom{.}t_1$ }
& 
&\tiny{$\phantom{.}t_2$ }
&
&\tiny{$\phantom{.}t_0$}
& 
&\tiny{$\phantom{.}t_{1}$ }
& 
&\tiny{$\phantom{.}t_2$}  \\[2.5pt]

\end{tabular}

\vspace{-0.475cm}

\begin{center}
  \begin{tikzpicture}[>={Stealth[inset=0pt,length=5pt,angle'=45]}, scale=0.875]
    \tikzset{every node/.style={font=\fontsize{5.5}{2}\selectfont}}
    \node[white] at (15.25, 0.0) {.};

    \begin{scope}[xshift=-1.7cm, yshift=-0.075cm]
      \node at (-0.1, 0.1)   {\frame{\includegraphics[width=1.5cm, decodearray={0.0 1.25 0 1.25 0.0 1.25}]{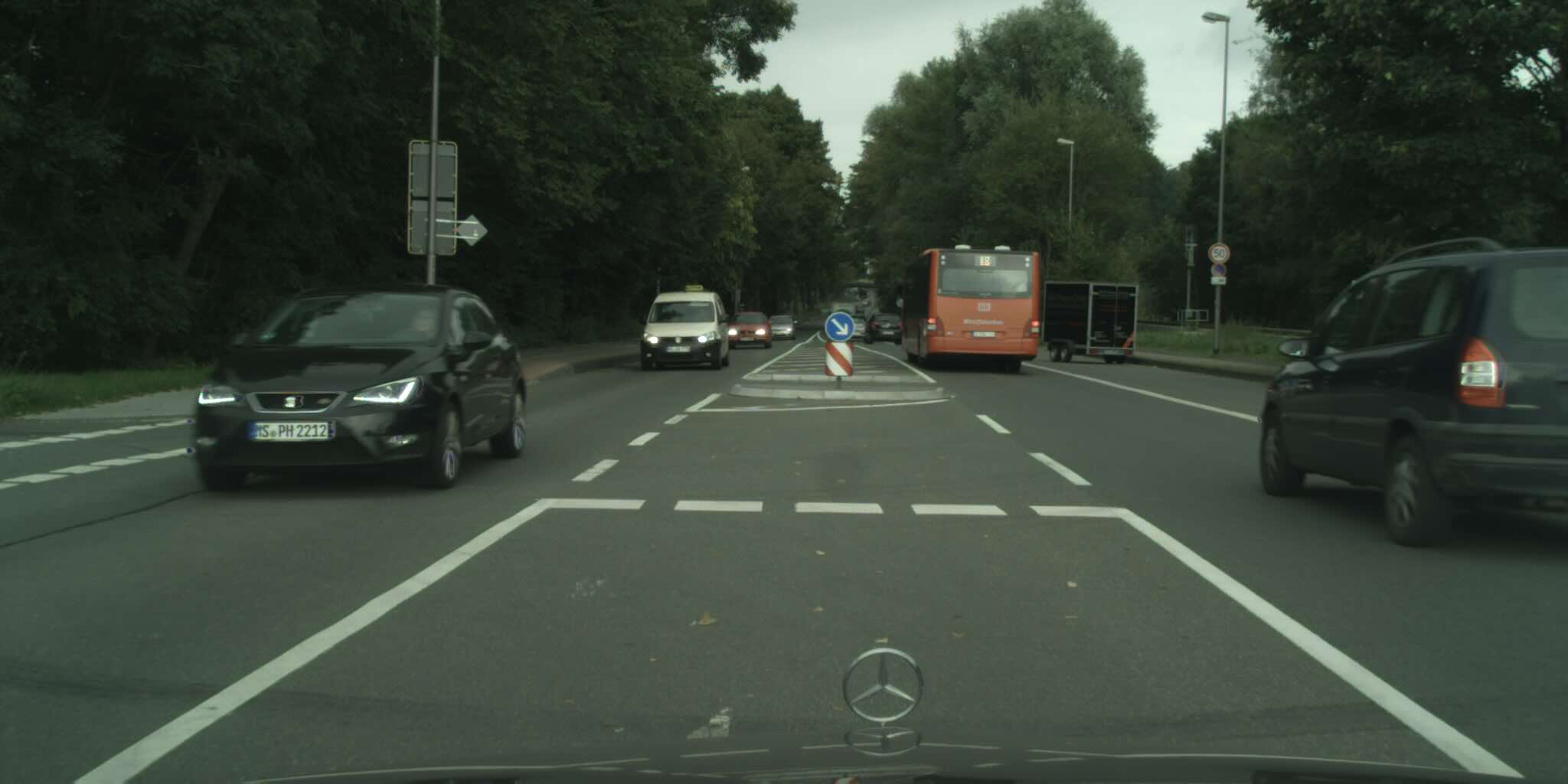}}};
      \node at (-0.05, 0.05) {\frame{\includegraphics[width=1.5cm, decodearray={0.0 1.25 0 1.25 0.0 1.25}]{artwork/teaser/data/munster_000118_000005_leftImg8bit.jpg}}};
      \node at (0, 0)        {\frame{\includegraphics[width=1.5cm, decodearray={0.0 1.25 0 1.25 0.0 1.25}]{artwork/teaser/data/munster_000118_000005_leftImg8bit.jpg}}};
    \end{scope}
    \node[white] at (-1.7, -0.05) {\textbf{\contour{black}{Input Video}}\vphantom{p}};

    \draw[thick, ->] (-0.775, 0.0) -- (-0.375, 0.0);
    \node at (0.6, 0.15)  {\frame{\includegraphics[width=1.5cm]{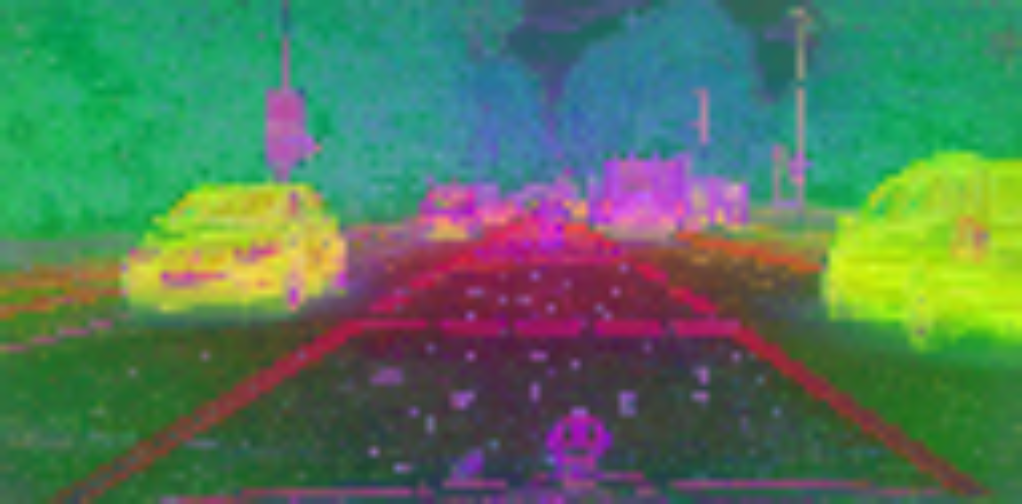}}};
    \node at (0.65, 0.10)  {\frame{\includegraphics[width=1.5cm]{artwork/teaser/data/munster_000118_000005_leftImg8bit_features.png}}};
    \node at (0.7, 0.05)  {\frame{\includegraphics[width=1.5cm]{artwork/teaser/data/munster_000118_000005_leftImg8bit_features.png}}};
    \node[white] at (0.0, 0.1) {\rotatebox[origin=c]{90}{\textbf{\contour{black}{Features}}}};
    \node at (1.35, 0.05)  {\frame{\includegraphics[width=1.5cm]{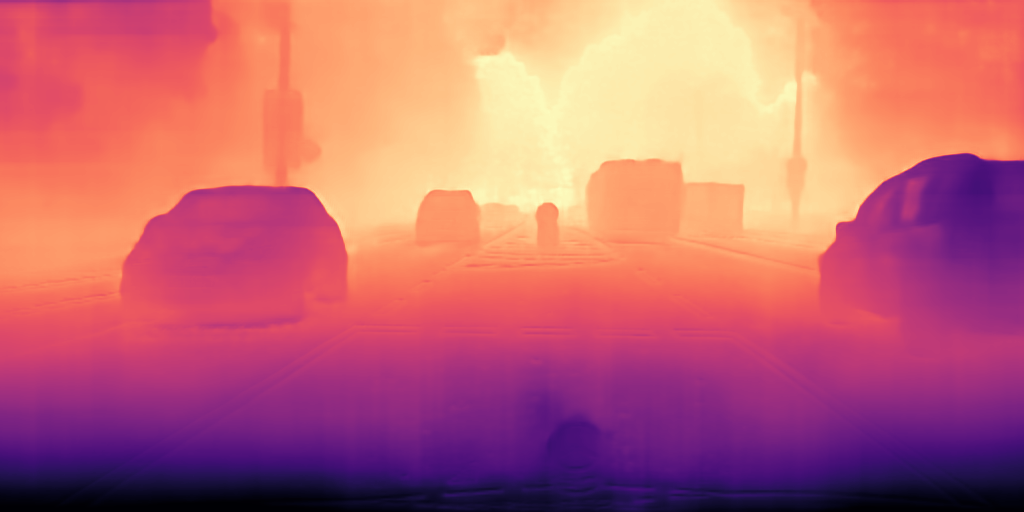}}};
    \node at (1.4, 0.0)  {\frame{\includegraphics[width=1.5cm]{artwork/teaser/data/munster_000118_000005_leftImg8bit_disparity.png}}};
    \node at (1.45, -0.05)  {\frame{\includegraphics[width=1.5cm]{artwork/teaser/data/munster_000118_000005_leftImg8bit_disparity.png}}};
    \node[white] at (0.75, -0.05) {\rotatebox[origin=c]{90}{\textbf{\contour{black}{Depth}}}};
    \node at (2.1, -0.05) {\frame{\includegraphics[width=1.5cm]{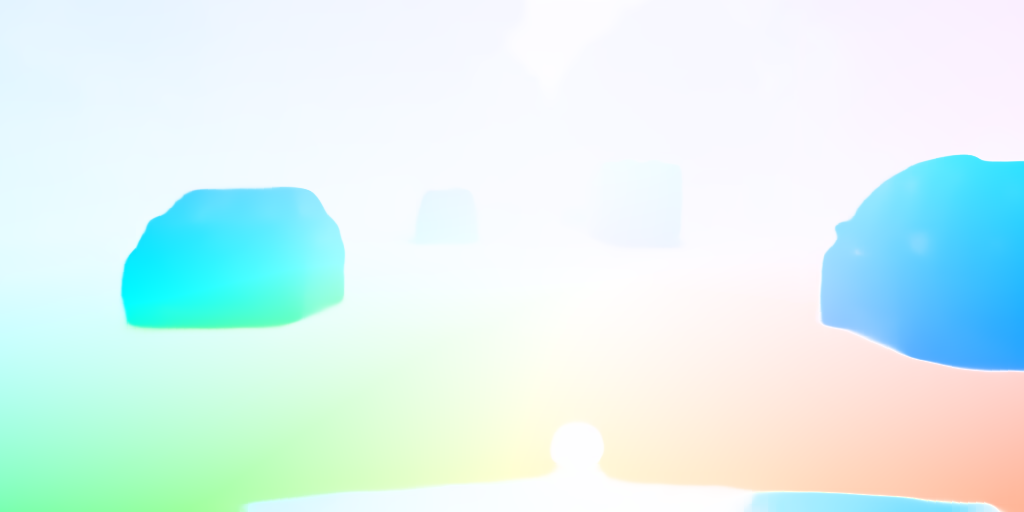}}};
    \node at (2.15, -0.10) {\frame{\includegraphics[width=1.5cm]{artwork/teaser/data/munster_000118_000005_leftImg8bit_optical_flow_forward.png}}};
    \node at (2.2, -0.15) {\frame{\includegraphics[width=1.5cm]{artwork/teaser/data/munster_000118_000005_leftImg8bit_optical_flow_forward.png}}};
    \node[black] at (1.5, -0.15) {\rotatebox[origin=c]{90}{\textbf{Motion}}};

    \draw[thick, ->] (3.15, 0) -- (3.55, 0);
    \draw[black, rounded corners=1, fill=gray!10] (3.65, 0.05)  rectangle ++(2.05, 0.4565) node[pos=.5] {\textbf{Instance Labeling}};
    \draw[black, rounded corners=1, fill=gray!35] (3.65,-0.05)  rectangle ++(2.05,-0.4565) node[pos=.5] {\textbf{Semantic Labeling}};
    \draw[black, rounded corners=1, fill=gray!70] (5.8,0.5065)  rectangle ++(1.0,-1.011) node[pos=.5] {\shortstack{\textbf{Temp.}\\\textbf{Tracking}}};
    \draw[thick, ->] (6.9, 0) -- (7.3, 0);

    \begin{scope}[xshift=8.3cm, yshift=-0.075cm]
      \node at (-0.1, 0.1)   {\frame{\includegraphics[width=1.5cm, decodearray={0.0 1.25 0 1.25 0.0 1.25}]{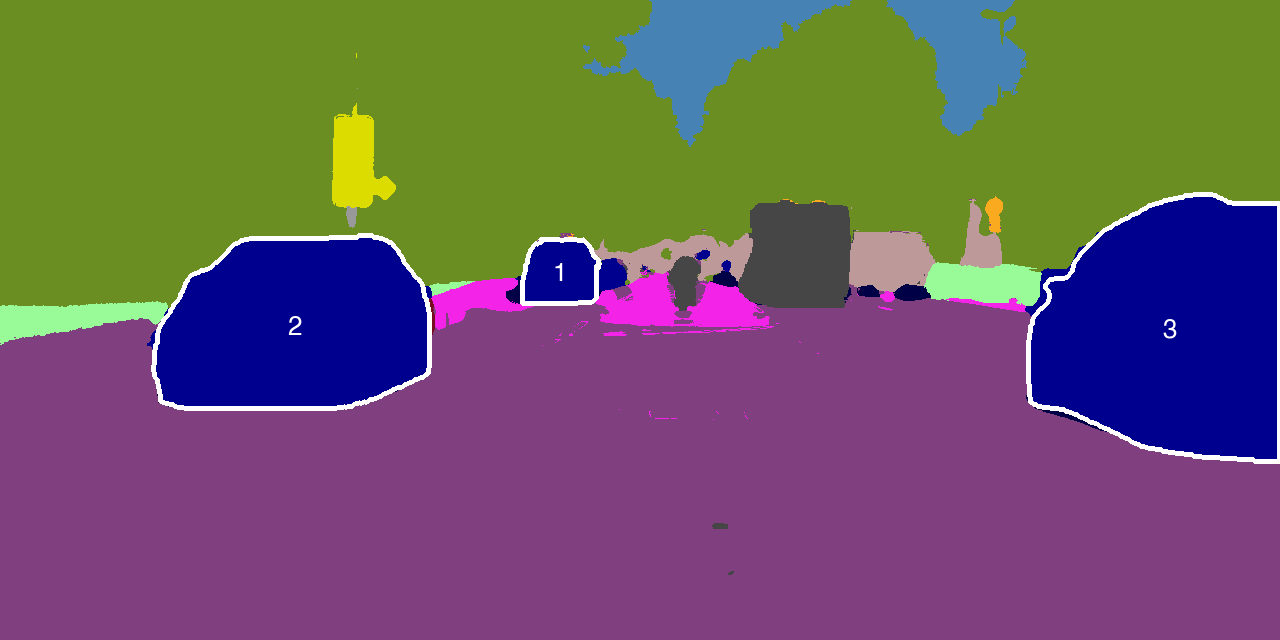}}};
      \node at (-0.05, 0.05) {\frame{\includegraphics[width=1.5cm, decodearray={0.0 1.25 0 1.25 0.0 1.25}]{artwork/teaser/data/munster_000118_000005_leftImg8bit_pred.png}}};
      \node at (0, 0)        {\frame{\includegraphics[width=1.5cm, decodearray={0.0 1.25 0 1.25 0.0 1.25}]{artwork/teaser/data/munster_000118_000005_leftImg8bit_pred.png}}};
    \end{scope}
    \node[white, align=center, text width=1.5cm] at (8.3, 0) {
    \shortstack{
    \textbf{\contour{black}{Panoptic}}\\[-2pt]
    \textbf{\contour{black}{Pseudo-Labels}}
    }
    };

    \draw[thick, ->] (9.25, 0) -- (9.65, 0);
    \draw[fill=white, draw=none] (10.15, 0) circle [radius=0.2cm] node {\scriptsize$\!\!\mathcal{L}_{\text{VDrop}}$};
    \draw[fill=white, white]     (10.9, 0) circle [radius=0.2cm];
    \draw[thick, <-] (10.55, 0) -- (10.95, 0);

    \begin{scope}[xshift=-0.05cm]    
    \pic [local bounding box=our_model] at (12.25, 0) {double_trapezium_small};
    \node[text=black, xshift=-0.025cm] at (our_model) {\textbf{VPS Model}};

    \draw[thick, <-] (13.45, 0) -- (13.85, 0);

    \begin{scope}[xshift=14.925cm, yshift=-0.075cm]
      \node at (-0.1, 0.1)   {\frame{\includegraphics[width=1.5cm, decodearray={0.0 1.25 0 1.25 0.0 1.25}]{artwork/teaser/data/munster_000118_000005_leftImg8bit.jpg}}};
      \node at (-0.05, 0.05) {\frame{\includegraphics[width=1.5cm, decodearray={0.0 1.25 0 1.25 0.0 1.25}]{artwork/teaser/data/munster_000118_000005_leftImg8bit.jpg}}};
      \node at (0, 0)        {\frame{\includegraphics[width=1.5cm, decodearray={0.0 1.25 0 1.25 0.0 1.25}]{artwork/teaser/data/munster_000118_000005_leftImg8bit.jpg}}};
    \end{scope}
    \node[white] at (14.925, -0.05) {\textbf{\contour{black}{Input Video}}\vphantom{p}};

    \end{scope}
  \end{tikzpicture}
\end{center}

\vspace{-2.5em}

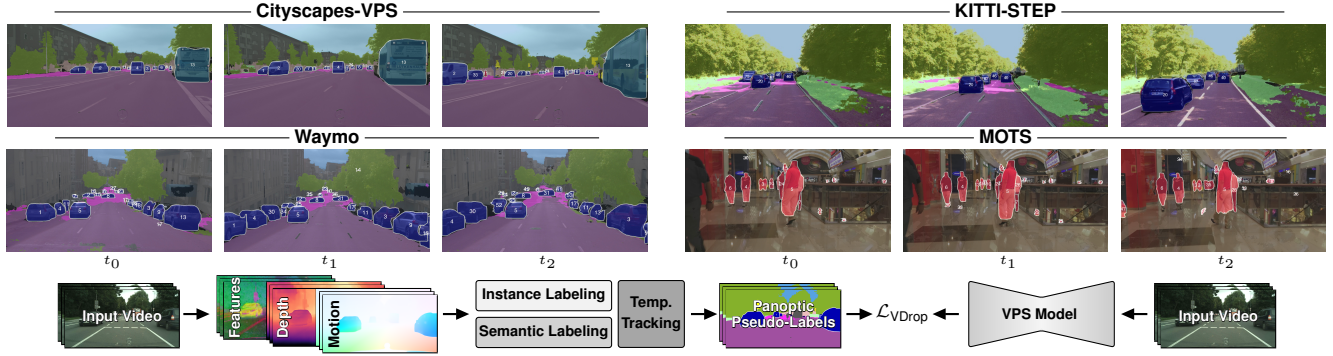
\captionof{figure}{\textbf{Results and overview of our unsupervised video panoptic segmentation approach \MethodName.} \textit{Top:} Panoptic video predictions by \MethodName across four datasets. \textit{Bottom:} We use self-supervised representations, motion, and depth cues from monocular videos to generate scene-centric panoptic video pseudo-labels and train a video panoptic segmentation model using a novel Video DropLoss.
\label{fig:teaser}}
\vspace{1.5\baselineskip}
}]
\begin{abstract}
Video panoptic segmentation (VPS) aims to jointly detect, segment, and track all objects while partitioning the video into semantically consistent regions. We introduce the task setting of \emph{unsupervised} VPS, omitting any human supervision. Existing unsupervised scene understanding works mainly focused on image segmentation tasks; the video domain remains underexplored. We propose \MethodName, the first unsupervised VPS approach. \MethodName generates temporally consistent panoptic video pseudo-labels from scene-centric videos by exploiting unsupervised depth, motion, and visual cues. Training on these pseudo-labels using a novel Video DropLoss yields an accurate, unsupervised VPS model. To benchmark progress, we introduce a comprehensive evaluation protocol and four competitive baselines, extending state-of-the-art unsupervised panoptic image and instance video segmentation models to VPS. \MethodName outperforms all baselines and demonstrates strong label-efficient learning. With \MethodName, our evaluation protocol, and baselines, we provide a strong foundation for future research on unsupervised VPS.
\end{abstract}
\thispagestyle{firststyle}
\section{Introduction}
Video Panoptic Segmentation (VPS)~\cite{Kim:2020:VPS, Weber:2021:STE} is a holistic scene understanding task that extends panoptic segmentation~\cite{Kirillov:2019:PS} from the spatial to the spatio-temporal domain, unifying instance and semantic segmentation over time. Specifically, VPS aims to detect, segment, classify, and temporally associate individual object instances, while also assigning each pixel a semantic category.
This comprehensive segmentation task enables parsing of complex, dynamic real-world environments and has a wide range of applications, such as autonomous driving, robotics, video editing, and medical imaging~\cite[see][for an overview]{Zhou:2024:ISF, Xie:2025:SVS}.

Advances in panoptic video understanding have been driven by supervised learning, relying on significant amounts of human-annotated data for training~\cite{Kim:2020:VPS, Weber:2021:STE, Cheng:2021:VMF, Li:2022:VKN, Qiao:2021:VDL, Elharrouss:2021:PSR, Shin:2024:VPS}. However, acquiring dense pixel-level instance and semantic annotations for images is highly resource intensive~\cite{Cordts:2016:CDS}. Extending labeling efforts to the temporal domain poses even more challenges, including limited scalability and label quality~\cite{Weber:2021:STE, Choudhuri:2023:CAR, Woo:2021:LAS}. Despite the success of densely annotated large-scale datasets, such as SA-1B~\cite{Kirillov:2023:SAM}, there is a natural interest in more efficient and scalable annotation-free alternatives~\cite{He:2022:MAE, Wang:2024:USA, Hahn:2025:UPS, Simeoni:2025:DIN}.

Unsupervised learning has emerged as a powerful paradigm, showcasing significant progress in scene understanding tasks such as unsupervised semantic~\cite{Hamilton:2022:USS, Sick:2024:USS, Hahn:2024:BUS}, instance~\cite{Wang:2023:CAL, Wang:2024:USA, Sick:2025:CCS}, and panoptic segmentation~\cite{Niu:2024:UUI, Hahn:2025:UPS}. 
Among these, U2Seg~\cite{Niu:2024:UUI} established the first approach for unsupervised image panoptic segmentation by combining semantic pseudo-labels from STEGO~\cite{Hamilton:2022:USS} and instance masks from CutLER's MaskCut component~\cite{Wang:2023:CAL}. CUPS~\cite{Hahn:2025:UPS} extended the paradigm to scene-centric%
\footnote{Scene-centric imagery captures complex environments with multiple interacting objects, as in Cityscapes~\cite{Cordts:2016:CDS}, whereas object-centric imagery typically depicts a single and isolated object, as in ImageNet~\cite{Russakovsky:2015:ILS}.}
imagery by using motion and depth cues from real-world \emph{stereo video} to generate panoptic pseudo-labels for \emph{images}, overcoming the need for object-centric imagery required for training U2Seg. These advances have focused on unsupervised \emph{image} segmentation. 
In contrast, the unsupervised panoptic segmentation of \emph{videos} remains underexplored, while offering broader applications for dynamic real-world environments and a more natural human perception of the world by also perceiving the temporal domain. We therefore introduce the task setting of \emph{unsupervised video panoptic segmentation}, which aims to panoptically segment videos without any human supervision.

To approach unsupervised VPS for the first time, we introduce \textbf{\MethodName}: scene-\textbf{C}entric \textbf{U}nsupervised \textbf{Video} \textbf{P}anoptic \textbf{S}egmentation. %
While building on CUPS~\cite{Hahn:2025:UPS}, which addresses unsupervised \emph{image} panoptic segmentation, \MethodName directly produces temporally consistent \emph{video} panoptic pseudo-labels. Additionally, our VPS pseudo-labeling method operates solely on monocular videos as input, thereby overcoming the need for stereo video during VPS training. To enable effective training on our pseudo-labels, we introduce a Video DropLoss and self-enhanced video copy-paste augmentation. For assessing the accuracy of \MethodName, we establish four competitive baselines, built using CUPS~\cite{Hahn:2025:UPS}, U2Seg~\cite{Niu:2024:UUI}, and VideoCutLER~\cite{Wang:2024:VCS}. 
\MethodName, together with the proposed baselines and evaluation protocol, forms a foundation for future work on unsupervised panoptic video understanding.

Specifically, we make the following contributions:
\emph{(i)}~We introduce the task setting of unsupervised video panoptic segmentation and propose a unified evaluation protocol spanning four established VPS datasets. 
To enable comparison, we extend the Segmentation and Tracking Quality (STQ) to the unsupervised setting by incorporating pseudo-semantic matching.
Moreover, we construct four competitive VPS baselines that combine state-of-the-art unsupervised semantic, video instance, and panoptic image segmentation models with unsupervised tracking.
\emph{(ii)}~We generate high-quality video panoptic pseudo-labels solely from monocular scene-centric videos using self-supervised visual, depth, and motion cues. Using a novel Video DropLoss and self-enhanced video copy-paste augmentation, we train on our pseudo-labels, leading to the first unsupervised VPS approach.
\emph{(iii)}~\MethodName consistently outperforms all unsupervised baselines across a wide range of scene-centric video datasets. Additionally, we show that \MethodName provides a strong foundation for approaching VPS using label-efficient learning.

\section{Related Work \label{sec:related_work}}

Unsupervised segmentation methods have been shaped by advances in self-supervised learning (SSL) and unsupervised low-level vision, particularly in motion and depth estimation. We first review these developments before discussing unsupervised segmentation approaches.

\inparagraphnospace{Self-supervised representation learning} aims to learn expressive and transferable visual representations from unlabeled data~\cite{Ericsson:2022:SSL}. 
A variety of pretext tasks have been proposed to achieve this~\cite{Albelwi:2022:SSL, Ericsson:2022:SSL}, enabling feature extractors that generalize across downstream tasks~\cite{Oquab:2023:DLR, Simeoni:2025:DIN}.
The advent of Vision Transformers (ViTs)~\cite{Dosovitskiy:2021:AIW} has further shaped SSL by facilitating large-scale training and enabling novel pretext designs~\cite{Zhou:2022:IBT, He:2022:MAE}. 
Contemporary methods typically optimize ViTs through contrastive learning~\cite{Bachman:2019:LRM, Chen:2020:ISL, He:2020:MCU, Chen:2021:SSL}, negative-free objectives~\cite{Bardes:2022:VIL, Caron:2021:EPS, Chen:2021:SSR, Grill:2020:BYL}, clustering~\cite{Asano:2020:SLC, Caron:2018:DCL, Caron:2020:ULV, Venkataramanan:2025:FRA}, masked modeling~\cite{He:2022:MAE, Gupta:2023:SMA, Nguyen:2024:RMA}, or a combination of these~\cite{Zhou:2022:IBT, Oquab:2023:DLR, Simeoni:2025:DIN}. 
Recent SSL frameworks, such as the DINO family~\cite{Caron:2021:EPS, Oquab:2023:DLR, Simeoni:2025:DIN}, provide semantically rich, dense features, suited for unsupervised segmentation~\cite{Hamilton:2022:USS, Wang:2023:CAL}.

\inparagraphnospace{Unsupervised optical flow} aims to estimate apparent motion directly from video without ground truth~\cite{Ahmadi:2016:UFL, Yu:2016:UFL}. While classical formulations were inherently unsupervised~\cite{Horn:1981:DOF, Lucas:1981:IIR, Barron:1994:POT}, early deep learning approaches relied on synthetic datasets to provide supervision~\cite{Dosovitskiy:2015:FN, Mayer:2016:OFD, Sun:2018:PWC}. Inspired by traditional formulations and motivated by the synthetic-to-real gap, deep learning-based unsupervised optical flow has been introduced~\cite{Yu:2016:UFL, Ren:2017:UFL, Ahmadi:2016:UFL, Meister:2018:ULO, Jonschkowski:2020:UFL}. Recent unsupervised deep optical flow methods provide accurate motion estimation, efficient inference, and strong generalization across diverse real-world domains~\cite{Stone:2021:SST, Marsal:2023:BFL, Lifshitz:2024:UOP}.

\inparagraphnospace{Unsupervised monocular depth estimation} aims to estimate depth of monocular imagery by learning from stereo images or monocular videos~\cite{Godard:2017:UDE, Garg:2016:UDE, Zhou:2017:ULD}. Learning depth from monocular videos is done by novel-view synthesis and photometric consistency~\cite{Zhou:2017:ULD, Vijayanarasimhan:2017:SNL}. Novel-view synthesis, however, assumes a static scene and breaks for dynamics~\cite{Yang:2021:LTS, Godard:2019:DIS, Sun:2023:DDF}. Recent approaches use auto-masking~\cite{Godard:2019:DIS}, semantic/instance cues~\cite{Feng:2022:DOM, Guizilini:2020:SGD, Klingner:2020:SSD, Casser:2019:DPS, Casser:2019:UMD, Lee:2021:LMD, Wimbauer:2021:MRC}, or multi-view~\cite{Woo:2024:PRD, Feng:2022:DOM, Watson:2021:TOP} to compensate for dynamic objects. Other methods, such as DynamoDepth~\cite{Sun:2023:DDF}, jointly learn depth, motion, and/or motion segmentation, decomposing the scene into static and dynamic parts~\cite{Li:2021:UMD, Hui:2022:RMD}.

\inparagraphnospace{Unsupervised instance segmentation} aims to detect and segment objects in images without human supervision~\cite{Simeoni:2024:UOL}. Recent approaches~\cite{Wang:2022:FLS, Gansbeke:2022:DOM, Wang:2023:CAL, Wang:2024:USA, Sun:2024:MLM} train class-agnostic detectors using pseudo-labels derived from SSL features of object-centric imagery. TokenCut~\cite{Wang:2022:TSO} obtains foreground masks from DINO features using normalized cuts~\cite{Shi:2000:NCI}. CutLER~\cite{Wang:2023:CAL} extends this by iteratively cutting multiple pseudo-masks per image, and is further improved by \cite{Wang:2024:USA, Arica:2024:CEU, Sick:2025:CCS}. A complementary direction exploits motion cues for object discovery~\cite{Yang:2021:DYS, Liu:2021:TEO, Choudhury:2022:GWM, Karazija:2022:UMS, Safadoust:2023:MOD, Sun:2024:MLM, Sommer:2022:SCS, Gong:2025:MRD}. 
Recently, unsupervised extensions to video have emerged. VideoCutLER~\cite{Wang:2024:VCS} trains on synthetic videos from image pseudo-masks, FlowCut~\cite{Sari:2025:FUV} enforces motion-based temporal consistency, and AutoQ-VIS~\cite{Lu:2025:AQI} improves pseudo-labels via automatic quality assessment.

\inparagraphnospace{Unsupervised semantic segmentation} aims to divide images into semantically meaningful regions without any human annotations. Early deep learning approaches~\cite{Ji:2019:IIC, Cho:2021:PUS, Harb:2021:IFS} used representation learning, encouraging embeddings to capture dense semantic similarity. 
Leveraging self-supervised DINO~\cite{Caron:2021:EPS} features as an inductive bias, STEGO~\cite{Hamilton:2022:USS} distills and clusters features to obtain unsupervised semantic segmentations. Building on the STEGO framework, subsequent methods~\cite{Seong:2023:LHP, Kim:2024:EAL, Sick:2024:USS, Hahn:2024:BUS, Jevtic:2025:SCD} refine the distillation and probing process. Other unsupervised segmentation approaches~\cite{Tian:2024:DAS, Namekata:2024:EMR, Couairon:2024:DCT} alternatively use vision-language diffusion features. To the best of our knowledge, there are no extensions of unsupervised semantic segmentation methods to video to date.

\inparagraphnospace{Unsupervised panoptic segmentation} has recently emerged as a natural next step, following advances in unsupervised semantic and instance segmentation. While panoptic segmentation of images and videos has been extensively studied in the supervised setting \cite[see][for an overview]{Elharrouss:2021:PSR, Zhou:2024:ISF}, we are only aware of two unsupervised image panoptic segmentation approaches, U2Seg~\cite{Niu:2024:UUI} and CUPS~\cite{Hahn:2025:UPS}. U2Seg combines CutLER's MaskCut~\cite{Wang:2023:CAL} and STEGO~\cite{Hamilton:2022:USS} to create pseudo-labels for panoptic training, but inherits MaskCut’s object-centric bias, significantly limiting accuracy on scene-centric data~\cite{Hahn:2025:UPS}. CUPS overcomes this by grouping unsupervised scene flow from stereo into rigid instances~\cite{Sommer:2022:SCS} and combining these with unsupervised semantics~\cite{Sick:2024:USS} to train a panoptic network. In our work, we employ both U2Seg and CUPS in competitive baselines and propose the first approach to directly perform unsupervised video panoptic segmentation. While we, similar to CUPS~\cite{Hahn:2025:UPS}, use self-supervised representations, motion, and depth cues, \MethodName requires only monocular video for VPS pseudo-labeling, captures non-rigid instance motions, and directly generates panoptic video pseudo-labels.

\section{Method: Unsupervised VPS}\label{sec:method}
\emph{First}, we generate panoptic video pseudo-labels (\cf \cref{subsec:pseudo_label_generation,fig:pseudo_label_generation}) from monocular videos.
\emph{Second}, we train a VPS model (\cf \cref{subsec:learn_vups}) using these pseudo-labels, a novel Video DropLoss, and self-enhanced video copy-paste augmentations, leading to the first unsupervised VPS model. \emph{Third}, to enable evaluation of \MethodName and future approaches, we present an evaluation protocol for the unsupervised VPS setting (\cf \cref{subsec:eval_vups}).

\subsection{Generating VPS pseudo-labels}
\label{subsec:pseudo_label_generation}
\begin{figure*}[t]
    \centering   
    \vspace{-0.2em} %
    \input{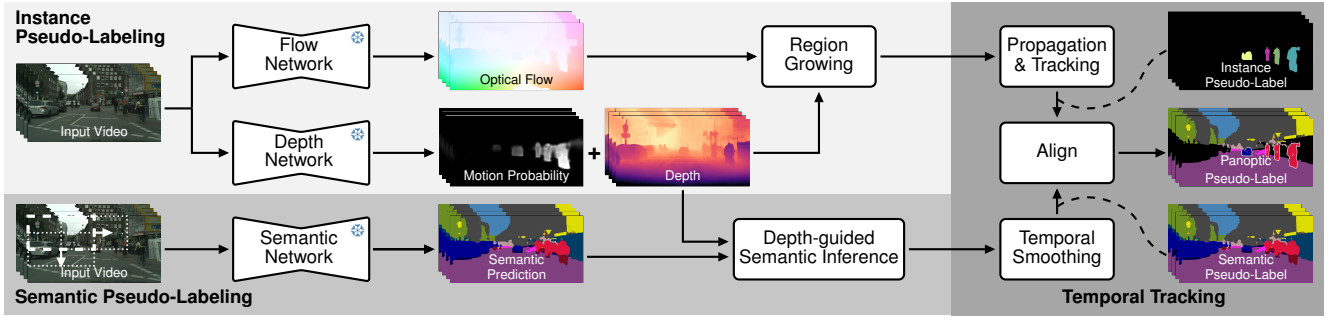}
    \vspace{-0.7em}
    \caption{\textbf{\MethodName pseudo-label generation.} \emph{Instance pseudo-labeling} applies motion-based region growing segmentation using unsupervised optical flow from SMURF~\cite{Stone:2021:SST} and depth from DynamoDepth~\cite{Sun:2023:DDF}. \emph{Semantic pseudo-labeling} uses a $k$-means clustering prediction of distilled DINO features \cite{Caron:2021:EPS}, combined with a depth-guided inference \cite{Hahn:2025:UPS}. \emph{Temporal tracking} propagates and tracks the instance pseudo-labels, temporally smoothes the semantic pseudo-labels, and aligns the two signals into video panoptic pseudo-labels. \label{fig:pseudo_label_generation}}
    \vspace{-0.5em}
\end{figure*}

To generate temporally coherent panoptic video pseudo-labels, we adopt a bottom-up strategy (\cf \cref{fig:pseudo_label_generation}). Initially, we produce semantic and instance pseudo-labels for individual frames, which are then refined through temporal consistency processing along the video sequence.

\inparagraph{From motion and depth to instance pseudo-labels.} 
Drawing inspiration from Gestalt principles~\cite{Wertheimer:1912:ESU, Koffka:1935:PGP, Kohler:1967:GES}, we adopt the common fate, proximity, and similarity principle---neighborhoods that move together belong together---to derive class-agnostic instance pseudo-masks from monocular videos.
Accordingly, we defined objects as entities capable of moving. %
We obtain \emph{per-frame} instance pseudo-labels across an entire video clip as follows. 
Given two consecutive monocular frames, we obtain unsupervised optical flow $\mathbf{f}\in\mathbb{R}^{2\times \rm H \times \rm W}$ using SMURF~\cite{Stone:2021:SST}. Monocular depth $\mathbf{d}\in\mathbb{R}^{\rm H \times \rm W}$ is estimated by DynamoDepth~\cite{Sun:2023:DDF}. Alongside depth, DynamoDepth also estimates dense motion probabilities $\mathbf{m}\in [0,1]^{\rm H \times \rm W}$, decomposing the scene into static ($m_{h,w}\to 0$) and dynamic regions ($m_{h,w}\to 1$). 

We employ a variant of region growing~\cite{Adams:1994:REG, Haralick:1985:IST} to extract a variable number of instance pseudo-masks. Specifically, we threshold $\mathbf{m}$ at $\alpha=0.15$ to obtain instance seeds. Next, we iteratively merge pixels within a Chebyshev neighborhood $r$ based on their relative depth and flow difference. In particular, for pixel $\mathbf{x} = (h, w)$ with $m_{\mathbf{x}} > \alpha$, we merge pixels within the Chebyshev neighborhood $\mathcal{N}_{r}(\mathbf{x})=\{\mathbf{y}\ |\  ||\mathbf{y}-\mathbf{x}||_{\infty} \leq r\,\wedge\,m_{\mathbf{y}} > \alpha,\, \mathbf{y} \neq \mathbf{x}\}$ to $\mathbf{x}$ if
\begin{equation}
    \frac{|d_{\mathbf{x}} - d_{\mathbf{y}}|}{|d_{\mathbf{x}}|}<\tau_{d}\quad\text{and}\quad\frac{||f_{\mathbf{x}} - f_{\mathbf{y}}||_{2}}{||f_{\mathbf{x}}||_{2}}<\tau_{f},
\end{equation}
with $\mathbf{y}\in\mathcal{N}_{r}(\mathbf{x})$. Merging proceeds iteratively until convergence and can be parallelized for efficiency. 
The resulting set of $l$ class-agnostic pseudo-instance masks $\mathbf{M} \in \{0,1\}^{l\times \rm H \times \rm W}$ groups pixels that share consistent relative depth and motion. Unlike the rigid-motion pseudo-labeling in CUPS~\cite{Hahn:2025:UPS}, we do not assume rigidity but exploit smoothness, enabling us to also capture non-rigidly moving instances, such as pedestrians in motion (\cf \cref{fig:qualitative_pseudolabels}).

\inparagraph{From SSL features to semantic pseudo-labels.\label{sec:semanticpseudolabeling}}
We derive an unsupervised semantic segmentation model $\mathcal{S}$ by distilling DINO~\cite{Caron:2021:EPS} features into a lower-dimensional embedding via a contrastive objective, leveraging monocular depth as an auxiliary cue. 
Clustering with stochastic cosine-distance $k$\nobreakdash-means yields
$\mathcal{S}: \mathbb{R}^{3\times \rm H\times \rm W} \rightarrow \{0,1\}^{\rm c_p\times \rm H\times \rm W}$,
mapping an input image $\mathbf{I}$ to dense semantic pseudo-labels with $\rm c_p$ semantic pseudo-classes, consistent across the entire dataset.
While unsupervised semantic segmentation approaches typically operate at low resolutions (\eg, $\text{\num{320}}^{\text{\num{2}}}$), close to that used for SSL pre-training \cite{Caron:2021:EPS}, we use depth-guided semantic inference \cite{Hahn:2025:UPS} to obtain high-resolution semantic predictions.
Specifically, we infer a semantic prediction $\mathbf{P}^\text{low}$ at lower resolution and $\mathbf{P}^\text{high}$ at a higher resolution using sliding-window inference. $\mathbf{P}^\text{low}$ captures coarse, near-field semantics and $\mathbf{P}^\text{high}$ preserves fine details via sliding-window inference and soft aggregation.
$\mathbf{P}^\text{low}$ is upsampled to the resolution of $\mathbf{P}^\text{high}$ and both are fused with a per-pixel depth weight $\alpha_{h,w} = (d_{h,w} + 1)^{-1}$ using the monocular depth prediction $\mathbf{d}$ from DynamoDepth:%
\begin{equation}
    \mathbf{P}^\ast = \boldsymbol{\alpha} \odot \mathbf{P}^\text{low} + (1-\boldsymbol{\alpha}) \odot \mathbf{P}^\text{high}.
\end{equation}%
We further apply regularized Frank-Wolfe inference \cite{Huu:2021:RFD} for dense CRFs \cite{Kraehenbuehl:2011:EIF}, enabling fast spatial regularization.
Building on~\cite{Hahn:2025:UPS}, we adapt DepthG~\cite{Sick:2024:USS} retrained with the monocular depth from DynamoDepth~\cite{Sun:2023:DDF}, ensuring consistency with our unsupervised, monocular setting.

\inparagraph{From image to video pseudo-labels.}
A key component of our pseudo-label generation is the temporal processing and %
fusion of frame-wise semantic and instance information.

\emph{Instance propagation and tracking} extends the frame-wise, class-agnostic instance pseudo-labels to the video domain using
optical-flow-based mask propagation and IoU-based association.
Given three consecutive frames $\mathbf{I}_{t-1}$, $\mathbf{I}_{t}$, and $\mathbf{I}_{t+1}$,
we estimate the forward and backward optical flows
$\mathbf{f}^{\rm fw}_{t-1,t}$, $\mathbf{f}^{\rm bw}_{t-1,t}$ and
$\mathbf{f}^{\rm fw}_{t,t+1}$, $\mathbf{f}^{\rm bw}_{t,t+1}$, using SMURF.
We then perform backward warping of the instance pseudo-labels
$\mathbf{M}_{t}$ and $\mathbf{M}_{t+1}$ to their respective previous frames, resulting in $\hat{\mathbf{M}}_{t \veryshortrightarrow t-1}$ and $\hat{\mathbf{M}}_{t+1 \veryshortrightarrow t}$
while ignoring occluded pixels identified via forward–backward flow consistency~\cite{Sundaram:2010:OCC}.
To match the instance IDs between frame $t-1$ and $t$, we compute an IoU cost matrix between
all instance masks in $\mathbf{M}_{t-1}$ and $\hat{\mathbf{M}}_{t \veryshortrightarrow t-1}$.
Hungarian matching is applied to all pairs with $\mathrm{IoU} > \tau_{m}=0.4$,
and the resulting associations are used to update the instance IDs in $\mathbf{M}_{t}$.
For instances in $\mathbf{M}_{t-1}$ without a match, we attempt recovery using $\hat{\mathbf{M}}_{t+1 \veryshortrightarrow t}$ under the same threshold $\tau_{m}$,
thereby resurrecting instances lost in frame $t$ by warping back masks from $t+1$.
The remaining masks in $\mathbf{M}_{t+1 \veryshortrightarrow t}$ are assigned new IDs,
and the temporal window advances by one frame until the end of the video clip.
Finally, we filter out short-lived instances that appear in less than \num{2} frames of the clip.

\emph{Temporal semantic smoothing} enforces temporal consistency of semantic pseudo-labels by aggregating neighboring predictions.
For each frame $t$, we obtain $\hat{\mathbf{P}}_{t-1\veryshortrightarrow t}$ and $\hat{\mathbf{P}}_{t+1\veryshortrightarrow t}$, the warped pseudo-labels from adjacent frames using flow.
The temporally smoothed label $\tilde{\mathbf{P}}_{t}$ is obtained via pixel-wise majority vote over 
$\{\hat{\mathbf{P}}_{t-1\veryshortrightarrow t},\, \mathbf{P}_{t}^*,\, \hat{\mathbf{P}}_{t+1\veryshortrightarrow t}\}$ 
using a three-frame sliding window.

\emph{Aligning semantic and instance pseudo-labels} per video clip results in the final video panoptic pseudo-labels.
We align the semantic and instance signals by assigning a consistent semantic pseudo-class to all masks of an instance ID across an entire clip, determined by a majority vote over all semantic pseudo-labels within the instance masks. 

Once all video panoptic pseudo-labels are obtained, we aim to retrieve the split of the semantic pseudo-classes into pseudo ``thing'' and ``stuff'' classes.
We aggregate pixel distributions across all clips by computing the ratio of each semantic pseudo-class frequency within the instance masks relative to its overall frequency.
We designate semantic pseudo-classes with a high ratio above a threshold $\psi^{\rm ts}$ as ``thing'', and those below as ``stuff''.

\subsection{Learning from VPS pseudo-labels\label{subsec:learn_vups}}
Using our panoptic video pseudo-labels, we aim to train a model to perform unsupervised VPS. 
In particular, given an input video of $\rm T$ frames, the model predicts 
a panoptic video segmentation $\mathbf{P}=(\mathbf{S},\mathbf{R})$, composed of the predicted pseudo-classes $\mathbf{S}\in\{1, 2,\ldots,c_{\text{p}}\}^{\rm T \times \rm H\times W}$ and $n_{p}$ binary video instance masks $\mathbf{R}\in\{0, 1\}^{n_{p} \times \rm T \times \rm H\times W}$ for ``thing'' object instances. 
Since our pseudo-labels capture only moving “thing’’ instances (\eg, moving cars), we train the VPS model sparsely to generalize to static objects (\eg, parked cars). We introduce a Video DropLoss, extending the DropLoss~\cite{Wang:2023:CAL} to video, and a self-enhanced video copy-paste augmentation to improve small-object detection.

\inparagraph{Video DropLoss.} Given two consecutive video frames of our pseudo-labeled clips, we infer ``thing'' video instance detections $\mathbf{D}_{j}$ (masks \& semantic class) with their tracking latent representation $\mathbf{E}_{j}$ from our model. Given a sparse set of pseudo ``thing'' video instance labels $\hat{\mathbf{D}}_{i}$ (masks \& pseudo-class, derived from $\mathbf{M}$ \& $\tilde{\mathbf{P}}$) and their track $\hat{{\rm id}}_{i}$, we supervise ``thing'' detections with our Video DropLoss: %
\begin{equation}\label{eq:droploss}
     \mathcal{L}_{\text{VDrop}}=\mathds{1}{\mkern-1.5mu}\big(\operatorname{IoU}_j^{\max}>\tau_{\operatorname{IoU}}\big)\mathcal{L}_{\text{d}}(\mathbf{D}_{j}, \hat{\mathbf{D}}_{i})\mathcal{L}_{\text{t}}(\mathbf{E}_{j}, \hat{{\rm id}}_{i}),
\end{equation}
where $\mathcal{L}_{\text{t}}$ denotes the tracking loss~\cite{Yang:2019:VIS} and $\mathcal{L}_{\text{d}}$ the ``thing'' detection loss~\cite{Kirillov:2019:PS}. This Video DropLoss pseudo-supervises only ``thing'' instance predictions $\mathbf{D}_{j}$ and their tracking representation $\mathbf{E}_{j}$ that sufficiently overlap with a pseudo-instance $\hat{\mathbf{D}}_{i}$ (\ie $\operatorname{IoU}_j^{\max}>\tau_{\operatorname{IoU}}$). Our Video DropLoss enables learning from our sparse pseudo-labels while providing the freedom to 
predict objects and their tracks that are not covered by our pseudo-labels (\eg, static objects). Semantics of ``stuff'' regions are supervised using a standard cross-entropy loss.

\inparagraph{Self-enhanced video copy-paste augmentation.} To improve the ``thing'' detection and tracking accuracy of the VPS model on small objects, we introduce a self-enhanced video copy-paste augmentation. Copy-pasting instance masks~\cite{Dwibedi:2017:CPA, Ghiasi:2021:CPA, Dvornik:2018:MVA} has been shown to be particularly effective when training with sparse pseudo-labels~\cite{Wang:2023:CAL, Sick:2025:CCS, Hahn:2025:UPS}. Instead of copy-pasting instance masks derived from the pseudo-labels onto another image for augmentation, CUPS~\cite{Hahn:2025:UPS} has shown that it is beneficial to derive the instance mask from the model's prediction itself. The intuition behind this is that the network gradually discovers more ``thing'' objects than captured by the pseudo-labels. We extend this idea to the video domain. In particular, given a training batch, we perform inference and extract confident ``thing'' video instances from the model's VPS prediction. We apply random scaling and horizontal flipping to the video instance masks and paste the augmented masks into clips of the training batch. We paste masks using random trajectories, ensuring diverse motion patterns. Finally, we train our model on the batch of augmented clips.

\subsection{Unsupervised VPS evaluation protocol\label{subsec:eval_vups}}
In the absence of supervision, our predicted semantic pseudo-classes do not align with the ground-truth semantic class IDs~\cite{Cho:2021:PUS, Hamilton:2022:USS, Seong:2023:LHP, Kim:2024:EAL, Sick:2024:USS, Hahn:2024:BUS, Hahn:2025:UPS, Jevtic:2025:SCD}. Therefore, a mapping between pseudo and ground-truth categories is required before using standard evaluation metrics. We present a simple, hyperparameter-free matching strategy for aligning the pseudo-classes while strictly preserving the separation between ``thing'' and ``stuff'' categories. %

Specifically, given a video of length $\rm T$, we obtain an unsupervised VPS prediction $\mathbf{P}=(\mathbf{S},\mathbf{R})$. Only for evaluation, we have given the ground-truth VPS label $\bar{\mathbf{P}}=(\bar{\mathbf{S}},\bar{\mathbf{R}})$, with the semantic ground truth $\bar{\mathbf{S}}\in\{1,2,\ldots,c_{\text{gt}}\}^{\rm T \times \rm H\times W}$ and the corresponding $n_{\text{gt}}$ binary video instance masks $\bar{\mathbf{R}}\in\{0, 1\}^{n_{\text{gt}} \times \rm T \times \rm H\times W}$. 

Panoptic segmentation~\cite{Kirillov:2019:PS, Kim:2020:VPS, Weber:2021:STE} distinguishes between ``thing'' categories for which instance masks are predicted and ``stuff'' categories for which only semantics are predicted. To adhere to this strict separation between both, we extract the set of semantic pseudo ``thing'' categories $\mathbb{S}^{\text{Th}}_{\text{p}}\subset\{1, \ldots,c_{\text{p}}\}$ (\ie, categories with video instance predictions) and semantic pseudo ``stuff'' categories $\mathbb{S}^{\text{St}}_{\text{p}}\subset\{1, \ldots,c_{\text{p}}\}$, with $\mathbb{S}^{\text{Th}}_{\text{p}}\cap\mathbb{S}^{\text{St}}_{\text{p}}=\emptyset$ and $\mathbb{S}^{\text{Th}}_{\text{p}}\cup\mathbb{S}^{\text{St}}_{\text{p}}=\{1, \ldots,c_{\text{p}}\}$. Similarly, we know the ground-truth semantic ``thing'' categories $\mathbb{S}^{\text{Th}}_{\text{gt}}\subset\{1, \ldots,c_{\text{gt}}\}$ and semantic ``stuff'' categories $\mathbb{S}^{\text{St}}_{\text{gt}}\subset\{1, \ldots,c_{\text{gt}}\}$, with $\mathbb{S}^{\text{Th}}_{\text{gt}}\cap\mathbb{S}^{\text{St}}_{\text{gt}}=\emptyset$ and $\mathbb{S}^{\text{Th}}_{\text{gt}}\cup\mathbb{S}^{\text{St}}_{\text{gt}}=\{1, \ldots,c_{\text{gt}}\}$. For each category type, we construct a cost matrix $\mathbf{A}^{\text{Th}}\in\mathbb{N}^{|\mathbb{S}_{\text{p}}^{\text{Th}}|\times |\mathbb{S}_{\text{gt}}^{\text{Th}}|}$ and 
$\mathbf{A}^{\text{St}}\in\mathbb{N}^{|\mathbb{S}_{\text{p}}^{\text{St}}|\times |\mathbb{S}_{\text{gt}}^{\text{St}}|}$ that accumulates the number of overlapping pixels between every pseudo and ground-truth class across all videos in the validation set. We independently apply Hungarian matching~\cite{Kuhn:1955:HUN} to both matrices, maximizing pixel overlap, and establish an initial 
correspondence by matching one ground-truth class with a pseudo-class.
If there exist more pseudo than ground-truth classes, unmatched pseudo-classes are assigned to the ground-truth class with the highest overlap. 

After alignment, we follow the established protocol by Weber \etal~\cite{Weber:2021:STE} from the supervised literature. 
In particular, we compute the Segmentation and Tracking Quality (STQ), composed of the Association Quality (AQ) and Segmentation Quality (SQ). STQ measures accuracy on full videos at the pixel level, requires no threshold-based matching for validating video instance detections, and considers both precision and recall, different from other VPS metrics~\cite{Weber:2021:STE, Kim:2020:VPS, Hurtado:2020:MOP}. More details and a discussion on other VPS metrics are provided in the supplement.

\begin{table*}[t]
    \centering
    \caption{\textbf{Unsupervised VPS on Cityscapes-VPS val.} We compare \MethodName to our unsupervised VPS baselines, using STQ, AQ, and SQ (all in \%, $\uparrow$). \MethodName achieves state-of-the-art accuracy on Cityscapes-VPS val. $\dagger$ denotes CUPS retrained using monocular videos.}\label{tab:uvps_cs}
    \vspace{-0.5em}
    \footnotesize\sisetup{table-number-alignment=center}
\setlength{\tabcolsep}{7.5pt}
\renewcommand{\arraystretch}{0.875}
\begin{tabularx}{\textwidth}{>{\hspace{-\tabcolsep}\raggedright\columncolor{white}[\tabcolsep][\tabcolsep]}lYYZZZ}
    \toprule
    {\textbf{Method}} & {\textbf{Training data}} & {\textbf{Pseudo-classes}} & {\textbf{STQ}} & {\textbf{AQ}} & {\textbf{SQ}} \\
    \midrule
    \textcolor{tud0c}{Supervised~\cite{Kirillov:2019:PFP, Yang:2019:VIS}} & \textcolor{tud0c}{Cityscapes \& Cityscapes-VPS} & \textcolor{tud0c}{--} & \color{tud0c}42.0 & \color{tud0c}27.0 & \color{tud0c}65.3 \\
    \midrule
    {DepthG~\cite{Sick:2024:USS}~+~VideoCutLER~\cite{Wang:2024:VCS}} & Cityscapes \& ImageNet & 27 & 9.9 & 3.4	& 28.2 \\
    {U2Seg~\cite{Niu:2024:UUI}~+~SORT~\cite{Bewley:2016:SOR}} & COCO \& ImageNet & 800 + 27 & 11.4 & 5.6 & 23.0 \\
    {CUPS~\cite{Hahn:2025:UPS}~+~SORT~\cite{Bewley:2016:SOR}} & Cityscapes (stereo videos) & 27 & 20.6 & 13.3	& 31.8 \\
    {CUPS\textsuperscript{$\dagger$} \cite{Hahn:2025:UPS}~+~SORT~\cite{Bewley:2016:SOR}} & Cityscapes (monocular videos) & 27 & 17.8 & 10.6	& 29.9 \\
    \midrule
    \rowcolor{tud0c!20} 
    {\MethodName \textit{(Ours)}} & Cityscapes (monocular videos) & 27 & \bfseries 22.2	& \bfseries 15.3	& \bfseries 32.3 \\
    \bottomrule
\end{tabularx}

    \vspace{-0.4em}
\end{table*}%
\begin{figure*}[t]
    \centering
    \input{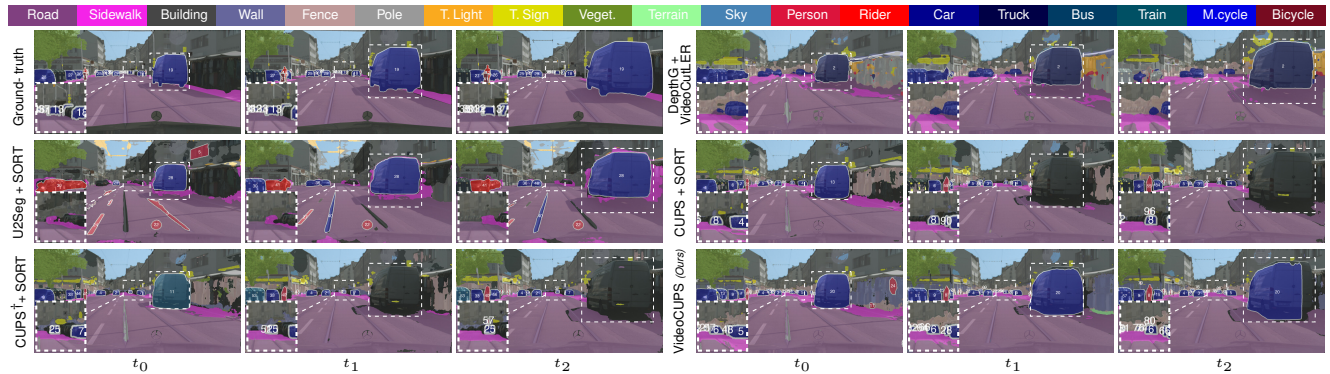}
    \vspace*{-1.2em}
    \caption{\textbf{Qualitative unsupervised VPS examples.} We compare \MethodName to our baselines DepthG~\cite{Sick:2024:USS} +~VideoCutLER~\cite{Wang:2024:VCS}, U2Seg~\cite{Niu:2024:UUI}~+~SORT, CUPS~\cite{Hahn:2025:UPS}~+~SORT~\cite{Bewley:2016:SOR}, and CUPS\textsuperscript{\textdagger}~\cite{Hahn:2025:UPS}~+~SORT~\cite{Bewley:2016:SOR} on Cityscapes-VPS val. We highlight regions of improvement.}
    \label{fig:qualitative}
    \vspace{-0.5em}
\end{figure*}%

\section{Experiments \label{sec:experiments}}
We evaluate the unsupervised VPS accuracy of \MethodName within its training domain and its generalization (\cref{subsec:uvpsresults}). To assess \MethodName's accuracy, we also report four baselines. Next, we provide label-efficient learning results (\cref{subsec:lelresults}). Finally, we analyze the impact of our core components (\cref{subsec:analysis}). Additional results are in the supplement.

\inparagraph{Datasets.} We train \MethodName on video pseudo-labels generated from the Cityscapes training sequences (\num{2975} clips of \num{30} frames each) and evaluate it on the Cityscapes-VPS val set~\cite{Kim:2020:VPS}. To assess generalization, we conduct cross-domain evaluations on KITTI-STEP~\cite{Weber:2021:STE} and Waymo~\cite{Sun:2020:SPA, Mei:2022:WAY}, and further test out-of-domain (OOD) generalization on MOTS~\cite{Voigtlaender:2019:MMO}. 
While Cityscapes-VPS, KITTI-STEP, and Waymo focus on understanding driving scenes, MOTS addresses human-centric segmentation and tracking in indoor and outdoor settings.
For all cross-domain datasets, we ensure compatibility of their label spaces with the Cityscapes category definitions through matching (\cf \cref{subsec:eval_vups}). Note, we ignore extremely small instances in Waymo; more details are in the supplement.

\inparagraph{Evaluation details.} We follow the evaluation protocol outlined in \cref{subsec:eval_vups} and report the Segmentation and Tracking Quality (STQ)~\cite{Weber:2021:STE} alongside the Association Quality (AQ) and Segmentation Quality (SQ), all in \%.%

\inparagraph{Implementation details.} We generate video pseudo-labels using $c_{\text{p}} = 27$ pseudo-classes, following CUPS~\cite{Hahn:2025:UPS}. To adhere to our purely unsupervised and monocular setup, we retrain DynamoDepth~\cite{Sun:2023:DDF} with a DINO ResNet-18~\cite{He:2016:DRL, Caron:2021:EPS}, instead of an ImageNet-supervised ResNet-18, and also retrain DepthG with monocular depth from DynamoDepth. Our region growing uses $\tau_{d}=\text{\num{0.02}}$, $\tau_{f}=\text{\num{0.04}}$, and $r=\text{\num{8}}$. To ensure fairness to our baselines U2Seg~\cite{Niu:2024:UUI} and CUPS~\cite{Hahn:2025:UPS}, which employ a Panoptic Cascade Mask R-CNN~\cite{Cai:2018:CRC, Kirillov:2019:PFP}, we use the closest video extension Panoptic Cascade MaskTrack R-CNN~\cite{Cai:2018:CRC, Kirillov:2019:PFP, Yang:2019:VIS} with a DINO ResNet-50~\cite{Caron:2021:EPS, He:2016:DRL, Wang:2023:CAL}. We train using AdamW~\cite{Loshchilov:2018:ADW}, our self-enhanced video copy-paste augmentation, and Video DropLoss (with $\tau_{\operatorname{IoU}}=\text{\num{0.5}}$) for eight epochs. 
We refer to the supplement for further details.

\begin{table*}[t]
    \centering
    \caption{\textbf{Generalization results.} Video panoptic segmentation results, comparing \MethodName to our unsupervised VPS baselines, using STQ, AQ, and SQ (all in \%, $\uparrow$). We evaluate generalization to the Waymo and KITTI-STEP datasets as well as to the OOD dataset MOTS. \MethodName consistently outperforms all of the proposed baselines. $\dagger$ denotes CUPS retrained using monocular videos.}\label{tab:uvps_gen}
    \vspace{-0.5em}
    \footnotesize\sisetup{table-number-alignment=center}
\setlength{\tabcolsep}{7.0pt}
\renewcommand{\arraystretch}{0.875}
\begin{tabularx}{\textwidth}{
    >{\hspace{-\tabcolsep}\raggedright\columncolor{white}[\tabcolsep][\tabcolsep]}X
    ZZZl
    ZZZl
    ZZZ 
}
    \toprule
    & \multicolumn{3}{c}{\textbf{KITTI-STEP}} & \hphantom{SVB} & \multicolumn{3}{c}{\textbf{Waymo}} & \hphantom{SVB} & \multicolumn{3}{c}{\textbf{MOTS} \textit{(OOD)}} \\
    \cmidrule(l{0.0em}r{0.0em}){2-4} \cmidrule(l{0.0em}r{0.0em}){6-8} \cmidrule(l{0.0em}r{0.0em}){10-12}
    \multirow{-2}{*}{\vspace{0.5em}\textbf{Method}} 
        & \textbf{STQ} & \textbf{AQ} & \textbf{SQ} &
        & \textbf{STQ} & \textbf{AQ} & \textbf{SQ} &
        & \textbf{STQ} & \textbf{AQ} & \textbf{SQ} \\
    \midrule
    \textcolor{tud0c}{Supervised~\cite{Kirillov:2019:PFP, Yang:2019:VIS}} & \color{tud0c} 53.9 & \color{tud0c} 59.9 & \color{tud0c} 48.4 & & \color{tud0c} 22.3 & \color{tud0c} 12.6 & \color{tud0c} 39.4 & & \color{tud0c} 20.5 & \color{tud0c} 12.7	& \color{tud0c} 33.1 \\
    \midrule
    {DepthG~\cite{Sick:2024:USS}~+~VideoCutLER~\cite{Wang:2024:VCS}} & 13.2	& 8.7	& 20.1 & & 7.9	& 2.6	& 23.9 & & 14.5	& 6.8	& 30.7\\
    {U2Seg~\cite{Niu:2024:UUI}~+~SORT~\cite{Bewley:2016:SOR}}        & 24.0	& 21.1	& 27.2 & & 10.4	& 4.8	& 22.6 & & 14.9	& 7.2	& 30.8 \\
    {CUPS~\cite{Hahn:2025:UPS}~+~SORT~\cite{Bewley:2016:SOR}} & 34.2 & 37.7 & 31.1 & & 17.5 & 9.9 & 30.8 & & 16.7 & 10.4 & 27.0\\
    {CUPS\textsuperscript{$\dagger$} \cite{Hahn:2025:UPS}~+~SORT~\cite{Bewley:2016:SOR}} & 32.9 & 35.4 & 30.5 & & 16.6 & 9.3 & 29.8 & & 14.9 & 7.8 & 28.3 \\
    \midrule
    \rowcolor{tud0c!20} {\MethodName \textit{(Ours)}} & \bfseries 37.3 & \bfseries 43.6 & \bfseries 32.0 & &  \bfseries 18.4 & \bfseries 10.7 & \bfseries 31.6 & & \bfseries 18.6 & \bfseries 10.5 & \bfseries 33.0 \\
    \bottomrule
\end{tabularx}%

    \vspace{-0.8em}
\end{table*}

\begin{figure*}[t]
    \centering
    \pgfdeclarelayer{foreground}
\pgfdeclarelayer{background}
\pgfsetlayers{background,main,foreground}

\begin{tikzpicture}[clip, every node/.style={font=\sffamily\fontsize{7}{2}}, >={Stealth[inset=0pt,length=1.25pt,angle'=45]}]
  \tikzset{every picture/.style={/utils/exec={\sffamily\fontsize{7}{2}}}}
  \tikzset{every major tick/.append style={line width=.5pt, major tick length=3.5pt, gray!85}}

  \pgfkeys{/pgfplots/CommonAxisStyle/.style={
      enlargelimits=false,
      clip=true,
      height=3.4cm,
      width=0.378\textwidth,
      grid=both,
      xtick pos=bottom,
      ylabel shift=-3.8pt,
      ytick pos=left,
      grid style={line width=.1pt, draw=white, dash pattern=on 1pt off 1pt},
      major grid style={line width=.2pt,draw=gray!65},
      minor tick num=1,
      xmin=-2,
      xmax=117, %
      xtick={0,5,10,20,40,60,100,115},
      xticklabels={\tiny{\hspace{-2pt}\shortstack{w/o\\[-1.5pt]fine-tune}}, 5\vphantom{6}, 10, 20, 40, 60, 100, 224},
      ticklabel style = {font=\fontsize{7}{2}},
  }}

  \begin{groupplot}[group style={group size=3 by 1, horizontal sep=8mm}, CommonAxisStyle]

    \nextgroupplot[
      ylabel=STQ,
      ymin=20, ymax=44,
      ytick={10,20,30,40,50},
      yticklabels={10,20,30,40,50},
        after end axis/.code={
    \draw[white, line width=1.95mm] (rel axis cs:0.925,-0.05) -- (rel axis cs:0.925,1.05);
    \node[font=\sffamily\small, fill=white, inner sep=0.1pt]
      at (rel axis cs:0.925,0.0) {//};
    \node[font=\sffamily\small, fill=white, inner sep=0.1pt]
      at (rel axis cs:0.925,1.0) {//};
  },
      ]

      \addplot[color=gray, mark=star, mark size=1.9pt, mark options={solid}, dash pattern=on 4pt off 2.45pt, semithick] table[x=n,y=stq] {supervised.dat};\label{pgfplots:supervised};

        \addplot[color=gray, dash pattern=on 4pt off 2.45pt, semithick, mark=diamond*, mark options={solid}, mark size=1.5pt,
        error bars/.cd, y dir=both, y explicit]
        table[x=n,y=stq, y error=stqstd] {supervisedrand.dat}; \label{pgfplots:supervisedrand}
      \addplot[color=tud6b, semithick, mark=*, mark size=0.9pt,
        error bars/.cd, y dir=both, y explicit]
        table[x=n,y=stq, y error=stqstd] {vcups.dat}; \label{pgfplots:vcups}
      \addplot[color=tud10a, semithick, mark=square*, mark size=0.9pt,
        error bars/.cd, y dir=both, y explicit]
        table[x=n,y=stq, y error=stqstd] {dino.dat}; \label{pgfplots:dino}

        \begin{pgfonlayer}{foreground}
            \draw[<->] (axis cs:97, 40.7) -- (axis cs:97, 42.0) node[midway, left=1pt, fill=white, fill opacity=0.65, text opacity=1.0, font=\sffamily\tiny, inner sep=0.3pt] {-1.3\%};
            \draw[dashed, black, dash pattern=on 1pt off 1pt] (axis cs:97, 40.7) -- (axis cs:100, 40.7);
            \draw[dashed, black, dash pattern=on 1pt off 1pt] (axis cs:97, 42.0) -- (axis cs:100, 42.0);
            \draw[<->] (axis cs:103, 40.7) -- (axis cs:103, 38.1) node[midway, right=1pt, fill=white, fill opacity=0.65, text opacity=1.0, font=\sffamily\tiny, inner sep=0.3pt] {\dlt{+2.6\%}};
            \draw[dashed, black, dash pattern=on 1pt off 1pt] (axis cs:103, 40.7) -- (axis cs:100, 40.7);
            \draw[dashed, black, dash pattern=on 1pt off 1pt] (axis cs:103, 38.1) -- (axis cs:100, 38.1);
            \draw[<->] (axis cs:13, 32.5) -- (axis cs:13, 27.9) node[midway, right=1pt, yshift=3.2pt, fill=white, fill opacity=0.65, text opacity=1.0, font=\sffamily\tiny, inner sep=0.3pt] {\dlt{+4.6\%}};
            \draw[dashed, black, dash pattern=on 1pt off 1pt] (axis cs:13, 32.5) -- (axis cs:10, 32.5);
            \draw[dashed, black, dash pattern=on 1pt off 1pt] (axis cs:13, 27.9) -- (axis cs:10, 27.9);
        \end{pgfonlayer}

    \nextgroupplot[
      ylabel=AQ,
      xlabel={Fraction of Cityscapes-VPS~\cite{Kim:2020:VPS} train labels used for training in \%.},
      xlabel shift=-1.0pt,
      ymin=14, ymax=28.205,
      ytick={10,15,20,25,30},
      yticklabels={10,15,20,25,30},
              after end axis/.code={
    \draw[white, line width=1.95mm] (rel axis cs:0.925,-0.05) -- (rel axis cs:0.925,1.05);
    \node[font=\sffamily\small, fill=white, inner sep=0.1pt]
      at (rel axis cs:0.925,0.0) {//};
    \node[font=\sffamily\small, fill=white, inner sep=0.1pt]
      at (rel axis cs:0.925,1.0) {//};
  },
    ]
        \addplot[color=gray, dash pattern=on 4pt off 2.45pt, semithick, mark=star, mark options={solid}, mark size=1.65pt,
        error bars/.cd, y dir=both, y explicit]
        table[x=n,y=aq, y error=aqstd] {supervised.dat};
        \addplot[color=gray, dash pattern=on 4pt off 2.45pt, semithick, mark options={solid}, mark=diamond*, mark size=1.5pt,
        error bars/.cd, y dir=both, y explicit]
        table[x=n,y=aq, y error=aqstd] {supervisedrand.dat};
      
      \addplot[color=tud6b, semithick, mark=*, mark size=0.9pt,
        error bars/.cd, y dir=both, y explicit]
        table[x=n,y=aq, y error=aqstd] {vcups.dat};
      \addplot[color=tud10a, semithick, mark=*, mark size=0.9pt,
        error bars/.cd, y dir=both, y explicit]
        table[x=n,y=aq, y error=aqstd] {dino.dat};

    \begin{pgfonlayer}{foreground}
            \draw[-] (axis cs:97, 26.7) -- (axis cs:97, 27.0) node[midway, left=1pt, fill=white, fill opacity=0.65, text opacity=1.0, font=\sffamily\tiny, inner sep=0.3pt] {-0.3\%};
            \draw[dashed, black, dash pattern=on 1pt off 1pt] (axis cs:97, 26.7) -- (axis cs:100, 26.7);
            \draw[dashed, black, dash pattern=on 1pt off 1pt] (axis cs:97, 27.0) -- (axis cs:100, 27.0);
            \draw[<->] (axis cs:103, 27.0) -- (axis cs:103, 24.7) node[midway, right=1pt, fill=white, fill opacity=0.65, text opacity=1.0, font=\sffamily\tiny, inner sep=0.3pt] {\dlt{+2.3\%}};
            \draw[dashed, black, dash pattern=on 1pt off 1pt] (axis cs:103, 27.0) -- (axis cs:100, 27.0);
            \draw[dashed, black, dash pattern=on 1pt off 1pt] (axis cs:103, 24.7) -- (axis cs:100, 24.7);
            \draw[<->] (axis cs:13, 23.3) -- (axis cs:13, 19.4) node[midway, right=1pt, yshift=1.0pt, fill=white, fill opacity=0.65, text opacity=1.0, font=\sffamily\tiny, inner sep=0.3pt] {\dlt{+3.9\%}};
            \draw[dashed, black, dash pattern=on 1pt off 1pt] (axis cs:13, 23.3) -- (axis cs:10, 23.3);
            \draw[dashed, black, dash pattern=on 1pt off 1pt] (axis cs:13, 19.4) -- (axis cs:10, 19.4);
        \end{pgfonlayer}

    \nextgroupplot[
      ylabel=SQ,
      ymin=29, ymax=68.7,
      ytick={20,30,40,50,60,70},
      yticklabels={20,30,40,50,60,70},
              after end axis/.code={
    \draw[white, line width=1.95mm] (rel axis cs:0.925,-0.05) -- (rel axis cs:0.925,1.05);
    \node[font=\sffamily\small, fill=white, inner sep=0.1pt]
      at (rel axis cs:0.925,0.0) {//};
    \node[font=\sffamily\small, fill=white, inner sep=0.1pt]
      at (rel axis cs:0.925,1.0) {//};
  },
    ]
    \addplot[color=gray, dash pattern=on 4pt off 2.45pt, semithick, mark=star, mark options={solid}, mark size=1.65pt,
        error bars/.cd, y dir=both, y explicit]
        table[x=n,y=sq, y error=sqstd] {supervised.dat};
    \addplot[color=gray, dash pattern=on 4pt off 2.45pt, semithick, mark=diamond*, mark options={solid}, mark size=1.5pt,
        error bars/.cd, y dir=both, y explicit]
        table[x=n,y=sq, y error=sqstd] {supervisedrand.dat};
      \addplot[color=tud6b, semithick, mark=*, mark size=0.9pt,
        error bars/.cd, y dir=both, y explicit]
        table[x=n,y=sq, y error=sqstd] {vcups.dat};
      \addplot[color=tud10a, semithick, mark=*, mark size=0.9pt,
        error bars/.cd, y dir=both, y explicit]
        table[x=n,y=sq, y error=sqstd] {dino.dat};

    \begin{pgfonlayer}{foreground}
            \draw[<->] (axis cs:97, 62.2) -- (axis cs:97, 65.3) node[midway, left=1pt, fill=white, fill opacity=0.65, text opacity=1.0, font=\sffamily\tiny, inner sep=0.3pt] {-3.1\%};
            \draw[dashed, black, dash pattern=on 1pt off 1pt] (axis cs:97, 62.2) -- (axis cs:100, 62.2);
            \draw[dashed, black, dash pattern=on 1pt off 1pt] (axis cs:97, 65.3) -- (axis cs:100, 65.3);
            \draw[<->] (axis cs:103, 62.2) -- (axis cs:103, 58.7) node[midway, right=1pt, fill=white, fill opacity=0.65, text opacity=1.0, font=\sffamily\tiny, inner sep=0.3pt] {\dlt{+3.5\%}};
            \draw[dashed, black, dash pattern=on 1pt off 1pt] (axis cs:103, 62.2) -- (axis cs:100, 62.2);
            \draw[dashed, black, dash pattern=on 1pt off 1pt] (axis cs:103, 58.7) -- (axis cs:100, 58.7);
            \draw[<->] (axis cs:13, 45.4) -- (axis cs:13, 40.3) node[midway, right=1pt, yshift=-1.0pt, fill=white, fill opacity=0.65, text opacity=1.0, font=\sffamily\tiny, inner sep=0.3pt] {\dlt{+5.1\%}};
            \draw[dashed, black, dash pattern=on 1pt off 1pt] (axis cs:13, 45.4) -- (axis cs:10, 45.4);
            \draw[dashed, black, dash pattern=on 1pt off 1pt] (axis cs:13, 40.3) -- (axis cs:10, 40.3);
        \end{pgfonlayer}
    
  \end{groupplot}

  \node[draw,fill=white, inner sep=2.0pt, anchor=south] at ([yshift=1.5mm]group c2r1.north) {\fontsize{7}{2}\sffamily \MethodName~\textit{(Ours)}~\ref*{pgfplots:vcups} \quad \quad DINO \ref*{pgfplots:dino} \quad \quad Supervised (random init.)~\ref*{pgfplots:supervisedrand} \quad \quad Supervised (Cityscapes pre-training init.)~\ref*{pgfplots:supervised}} ;
\end{tikzpicture}
    \vspace*{-0.6em}
    \caption{\textbf{Label-efficient learning.} We fine-tune \MethodName{} and a DINO-initialized model on varying fractions of labeled Cityscapes-VPS train clips and report STQ, AQ, \& SQ (all in~\%,~$\uparrow$) on Cityscapes-VPS val. We also report models trained on the full Cityscapes-VPS train set, \emph{without} any pre-training (rand.\ init.) and \emph{with} supervised image pre-training on Cityscapes. \SI{224}{\%} denotes using both Cityscapes \& Cityscapes-VPS. For training on Cityscapes-VPS subsets, we report the average and standard deviation over three different subsets.} \label{fig:label_efficient_learning}
    \vspace{-0.5em}
\end{figure*}
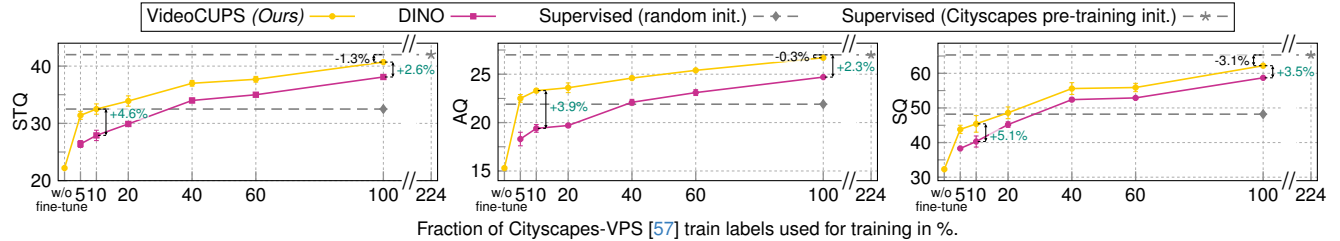%

\begin{figure*}[t]
    \centering
    \input{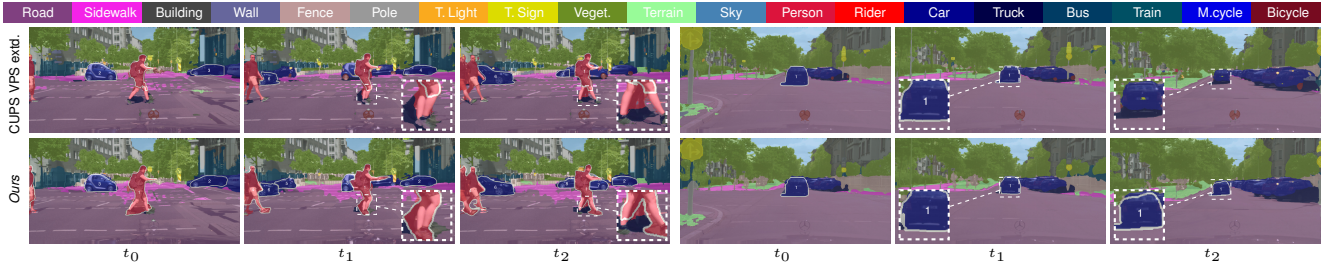}
    \vspace*{-1.1em}
    \caption{\textbf{Qualitative pseudo-label examples.} We compare \MethodName pseudo-labels to CUPS pseudo-labels extended to video with our approach. 
    While CUPS benefits from stereo cues for improved depth, resulting in better semantics, our monocular pseudo-labels discover more instances, maintain longer tracks \emph{($t_1$ \& $t_2$; right)}, and capture more non-rigid motion \emph{($t_1$; left)}. We visualize matched pseudo-classes.
    \label{fig:qualitative_pseudolabels}}
    \vspace{-2.7em}
\end{figure*}%

\inparagraph{Unsupervised VPS baselines.}
As there are no existing unsupervised VPS approaches, we construct four competitive baselines. \emph{DepthG~+~VideoCutLER} combines the unsupervised semantic segmentation approach DepthG~\cite{Sick:2024:USS} with the class-agnostic video instance segmentation method VideoCutLER~\cite{Wang:2024:VCS}. We adopt the ``thing''/``stuff'' separation and the semantic-instance fusion scheme from our pseudo-labeling. Since running VideoCutLER on long videos leads to memory exhaustion, we split them into clips of \num{30} frames with a \num{5}-frame temporal overlap. Instance IDs are aligned across clips using IoU overlap. \emph{U2Seg~\cite{Niu:2024:UUI}~+~SORT} and  \emph{CUPS~\cite{Hahn:2025:UPS}~+~SORT} combine existing state-of-the-art approaches to unsupervised panoptic image segmentation with SORT~\cite{Bewley:2016:SOR}, a well-established unsupervised multi-object tracker. SORT assigns temporally consistent IDs to the ``thing'' detections of the respective model. We use the proposed hyperparameters by Bewley~\etal~\cite{Bewley:2016:SOR}. As CUPS utilizes stereo video for training, we also provide a monocular variant of CUPS using monocular depth from DynamoDepth to assess the impact of using stereo cues. This variant is denoted as \emph{CUPS\textsuperscript{$\dagger$}~+~SORT}.

\inparagraph{Supervised upper bound.}
To contextualize our unsupervised results, we train a supervised equivalent of \MethodName. Following the protocol in supervised VPS~\cite{Kim:2020:VPS, Weber:2021:STE}, we initialize with a pre-trained backbone (DINO~\cite{Caron:2021:EPS}) and pre-train on Cityscapes~\cite{Cordts:2016:CDS} panoptic image annotations. Next, we fine-tune for VPS on Cityscapes-VPS~\cite{Kim:2020:VPS}. %

\subsection{Unsupervised VPS results} \label{subsec:uvpsresults}

\paragraph{In-domain results.} In \cref{tab:uvps_cs}, we compare \MethodName{} against our proposed baselines DepthG~+~VideoCutLER, U2Seg~+~SORT, and CUPS~+~SORT (w/ and w/o stereo) on the Cityscapes-VPS validation set. \MethodName{} significantly outperforms DepthG~+~VideoCutLER and U2Seg~+~SORT, increasing STQ by \SI{+12.3}{\%} and \SI{+10.8}{\%} points, respectively. We attribute the lower STQ of both baselines to their training on object-centric data. Instead, \MethodName{} can train directly on scene-centric videos.
CUPS~+~SORT requires \emph{stereo} video for video training, limiting applicability. While being \emph{monocular}, \MethodName{} reaches an STQ of \SI{22.2}{\%}, outperforming the stereo CUPS~+~SORT baselines (\SI{20.6}{\%} STQ). In comparison to the monocular variant, CUPS\textsuperscript{$\dagger$}~+~SORT, \MethodName{} leads to an improved STQ of \SI{4.4}{\%}. This demonstrates that our pseudo-labeling approach leverages monocular and scene-centric video more effectively, while CUPS requires stereo cues to achieve competitive results.
These findings are also reflected in the qualitative comparison in \cref{fig:qualitative}.

\inparagraph{Domain generalization results.} In \cref{tab:uvps_gen}, we assess cross-domain generalization on KITTI-STEP, Waymo, and MOTS (OOD). \MethodName{} consistently outperforms all four baselines across datasets, achieving improvements of up to \SI{3.3}{\percent}~STQ on KITTI-STEP. We observe that STQ is higher for KITTI-STEP compared to Waymo and MOTS, as fewer instances need to be detected and tracked~\cite{Mei:2022:WAY, Weber:2021:STE}. These results showcase that unsupervised training generalizes effectively across domains. While the supervised model is still more accurate than unsupervised approaches, we observe that supervised learning is more susceptible to domain shifts, particularly on Waymo and MOTS.

\subsection{Label-efficient learning}\label{subsec:lelresults}
Achieving high-quality video panoptic segmentation ultimately depends on adapting to a predominantly human-defined semantic taxonomy, which remains beyond the reach of fully unsupervised approaches (\cf \cref{tab:uvps_cs} \& \ref{tab:uvps_gen}). A promising direction is unsupervised pre-training to acquire robust spatio-temporal and segmentation priors, followed by fine-tuning on a small set of labelled examples. This approach enables efficient adaptation to human-defined tasks while minimizing the need for extensive annotations.

In \cref{fig:label_efficient_learning}, we explore this scenario by comparing the unsupervised \MethodName{}-initialized model to the same architecture initialized with DINO~\cite{Caron:2021:EPS} and trained with varying fractions of Cityscapes-VPS labels. We also report our supervised upper bound, using supervised panoptic image pre-training on Cityscapes and full Cityscapes-VPS fine-tuning. To assess the impact of the supervised image pre-training, we also report a randomly initialized model (with He init.~\cite{He:2015:DDR}) trained on the full Cityscapes-VPS training set. Note that the Cityscapes and Cityscapes-VPS training splits are disjoint, containing \num{2975} images\,/\,\num{2975} labels and \num{400} clips\,/\,\num{2400} labels, respectively.

Fine-tuning \MethodName{} with different fractions of VPS labels consistently outperforms the DINO pre-trained model. In particular, when using \SI{10}{\%} of Cityscapes-VPS labels, \MethodName{} improves by \SI{4.6}{\%} STP over DINO. While the delta reduces for larger fractions of annotations, \MethodName{} still outperforms DINO by \SI{3.5}{\%} STQ when using \SI{100}{\%} of labels. In comparison to the randomly initialized supervised model trained on \SI{100}{\%} of the Cityscapes-VPS labels, \MethodName{} requires only \SI{10}{\%} of the labels to reach the same STQ. Training \MethodName on all Cityscapes-VPS labels, almost closes the gap to the supervised model trained on \emph{both} Cityscapes \& Cityscapes-VPS (\SI{38.1}{\percent} \vs \SI{40.4}{\percent} STQ), despite the latter using \SI{124}{\percent} more labels. These results show that our unsupervised training is a strong initialization for learning with limited labels.

\begin{table}[t]
    \centering
    \caption{\textbf{Video pseudo-label generation ablation}, analyzing the contribution of individual components, using STQ, AQ, and SQ (all in \%, $\uparrow$) for pseudo-labels generated on Cityscapes-VPS val.\label{tab:pseudo_ablation}}
    \vspace{-0.3em}
    \footnotesize\sisetup{table-number-alignment=center}
\setlength{\tabcolsep}{3pt}
\renewcommand{\arraystretch}{0.875}
\begin{tabularx}{\columnwidth}{>{\hspace{-\tabcolsep}\raggedright\columncolor{white}[\tabcolsep][\tabcolsep]}lYZZZ}
	\toprule
    \textbf{Pseudo-label configuration} & \textbf{Mono.} & \textbf{STQ} & \textbf{AQ} & \textbf{SQ} \\
    \midrule  
    Vanilla semantics + region growing instances         & \dlt{\cmark} & 9.3 & 2.6 & 32.5 \\
    \;+ Depth-guided semantic inference~\cite{Hahn:2025:UPS} & \dlt{\cmark} & 9.4 & 2.7 & 32.6  \\
    \;+ Instance propagation \& tracking    & \dlt{\cmark} & 12.0 & 4.4 & 32.4  \\
    \rowcolor{tud0c!20} \;+ Temporal semantic smoothing \textit{(full config.)} & \dlt{\cmark} & \bfseries 12.1 & \bfseries 4.5 & 32.3 \\
    \midrule 
    Video-extended CUPS pseudo labels~\cite{Hahn:2025:UPS} & {\textcolor{tud9c}{\xmark}} & 11.6 & 3.9 & \bfseries 35.0 \\
	\bottomrule
\end{tabularx}

    \vspace{-0.5em}
\end{table}

\subsection{Analyzing \MethodName{}} \label{subsec:analysis}

\inparagraph{\MethodName{} pseudo-label analysis. \label{sec:pseudo}} \Cref{tab:pseudo_ablation} presents an ablation of our pseudo-label generation, evaluating the contribution of each core component on Cityscapes-VPS val. Starting from combining the unsupervised semantic prediction of DepthG (vanilla semantics) with region-growing object proposals, we incrementally add depth-guided semantic inference~\cite{Hahn:2025:UPS}, instance propagation, and temporal semantic smoothing. Each component contributes to improving the final STQ of our pseudo-labels, while our instance propagation aids the most. Temporal semantic smoothing results in only a minor increase in STQ. We attribute this partly to the limited temporal quality of the Cityscapes-VPS labels, as noted by Zhou \etal~\cite{Zhou:2022:IBT} and Woo \etal~\cite{Woo:2021:LAS}.

For reference, we also compare against CUPS pseudo-labels generated using stereo video. In particular, we use the CUPS panoptic image pseudo-labels and extend these by our tracking and temporal smoothing (\cf \cref{subsec:pseudo_label_generation}) to obtain temporally consistent video pseudo-labels. Despite the absence of strong stereo cues, our purely monocular pseudo-labels achieve a higher STQ (\SI{12.1}{\percent} \vs \SI{11.6}{\percent}). Only in SQ, the stereo pseudo-labels from CUPS improve over our monocular pseudo-labels. We attribute this to the lower-quality depth cues of our monocular approach, resulting in weaker depth-guided semantic inference. As a qualitative reference, we provide examples of our \MethodName pseudo-labels as well as our video extension of the CUPS pseudo-labels in \cref{fig:qualitative_pseudolabels}.

\begin{table}[t]
    \centering
    \caption{\textbf{\MethodName training ablation}, analyzing the contribution of our core training components, using STQ, AQ, and SQ (all in \%, $\uparrow$) on Cityscapes-VPS val.\label{tab:training_ablation}}
    \vspace{-0.3em}
    \footnotesize\sisetup{table-number-alignment=center}
\setlength{\tabcolsep}{2.75pt}
\renewcommand{\arraystretch}{0.875}
\begin{tabularx}{\columnwidth}{>{\hspace{-\tabcolsep}\raggedright\columncolor{white}[\tabcolsep][\tabcolsep]}XcZZZ}
	\toprule
    \textbf{Training configuration} & \textbf{STQ} & \textbf{AQ} & \textbf{SQ} \\
    \midrule  
    Vanilla training & 17.8 & 10.0 & 31.8 \\
    \;+ Video DropLoss ($\mathcal{L}_{\text{VDrop}}$) & 21.5 & 14.4 & 32.1 \\
    \;+ Video copy-paste augmentation & 21.7 & 14.8 & 31.8 \\
    \rowcolor{tud0c!20}\;+ Self-enhance copy-paste augmentation \textit{(full config.)} & \bfseries 22.2 & \bfseries 15.3 & \bfseries 32.3 \\
	\bottomrule
\end{tabularx}

    \vspace{-0.5em}
\end{table}

\inparagraph{\MethodName{} training analysis.}
In \cref{tab:training_ablation}, we analyze the contribution of individual training components on Cityscapes-VPS. Starting from a vanilla training setup, adding our Video DropLoss improves STQ and AQ by mitigating instances missed by pseudo-labeling. Adding video copy-paste augmentation further improves STQ. Adding our self-enhanced copy-paste augmentation (full config.) achieves the highest STQ, aiding in the detection and tracking of small objects, as indicated by the improved AQ.

\section{Conclusion}
\label{sec:conclusion}

We introduced the task setting of \emph{unsupervised video panoptic segmentation} and defined a comprehensive evaluation protocol across multiple scene-centric datasets. Our proposed method, \MethodName, is the first to approach this problem, showcasing that unsupervised panoptic video understanding can be achieved entirely \emph{without} human supervision. \MethodName relies solely on monocular videos for VPS pseudo-labeling, removing the need for stereo. Compared with four proposed baselines built from state-of-the-art unsupervised panoptic image and video instance segmentation methods, \MethodName consistently outperforms these baselines across various scene-centric VPS datasets. We further demonstrate that \MethodName provides a strong initialization for learning from limited annotated VPS examples. Together, our task definition, evaluation protocol, baselines, and method establish a foundation for future research on unsupervised panoptic video understanding.

{\small \inparagraph{Acknowledgments.} This project was partially supported by the European Research Council (ERC) Advanced Grant SIMULACRON (grant agreement No.\ 884679), DFG project CR 250/26-1 ``4D-YouTube'', and GNI Project ``AICC''. This project was also partially supported by the ERC under the European Union’s Horizon 2020 research and innovation programme (grant agreement No.\ 866008). Additionally, this work has been co-funded by the LOEWE initiative (Hesse, Germany) within the emergenCITY center [LOEWE/1/12/519/03/05.001(0016)/72] and by the Deutsche Forschungsgemeinschaft (German Research Foundation, DFG) under Germany's Excellence Strategy (EXC 3066/1 ``The Adaptive Mind'', Project No.\ 533717223). Christoph Reich is supported by the Konrad Zuse School of Excellence in Learning and Intelligent Systems (\href{https://eliza.school}{ELIZA}) through the DAAD programme Konrad Zuse Schools of Excellence in Artificial Intelligence, sponsored by the German Federal Ministry of Education and Research. Christian Rupprecht is supported by an Amazon Research Award. Finally, we acknowledge the support of the European Laboratory for Learning and Intelligent Systems (ELLIS) and thank Simone Schaub-Meyer for insightful discussions.}
{
    \small
    \bibliographystyle{ieeenat_fullname}
    \bibliography{bibtex/short, bibtex/new, bibtex/external, bibtex/papers}
}

\clearpage
\setcounter{section}{0}
\renewcommand\thesection{\Alph{section}}
\setcounter{page}{1}
\pagenumbering{roman}
\twocolumn[{%
\renewcommand\twocolumn[1][]{#1}%
\maketitlesupplementary
{\large Christoph Reich\textsuperscript{\normalfont{}* 1,2,5,6}
\authorstep Oliver Hahn\textsuperscript{\normalfont{}* 2,3}
\authorstep Nikita Araslanov\textsuperscript{\normalfont{}\,1,5}
\authorstep Laura Leal-Taixé\textsuperscript{\normalfont{}\,3}\\[-0pt]
Christian Rupprecht\textsuperscript{\normalfont{}\,4}
\authorstep Daniel Cremers\textsuperscript{\normalfont{}\textdagger{} 1,5,6}
\authorstep Stefan Roth\textsuperscript{\normalfont{}\textdagger{} 2,6,7}\\[3pt]
\small{\textsuperscript{1}TU Munich\affiliationstep \textsuperscript{2}TU Darmstadt \affiliationstep 
\textsuperscript{3}NVIDIA \affiliationstep 
\textsuperscript{4}University of Oxford \affiliationstep \textsuperscript{5}MCML\affiliationstep \textsuperscript{6}ELIZA\affiliationstep \textsuperscript{7}hessian.AI\affiliationstep
\textsuperscript{*}equal contribution\affiliationstep
\textsuperscript{\textdagger}equal advising}\\[1pt]\small {\url{https://visinf.github.io/videocups}}
\vspace{0.75cm}
}
}]

\noindent{}In this supplement, we first provide additional implementation details, including dataset information, to aid reproducibility (\cref{sec:supp_reproduce}). We discuss our evaluation protocol and choice of metric in \Cref{sec:supp_eval}. Next, we provide additional quantitative and qualitative results and analysis (\cref{sec:supp_results}). Finally, we provide a comprehensive discussion on the limitations of \MethodName and outline potential future research directions (\cref{sec:supp_limit_future}).

\section{Reproducibility}\label{sec:supp_reproduce}

This section provides further information about the datasets used and details on our implementation to ensure reproducibility. To further ensure reproducibility and provide a foundation for future work on unsupervised VPS, our code is available at \url{https://github.com/visinf/cups/tree/main/videocups}.

\subsection{Datasets}

Here, we provide additional details on the datasets utilized for training and evaluation.

\inparagraphnospace{Cityscapes~\cite{Cordts:2016:CDS}} is a dataset of urban driving scenes composed of \num{5000} high-resolution images at \num{1024}$\times$\num{2048} pixels. The dataset is split into \num{2975} training, \num{500} validation, and \num{1525} test images, each annotated at the pixel level with panoptic labels. While Cityscapes provides different levels of semantic annotations (\num{27}, \num{19}, and \num{7} semantic categories), the Cityscapes evaluation protocol employs \num{19} categories for evaluation. These \num{19} categories are composed of \num{8} ``thing'' and \num{11} ``stuff'' categories. Every annotated training image is extracted from a \num{30}-frame video clip. Following prior work~\cite{Hahn:2025:UPS}, we utilize these \num{2975} training clips (\num{86275} video frames) for generating pseudo-labels and training.

\inparagraphnospace{Cityscapes-VPS~\cite{Kim:2020:VPS}} extends the Cityscapes dataset with panoptic video annotations. In particular, Cityscapes-VPS offers VPS annotations of the \num{500} Cityscapes validation clips. Cityscapes-VPS provides annotations for every fifth frame of each \num{30}-frame clip using \num{19} categories matching Cityscapes. The annotations are obtained using a semi-automated annotation process with human correction. Still, multiple works pointed out labeling errors~\cite{Zhou:2022:IBT, Woo:2021:LAS}. The \num{500} Cityscapes-VPS dataset provides a split into \num{400} training, \num{50} validation, and \num{50} test clips. We perform evaluation on the \num{50} validation video sequences, following the originally proposed setting.

\inparagraphnospace{KITTI-STEP~\cite{Weber:2021:STE}} provides panoptic video annotations for the KITTI-MOTS dataset~\cite{Voigtlaender:2019:MMO, kitti} and comprises \num{12} training, \num{9} validation, and \num{29} test videos. While Cityscapes-VPS provides more video clips and ``thing'' detections per frame, KITTI-STEP provides significantly longer video clips, on average \num{381} annotated frames per sequence, and longer tracks (average track length \num{51} frames). This is significantly longer than the \num{6} annotated frames of each Cityscapes-VPS clip. The semantic taxonomy of KITTI-STEP matches the \num{19}-class taxonomy of Cityscapes, however, provides fewer ``thing'' classes---only persons and cars are annotated instance-wise. To compensate for this during evaluation, we perform semantic matching using the ``thing'' and ``stuff'' separation of Cityscapes and ignore video instance predictions of semantic classes different than ``person'' and ``car''. For evaluation, we use the validation split.

\inparagraphnospace{Waymo~\cite{Mei:2022:WAY, Sun:2020:SPA}} comprises panoramic video panoptic annotations for \num{2860} clips, covering a broad range of street scenes under diverse conditions (\eg, night, rain, \etc). The dataset provides five camera views and is split into \num{2002} training, \num{286} validation, and \num{572} test clips. We use the forward-facing view with a resolution of \num{1080}\,$\times$\,\num{1920} pixels. Following established practice~\cite{Hahn:2025:UPS}, we map Waymo’s semantic labels to the Cityscapes taxonomy, resulting in 16 categories, and report results on the validation split. As Waymo includes a substantial number of very small ``thing'' detections, we remove instances with an average track size below \num{400} pixels. This mitigates the impact of extremely fine-grained annotations, which current unsupervised approaches cannot segment reliably, and ensures comparability to Cityscapes-VPS.

\inparagraphnospace{MOTS~\cite{Voigtlaender:2019:MMO}} is used to assess scenes-centric VPS accuracy on videos different from autonomous driving scenarios. We utilize the four MOTChallenge training video sequences, each composed of \num{2866} frames, for evaluation. These frames entail a resolution of \num{1080}$\,\times\,$\num{1920} or \num{480}$\,\times\,$\num{640} pixels. MOTS provides two annotated categories ``person'' and ``background'', while providing video instance segmentation for each ``person'' instance. We consider ``person'' as a ``thing'' category and ``background'' as a ``stuff'' category.

\subsection{Implementation details}
We implement \MethodName using PyTorch~\cite{pytorch} and train using PyTorch Lightning~\cite{pytorchl}. We utilized Detectron2~\cite{detectron2} for implementing Panoptic Cascade MaskTrack R-CNN and Kornia~\cite{kornia} for augmentations. Our implementation is partly built upon the code from previous work~\cite{Hamilton:2022:USS, Sick:2024:USS, Hahn:2025:UPS}.

\inparagraph{Pre-trained models.} Our full pipeline utilizes SMURF~\cite{Teed:2020:RAF}, Dynamo-Depth~\cite{Sun:2023:DDF}, and DepthG~\cite{Sick:2024:USS}. To ensure full compliance with our purely unsupervised and monocular setup, we retrain Dynamo-Depth and DepthG. While SMURF has already been trained using monocular, unlabeled videos, DepthG uses a supervised depth model, and Dynamo-Depth initializes training with an ImageNet~\cite{Russakovsky:2015:ILS} supervised backbone. In particular, we retrain Dynamo-Depth~\cite{Sun:2023:DDF} with a DINO ResNet-18~\cite{He:2016:DRL, Caron:2021:EPS}, instead of an ImageNet-supervised~\cite{Russakovsky:2015:ILS} ResNet-18. DepthG is retrained with the monocular depth estimates of our retrained Dynamo-Depth model.

\inparagraph{Pseudo-label generation.} We generate panoptic video pseudo-labels using $c_{\text{p}} = \text{\num{27}}$ pseudo-classes on the Cityscapes training sequences, following CUPS~\cite{Hahn:2025:UPS} and use a thing-stuff threshold of $\mathbf{\psi^{\rm ts}}=\text{\num{0.01}}$ (\cf \cref{sec:thingstuffanalysis}). Semantic pseudo-labeling uses the prediction of our retrained DepthG, and depth-guided semantic inference follows the same setting as proposed in~\cite{Hahn:2025:UPS}. We post-process pseudo-labels with a CRF~\cite{Kraehenbuehl:2011:EIF} using regularized Frank-Wolfe inference \cite{Huu:2021:RFD} and use the original hyperparameters. SMURF and our retrained Dynamo-Depth are used for pseudo-labeling. Our region growing uses $\alpha=\text{\num{0.15}}$, $\tau_{d}=\text{\num{0.02}}$, $\tau_{f}=\text{\num{0.04}}$, and $r=\text{\num{8}}$ (\cf \cref{tab:supp_regiongrowing_hyperparameter}). Instance propagation and tracking uses a sliding window of length three and an IoU-threshold of $\tau_m = \text{\num{0.4}}$ in Hungarian matching (\cf \cref{tab:trackingthresh}). Temporal semantic smoothing likewise uses a three-frame sliding window.

\inparagraph{Training and evaluation.} To ensure fairness to our baselines U2Seg~\cite{Niu:2024:UUI} and CUPS~\cite{Hahn:2025:UPS}, which employ a Panoptic Cascade Mask R-CNN~\cite{Cai:2018:CRC, Kirillov:2019:PFP}, we use the closest video extension, the Panoptic Cascade MaskTrack R-CNN~\cite{Cai:2018:CRC, Kirillov:2019:PFP, Yang:2019:VIS}. Following CutLER~\cite{Wang:2023:CAL}, U2Seg~\cite{Niu:2024:UUI}, and CUPS~\cite{Hahn:2025:UPS}, we utilize a ResNet-50~\cite{He:2016:DRL} backbone with DINO~\cite{Caron:2021:EPS} initialization, pre-trained self-supervised for two epochs on ImageNet~\cite{Russakovsky:2015:ILS}. Building on CUPS~\cite{Hahn:2025:UPS}, we train using AdamW~\cite{Loshchilov:2018:ADW} with a base learning rate of \num{2e-5}, our self-enhanced video copy-paste augmentation, and our Video DropLoss (with $\tau_{\operatorname{IoU}}=\text{\num{0.5}}$) for eight epochs. Our self-enhanced video copy-paste augmentation starts using model predictions after one epoch. During the first epoch, we copy-paste pseudo-labels. We paste between one and eight ``thing'' video detections into another video clip. Training was performed on four NVIDIA A100 GPUs (\SI{80}{GB}) using a batch size of \num{24}. Evaluation of \MethodName and our unsupervised baselines is performed using the native resolution of each dataset (\eg, \num{1024}\,$\times$\,\num{2048} for Cityscapes~\cite{Cordts:2016:CDS}).

\section{Unsupervised VPS Evaluation Protocol}\label{sec:supp_eval}

As we train in a fully unsupervised fashion, our model only predicts pseudo-classes. These need to be mapped to the ground-truth categories. For this, we presented a simple and hyperparameter-free approach in \cref{subsec:eval_vups}. After mapping pseudo-categories to ground-truth categories, we utilize the established Segmentation and Tracking Quality (STQ)~\cite{Weber:2021:STE}. In the following, we provide details on the STQ and discuss other VPS metrics.

\subsection{Segmentation and Tracking Quality}
After mapping pseudo-categories to ground-truth categories (\cf \cref{subsec:eval_vups}), we are equipped with the VPS predictions per clip $\mathbf{P}_{i}=(\check{\mathbf{S}}_{i},\mathbf{R}_{i})$, where $\check{\mathbf{S}}_{i}\in \{1,2,\ldots,c_{\text{gt}}\}^{\rm T \times \rm H\times W}$ denotes the matched semantic predictions obtained using the pseudo-semantics $\mathbf{S}$; $i$ is the clip index, and $\mathbf{R}_{i}\in\{0, 1\}^{n_{p} \times \rm T \times \rm H\times W}$ indicates the per-frame presence of $n_{p}$ predicted ``thing'' video instances. For evaluation, we use the ground-truth VPS labels $\bar{\mathbf{P}}_{i}=(\bar{\mathbf{S}}_{i},\bar{\mathbf{R}}_{i})$, with the semantic ground truth $\bar{\mathbf{S}}_{i}\in\{1,2,\ldots,c_{\text{gt}}\}^{\rm T \times \rm H\times W}$ and the corresponding $n_{\text{gt}}$ binary video instance masks $\bar{\mathbf{R}}_{i}\in\{0, 1\}^{n_{\text{gt}} \times \rm T \times \rm H\times W}$. The Segmentation and Tracking Quality is computed as
\begin{equation}
    \operatorname{STQ}=(\operatorname{AQ} \cdot \operatorname{SQ})^{\frac{1}{2}},
\end{equation}
where $\operatorname{AQ}$ is the Association Quality, and $\operatorname{SQ}$ is the Segmentation Quality. Specifically, the $\operatorname{AQ}$ measures how accurately instances were detected and tracked over time, while $\operatorname{SQ}$ measures how well pixel semantics were predicted, effectively decoupling segmentation and association.

More specifically, the Segmentation Quality, $\operatorname{SQ}$, is defined as the mean Intersection over Union over the ground-truth classes $c\in\{1,2,\ldots,c_{\text{gt}}\}$ computed as
\begin{equation}
    \operatorname{SQ}=\frac{1}{c_{\text{gt}}}\sum_{c\in\{1,\ldots,c_{\text{gt}}\}}\frac{\text{TP}_{c}}{\text{TP}_{c} + \text{FP}_{c} + \text{FN}_{c}},
\end{equation}
with
\begin{align}
    \mathrm{TP}_{c}&=\sum_{i,t,h,w}\mathds{1}\!\left[\check{\mathbf{S}}_{i,t,h,w}=c\right]\,\mathds{1}\!\left[\bar{\mathbf{S}}_{i,t,h,w}=c\right]\!, \\
    \mathrm{FP}_{c}&=\sum_{i,t,h,w}\mathds{1}\!\left[\check{\mathbf{S}}_{i,t,h,w}=c\right]\,\mathds{1}\!\left[\bar{\mathbf{S}}_{i,t,h,w}\neq c\right]\!, \\
    \mathrm{FN}_{c}&=\sum_{i,t,h,w}\mathds{1}\!\left[\check{\mathbf{S}}_{i, t,h,w}\neq c\right]\,\mathds{1}\!\left[\bar{\mathbf{S}}_{i,t,h,w}=c\right]\!,
\end{align}
computed over the temporal and spatial dimensions as well as all evaluated clips.

The Association Quality, $\operatorname{AQ}$, is computed using the ``thing'' video detections $\mathbf{R}_{i}$ and the ground truth $\bar{\mathbf{R}}_{i}$. First, the true positive $\operatorname{TPA}_{i}(g,p)$, false positive $\mathrm{FPA}_{i}(g,p)$, and false negative $\operatorname{FNA}_{i}(g,p)$ association areas for the predicted video instances $g\in\{1,2,\ldots,n_{p}\}$ and the ground-truth video instances $p\in\{1,2,\ldots,n_{\text{gt}}\}$ are computed per clip $i$ as
\begin{align}
    \operatorname{TPA}_{i}(g,p)&=\sum_{t,h,w}\bar{\mathbf{R}}_{i,g, t,h,w}\,\mathbf{R}_{i,p,t,h,w},\\
    \operatorname{FPA}_{i}(g,p)&=\sum_{t,h,w}\mathbf{R}_{i,p,t,h,w}-\operatorname{TPA}_{i}(g,p), \\
    \operatorname{FNA}_{i}(g,p)&=\sum_{t,h,w}\bar{\mathbf{R}}_{i,g,t,h,w}-\operatorname{TPA}_{i}(g,p).
\end{align}
Next, the pairwise association Intersection over Union $\operatorname{IoU}^{\text{A}}_{i}(g,p)$ is computed using
\begin{equation}
\operatorname{IoU}^{\text{A}}_{i}(g,p)\!=\!\frac{\operatorname{TPA}_{i}(g,p)}{\operatorname{TPA}_{i}(g,p)\!+\!\operatorname{FPA}_{i}(g,p)\!+\!\operatorname{FNA}_{i}(g,p)}.
\end{equation}
Finally, using $\operatorname{TPA}_{i}(g,p)$ and $\operatorname{IoU}^{\text{A}}_{i}(g,p)$, the Association Quality is computed as
\begin{equation}
    \operatorname{AQ}=\sum_{i}\frac{\displaystyle\sum_{g=1}^{n_{\mathrm{gt},i}}\frac{1}{\left\lvert \bar{\mathbf{R}}_{i,g} \right\rvert}\sum_{p=1}^{n_{p,i}}\operatorname{TPA}_{i}(g,p)\,\operatorname{IoU}^{\text{A}}_{i}(g,p)}{n_{\mathrm{gt}, i}},
\end{equation}
where $\left\lvert \bar{\mathbf{R}}_{i,g} \right\rvert$ denotes the total area (\ie, the number of pixels) of the ground-truth video instances $g$, $n_{\mathrm{gt}, i}$ the number of ground-truth instances of clip $i$, and $n_{p,i}$ the number of predicted instances of clip $i$.

\subsection{Discussion}
Existing work in the supervised domain offers alternative metrics for evaluating VPS, including the Video Panoptic Quality (VPQ)~\cite{Kim:2020:VPS} and Panoptic Tracking Quality (PTQ)~\cite{Hurtado:2020:MOP}. However, we adopt STQ as our primary metric in our proposed evaluation protocol for several reasons. \emph{First}, STQ provides a clear separation between segmentation and association quality, yielding more interpretable insights into model behavior. \emph{Second}, STQ evaluates entire videos at the per-pixel level and avoids the need for temporal windowing. In contrast, VPQ relies on fixed window sizes, which introduce sensitivity to the chosen hyperparameter and do not scale to full-length videos~\cite{vipseg, Weber:2021:STE}. \emph{Third}, STQ does not use a threshold-based matching of ``thing'' predictions, in contrast to both PTQ and VPQ, making it more robust across object scales and crowded scenes. For these reasons, we refrain from incorporating VPQ into our evaluation protocol, as STQ provides the most stable, interpretable, and hyperparameter-free evaluation. In doing so, we follow KITTI-STEP~\cite{Weber:2021:STE}, the 2D Video Panoptic Segmentation Challenge at CVPRW 2023, and the Waymo panoramic VPS dataset~\cite{Mei:2022:WAY}. For a more detailed discussion of different VPQ metrics, we refer to Weber~\etal~\cite{Weber:2021:STE}. 

Nevertheless, we report VPQ for \MethodName and all baselines on KITTI-STEP in \cref{tab:vpq} for completeness. \MethodName consistently outperforms all proposed baselines in both VPQ and STQ. We observe that the relative accuracy gap between the methods is smaller for VPQ than for STQ. We attribute this to the fundamental differences between the two metrics: VPQ is dominated by per-frame/per-window mask quality, whereas STQ explicitly captures both recognition and temporal consistency. Additionally, STQ does not penalize for the recovery/correcting of tracks \cite[see][for more details]{Weber:2021:STE}. Therefore, STQ better reflects progress in unsupervised VPQ.

\begin{table}[t]
    \centering
    \caption{\textbf{VPQ \vs STQ.} We compare \MethodName to our unsupervised VPS baselines using VPQ and STQ (all in \%, $\uparrow$) on KITTI-STEP. Both in VPQ and STQ, \MethodName outperforms the baselines. $\dagger$ denotes CUPS retrained using monocular videos.}
    \vspace{-0.3em}
    \footnotesize\sisetup{table-number-alignment=center}
\setlength{\tabcolsep}{15pt}
\renewcommand{\arraystretch}{0.875}
\begin{tabularx}{\columnwidth}{>{\hspace{-\tabcolsep}\raggedright\columncolor{white}[\tabcolsep][\tabcolsep]}Xc|ZZ}
	\toprule
    \textbf{Method} & {\textbf{VPQ}} & {\textbf{STQ}} \\
    \midrule
    {DepthG~\cite{Sick:2024:USS}~+~VideoCutLER~\cite{Wang:2024:VCS}} & 14.3 & 13.2 \\
    {U2Seg~\cite{Niu:2024:UUI}~+~SORT~\cite{Bewley:2016:SOR}} & 19.0 & 24.0 \\
    {CUPS~\cite{Hahn:2025:UPS}~+~SORT~\cite{Bewley:2016:SOR}} & 20.4 & 34.2 \\
    {CUPS\textsuperscript{$\dagger$} \cite{Hahn:2025:UPS}~+~SORT~\cite{Bewley:2016:SOR}} & 20.0 & 32.9 \\
    \midrule
    \rowcolor{tud0c!20} 
    {\MethodName \textit{(Ours)}} & \bfseries 21.1 & \bfseries 37.3 \\
	\bottomrule %
\end{tabularx}

    \vspace{-0.5em}
    \label{tab:vpq}
\end{table}

\section{Additional Results}\label{sec:supp_results}

Here, we provide additional qualitative and quantitative results extending our experiments in the main paper (\cf \cref{sec:experiments}).

\subsection{Pseudo-label thing-stuff threshold analysis \label{sec:thingstuffanalysis}}
The thing–stuff threshold $\psi^{\mathrm{ts}}$ introduced in \cref{subsec:pseudo_label_generation} partitions the semantic pseudo-classes into pseudo-\emph{thing} and pseudo-\emph{stuff} classes based on their frequency inside the instance masks across the training data. \Cref{tab:pseudo_thing_stuff} reports the effect of varying $\psi^{\mathrm{ts}}$ on pseudo-labels generated for the Cityscapes-VPS validation split. Very low thresholds assign many semantic pseudo-classes (\eg, \num{10} for $\psi^{\mathrm{ts}}=\num{0.0025}$) to the ``thing'' subset, leading to a form of over-clustering of true instance categories. Conversely, high thresholds reduce the number of thing pseudo-classes (\eg, \num{3} for $\psi^{\mathrm{ts}}=\num{0.03}$), which degrades results, as measured by the STQ. The best results are obtained at $\psi^{\mathrm{ts}}=\num{0.01}$, yielding five pseudo-thing classes and an STQ of \SI{12.1}{\%}.

\begin{table}[t]
    \centering
    \caption{\textbf{Pseudo-label thing-stuff threshold analysis.} We evaluate pseudo-labels generated on Cityscapes-VPS val using different values for the thing-stuff threshold $\psi^{\rm ts}$ using STQ (in $\%, \uparrow$).}
    \vspace{-0.3em}
    \footnotesize\sisetup{table-number-alignment=center}
\setlength{\tabcolsep}{9.0pt}
\renewcommand{\arraystretch}{0.885}
\begin{tabularx}{\columnwidth}{@{}Xccccccc}
	\toprule
    $\mathbf{\psi^{\rm ts}}\;\to$ & 0.0025 & 0.005 & \cellcolor{tud0c!20} 0.01 & 0.02 & 0.03 \\
    \midrule
    \bfseries STQ & 7.5  & 11.5  & \bfseries 12.1 & 11.3  & 10.7 \\
	\bottomrule
\end{tabularx}

    \label{tab:pseudo_thing_stuff}
\end{table}

\subsection{Tracking threshold analysis \label{sec:tracking thresh}}

In \cref{tab:trackingthresh}, we analyze the influence of the IoU-threshold $\tau_m$ used for tracking and instance propagation (\cf \Cref{subsec:pseudo_label_generation}). In particular, we generate pseudo-labels on the Cityscapes-VPS validation split and evaluate the pseudo-labels following the experimental setup from \cref{tab:pseudo_ablation}. Overall, our pseudo-labeling is robust to different $\tau_m$ values. Nonetheless, setting $\tau_m=0.4$ yields a slightly better STQ than \num{0.3} and \num{0.5}.

\begin{table}[t]
    \centering
    \caption{\textbf{Video pseudo-label instance propagation and tracking threshold $\tau_m$ analysis}, using different IoU thresholds evaluated on pseudo-labels generated on Cityscapes-VPS val, using STQ, AQ, and SQ (all in \%, $\uparrow$).}
    \vspace{-0.3em}
    \footnotesize\sisetup{table-number-alignment=center}
\setlength{\tabcolsep}{9pt}
\renewcommand{\arraystretch}{0.875}
\begin{tabularx}{\columnwidth}{>{\hspace{-\tabcolsep}\raggedright\columncolor{white}[\tabcolsep][\tabcolsep]}XZZZ}
	\toprule
    \textbf{Pseudo-label configuration} & \textbf{STQ} & \textbf{AQ} & \textbf{SQ} \\
    \midrule  
        $\tau_m =$ \num{0.3}    & 11.8 & 4.4 & 32.2  \\
        \rowcolor{tud0c!20} $\tau_m =$ \num{0.4}    & \bfseries 12.1 & \bfseries 4.5 & \bfseries 32.3  \\
        $\tau_m =$ \num{0.5}    & 12.0 & 4.4 & \bfseries 32.3  \\
	\bottomrule
\end{tabularx}

    \vspace{-0.5em}
    \label{tab:trackingthresh}
\end{table}

\begin{table}[t]
    \centering
    \caption{\textbf{Instance pseudo-labeling hyperparameter analysis.} We analyse our region growing hyperparameters (instance seed threshold $\alpha$, relative depth threshold $\tau_d$, relative flow threshold $\tau_f$, and neighborhood radius $\rm r$) on Cityscapes val and report STQ (in \%, $\uparrow$).}
    \label{tab:supp_regiongrowing_hyperparameter}
    \vspace{-0.3em}
    \footnotesize
\setlength{\tabcolsep}{1.25pt}
\begin{tabularx}{\columnwidth}{XZZZ|ZZZ|ZZZ|ZZZ@{}}
    \toprule \\[-10pt]
    & \multicolumn{3}{c|}{$\xleftarrow{}~\alpha~\xrightarrow{}$} & \multicolumn{3}{c|}{$\xleftarrow{}~\tau_{d}~\xrightarrow{}$} & \multicolumn{3}{c|}{$\xleftarrow{}~\tau_{f}~\xrightarrow{}$} & \multicolumn{3}{c}{$\xleftarrow{}~\rm r~\xrightarrow{}$} \\[1pt]
    & {0.05} & \cellcolor{tud0c!20}{0.15} & {0.25} & {0.01} & \cellcolor{tud0c!20}{0.02} & {0.04} & {0.01} & \cellcolor{tud0c!20}{0.04} & {0.07} & {2} & \cellcolor{tud0c!20}{8} & {14} \\
    \specialrule{0.5pt}{1.5pt}{2pt}
    \textbf{STQ} & \bfseries 12.1  & \bfseries 12.1 & 11.3 & 11.6  & \bfseries 12.1 & 12.0 & 12.0  & \bfseries 12.1 & 12.0 & \bfseries 12.1  & \bfseries 12.1 & \bfseries 12.1\\
    \bottomrule
\end{tabularx}

    \vspace{-0.5em}
    \label{tab:supp_instance_analysis}
\end{table}

\subsection{Instance pseudo-labeling analysis}

In \cref{tab:supp_instance_analysis}, we provide an analysis of our instance pseudo-labeling hyperparameters. We again generate pseudo-labels on the Cityscapes-VPS validation split and evaluate the pseudo-labels following the experimental setup from \cref{tab:pseudo_ablation}. We observe a robust behaviour of our pseudo-labeling \wrt the relative motion threshold $\tau_f$ and the neighbourhood radius $r$. Both the instance seed threshold $\alpha$ and the relative depth threshold $\tau_d$ still exhibit a relatively robust behaviour, while less robust than $\tau_f$ and $r$.

\subsection{Dynamic vs. static analysis}
\Cref{tab:supplement_dynamic_static} analyses the accuracy of \MethodName and our pseudo-labels on dynamic and static ``thing'' instances only. In particular, we utilize Cityscapes ground-truth motion masks~\cite{csmotionmasks} and compute STQ for moving ``thing'' instances (STQ\textsuperscript{D}) and static ``thing'' instances (STQ\textsuperscript{S}). When computing STQ\textsuperscript{D}, we ignore all ``stuff'' regions and static ``thing'' instances. Similarly, for STQ\textsuperscript{S} we ignore all ``stuff'' regions and dynamic ``thing'' instances. While our pseudo-labels only capture dynamic objects, \MethodName improves accuracy on \emph{both} static and dynamic objects. These results demonstrate the effectiveness of our Video DropLoss in enabling the network to detect and track objects missed by our pseudo-labels. Note that Cityscapes-VPS contains significantly more and smaller static than dynamic instances~\cite{csmotionmasks}, resulting in lower STQ\textsuperscript{S}, which is also observed for the supervised upper bound.

\subsection{Pseudo-labeling oracle}
In \cref{tab:supplement_pseudo_oracle}, we provide an oracle experiment by using supervised cues to generate pseudo labels. In particular, we use supervised depth~\cite{dav2}, flow~\cite{searaft}, and motion masks~\cite{csmotionmasks} for pseudo-labeling. These supervised cues significantly improve pseudo-label accuracy, demonstrating the potential benefit of more accurate unsupervised depth, flow, and motion segmentation to improve unsupervised VPS. Note that this only improves the moving-object masks while still using unsupervised semantics.

\begin{table}[t]
    \centering
    \caption{\textbf{Dynamic \vs static analysis.} We report STQ for dynamic (STQ\textsuperscript{D}) and static ``thing'' objects (STQ\textsuperscript{S}) only, ignoring stuff pixels. Both metrics in \% ($\uparrow$) on Cityscapes val.}
    \vspace{-0.3em}
    \footnotesize\sisetup{table-number-alignment=center}
\setlength{\tabcolsep}{5pt}
\begin{tabularx}{\columnwidth}{>{\hspace{-\tabcolsep}\raggedright\columncolor{white}[\tabcolsep][\tabcolsep]}XZZ@{}}
    \toprule
    \textbf{Approach} & \textbf{STQ\textsuperscript{D}} & \textbf{STQ\textsuperscript{S}} \\
    \specialrule{0.5pt}{1.5pt}{2pt}
    \textcolor{tud0c}{Supervised} & \color{tud0c}42.8 & \color{tud0c}28.6 \\
    \specialrule{0.5pt}{1.5pt}{2pt}
    Pseudo-labels & 16.8 & 5.9 \\
    \MethodName & \bfseries 23.9 & \bfseries 18.8 \\
    \bottomrule
\end{tabularx}

    \vspace{-0.5em}
    \label{tab:supplement_dynamic_static}
\end{table}

\begin{table}[t]
    \centering
    \caption{\textbf{Pseudo-labeling oracle.} We analyze pseudo-labels generated using supervised depth, flow \& motion masks and our unsupervised pseudo-labels on Cityscapes val. using STQ, AQ, and SQ (all in \%, $\uparrow$).}
    \vspace{-0.3em}
    \footnotesize\sisetup{table-number-alignment=center}
\setlength{\tabcolsep}{5pt}
\begin{tabularx}{\columnwidth}{>{\hspace{-\tabcolsep}\raggedright\columncolor{white}[\tabcolsep][\tabcolsep]}XZZZ@{}}
    \toprule
    \textbf{Pseudo-labels} & \textbf{STQ} & \textbf{AQ} & \textbf{SQ} \\
    \specialrule{0.5pt}{1.5pt}{2pt}  
    Supervised & \bfseries 17.3 & \bfseries 8.7 & \bfseries 34.3 \\ 
    Unsupervised \emph{(Ours)} & 12.1 & 4.5 & 32.3 \\
    \bottomrule
\end{tabularx}

    \vspace{-0.5em}
    \label{tab:supplement_pseudo_oracle}
\end{table}

\begin{table}[t]
    \centering
    \caption{\textbf{SSL features for semantic pseudo-labeling analysis.} We compare our modified version of DepthG~\cite{Sick:2024:USS} using DINO~\cite{Caron:2021:EPS} and DINOv3~\cite{Simeoni:2025:DIN}, evaluating semantic image segmentation using mIoU (in $\%, \uparrow$) for unsupervised clustering and supervised linear probing on Cityscapes val.}
    \vspace{-0.3em}
    \footnotesize\sisetup{table-number-alignment=center}
\setlength{\tabcolsep}{5pt}
\renewcommand{\arraystretch}{0.875}
\begin{tabularx}{\columnwidth}{>{\hspace{-\tabcolsep}\raggedright\columncolor{white}[\tabcolsep][\tabcolsep]}XZZ}
	\toprule
    \textbf{SSL-Features} & {\textbf{Unsupervised mIoU}} & {\color{tud0c}\textbf{Supervised mIoU}} \\
    \midrule
    {DINO~\cite{Caron:2021:EPS}} & 23.2  & \color{tud0c}28.6 \\
    {DINOv3~\cite{Simeoni:2025:DIN}} & 22.0  & \color{tud0c}41.0 \\
	\bottomrule
\end{tabularx}

    \vspace{-0.5em}
    \label{tab:sslfeats}
\end{table}

\subsection{Analysing SSL features for semantic pseudo-labeling \label{sec:sslfeats}}

We analyze the effect of different SSL feature representations on the unsupervised semantic segmentation component used for our pseudo-labeling (\cf \cref{sec:semanticpseudolabeling}). We experiment using our DepthG~\cite{Sick:2024:USS} variant, adapted to the unsupervised and monocular setting by replacing supervised depth with monocular predictions from Dynamo-Depth~\cite{Sun:2023:DDF}. Both \MethodName and the original DepthG employ DINO~\cite{Caron:2021:EPS} ViT-Base/8 features. We additionally evaluate DINOv3~\cite{Simeoni:2025:DIN} ViT-Base/16 features under the standard unsupervised semantic image segmentation protocol~\cite{Cho:2021:PUS, Hamilton:2022:USS, Seong:2023:LHP, Kim:2024:EAL, Sick:2024:USS, Hahn:2024:BUS} and report the mean Intersection over Union in \cref{tab:sslfeats}.

Despite stronger segmentation results from supervised linear probing, DINOv3 yields inferior unsupervised segmentation mIoU compared to DINO (\cf \cref{tab:sslfeats}). This aligns with prior observations for DINO compared to DINOv2~\cite{Hahn:2024:BUS}. We attribute the drop to the substantially larger patch sizes in DINOv2/v3, which result in a reduced spatial resolution. While the representations become more discriminative in a supervised setting, their coarse spatial granularity appears detrimental for unsupervised clustering. We use DINO(v1) features in our experiments to ensure a fair comparison with U2Seg~\cite{Niu:2024:UUI} and CUPS~\cite{Hahn:2025:UPS}.

\begin{table*}[t]
    \centering
    \caption{\textbf{Class-level results on Cityscapes-VPS val.} We compare \MethodName to the unsupervised VPS baselines, using the class-wise segmentation quality (SQ, in \%, $\uparrow$).
    $\dagger$\ denotes CUPS retrained using monocular videos.
    *\ denotes ``thing'' classes with spatio-temporal instance annotations. For reference, we also report the class-wise scores of \MethodName fine-tuned with \SI{10}{\%} of the Cityscapes-VPS annotations.}
    \vspace{-0.3em}
    \footnotesize\sisetup{table-number-alignment=center}
\setlength{\tabcolsep}{2.2pt}
\renewcommand{\arraystretch}{1.05}

\begin{tabularx}{\textwidth}{X *{19}{>{\centering\arraybackslash}Z}|Z}
    \toprule
    {\textbf{Method}} 
    & \cellcolor{road}\rotatebox{90}{\textcolor{white}{\textbf{Road}}}
    & \cellcolor{sidewalk}\rotatebox{90}{\textcolor{white}{\textbf{Sidewalk}}}
    & \cellcolor{building}\rotatebox{90}{\textcolor{white}{\textbf{Building}}}
    & \cellcolor{wall}\rotatebox{90}{\textcolor{white}{\textbf{Wall}}}
    & \cellcolor{fence}\rotatebox{90}{\textcolor{white}{\textbf{Fence}}}
    & \cellcolor{pole}\rotatebox{90}{\textcolor{white}{\textbf{Pole}}}
    & \cellcolor{trafficlight}\rotatebox{90}{\textcolor{white}{\textbf{Traffic~Light$\;$}}}
    & \cellcolor{trafficsign}\rotatebox{90}{\textcolor{white}{\textbf{Traffic~Sign}}}
    & \cellcolor{vegetation}\rotatebox{90}{\textcolor{white}{\textbf{Vegetation}}}
    & \cellcolor{terrain}\rotatebox{90}{\textcolor{white}{\textbf{Terrain}}}
    & \cellcolor{sky}\rotatebox{90}{\textcolor{white}{\textbf{Sky}}}
    & \cellcolor{person}\rotatebox{90}{\textcolor{white}{\textbf{Person\textsuperscript{*}}}}
    & \cellcolor{rider}\rotatebox{90}{\textcolor{white}{\textbf{Rider\textsuperscript{*}}}}
    & \cellcolor{car}\rotatebox{90}{\textcolor{white}{\textbf{Car\textsuperscript{*}}}}
    & \cellcolor{truck}\rotatebox{90}{\textcolor{white}{\textbf{Truck\textsuperscript{*}}}}
    & \cellcolor{bus}\rotatebox{90}{\textcolor{white}{\textbf{Bus\textsuperscript{*}}}}
    & \cellcolor{train}\rotatebox{90}{\textcolor{white}{\textbf{Train\textsuperscript{*}}}}
    & \cellcolor{motorcycle}\rotatebox{90}{\textcolor{white}{\textbf{Motorcycle\textsuperscript{*}}}}
    & \cellcolor{bicycle}\rotatebox{90}{\textcolor{white}{\textbf{Bicycle\textsuperscript{*}}}}
    & \rotatebox{90}{\textbf{Mean (SQ)}}
    \\
    
    \midrule
    {\textcolor{tud0c}{Supervised~\cite{Kirillov:2019:PFP}}} & \color{tud0c} 88.7 & \color{tud0c} 73.0 & \color{tud0c} 83.1 & \color{tud0c} 47.9 & \color{tud0c} 57.6 & \color{tud0c} 51.1 & \color{tud0c} 47.9 & \color{tud0c} 63.8 & \color{tud0c} 82.4 & \color{tud0c} 58.2 & \color{tud0c} 85.7 & \color{tud0c} 72.0 & \color{tud0c} 58.7 & \color{tud0c} 84.8 & \color{tud0c} 65.7 & \color{tud0c} 82.5 & \color{tud0c} 26.2 & \color{tud0c} 47.2 & \color{tud0c} 64.9 & \color{tud0c} 65.3 \\
    
    \midrule

    {DepthG~\cite{Sick:2024:USS}~+~VideoCutLER~\cite{Wang:2024:VCS}} &  85.4	&  17.8	& 67.5	& 1.3	& 9.5	& 8.5	& 7.2	& 24.9	& 81.3	& 25.0	& 78.0	& 53.6	& 0.0	& 74.6	& 2.0	& 0.0	& 0.0	& 0.0	& 0.0 & 28.2 \\
    {U2Seg~\cite{Niu:2024:UUI}~+~SORT~\cite{Bewley:2016:SOR}} & 77.4	& 0.4	& 50.6	& 7.2	& 1.2	& 1.0	& 0.4	& 0.1	& 77.7	& 22.5	& 74.9	& 15.1	& 2.9	& 48.0	& 2.7	& 32.8	& 0.5	& 4.5	& 17.8 & 23.0 \\
    {CUPS~\cite{Hahn:2025:UPS}~+~SORT~\cite{Bewley:2016:SOR}} & 81.1	& 16.2	& 63.2	& 1.3	& 4.9	& 30.6	& 17.6	& 40.7	& 80.9	& 36.3	& 81.9	& 47.9	& 0.0	& 61.1	& 0.0	& 23.3	& 0.0	& 0.0	& 18.2 & 31.8 \\
    {CUPS\textsuperscript{$\dagger$} \cite{Hahn:2025:UPS}~+~SORT~\cite{Bewley:2016:SOR}} & 81.3	& 13.2	& 62.2	& 1.5	& 5.0	& 30.3	& 19.7	& 38.1	& 81.1	& 30.1	& 81.9	& 45.4	& 0.0	& 62.8	& 0.0	& 12.4	& 0.0	& 0.0	& 2.3 & 29.9 \\
    \midrule
    \rowcolor{tud0c!20} {\MethodName \textit{(Ours)}} & 81.4	& 17.7	&  67.9	& 1.0	& 9.9	& 27.0	& 8.8	& 39.1	& 80.9	& 32.5	& 83.4	& 53.1	& 0.0	& 67.8	& 0.0	& 10.1	& 0.0	& 0.6	& 31.5 &  \bfseries 32.3\\
    \textcolor{tud0c}{\MethodName w/ \SI{10}{\%} VPS ann. \textit{(Ours)}} & \color{tud0c} 85.6 & \color{tud0c} 43.9 & \color{tud0c} 83.0 & \color{tud0c} 16.8 & \color{tud0c} 35.7 & \color{tud0c} 37.4 & \color{tud0c} 37.5 & \color{tud0c} 52.7 & \color{tud0c} 83.0 & \color{tud0c} 19.5 & \color{tud0c} 88.2 & \color{tud0c} 70.0 & \color{tud0c} 44.8 & \color{tud0c} 82.9 & \color{tud0c} 24.8 & \color{tud0c} 5.1 & \color{tud0c} 21.6 & \color{tud0c} 18.2 & \color{tud0c} 56.5 & \color{tud0c} 47.7 \\
    \bottomrule
\end{tabularx}

    \vspace{-0.5em}
    \label{tab:classscores}
\end{table*}

\subsection{Class-level analysis}

\Cref{tab:classscores} provides class-wise Segmentation Quality results of \MethodName and our baselines. Note that SQ only measures the segmentation accuracy, not detection and tracking accuracy. We observe that rare classes are still a significant challenge for \emph{all} unsupervised approaches. For example, the predictions of \MethodName only capture five out of the eight ``thing'' classes. CUPS~+~SORT and CUPS\textsuperscript{\textdagger}~+~SORT only capture four ``thing'' classes. Notably, while scoring a significantly lower overall SQ and aligning not well with ground-truth instances (\cf \cref{tab:uvps_cs}), U2Seg~+~SORT predicts all ``thing'' and ``stuff'' classes, most likely due to the significant overclustering with \num{827} pseudo-classes. On average, \MethodName outperforms our proposed baselines. In comparison to the supervised upper bound, missed classes account for most of the accuracy gap between supervised and unsupervised approaches, including \MethodName.
For frequent classes (\eg, ``Road'', ``Sky'', or ``Car''), our unsupervised VPS almost matches the results of the supervised upper bound.
Fine-tuning on just a few VPS annotations (\SI{10}{\%} of Cityscapes-VPS train) can adapt \MethodName to predict all semantic classes. 

\subsection{Qualitative results}
In addition to the Cityscapes-VPS qualitative results in \cref{subsec:uvpsresults}, we present further visual comparisons. We compare \MethodName to the proposed baselines DepthG~\cite{Sick:2024:USS}\ +\ VideoCutLER~\cite{Wang:2024:VCS}, U2Seg~\cite{Niu:2024:UUI}\ +\ SORT~\cite{Bewley:2016:SOR}, CUPS~\cite{Hahn:2025:UPS}\ +\ SORT, and CUPS\textsuperscript{\textdagger}~\cite{Hahn:2025:UPS}\ +\ SORT, across KITTI-STEP~\cite{Weber:2021:STE}, Waymo~\cite{Mei:2022:WAY, Sun:2020:SPA}, and MOTS~\cite{Voigtlaender:2019:MMO}. We also include qualitative out-of-domain (OOD) results on DAVIS~\cite{davis}. 
Importantly, we apply no post-processing to avoid confounding the evaluation. We deliberately do not filter small masks or discard short-lived instance tracks, as this would introduce additional inference-time hyperparameters.

\begin{figure*}[ht!]
    \centering
    \newcommand{\imgwidth}{0.32}
\newcommand{\dddviswidth}{0.182}

\tiny
\renewcommand{\arraystretch}{1.3}
\sffamily
\begin{tabularx}{0.997\textwidth}{*{19}{>{\centering\arraybackslash}X}}
    \cellcolor{road}\textcolor{white}{Road} &
    \cellcolor{sidewalk}\!\!\textcolor{white}{Sidewalk} &
    \cellcolor{building}\!\textcolor{white}{Building} &
    \cellcolor{wall}\textcolor{white}{Wall} &
    \cellcolor{fence}\textcolor{white}{Fence} &
    \cellcolor{pole}\textcolor{white}{Pole} &
    \cellcolor{trafficlight}\textcolor{white}{T.~Light} &
    \cellcolor{trafficsign}\textcolor{white}{T.~Sign} &
    \cellcolor{vegetation}\textcolor{white}{Veget.} &
    \cellcolor{terrain}\textcolor{white}{Terrain} &
    \cellcolor{sky}\textcolor{white}{Sky} &
    \cellcolor{person}\textcolor{white}{Person} &
    \cellcolor{rider}\textcolor{white}{Rider} &
    \cellcolor{car}\textcolor{white}{Car} &
    \cellcolor{truck}\textcolor{white}{Truck} & 
    \cellcolor{bus}\textcolor{white}{Bus} & 
    \cellcolor{train}\textcolor{white}{Train} &
    \cellcolor{motorcycle}\textcolor{white}{M.cycle} &
    \cellcolor{bicycle}\textcolor{white}{Bicycle}
\end{tabularx}

\vspace{1pt}

\scriptsize
\sffamily
\setlength{\tabcolsep}{1pt}
\renewcommand{\arraystretch}{1.0}

\begin{tabular}{
    >{\centering\arraybackslash}m{0.02\textwidth}
    >{\centering\arraybackslash}m{\imgwidth\textwidth}
    >{\centering\arraybackslash}m{\imgwidth\textwidth}
    >{\centering\arraybackslash}m{\imgwidth\textwidth}
}

\rotatebox[origin=lB]{90}{\hspace{-0.5em}\tiny{Ground-truth}}
& \includegraphics[width=\linewidth]{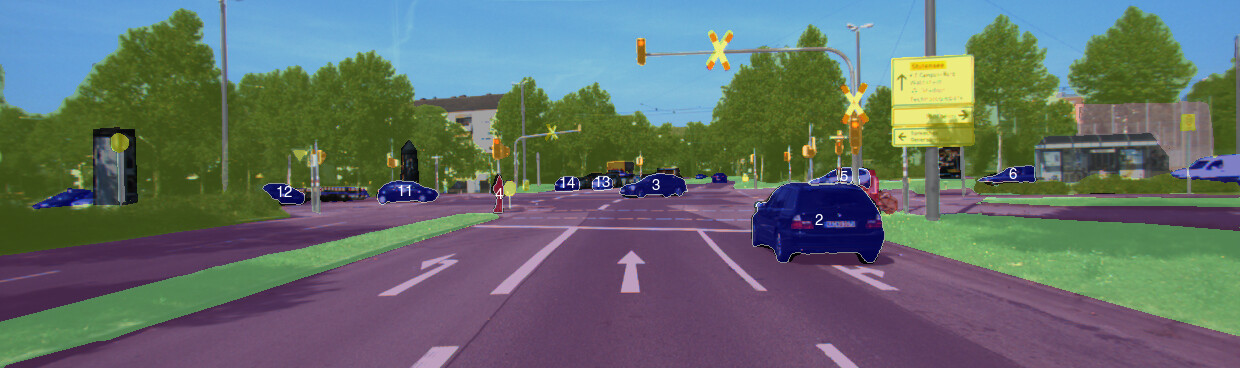} 
& \includegraphics[width=\linewidth]{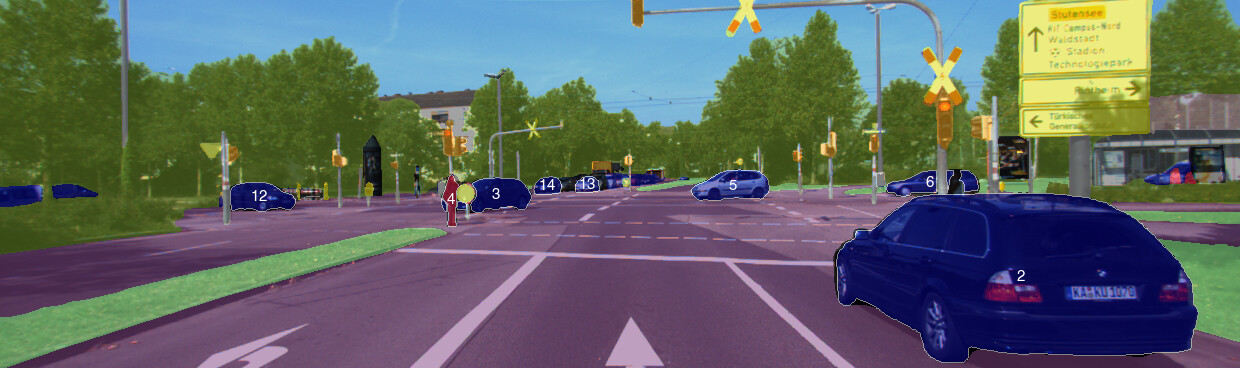} 
& \includegraphics[width=\linewidth]{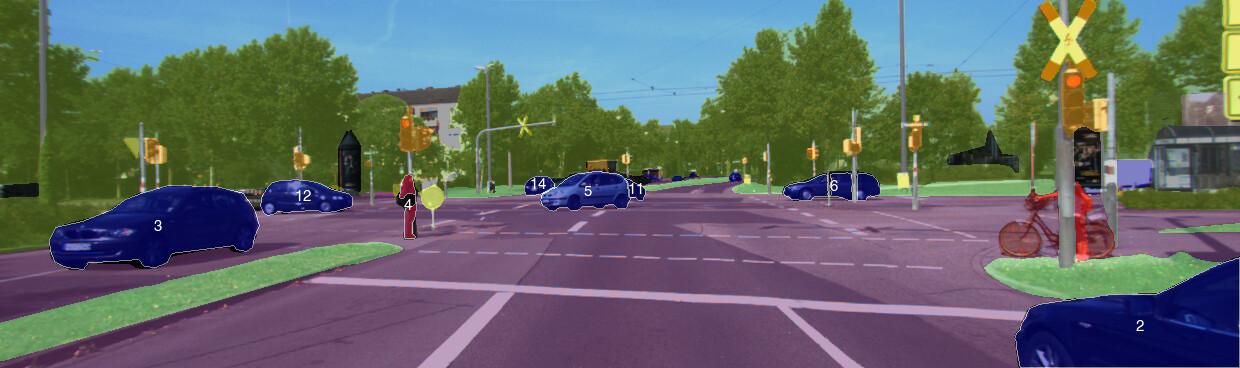} \\

{\rotatebox[origin=lB]{90}{\hspace{-0.3em}\tiny{\shortstack{DepthG\\[-2.5pt]+~VideoCutLER}}}}
& \includegraphics[width=\linewidth]{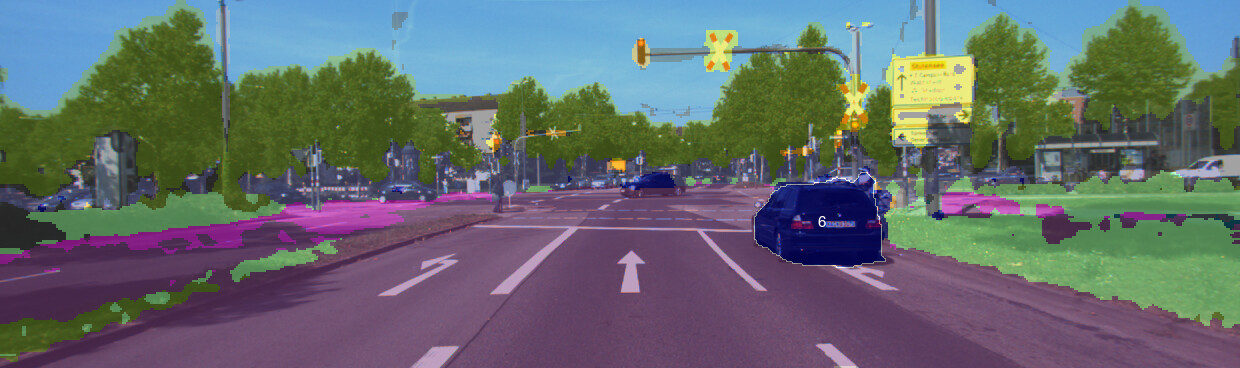} 
& \includegraphics[width=\linewidth]{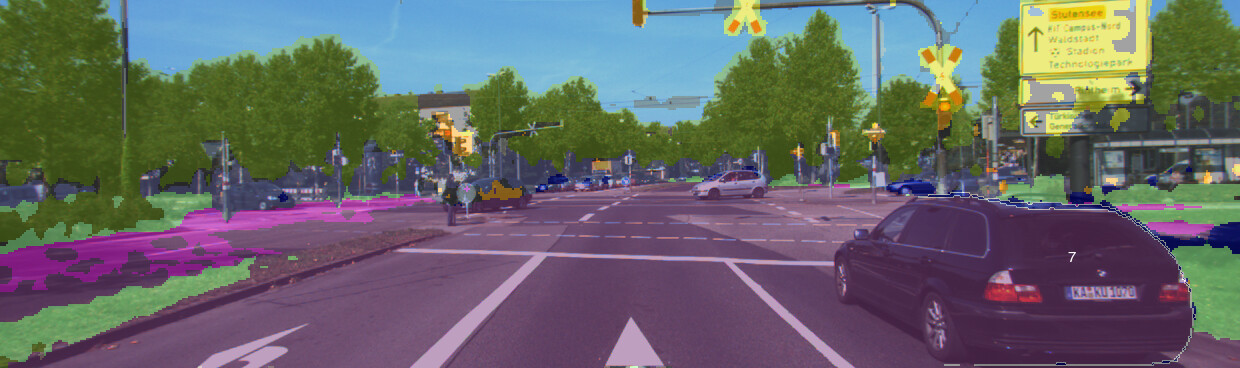} 
& \includegraphics[width=\linewidth]{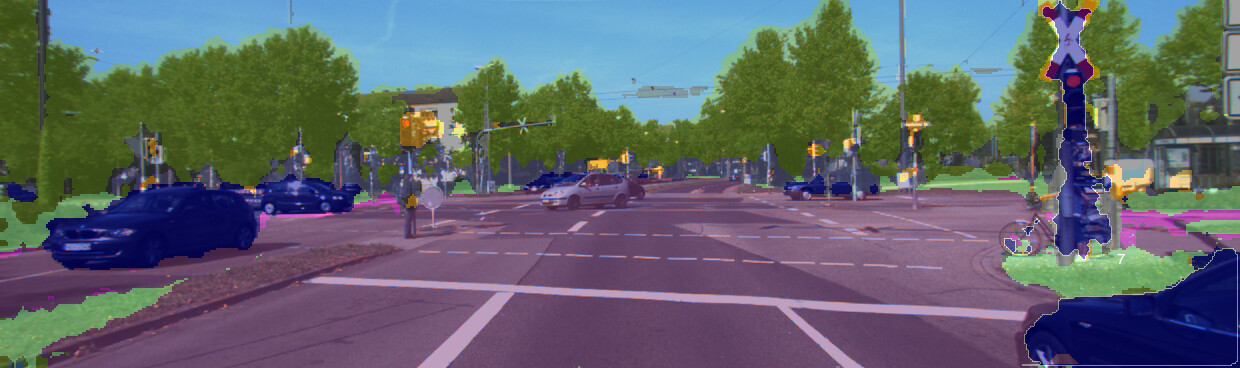} \\

\rotatebox[origin=lB]{90}{\hspace{-0.1em}\tiny{U2Seg~+~SORT}}
& \includegraphics[width=\linewidth]{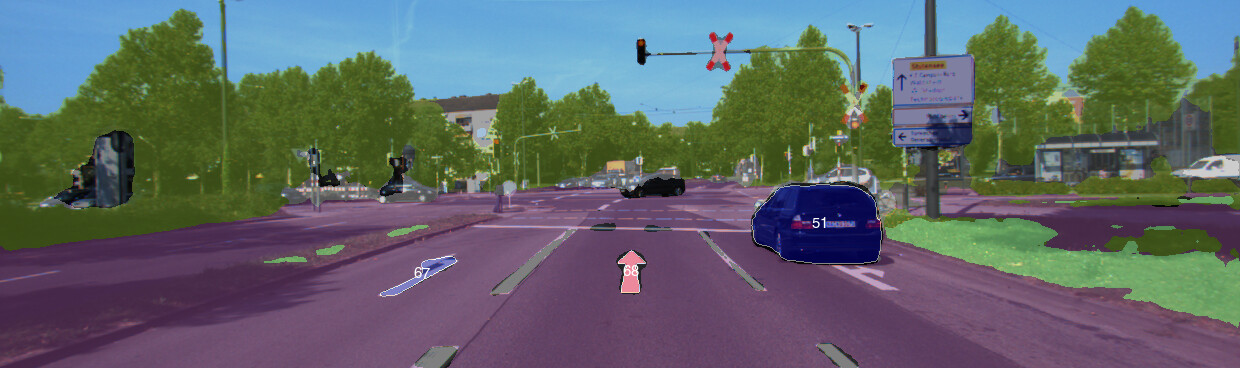} 
& \includegraphics[width=\linewidth]{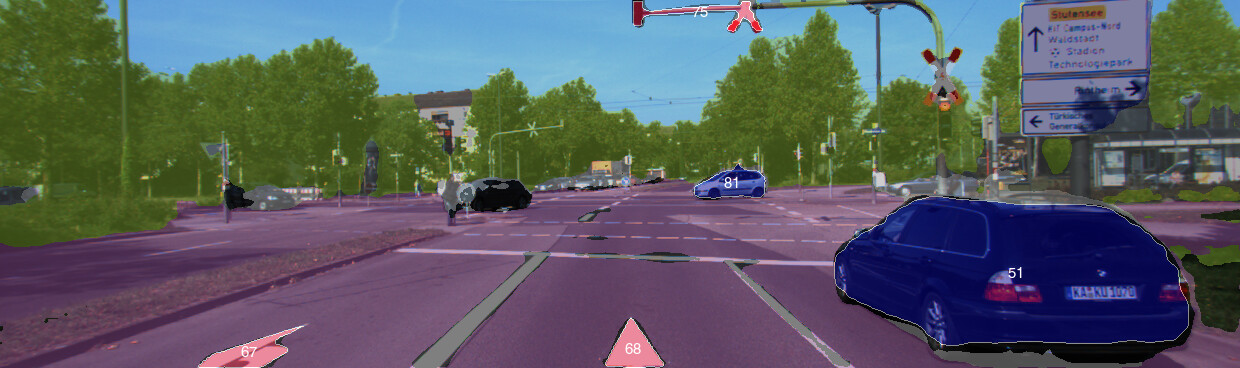} 
& \includegraphics[width=\linewidth]{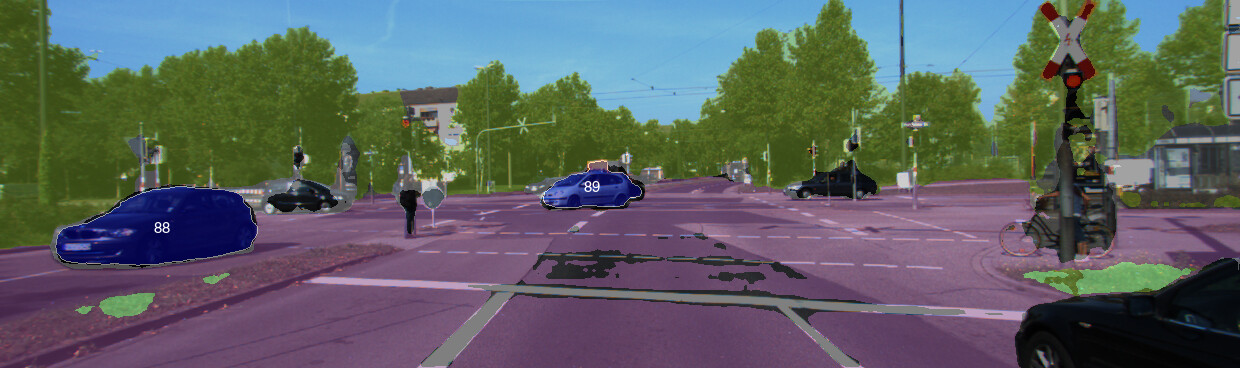}\\

\rotatebox[origin=lB]{90}{\hspace{-0.1em}\tiny{CUPS\phantom{j}+~SORT}}
& \includegraphics[width=\linewidth]{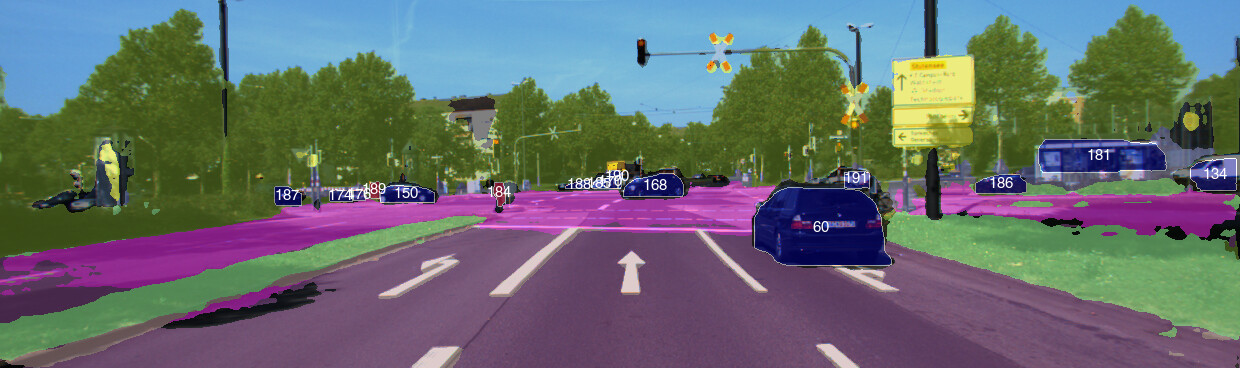} 
& \includegraphics[width=\linewidth]{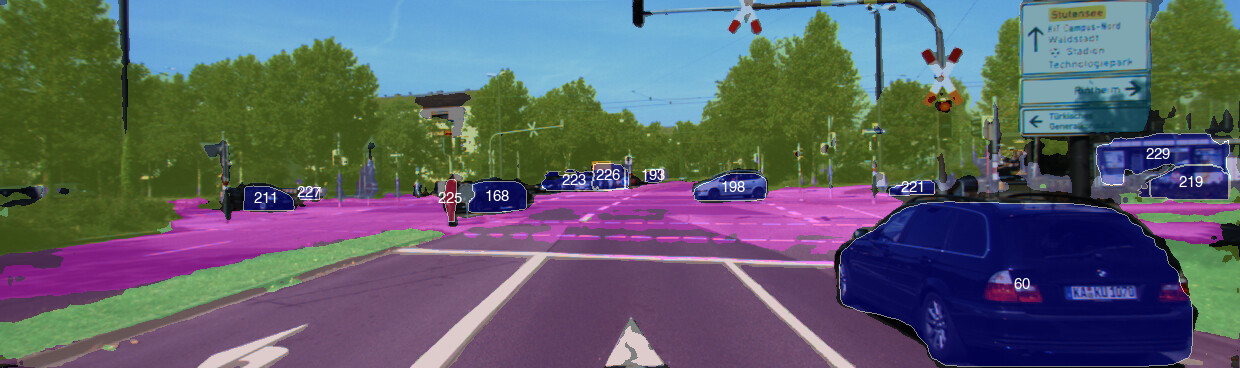} 
& \includegraphics[width=\linewidth]{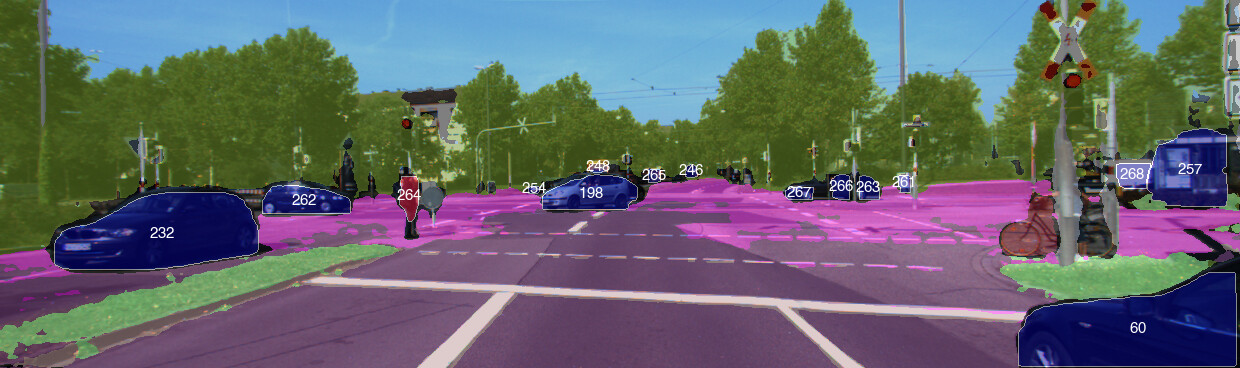}\\

\rotatebox[origin=lB]{90}{\hspace{-0.01em}\tiny{\shortstack{CUPS\textsuperscript{$\dagger$}\!\phantom{j}+~SORT\hphantom{g}}}}
& \includegraphics[width=\linewidth]{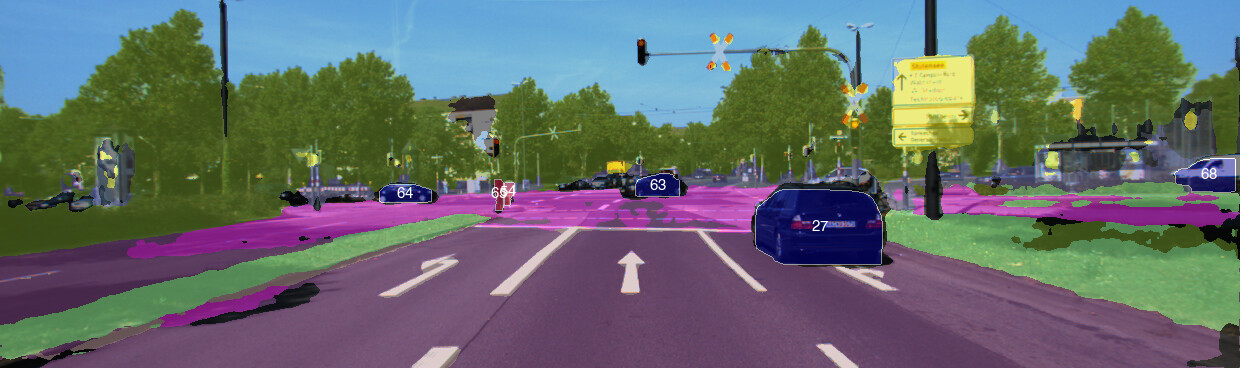} 
& \includegraphics[width=\linewidth]{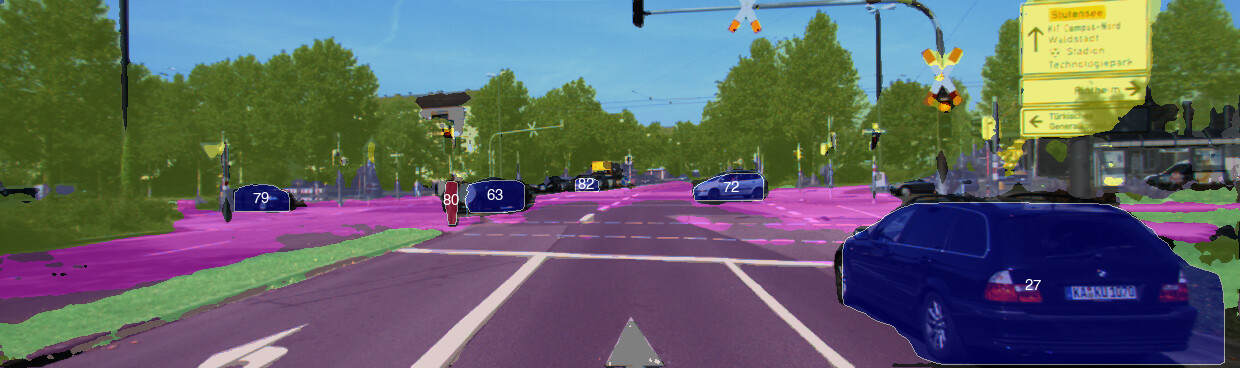} 
& \includegraphics[width=\linewidth]{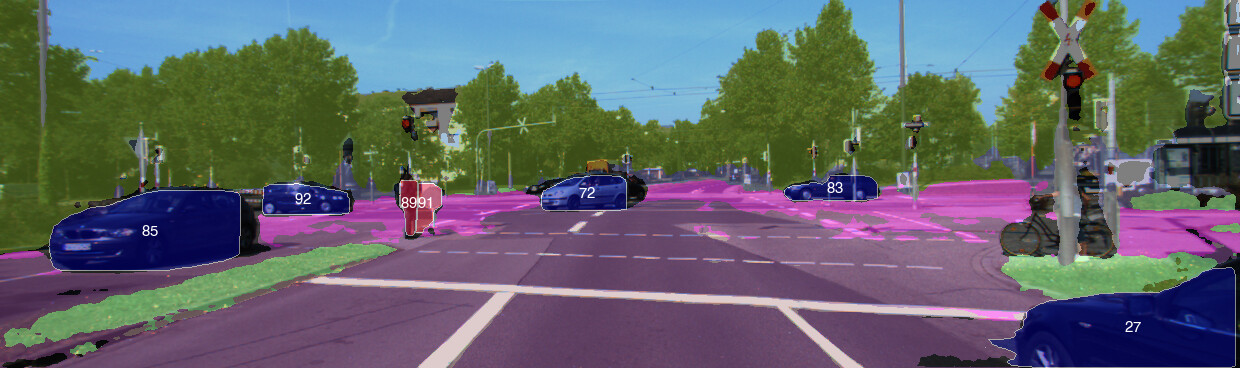}\\

\rotatebox[origin=lB]{90}{\hspace{-0.1em}\tiny{\MethodName}~\textit{(Ours)}}
& \includegraphics[width=\linewidth]{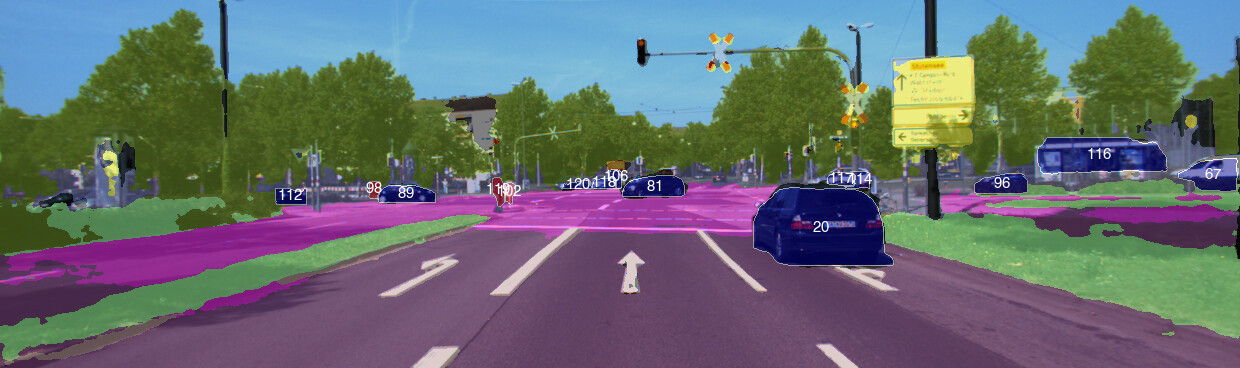} 
& \includegraphics[width=\linewidth]{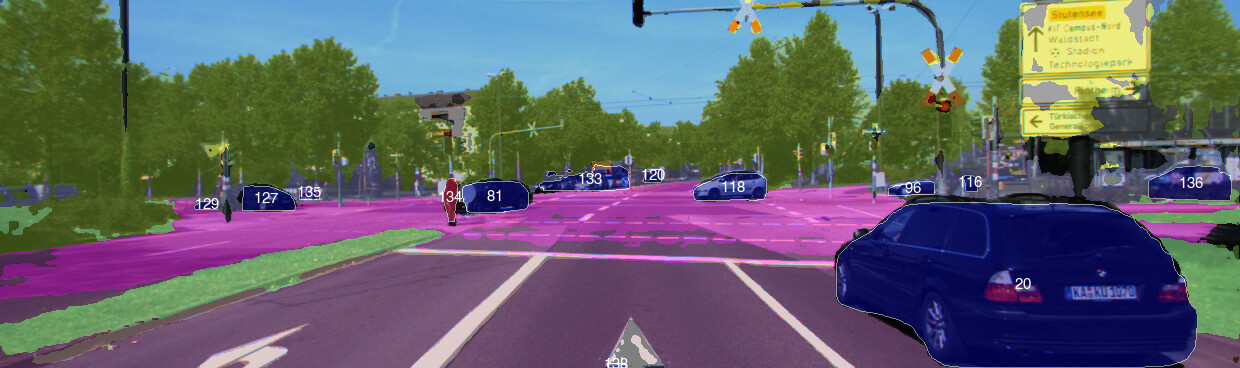} 
& \includegraphics[width=\linewidth]{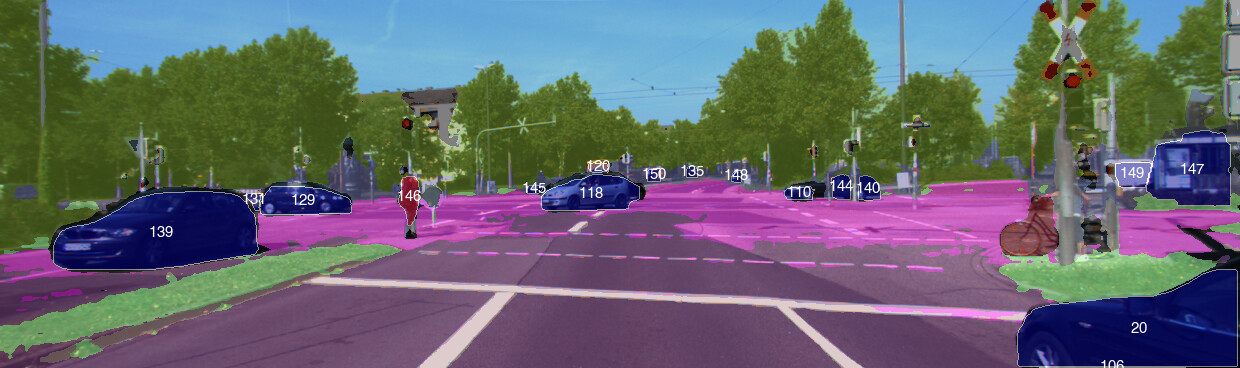}\\

\rotatebox[origin=lB]{90}{\hspace{-0.5em}\tiny{Ground-truth}}
& \includegraphics[width=\linewidth]{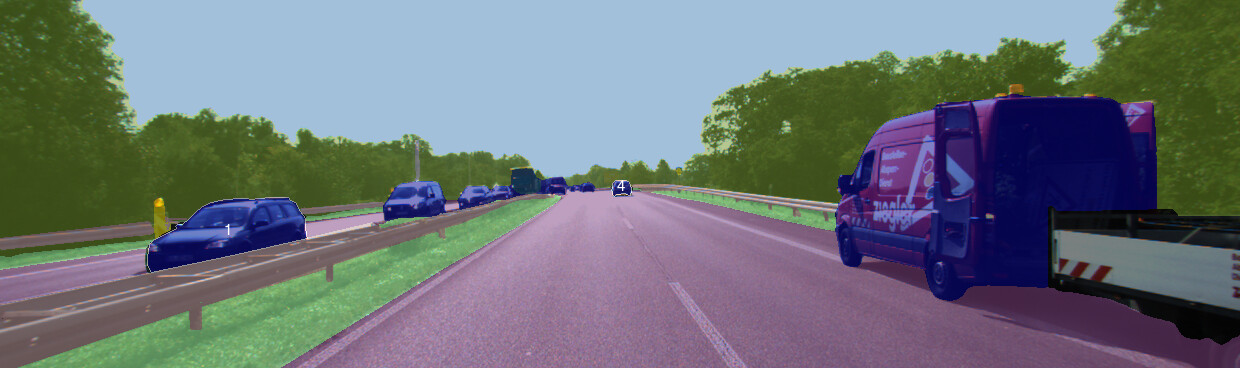} 
& \includegraphics[width=\linewidth]{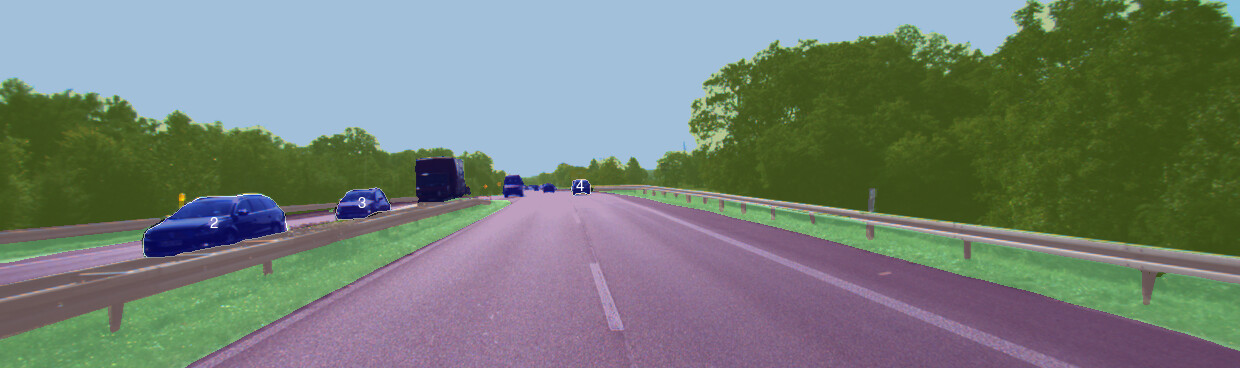}  
& \includegraphics[width=\linewidth]{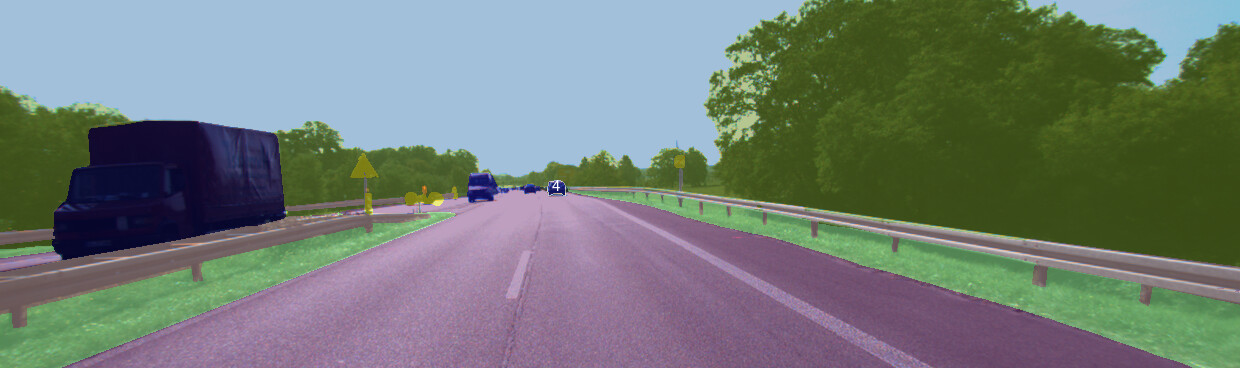}\\

{\rotatebox[origin=lB]{90}{\hspace{-0.3em}\tiny{\shortstack{DepthG\\[-2.5pt]+~VideoCutLER}}}}
& \includegraphics[width=\linewidth]{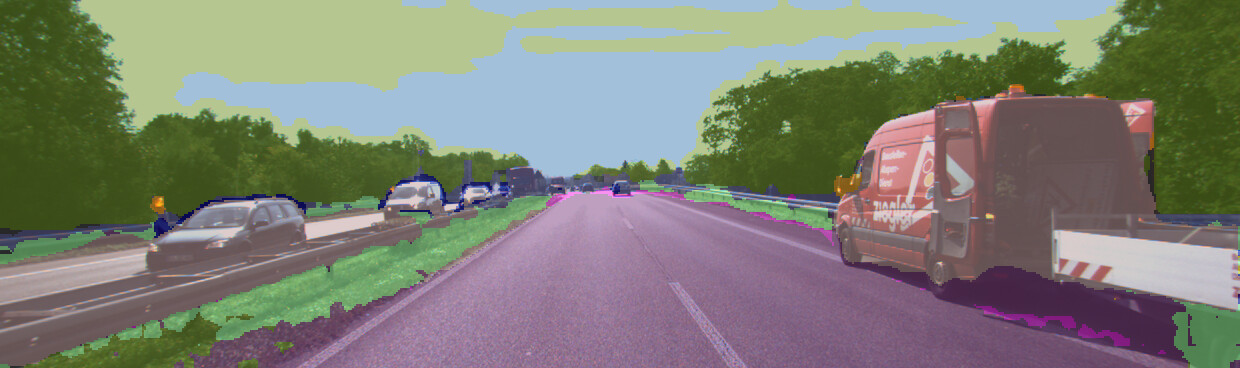} 
& \includegraphics[width=\linewidth]{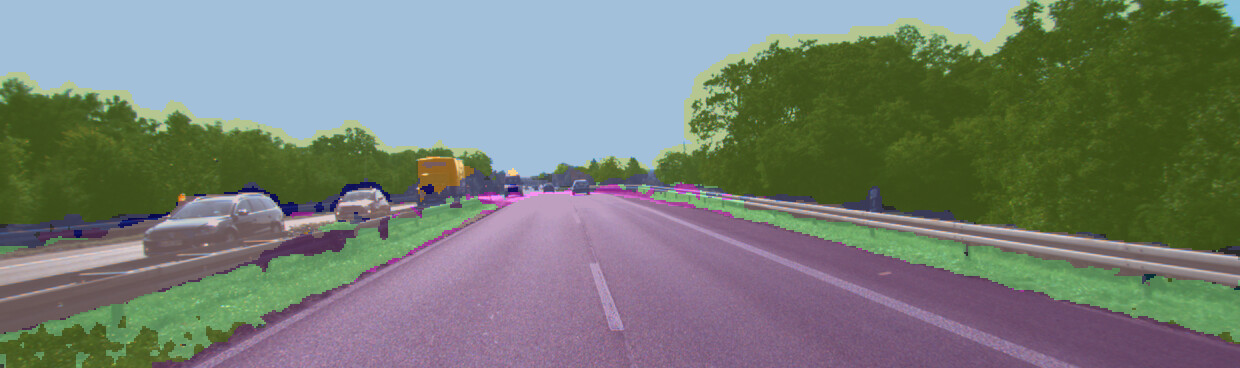}  
& \includegraphics[width=\linewidth]{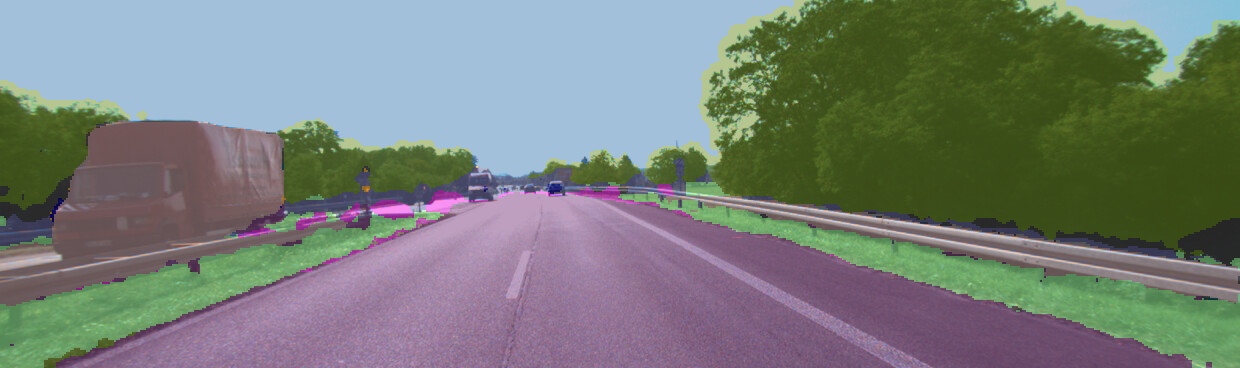}\\

\rotatebox[origin=lB]{90}{\hspace{-0.1em}\tiny{U2Seg~+~SORT}}
& \includegraphics[width=\linewidth]{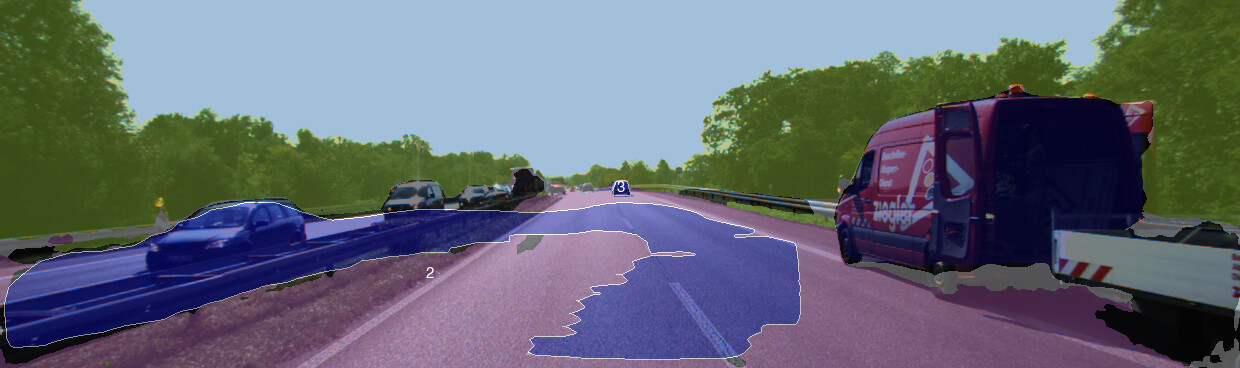} 
& \includegraphics[width=\linewidth]{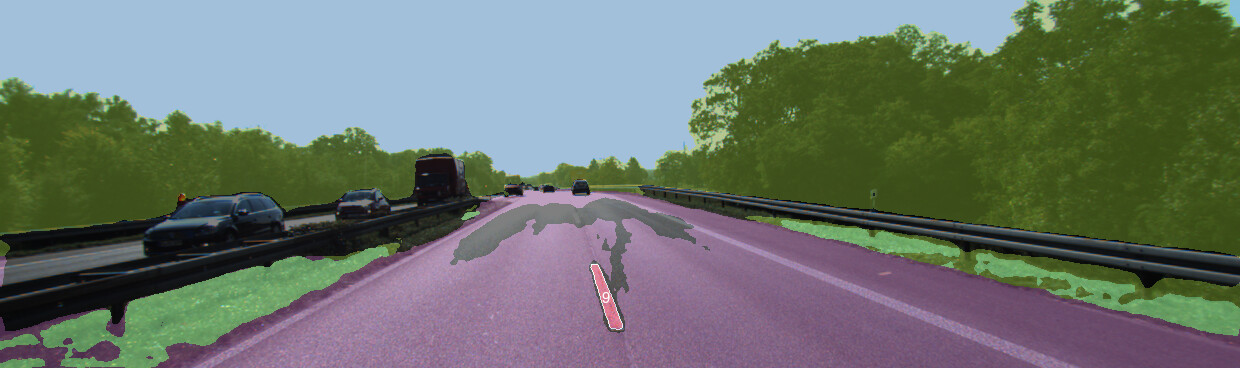}  
& \includegraphics[width=\linewidth]{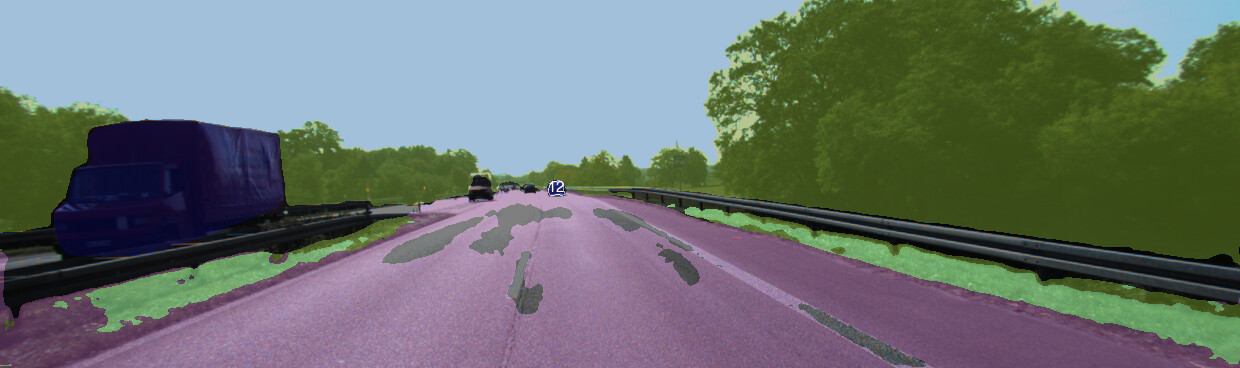}\\

\rotatebox[origin=lB]{90}{\hspace{-0.1em}\tiny{CUPS\phantom{j}+~SORT}}
& \includegraphics[width=\linewidth]{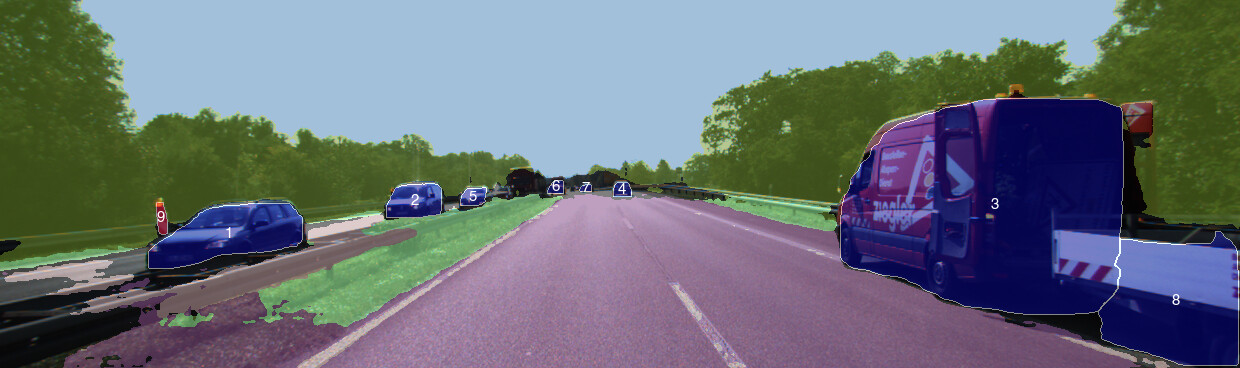} 
& \includegraphics[width=\linewidth]{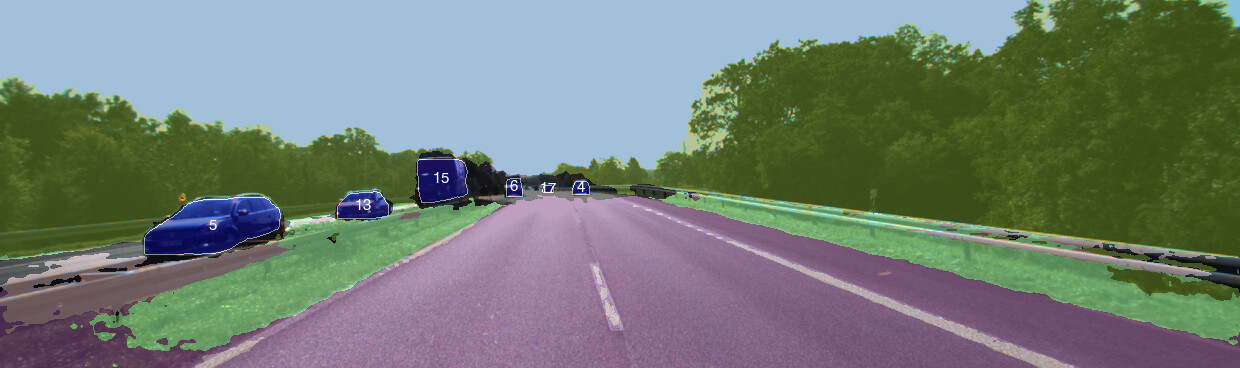}  
& \includegraphics[width=\linewidth]{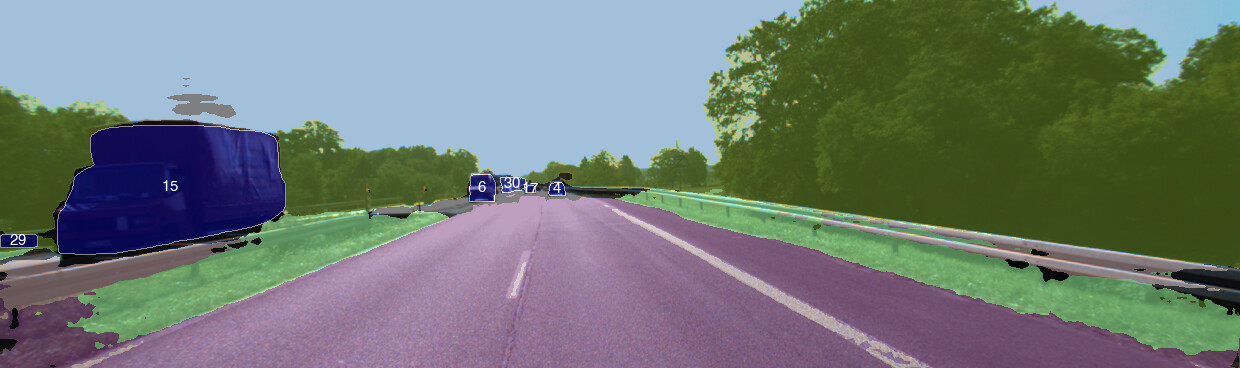}\\

\rotatebox[origin=lB]{90}{\hspace{-0.01em}\tiny{\shortstack{CUPS\textsuperscript{$\dagger$}\!\phantom{j}+~SORT\hphantom{g}}}}
& \includegraphics[width=\linewidth]{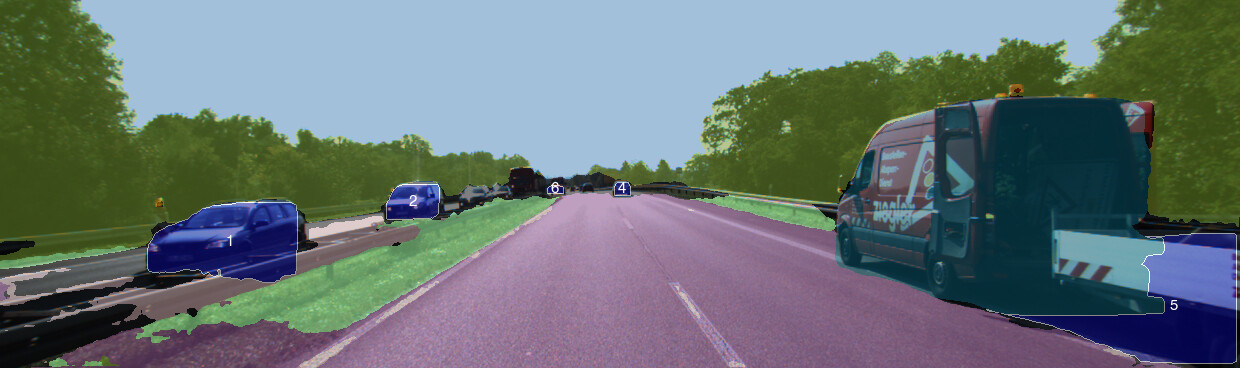} 
& \includegraphics[width=\linewidth]{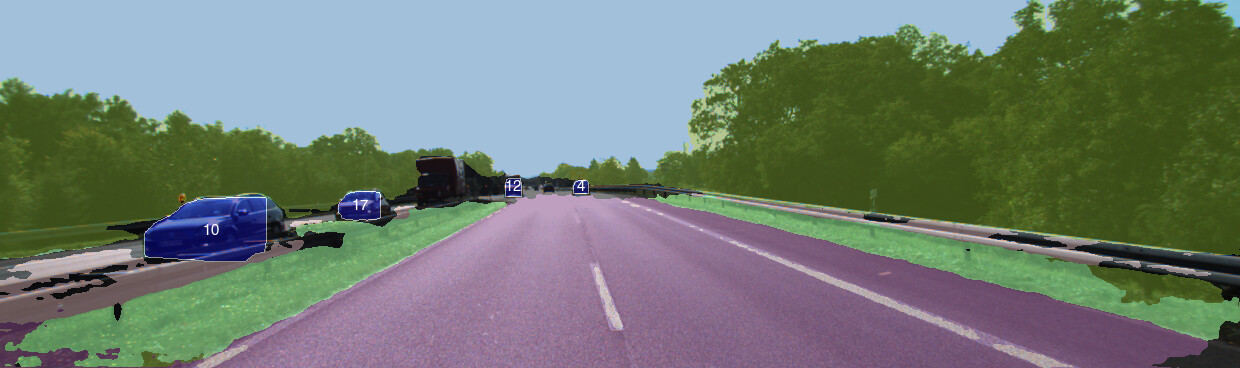}  
& \includegraphics[width=\linewidth]{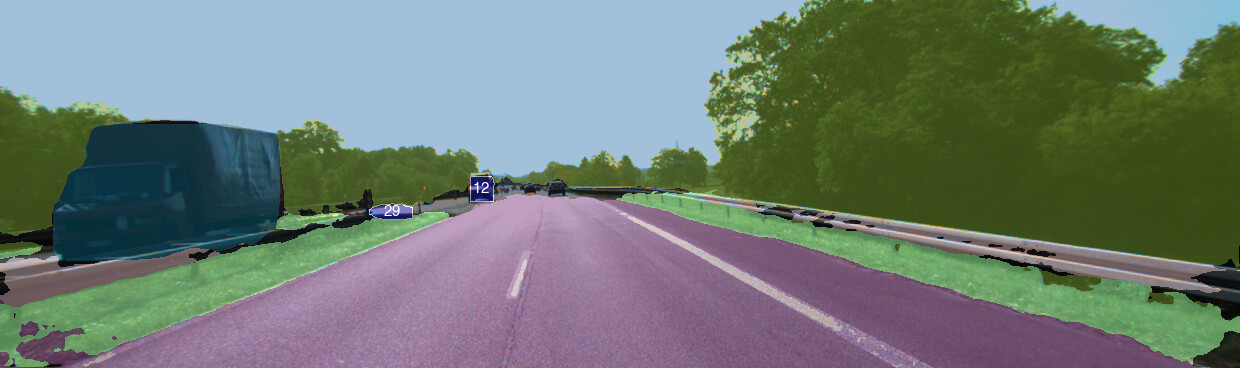}\\

\rotatebox[origin=lB]{90}{\hspace{-0.1em}\tiny{\MethodName}~\textit{(Ours)}}
& \includegraphics[width=\linewidth]{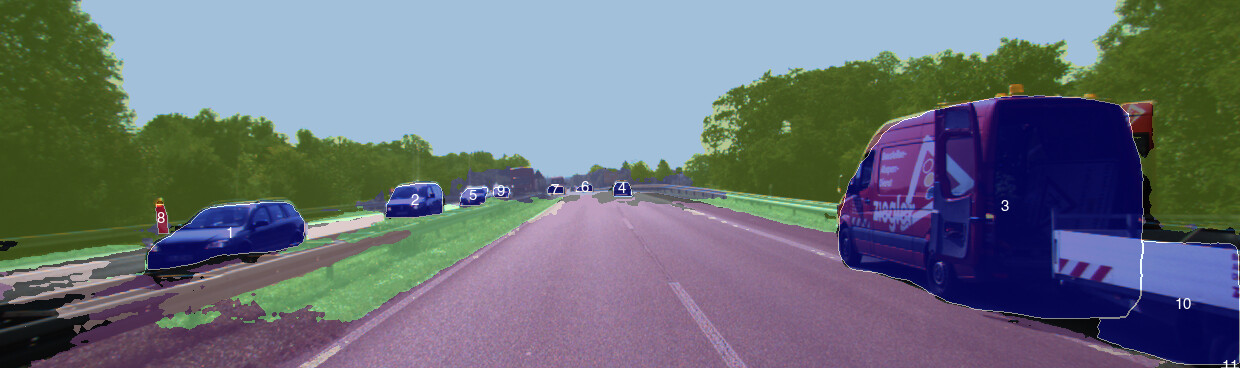} 
& \includegraphics[width=\linewidth]{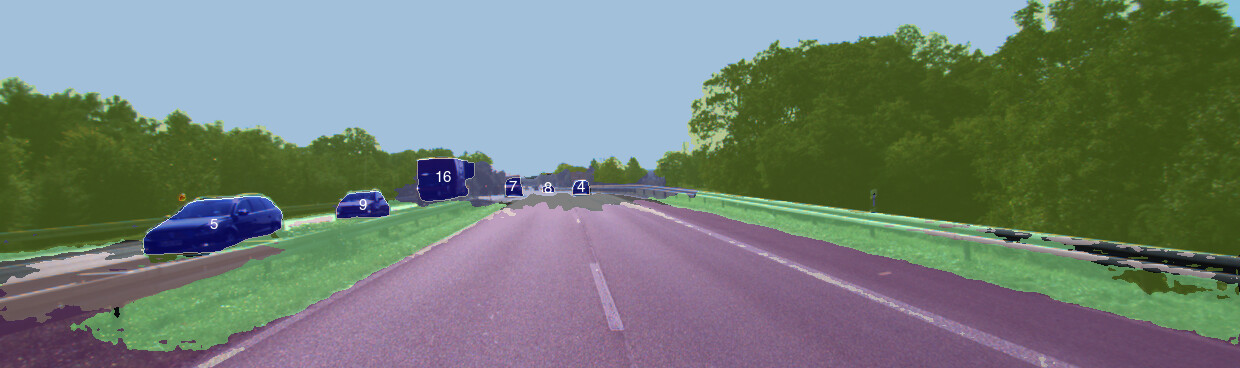}  
& \includegraphics[width=\linewidth]{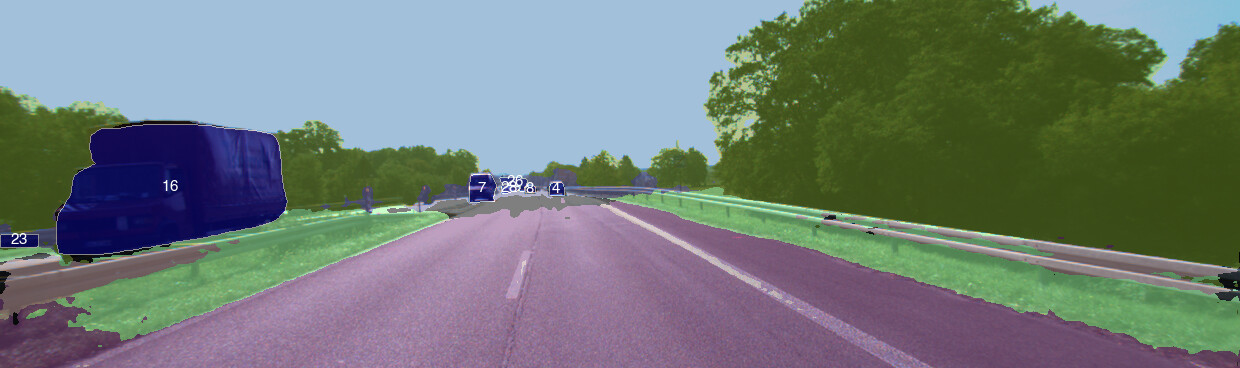}\\[-2.5pt]

&\tiny{$\phantom{.}t_0$}
&\tiny{$\phantom{.}t_1$} 
&\tiny{$\phantom{.}t_2$} \\

\end{tabular}

    \vspace{-1.2em}
    \caption{\textbf{KITTI-STEP---Qualitative unsupervised VPS examples.} We compare our proposed method \MethodName to the proposed baselines DepthG~\cite{Sick:2024:USS} +~VideoCutLER~\cite{Wang:2024:VCS}, U2Seg~\cite{Niu:2024:UUI}~+~SORT, CUPS~\cite{Hahn:2025:UPS}~+~SORT~\cite{Bewley:2016:SOR}, and CUPS\textsuperscript{\textdagger}~\cite{Hahn:2025:UPS}~+~SORT~\cite{Bewley:2016:SOR} on KITTI-STEP~\cite{Weber:2021:STE} val.\label{fig:qualitative_kitti}}
    \vspace{-0.5em}
\end{figure*}

\Cref{fig:qualitative_kitti} presents a qualitative comparison on KITTI-STEP. DepthG~+~VideoCutLER detects only a limited set of instances. U2Seg~+~SORT increases the number of predicted instances but frequently produces artifact-like instance predictions (\eg, erroneous arrow on the road; top example). CUPS~+~SORT yields a large number of instances with stable temporal identities, while its monocular variant, CUPS\textsuperscript{\textdagger}~+~SORT, misses several smaller background instances. In contrast, \MethodName consistently discovers both near and far objects, producing accurate masks and temporally robust tracks.

\begin{figure*}[ht!]
    \centering
    \newcommand{\imgwidth}{0.159}
\newcommand{\dddviswidth}{0.182}

\tiny
\renewcommand{\arraystretch}{1.3}
\sffamily
\begin{tabularx}{0.997\textwidth}{*{19}{>{\centering\arraybackslash}X}}
    \cellcolor{road}\textcolor{white}{Road} &
    \cellcolor{sidewalk}\!\!\textcolor{white}{Sidewalk} &
    \cellcolor{building}\!\textcolor{white}{Building} &
    \cellcolor{wall}\textcolor{white}{Wall} &
    \cellcolor{fence}\textcolor{white}{Fence} &
    \cellcolor{pole}\textcolor{white}{Pole} &
    \cellcolor{trafficlight}\textcolor{white}{T.~Light} &
    \cellcolor{trafficsign}\textcolor{white}{T.~Sign} &
    \cellcolor{vegetation}\textcolor{white}{Veget.} &
    \cellcolor{terrain}\textcolor{white}{Terrain} &
    \cellcolor{sky}\textcolor{white}{Sky} &
    \cellcolor{person}\textcolor{white}{Person} &
    \cellcolor{rider}\textcolor{white}{Rider} &
    \cellcolor{car}\textcolor{white}{Car} &
    \cellcolor{truck}\textcolor{white}{Truck} & 
    \cellcolor{bus}\textcolor{white}{Bus} & 
    \cellcolor{train}\textcolor{white}{Train} &
    \cellcolor{motorcycle}\textcolor{white}{M.cycle} &
    \cellcolor{bicycle}\textcolor{white}{Bicycle}
\end{tabularx}

\vspace{1pt}

\scriptsize
\sffamily
\setlength{\tabcolsep}{1pt}
\renewcommand{\arraystretch}{1.0}

\begin{tabular}{
    >{\centering\arraybackslash}m{0.017\textwidth}
    >{\centering\arraybackslash}m{\imgwidth\textwidth}
    >{\centering\arraybackslash}m{\imgwidth\textwidth}
    >{\centering\arraybackslash}m{\imgwidth\textwidth}
    >{\centering\arraybackslash}m{\imgwidth\textwidth}
    >{\centering\arraybackslash}m{\imgwidth\textwidth}
    >{\centering\arraybackslash}m{\imgwidth\textwidth}
}

\rotatebox[origin=lB]{90}{\hspace{-0.5em}\tiny{Ground-\phantom{j}truth}}
& \includegraphics[width=\linewidth]{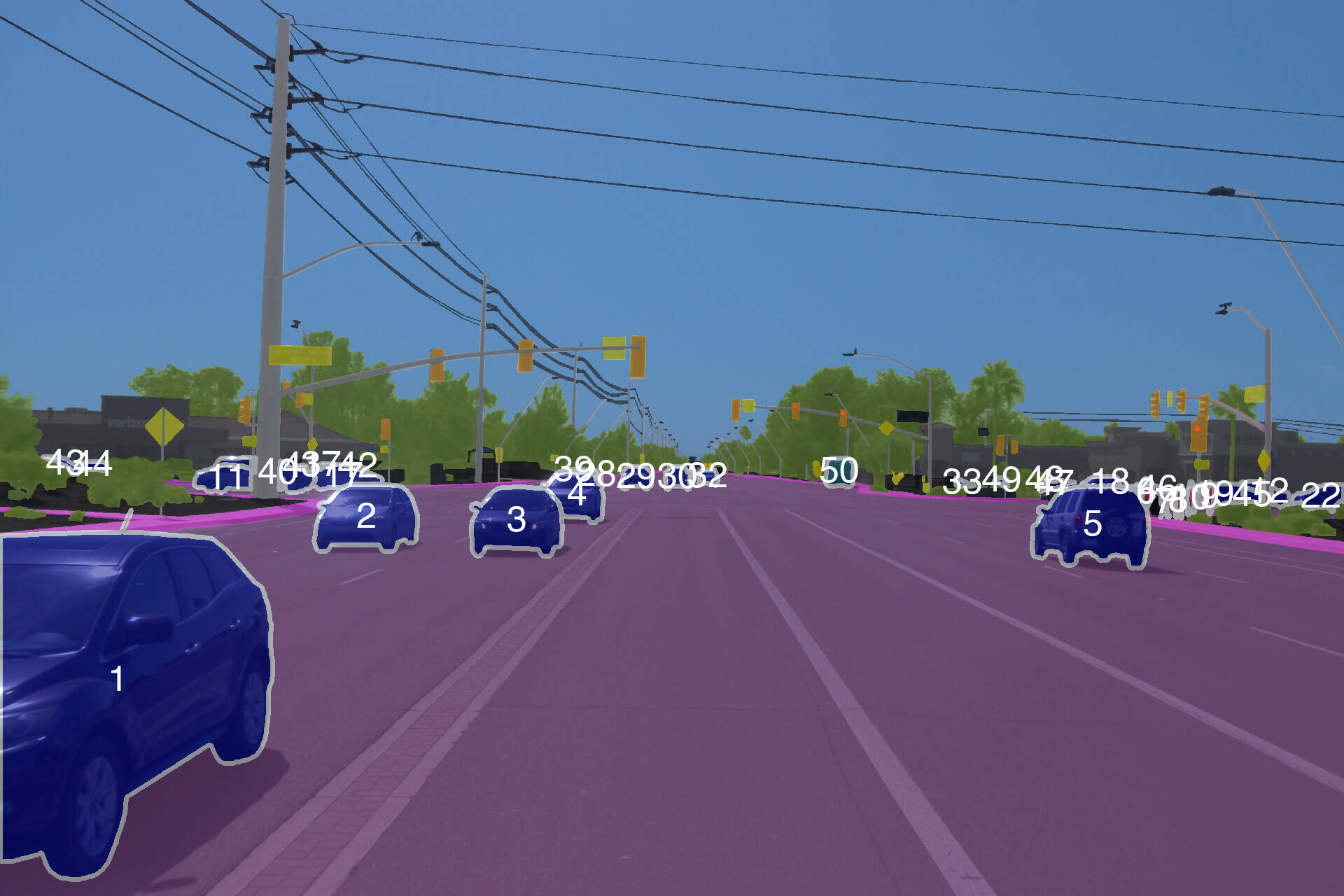} 
& \includegraphics[width=\linewidth]{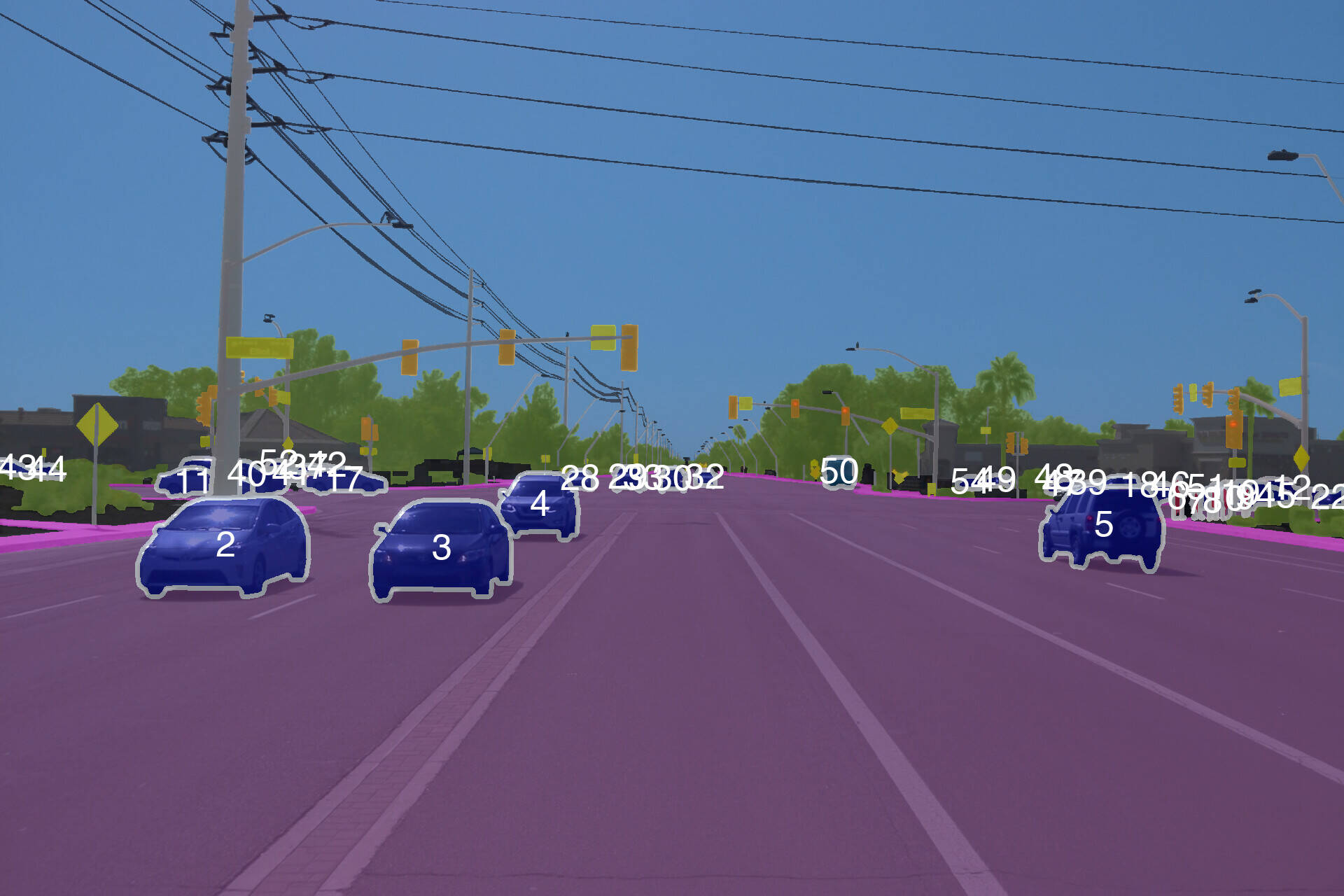}
& \includegraphics[width=\linewidth]{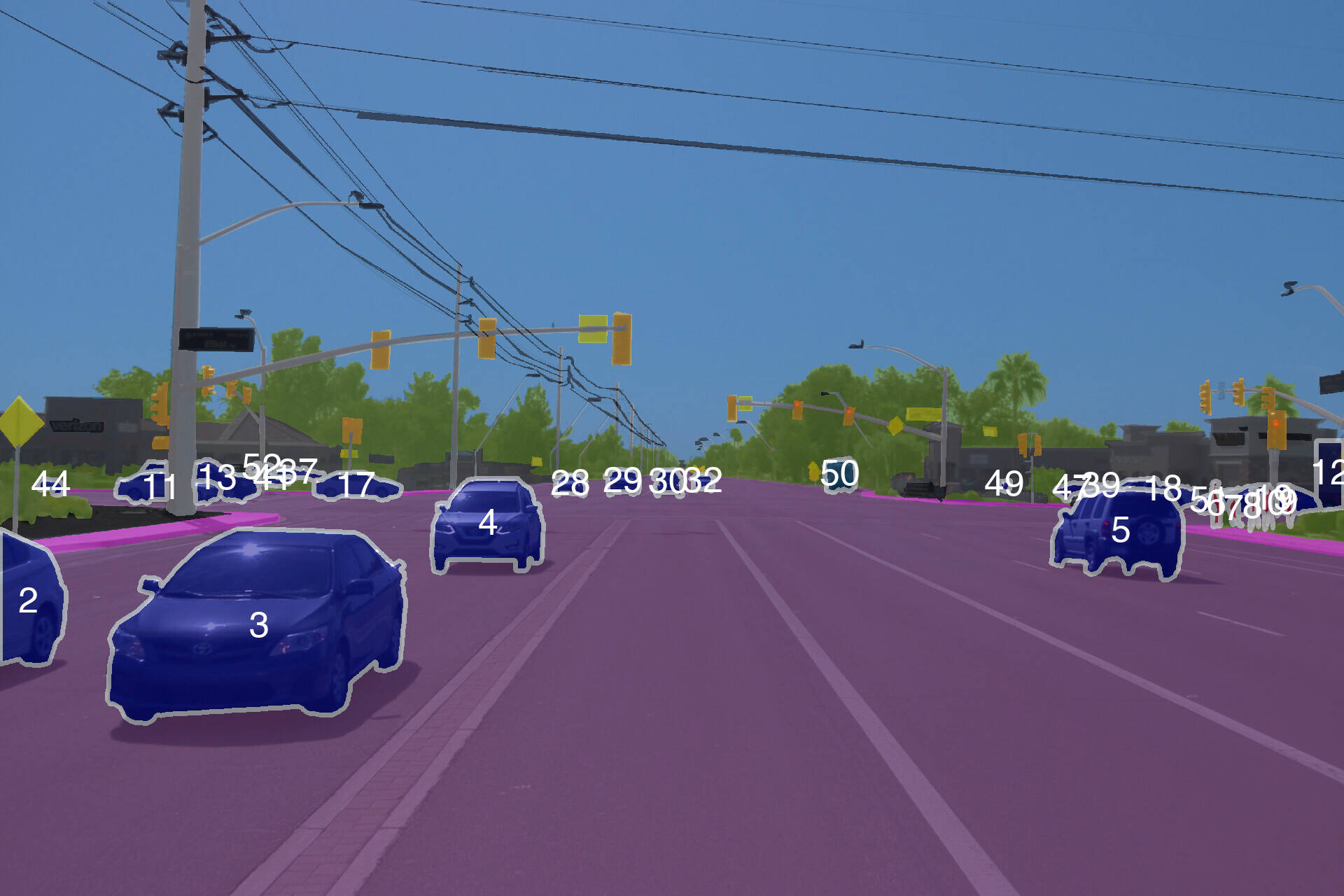}
& \includegraphics[width=\linewidth]{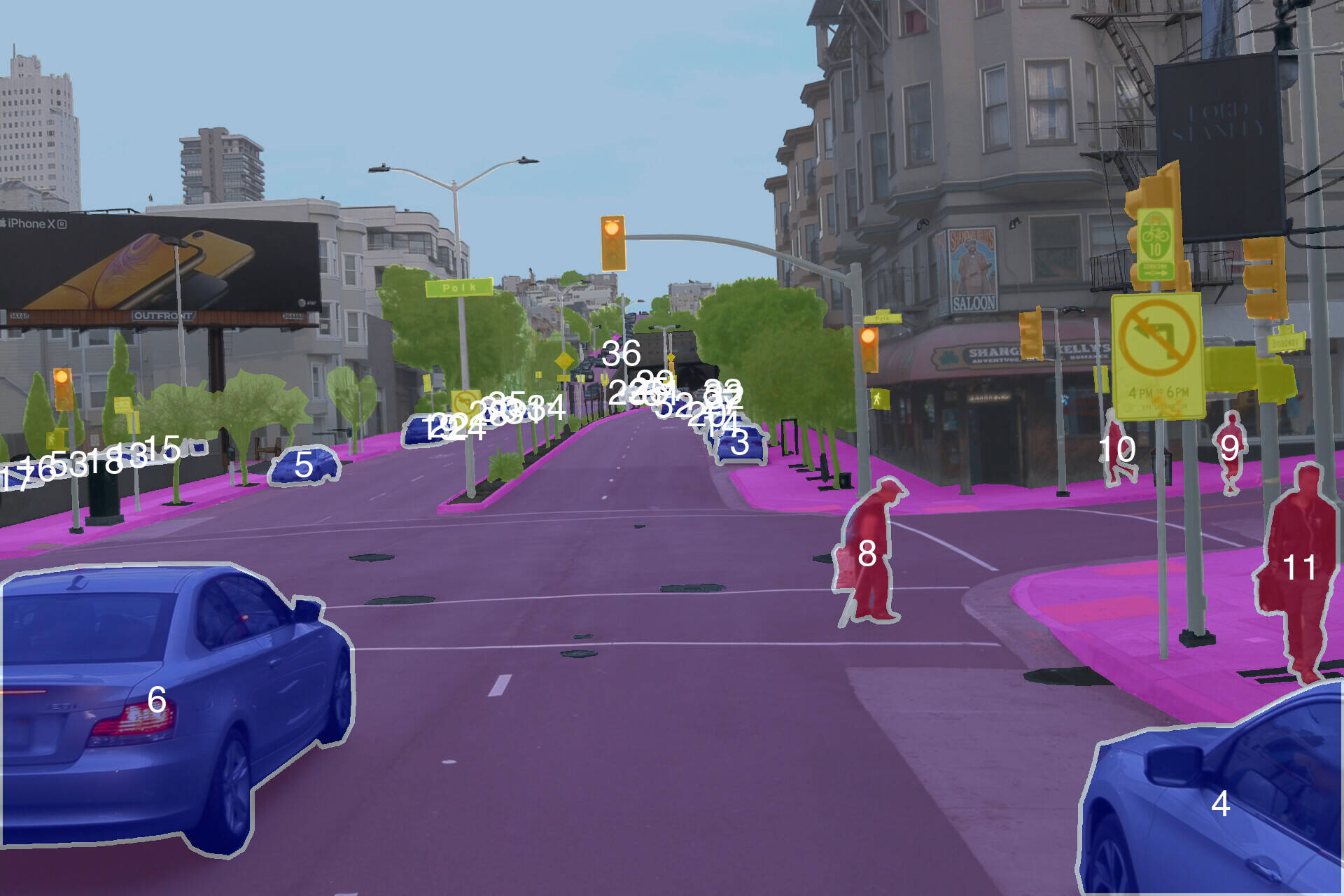}
& \includegraphics[width=\linewidth]{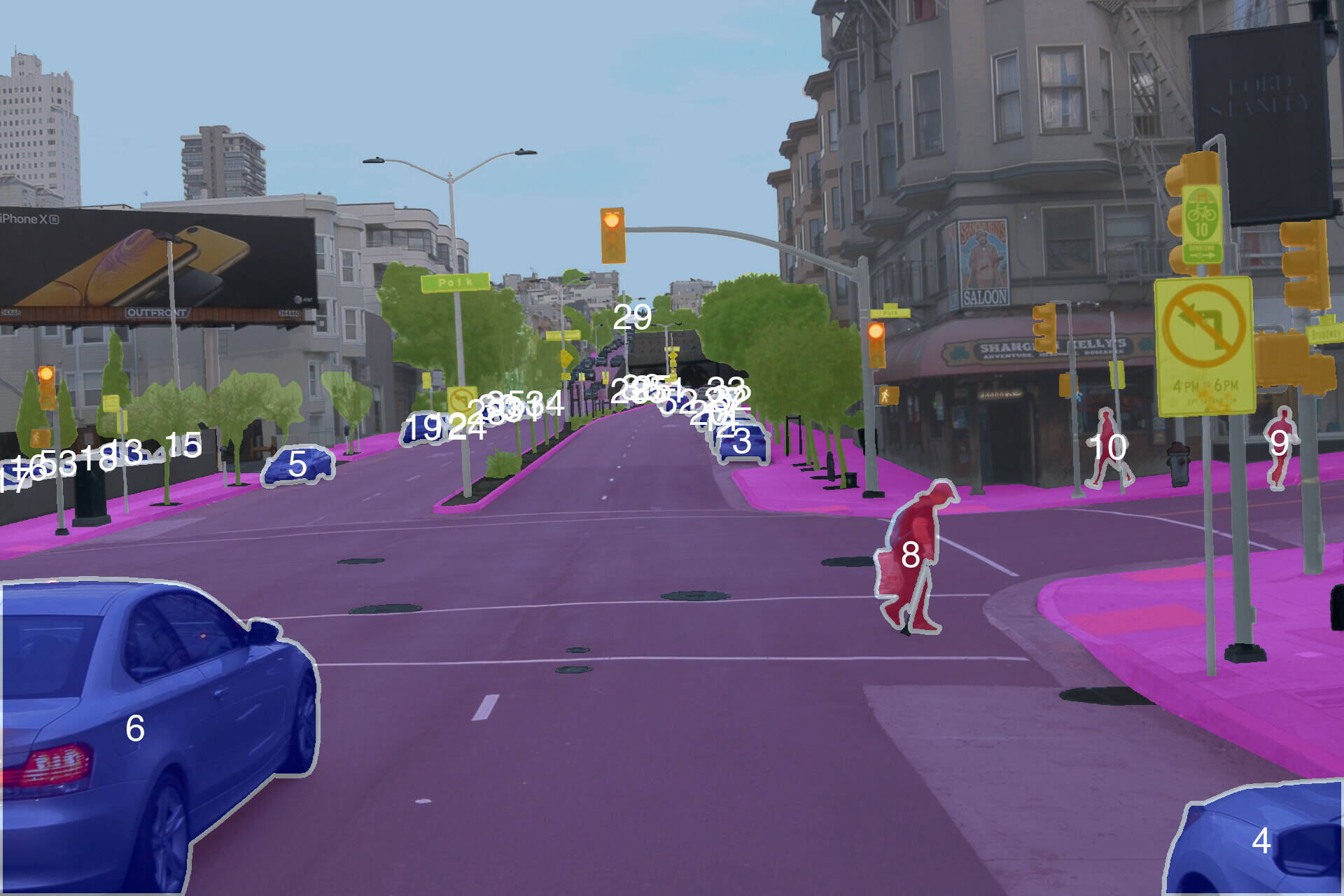}
& \includegraphics[width=\linewidth]{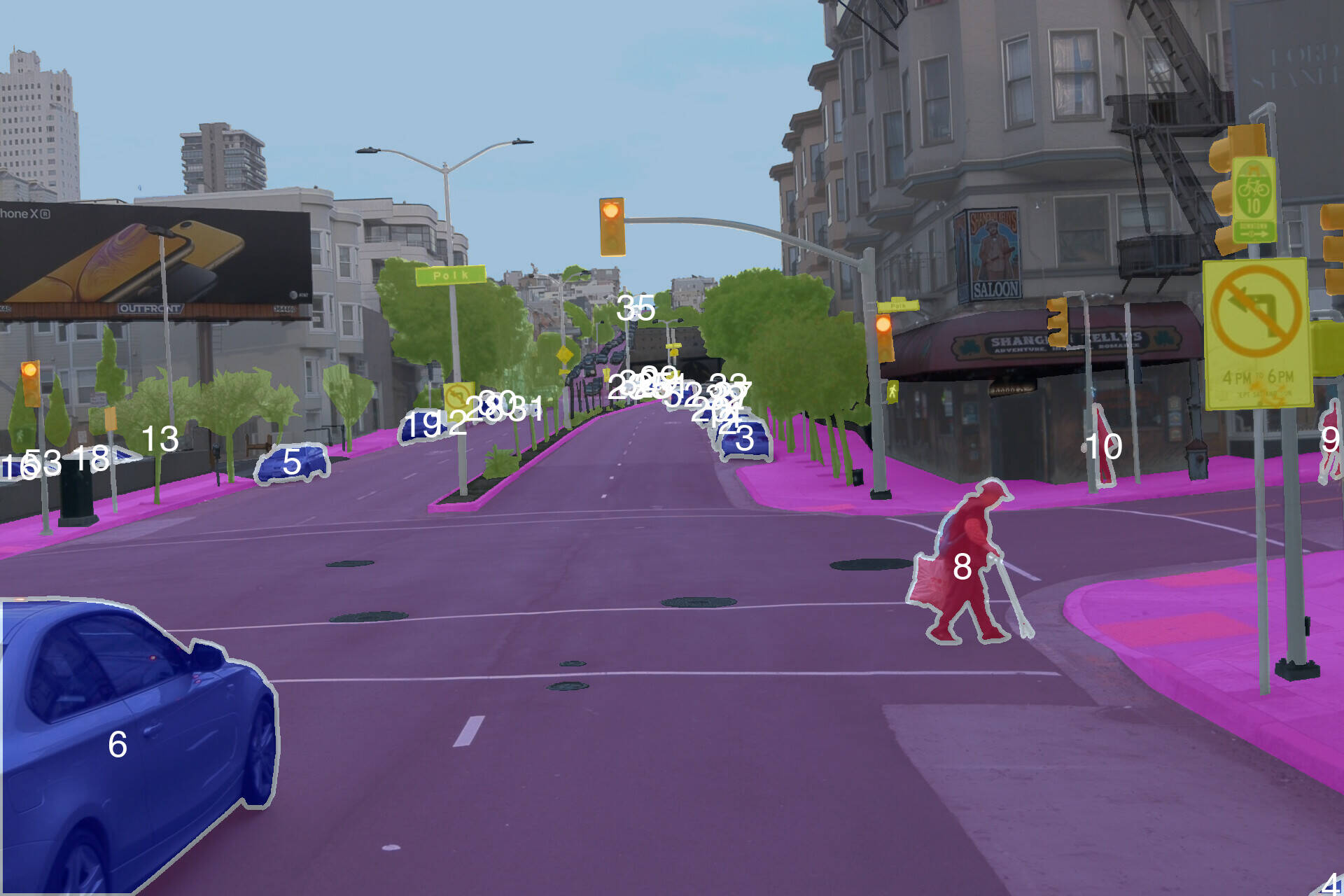} \\

{\rotatebox[origin=lB]{90}{\hspace{-0.3em}\tiny{\shortstack{DepthG~+\\[-3pt]VideoCutLER}}}}
& \includegraphics[width=\linewidth]{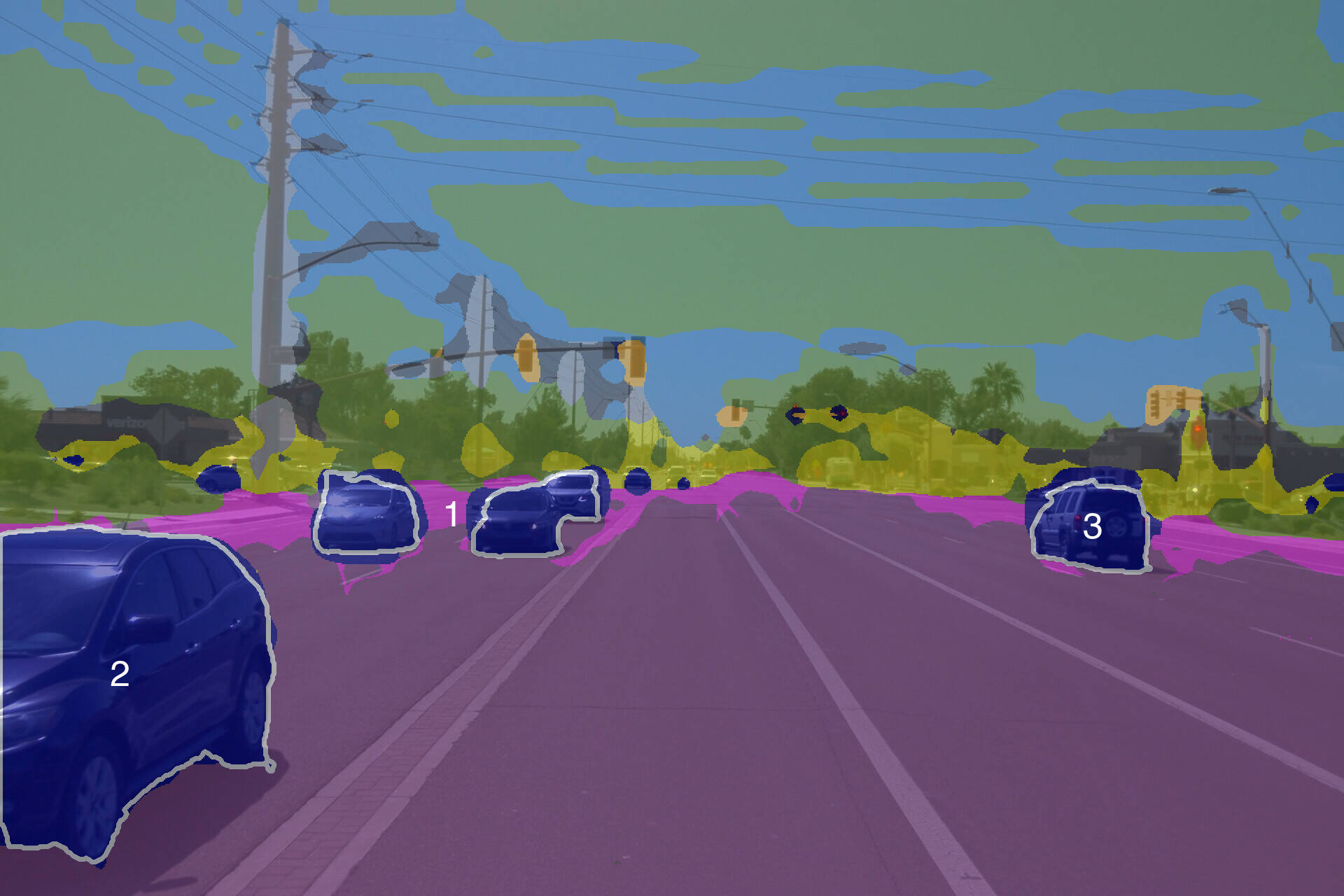} 
& \includegraphics[width=\linewidth]{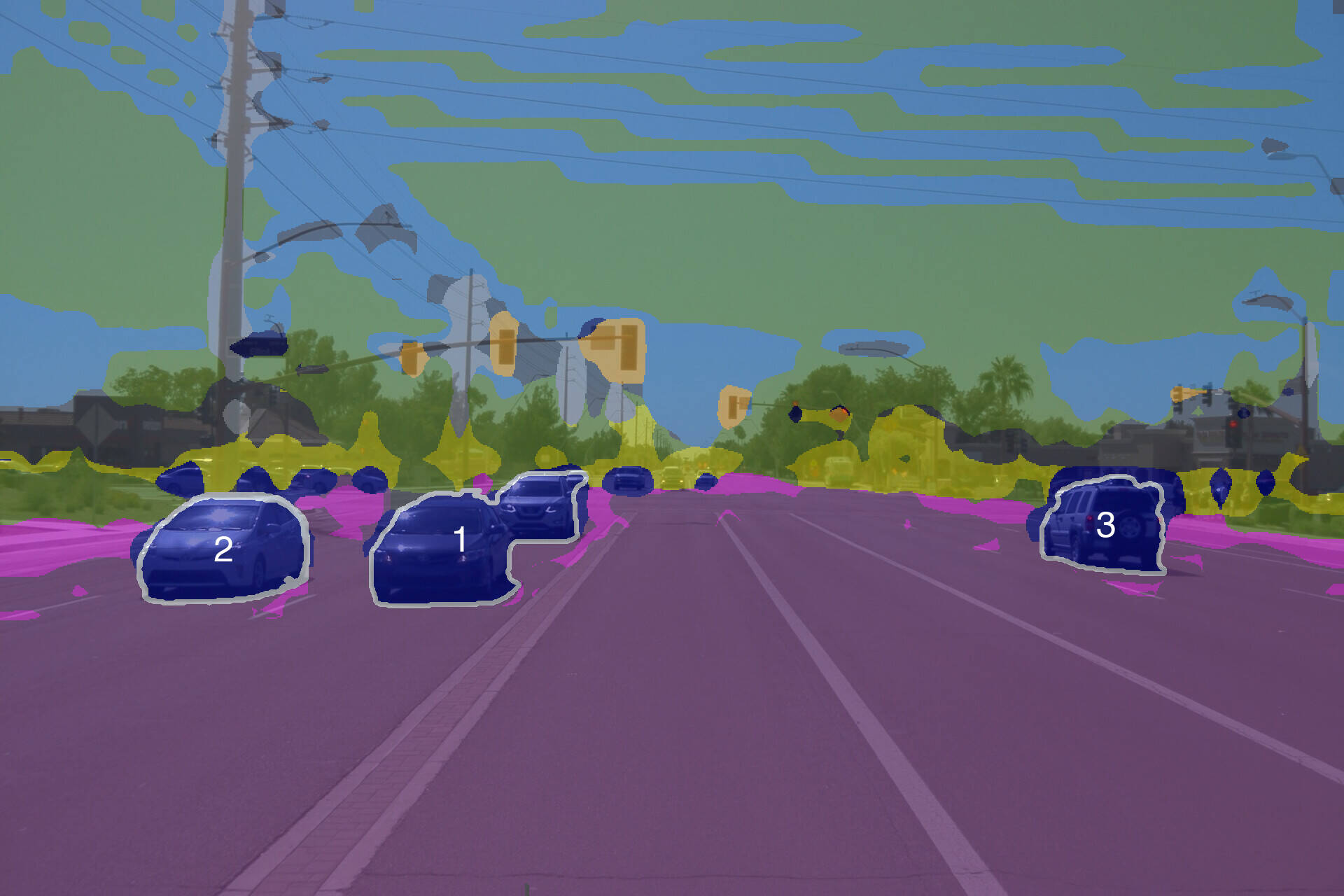}
& \includegraphics[width=\linewidth]{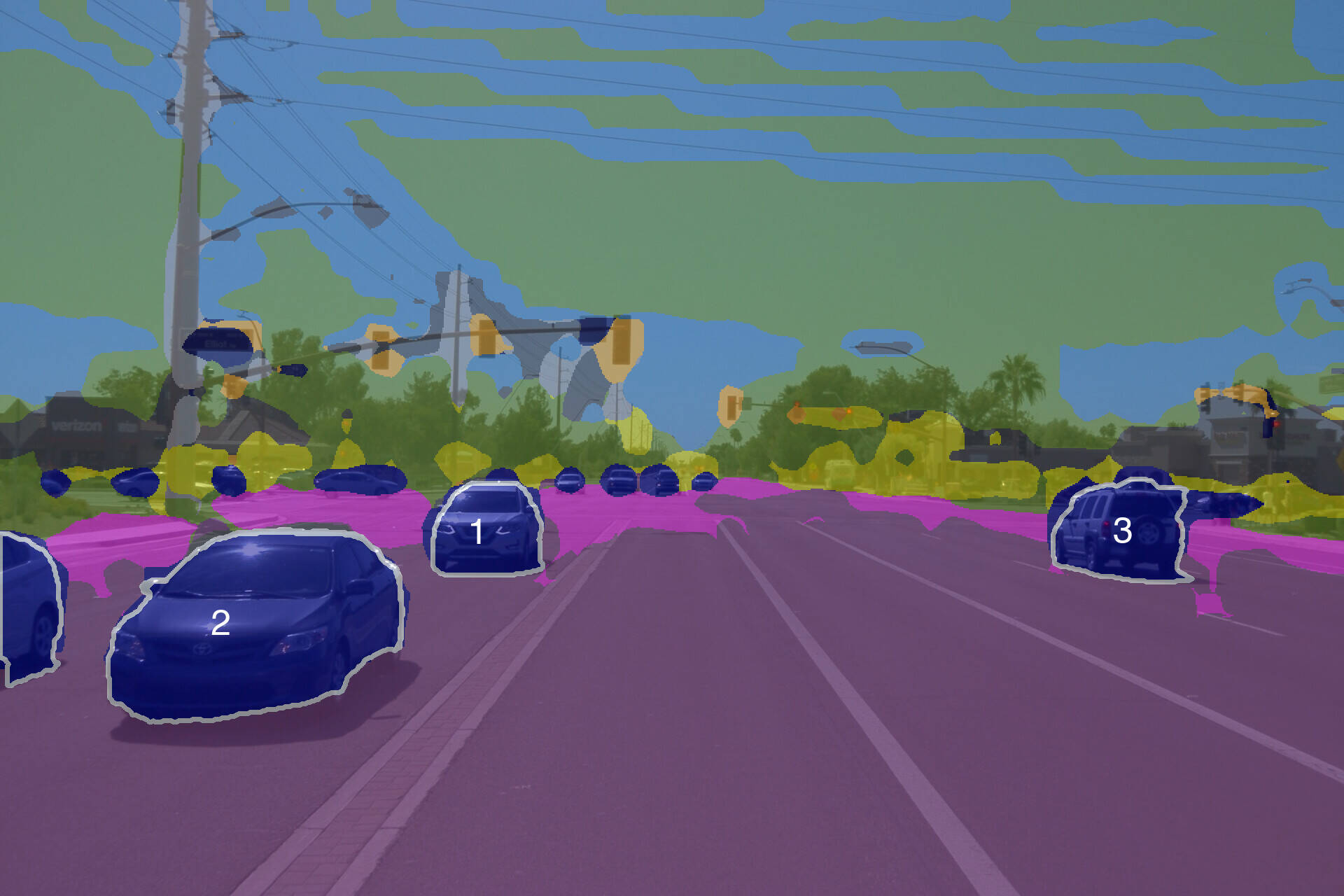}
& \includegraphics[width=\linewidth]{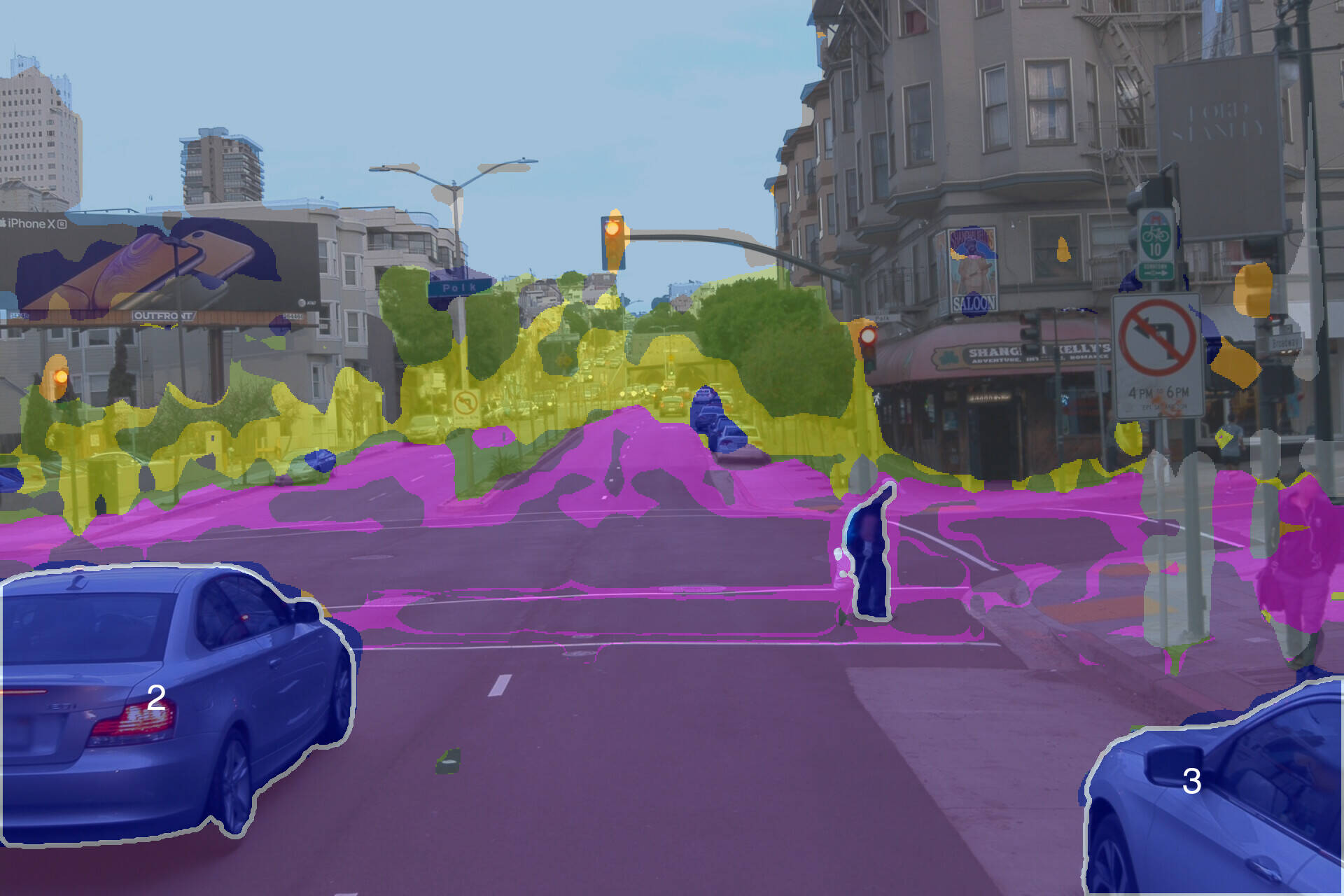}
& \includegraphics[width=\linewidth]{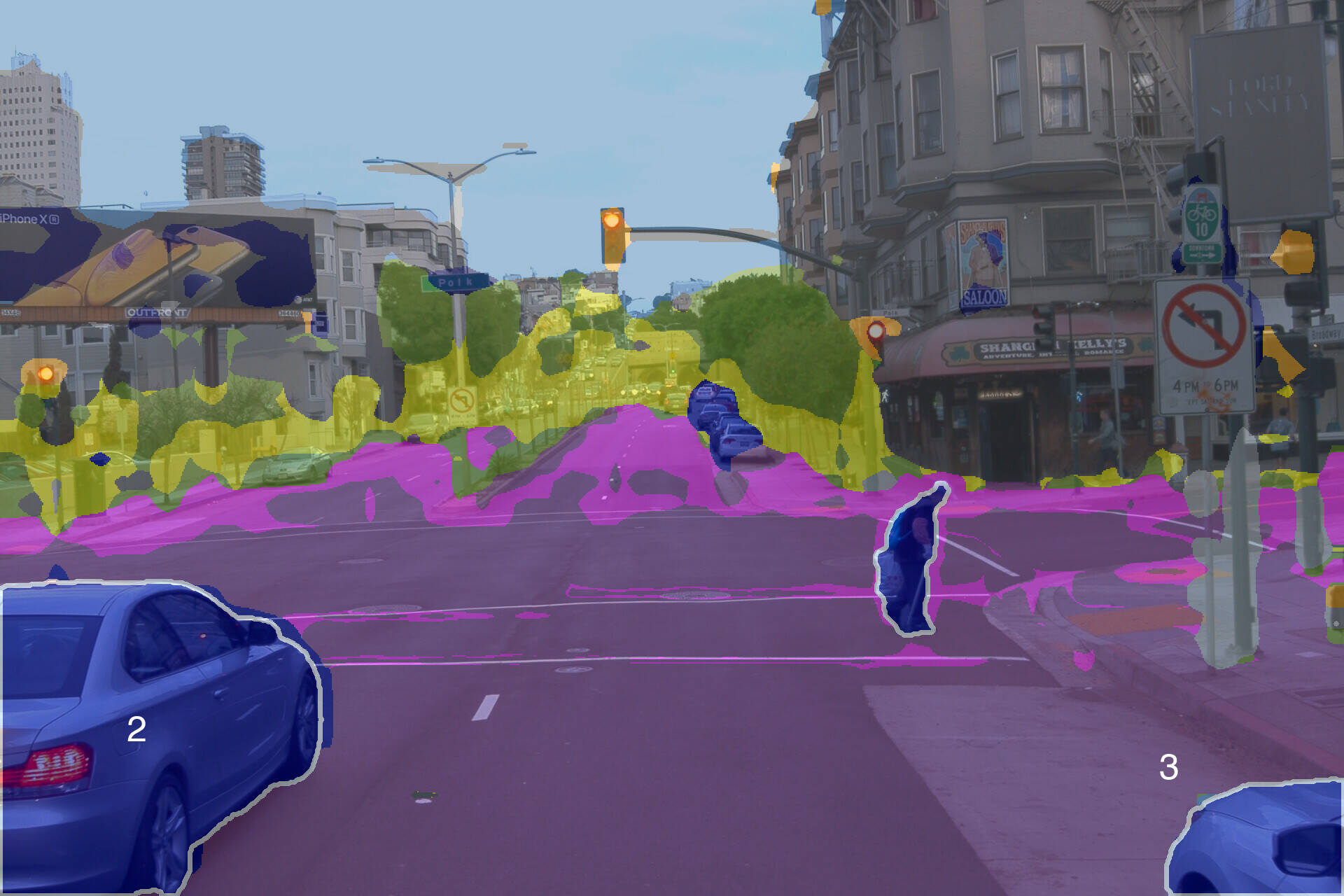}
& \includegraphics[width=\linewidth]{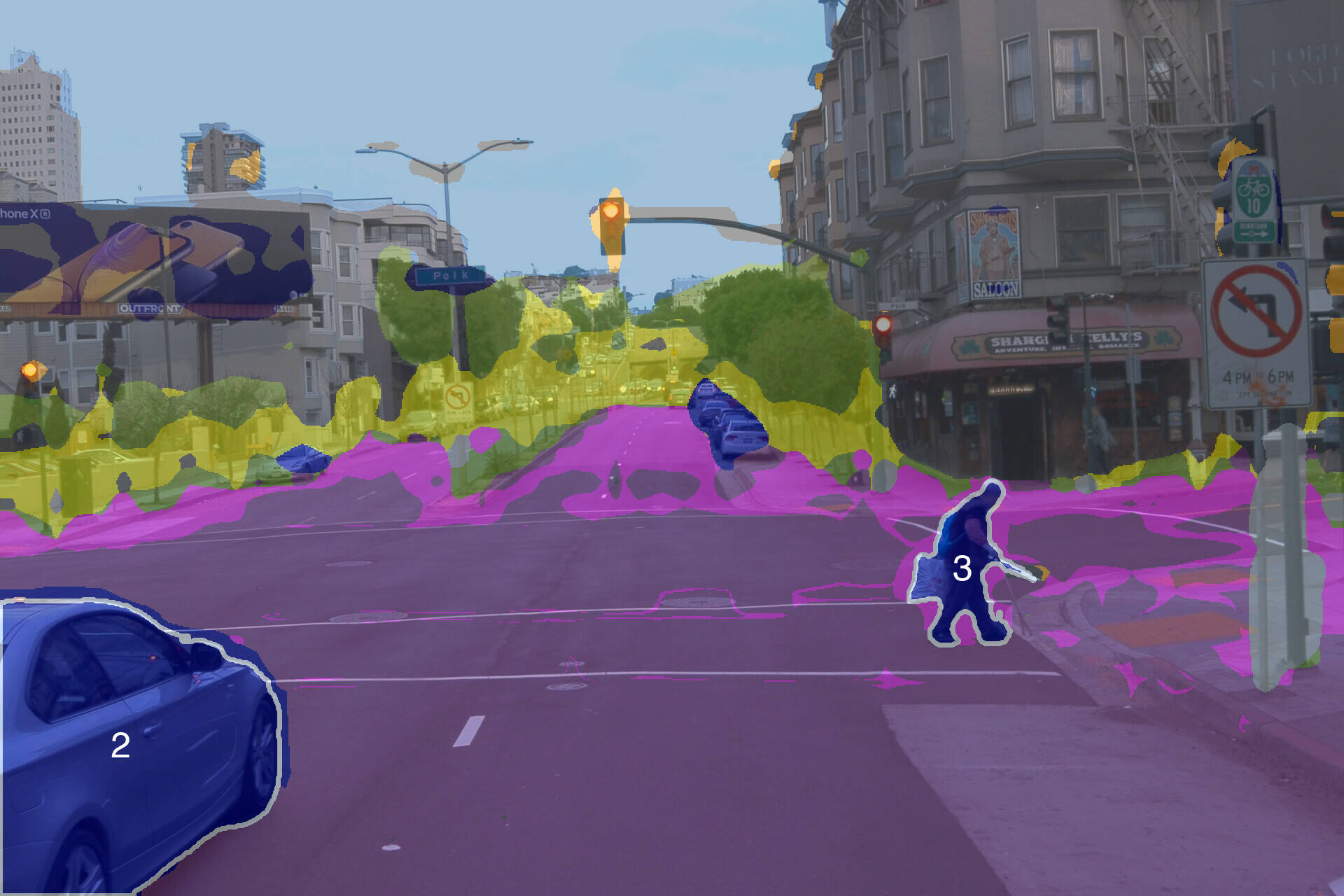} \\

\rotatebox[origin=lB]{90}{\hspace{-0.1em}\tiny{U2Seg~+~SORT}}
& \includegraphics[width=\linewidth]{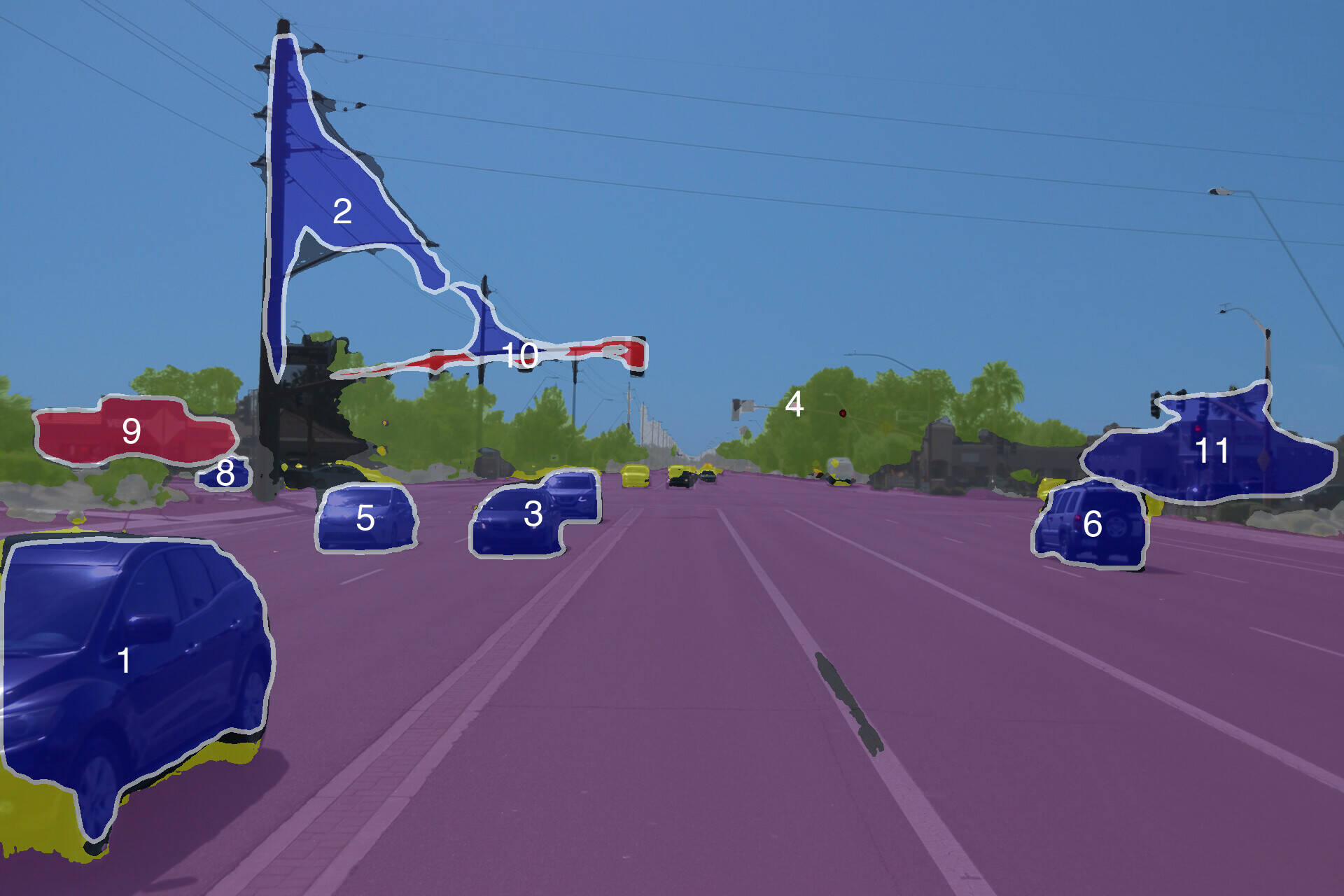} 
& \includegraphics[width=\linewidth]{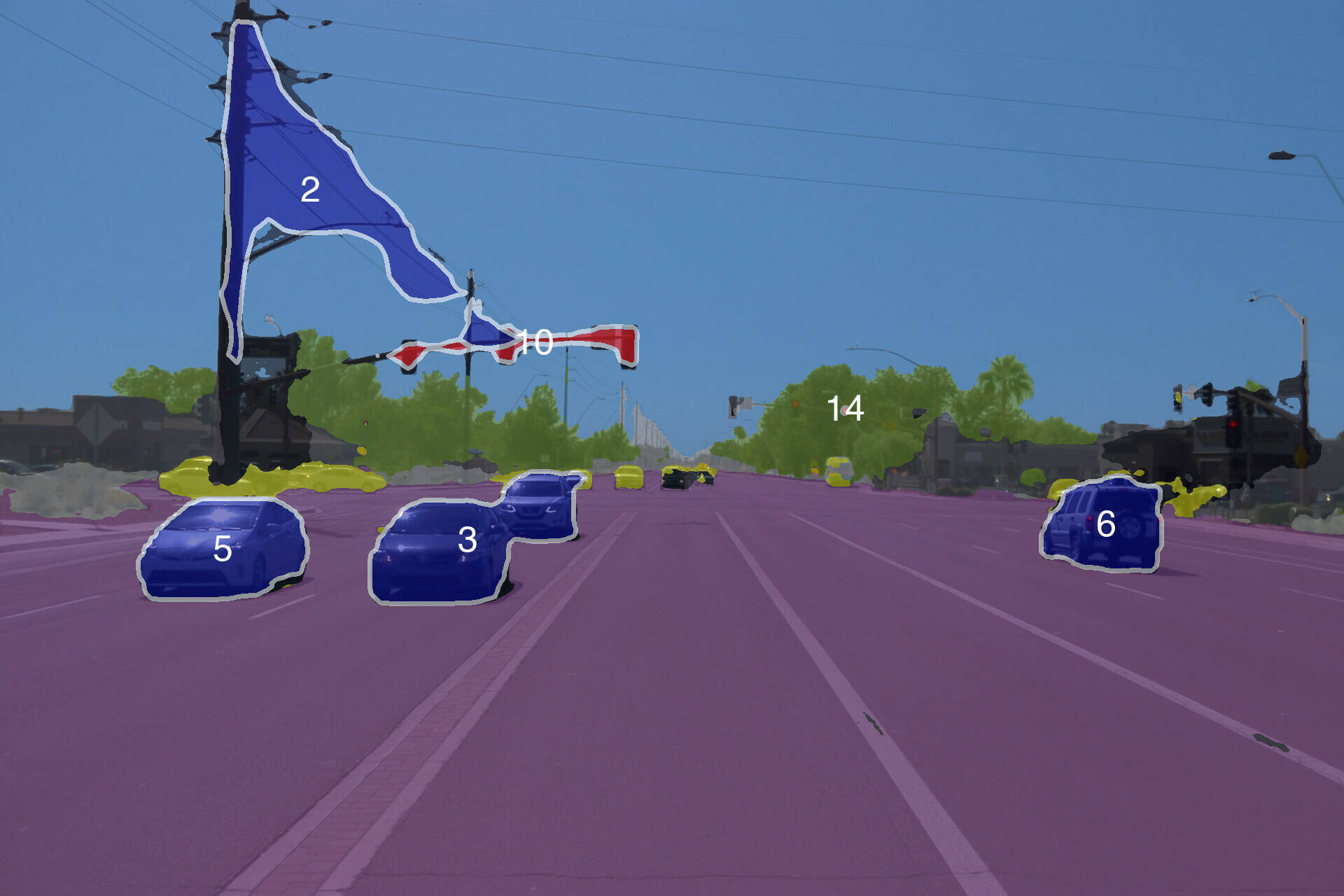} 
& \includegraphics[width=\linewidth]{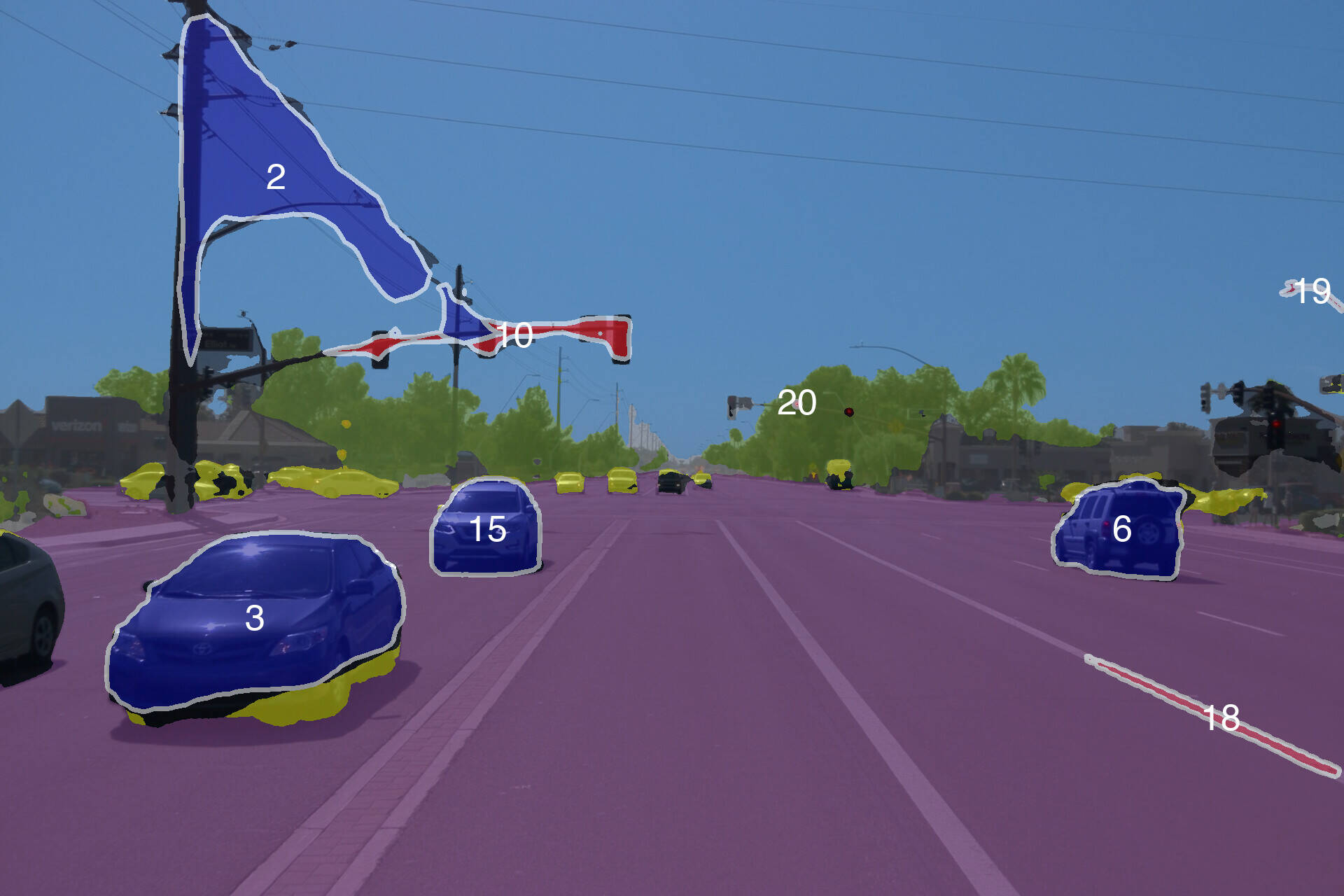} 
& \includegraphics[width=\linewidth]{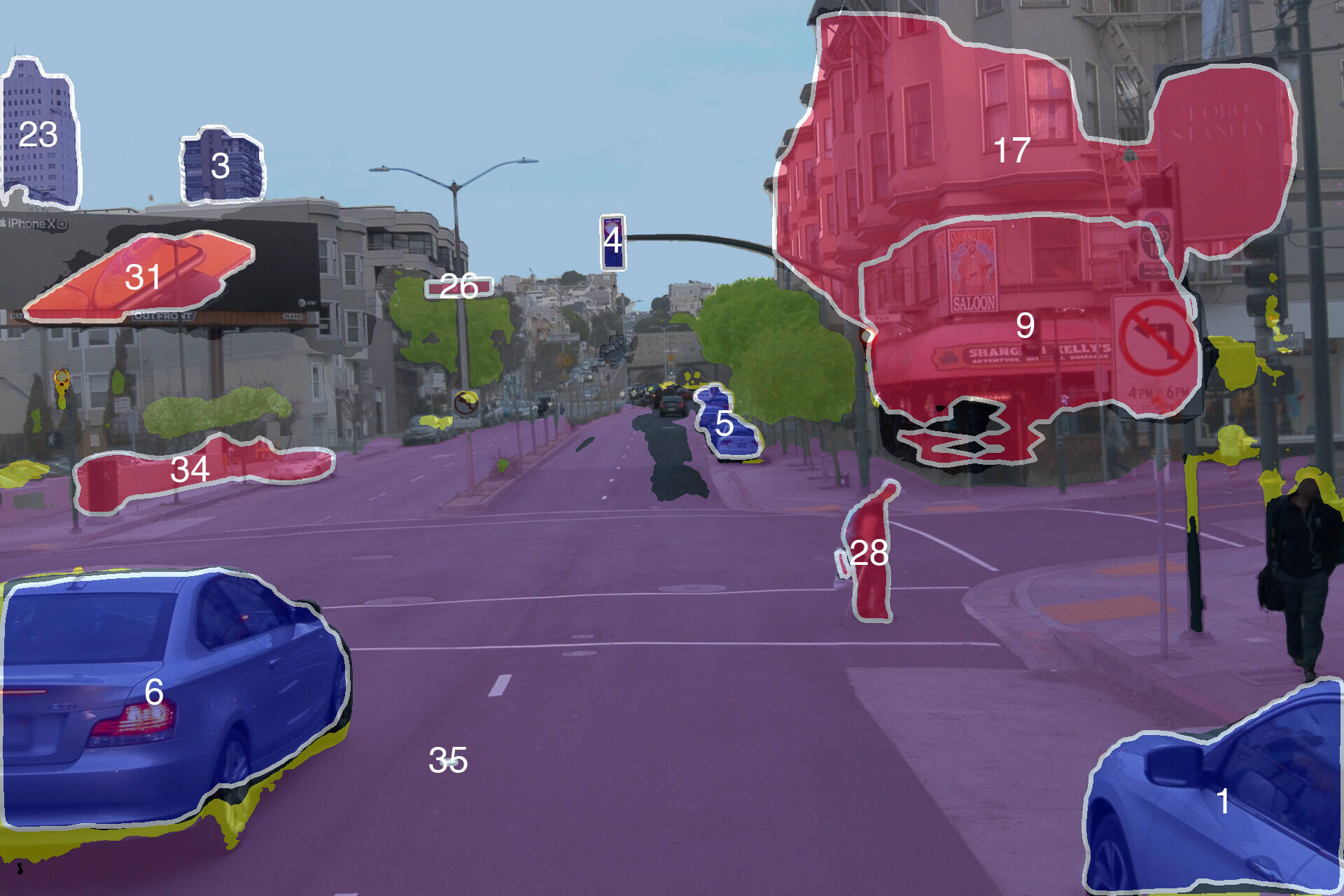} 
& \includegraphics[width=\linewidth]{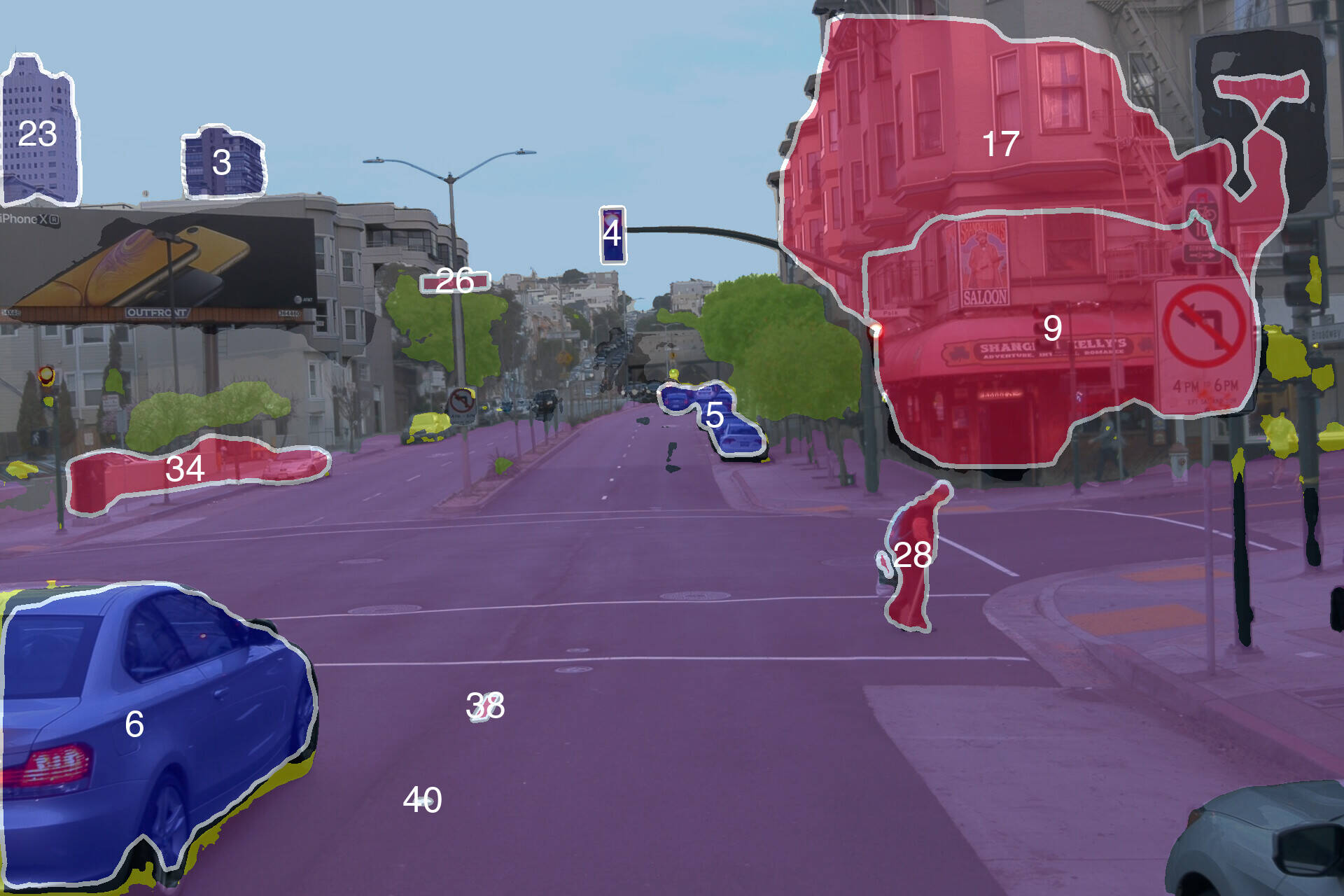} 
& \includegraphics[width=\linewidth]{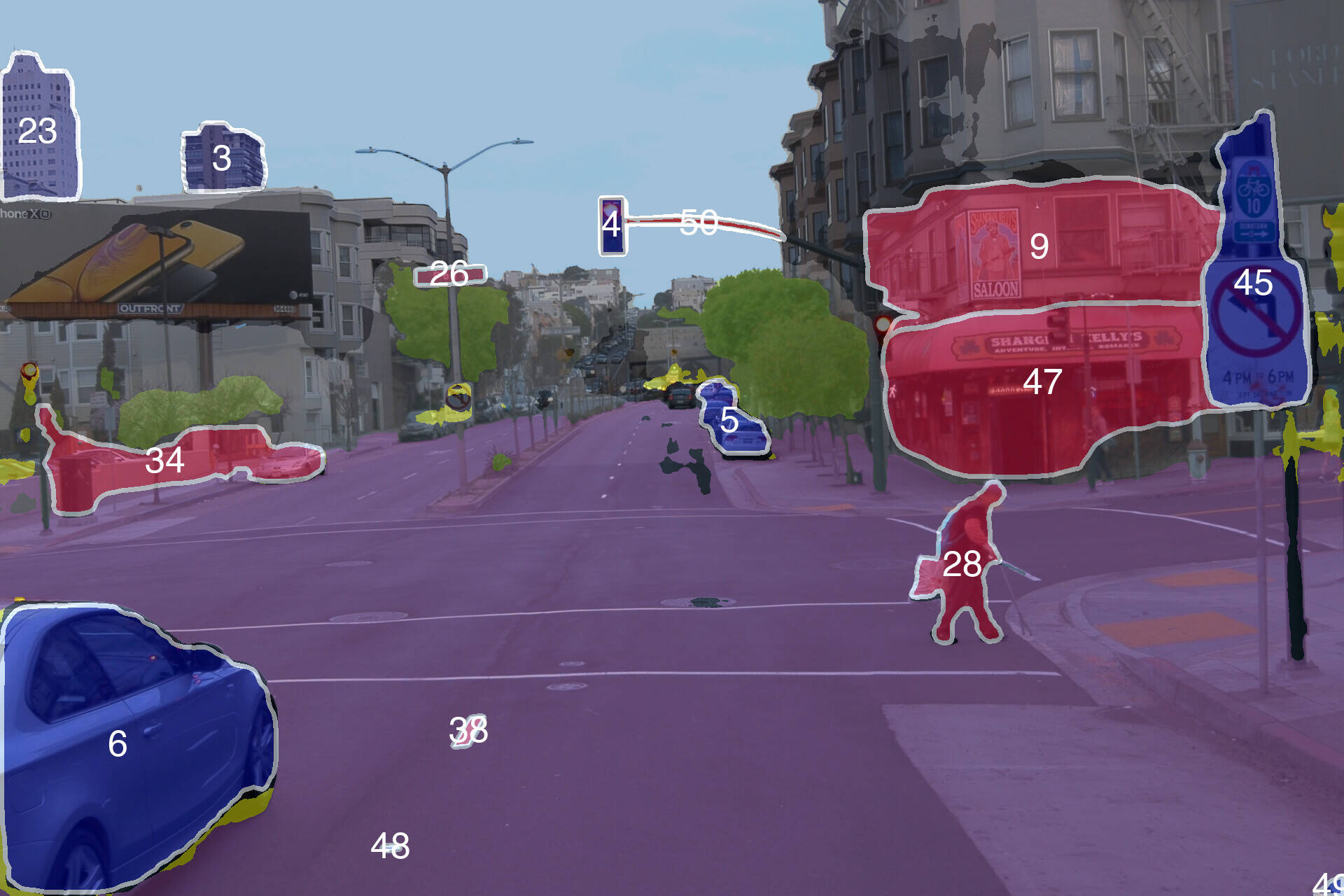}  \\

\rotatebox[origin=lB]{90}{\hspace{-0.1em}\tiny{CUPS~+~SORT}}
& \includegraphics[width=\linewidth]{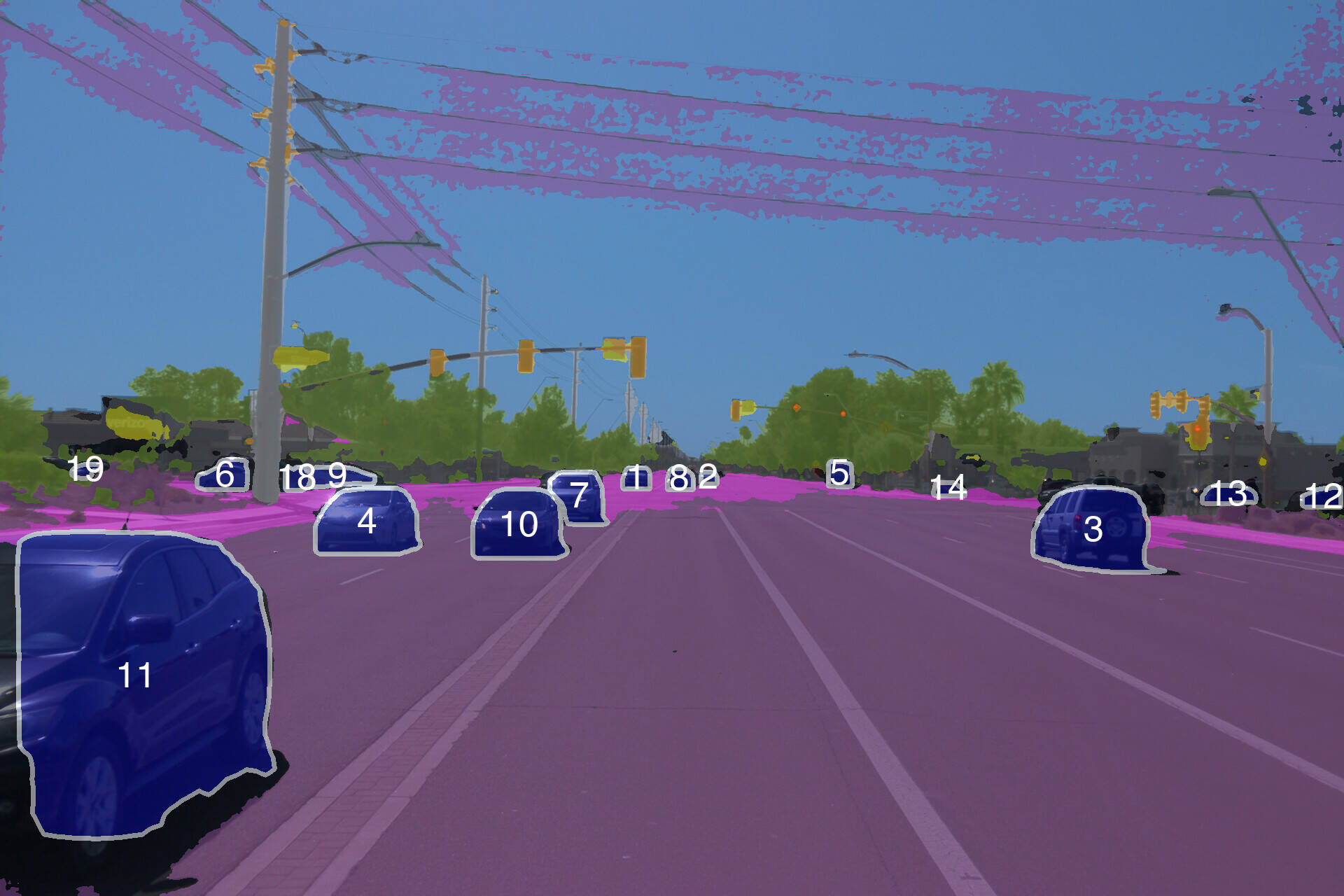}
& \includegraphics[width=\linewidth]{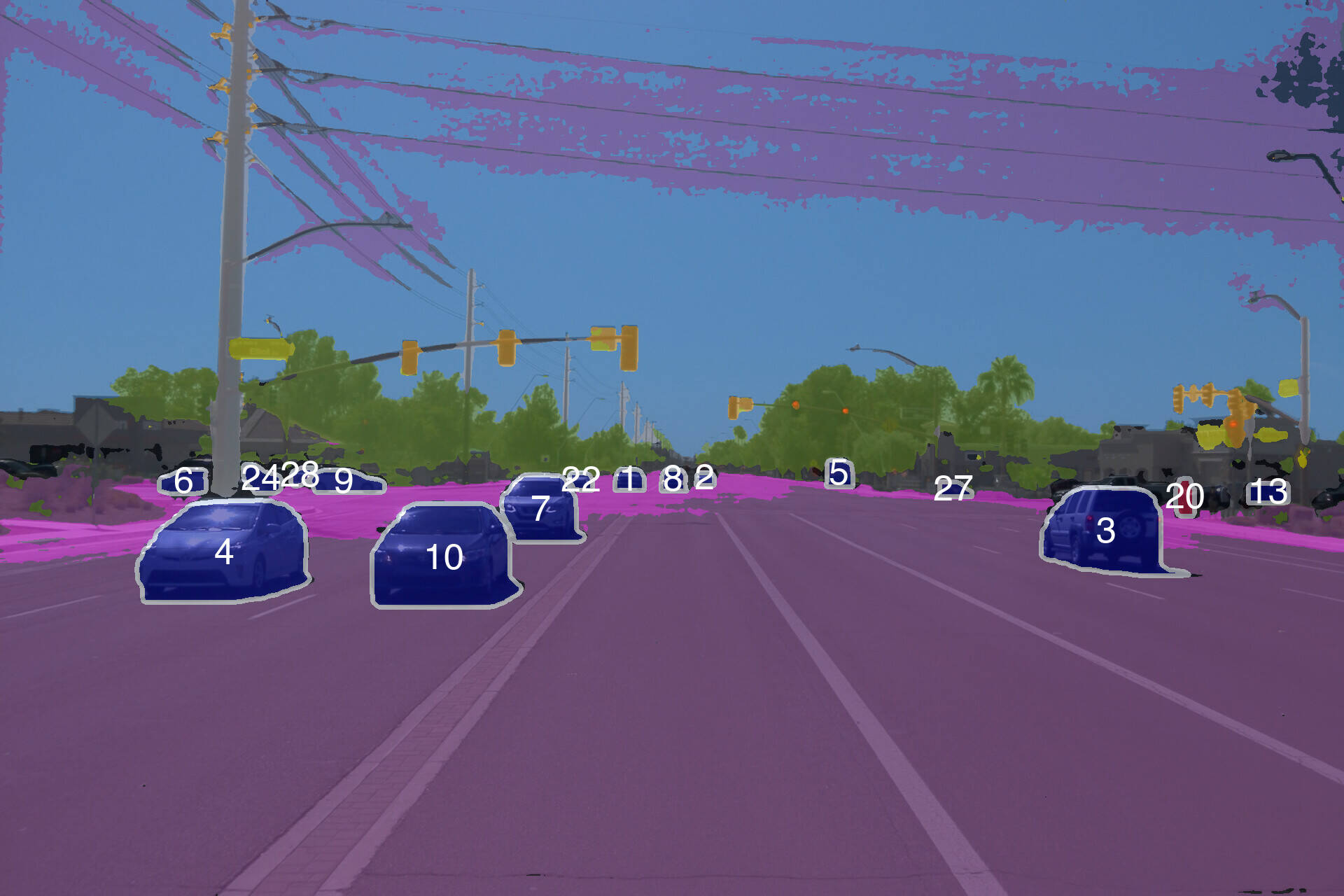}
& \includegraphics[width=\linewidth]{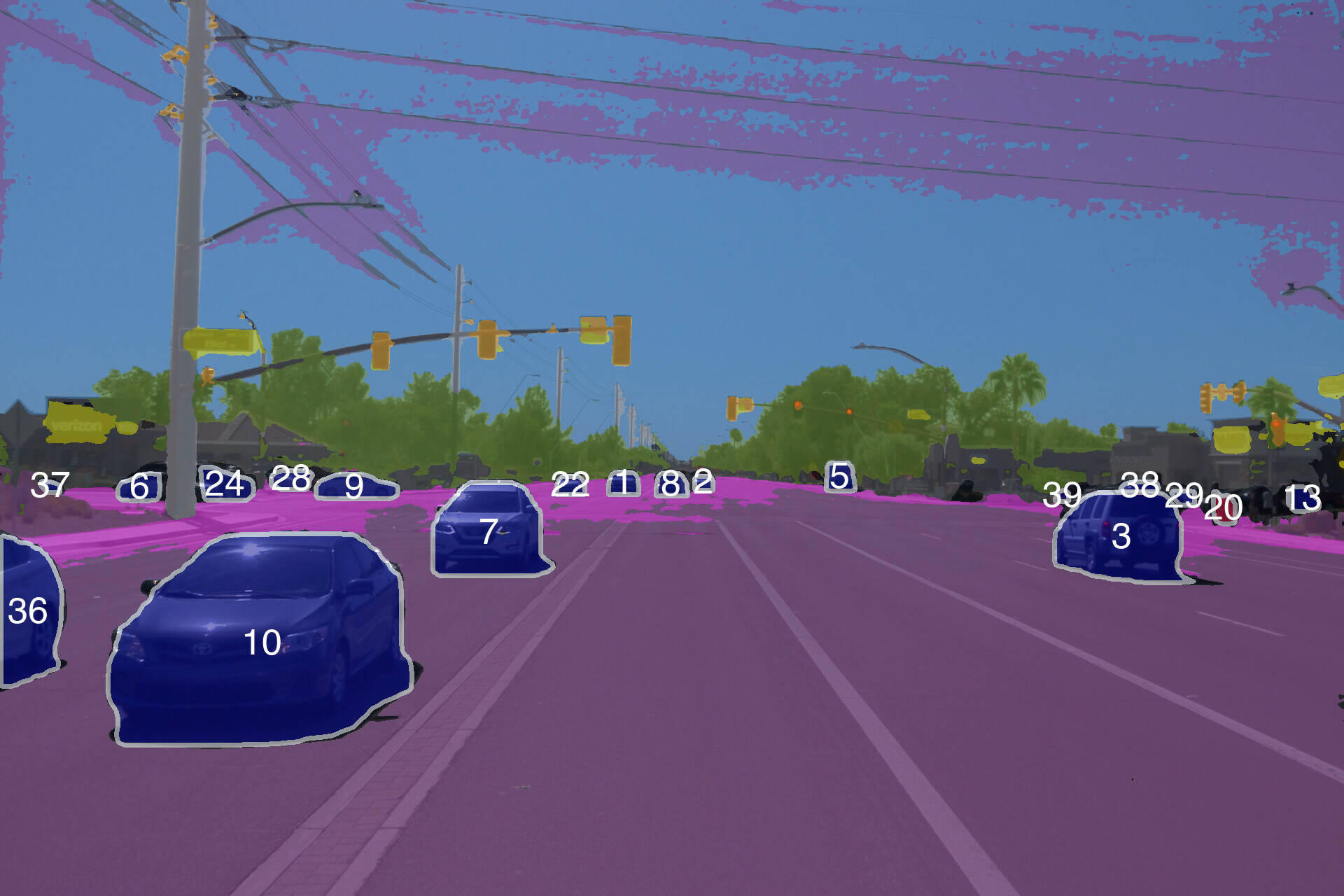}
& \includegraphics[width=\linewidth]{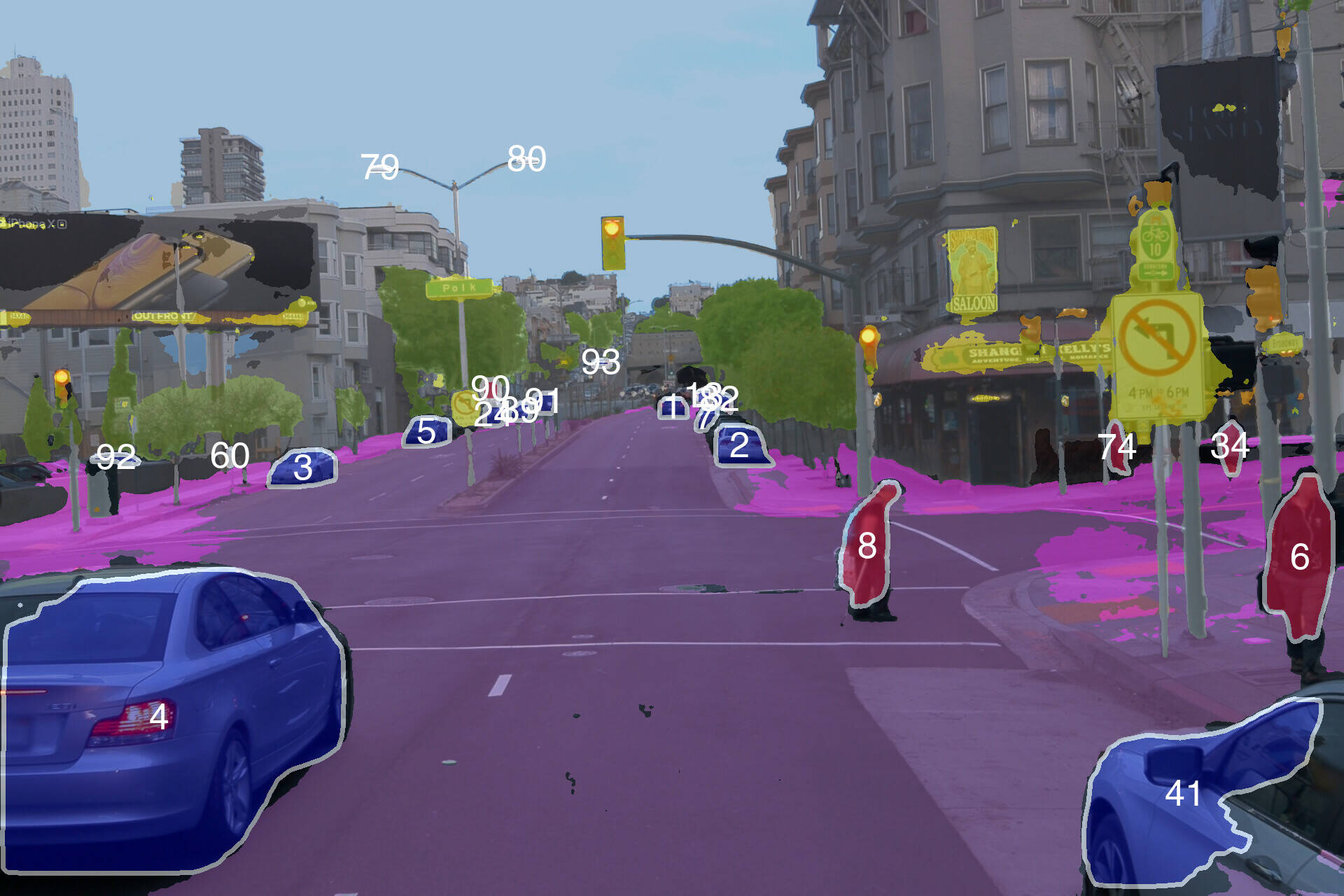}
& \includegraphics[width=\linewidth]{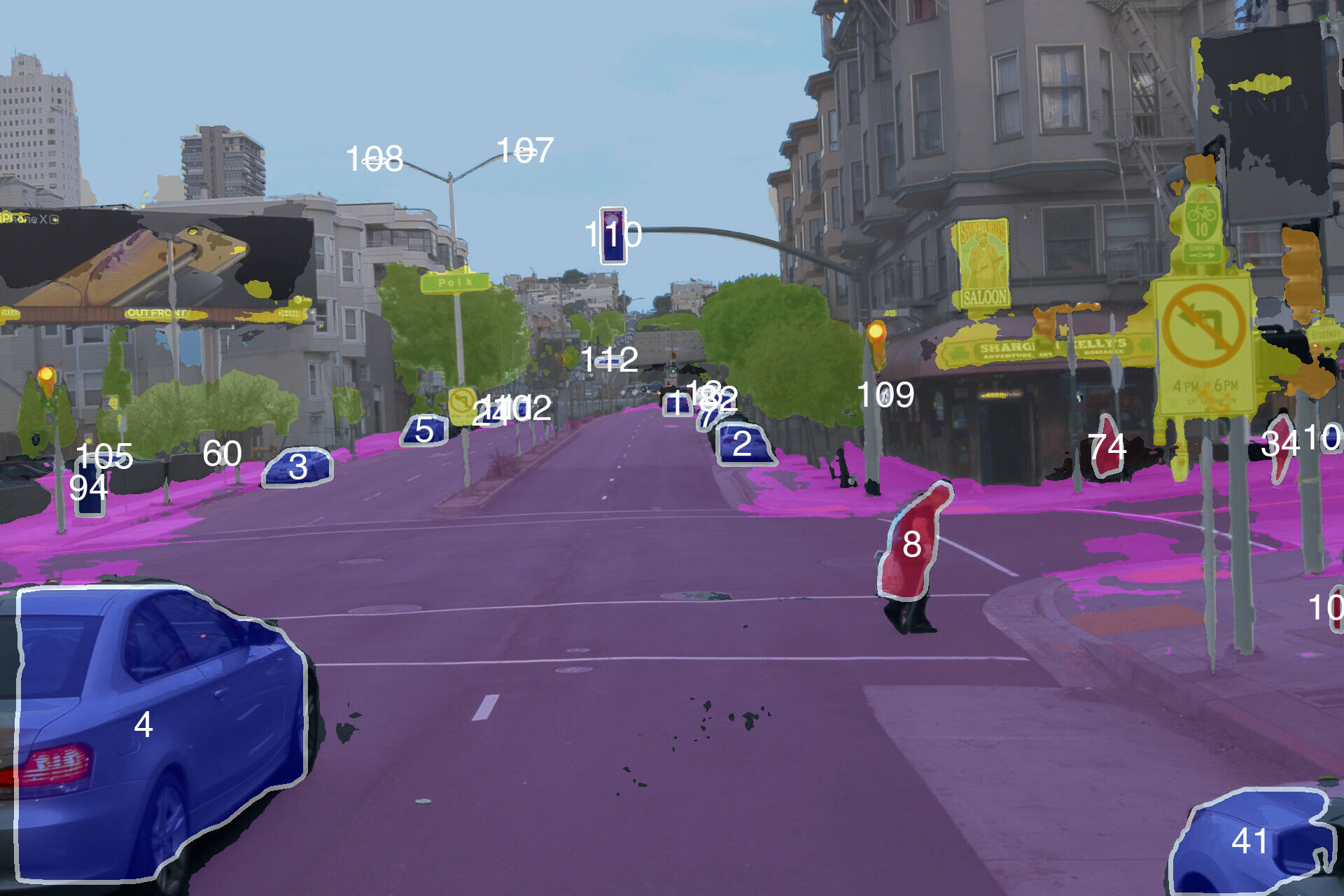}
& \includegraphics[width=\linewidth]{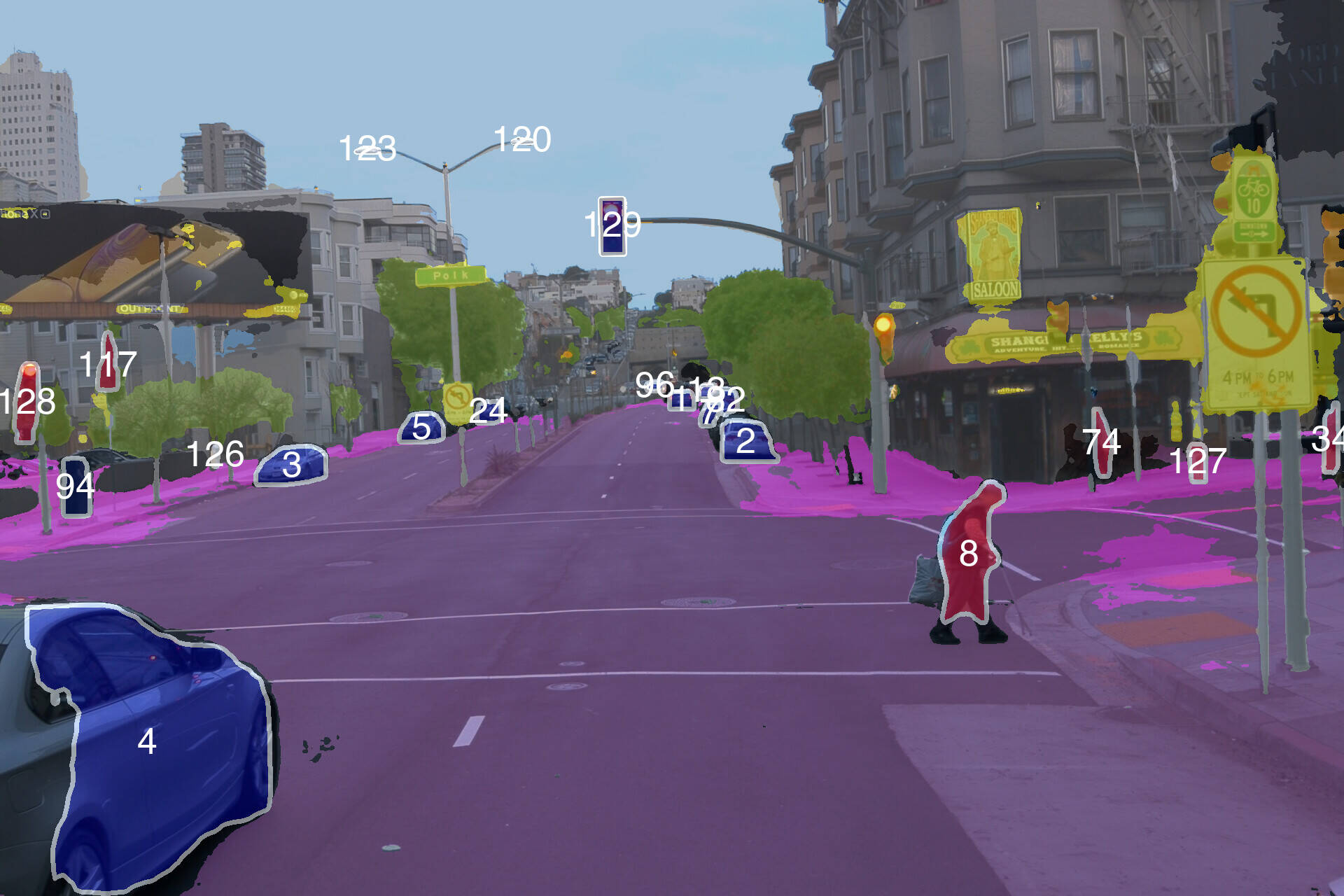} \\

\rotatebox[origin=lB]{90}{\hspace{-0.1em}\tiny{CUPS\textsuperscript{$\dagger$}\!+\phantom{j}SORT}}
& \includegraphics[width=\linewidth]{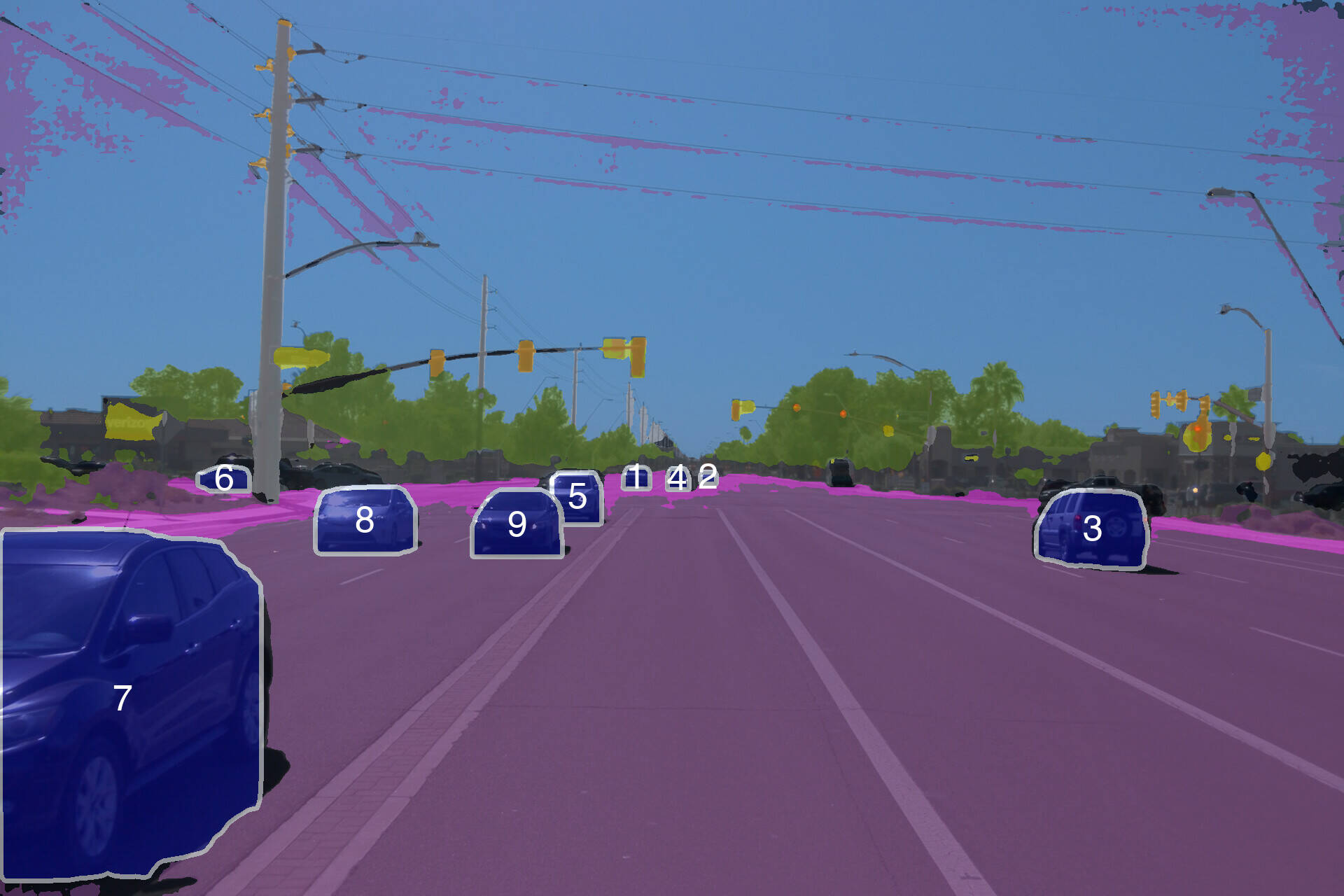} 
& \includegraphics[width=\linewidth]{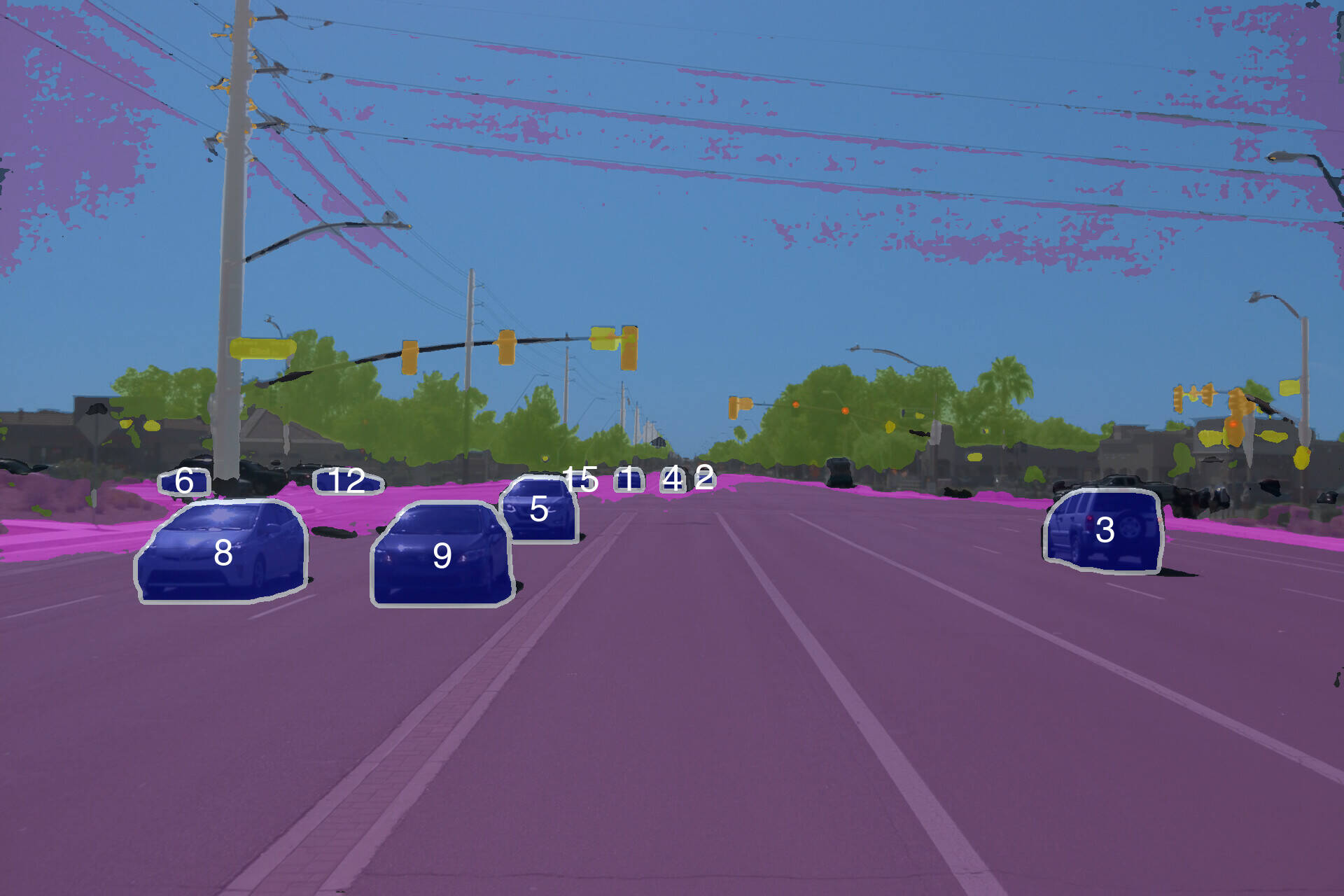} 
& \includegraphics[width=\linewidth]{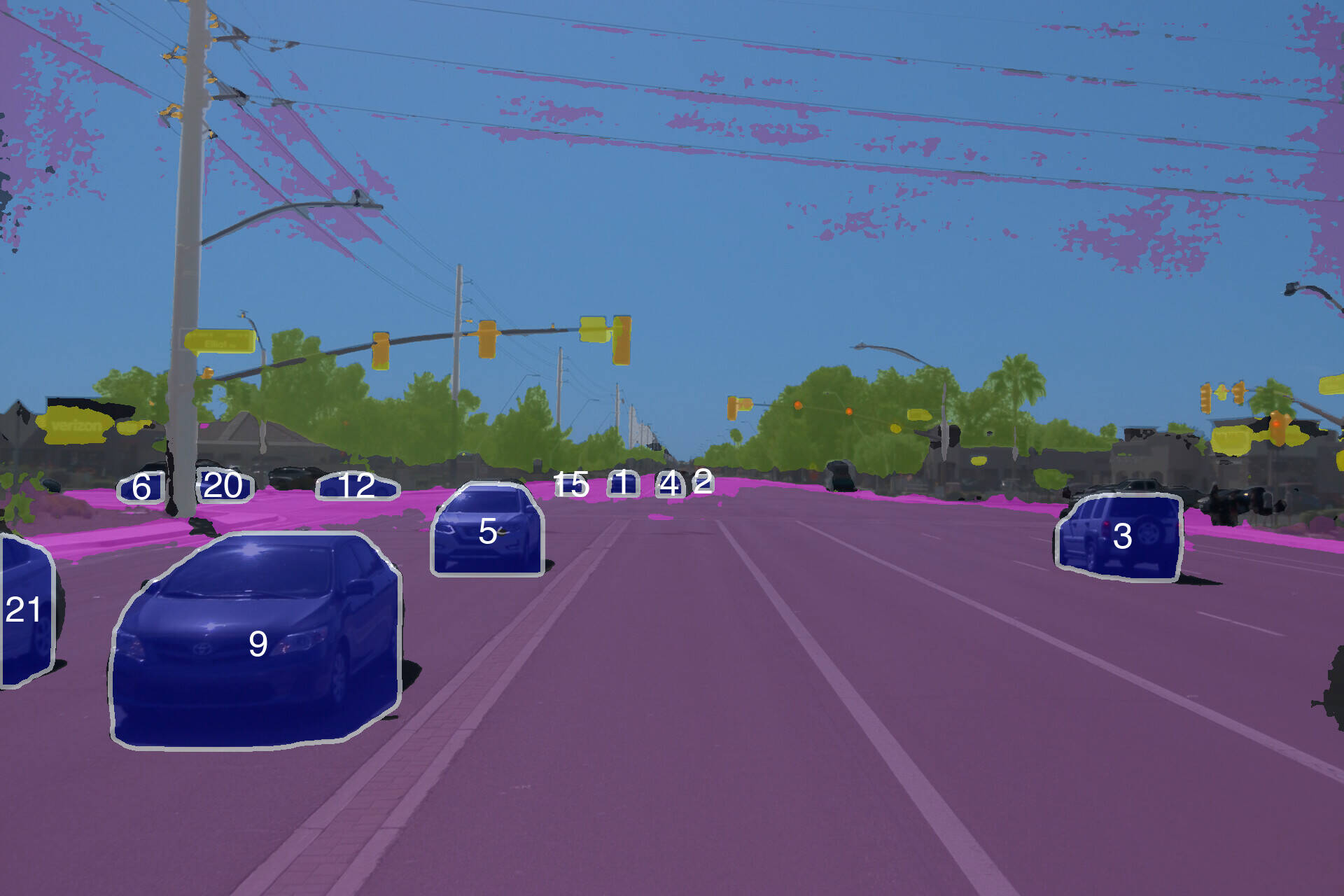} 
& \includegraphics[width=\linewidth]{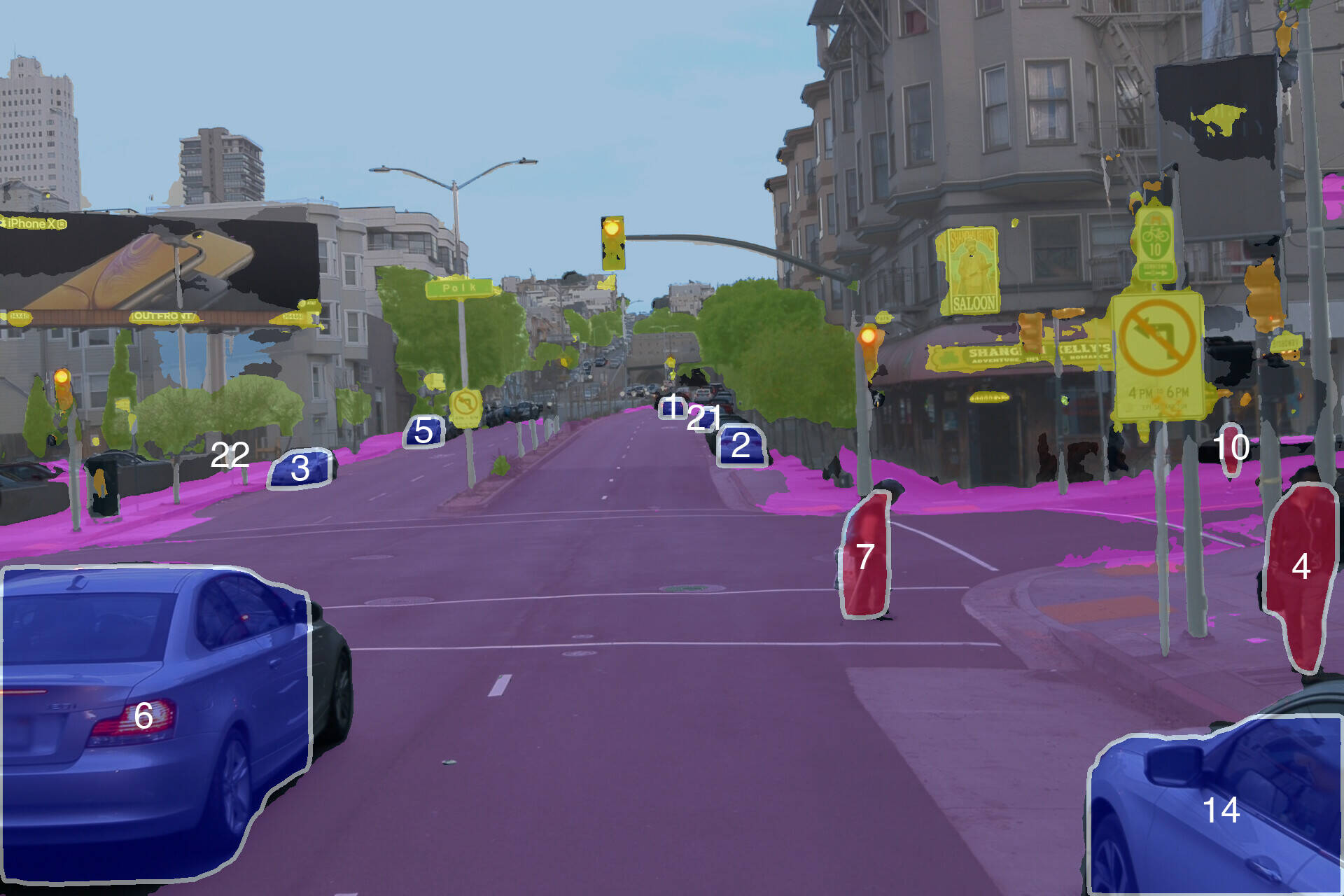} 
& \includegraphics[width=\linewidth]{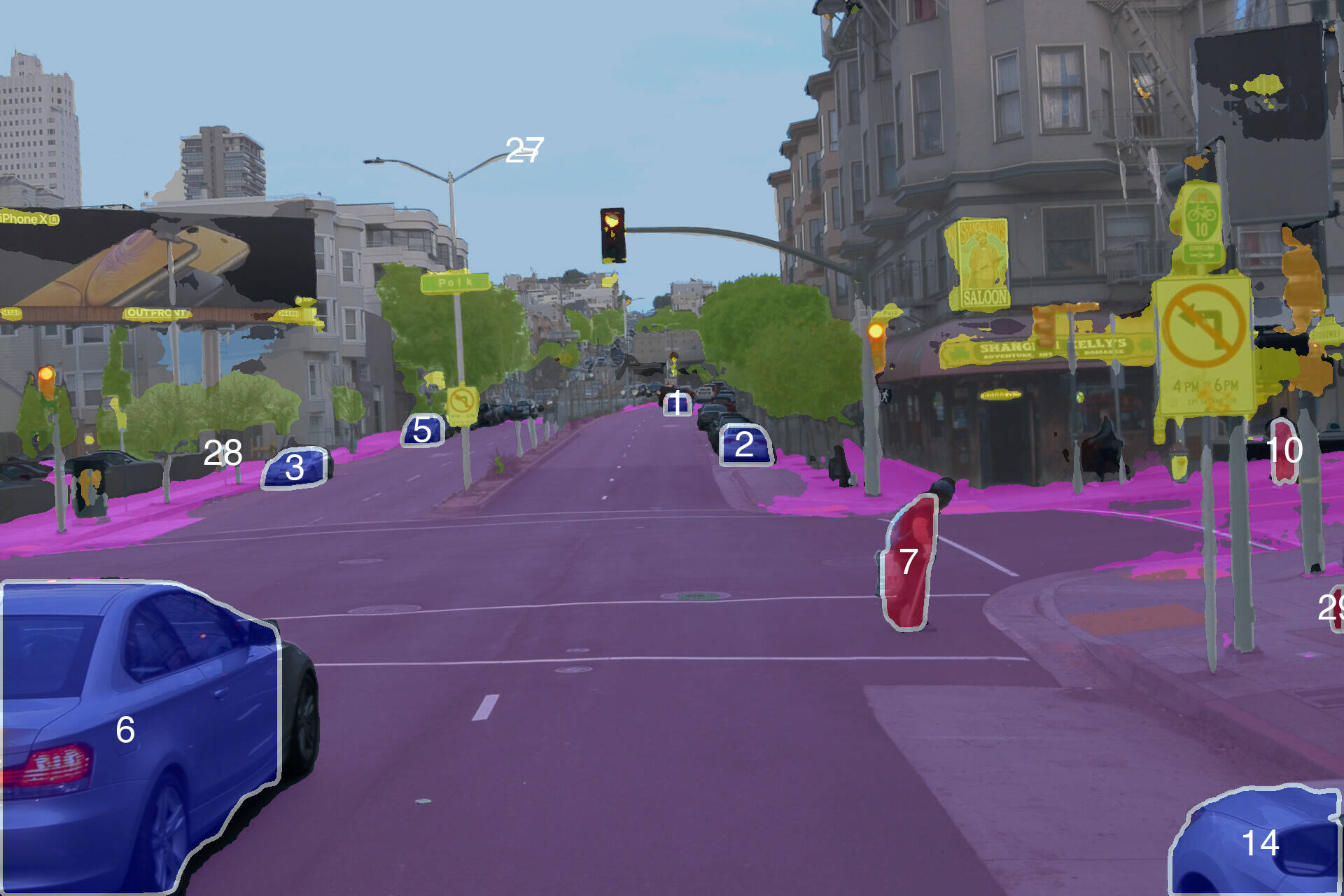} 
& \includegraphics[width=\linewidth]{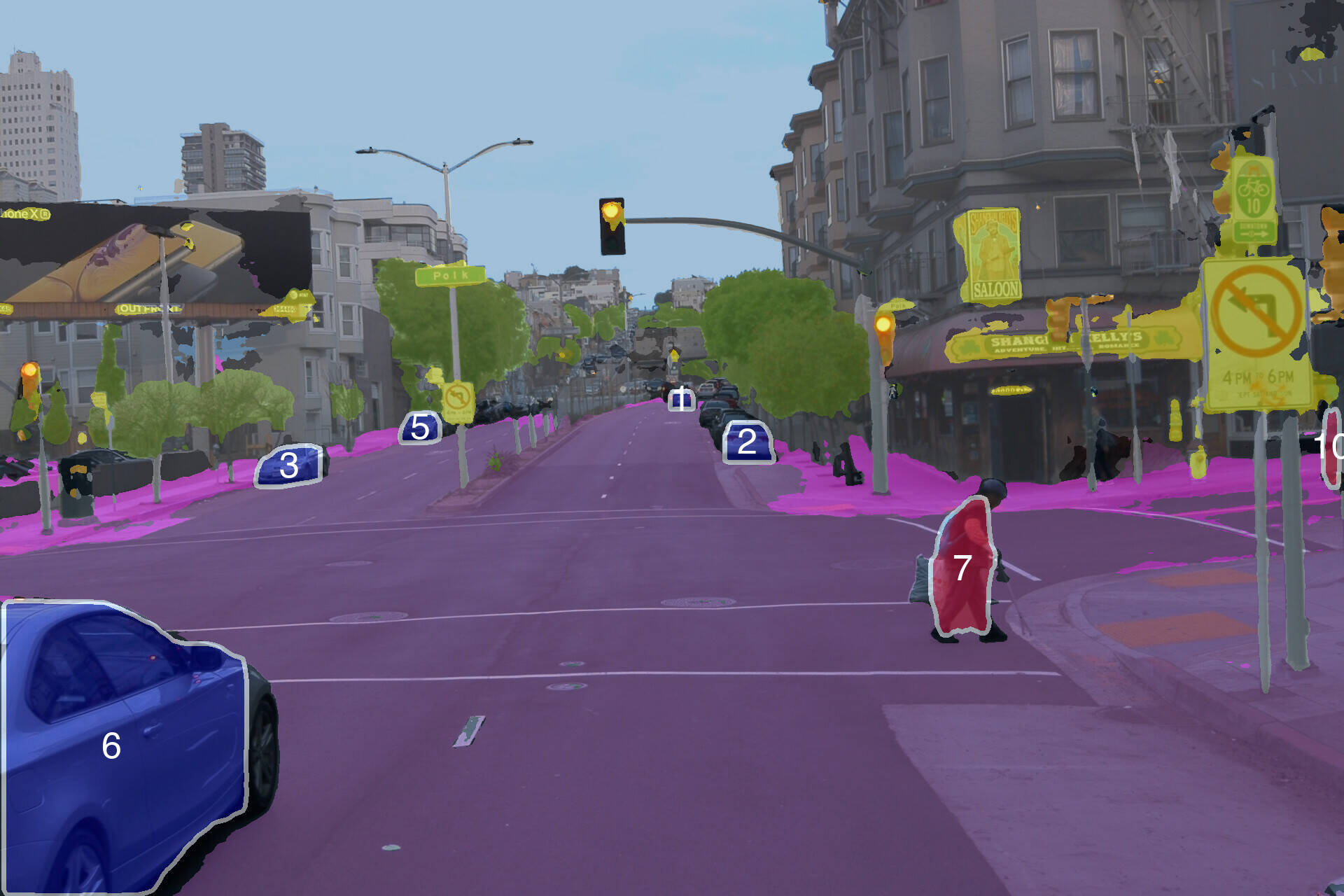} \\

\rotatebox[origin=lB]{90}{\hspace{-0.1em}\tiny{\MethodName~\textit{(Ours)}}}
& \includegraphics[width=\linewidth]{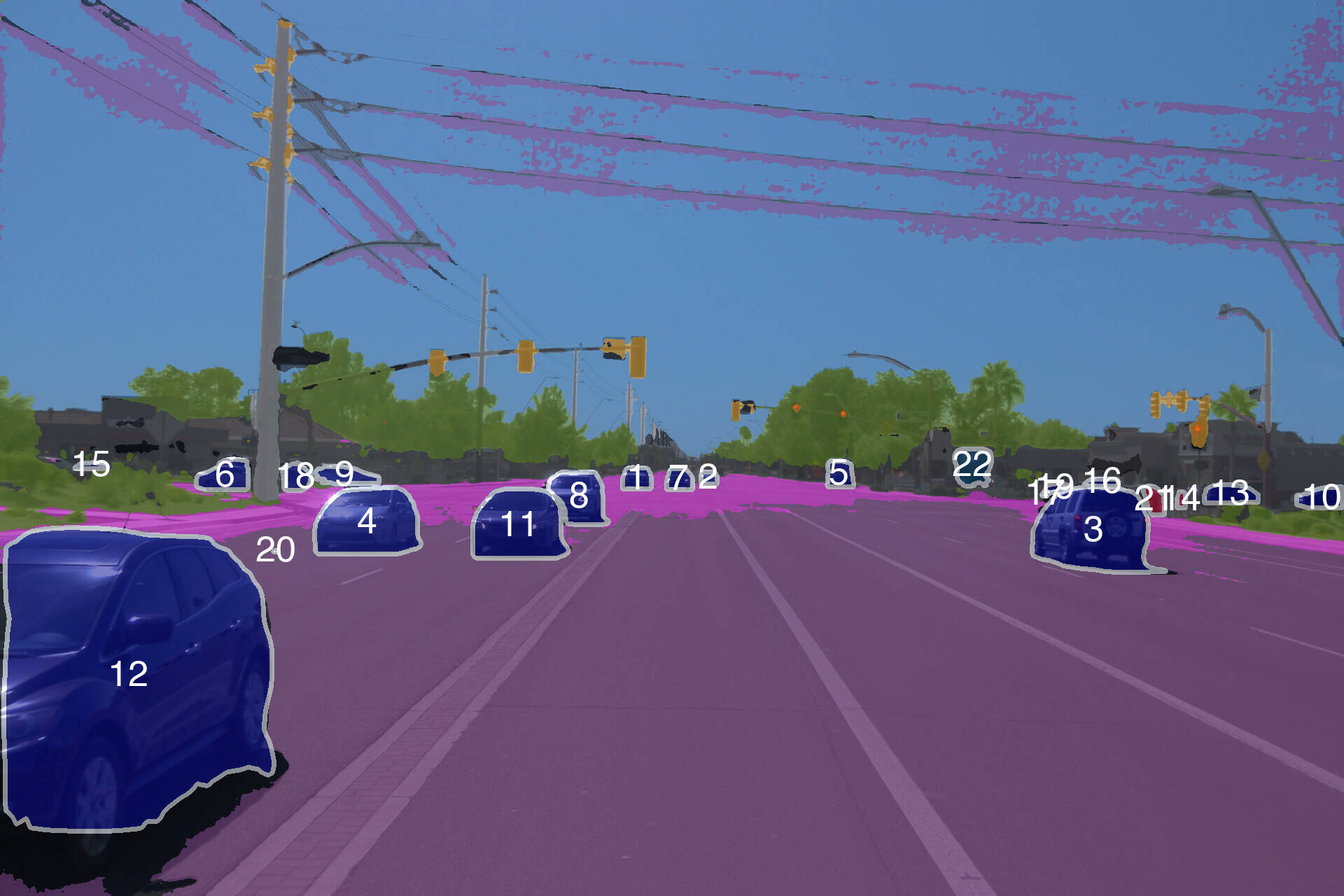} 
& \includegraphics[width=\linewidth]{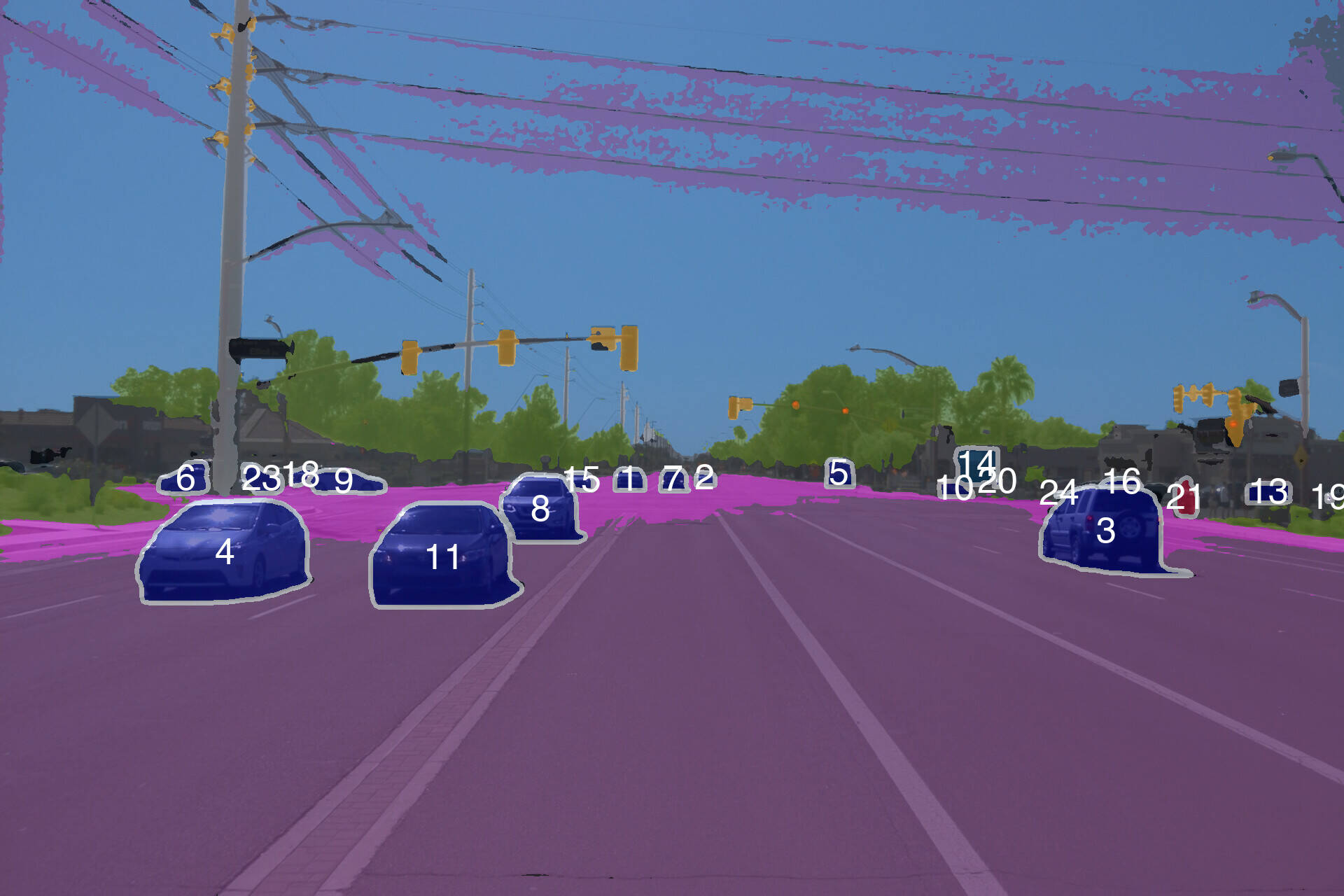}
& \includegraphics[width=\linewidth]{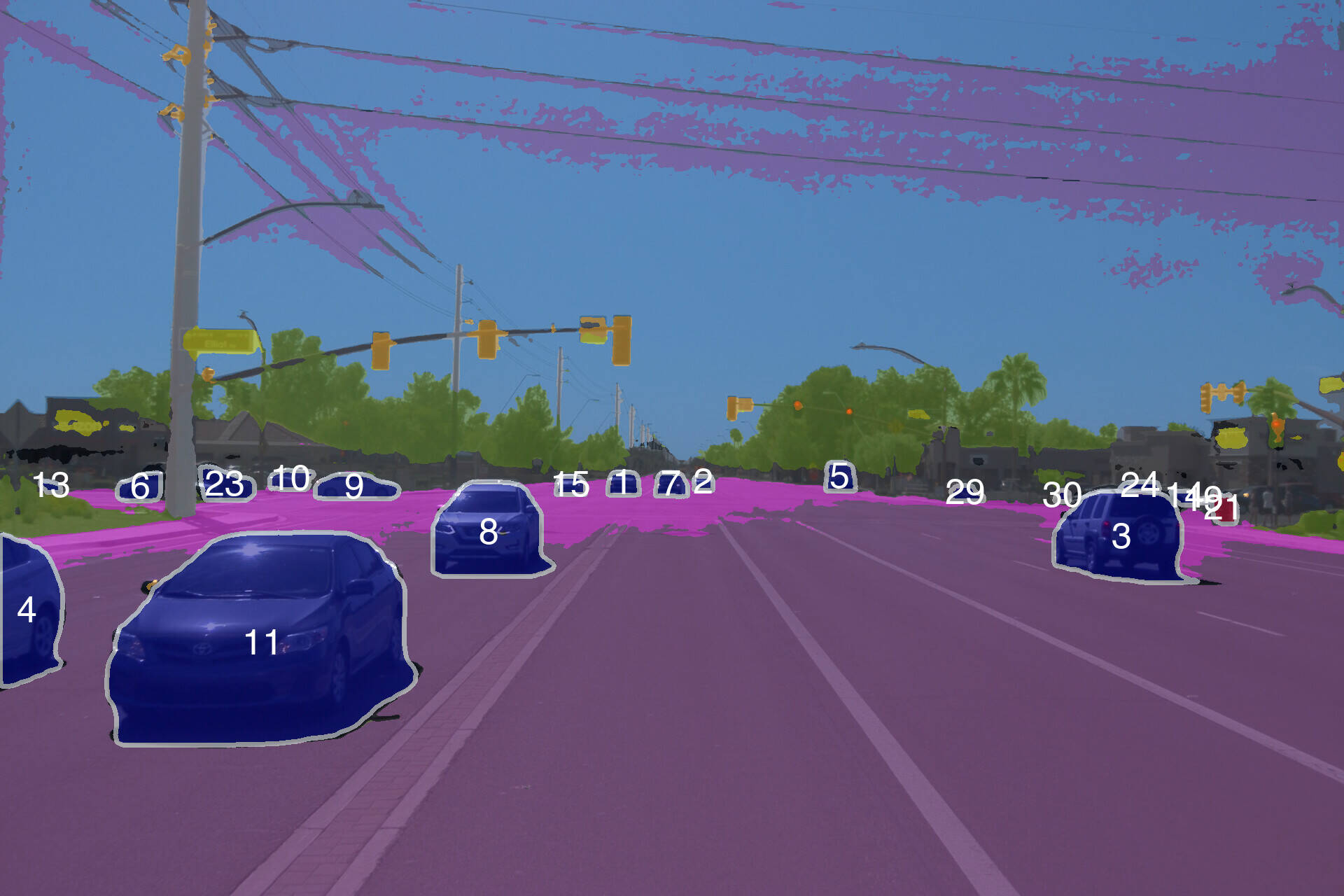}
& \includegraphics[width=\linewidth]{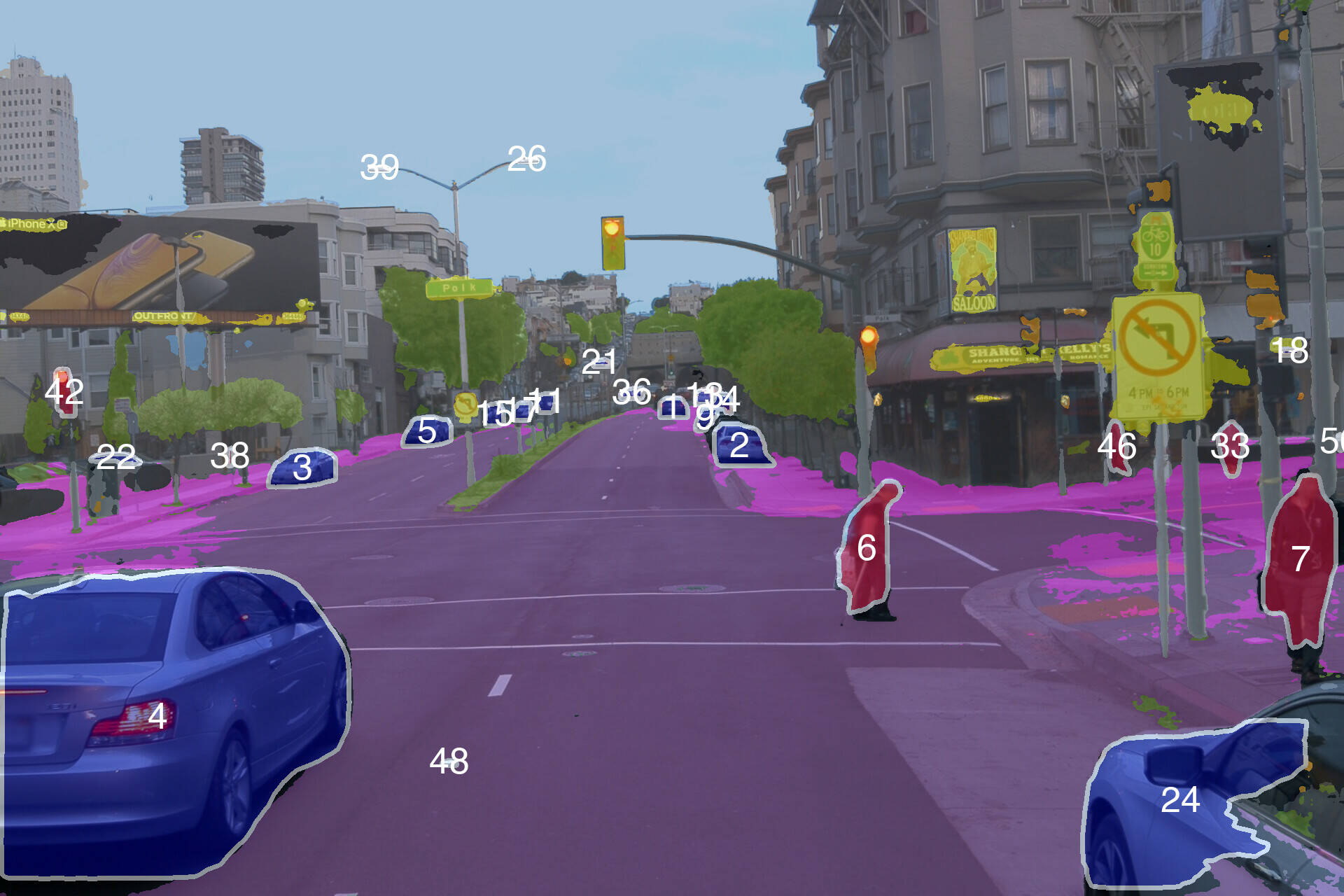}
& \includegraphics[width=\linewidth]{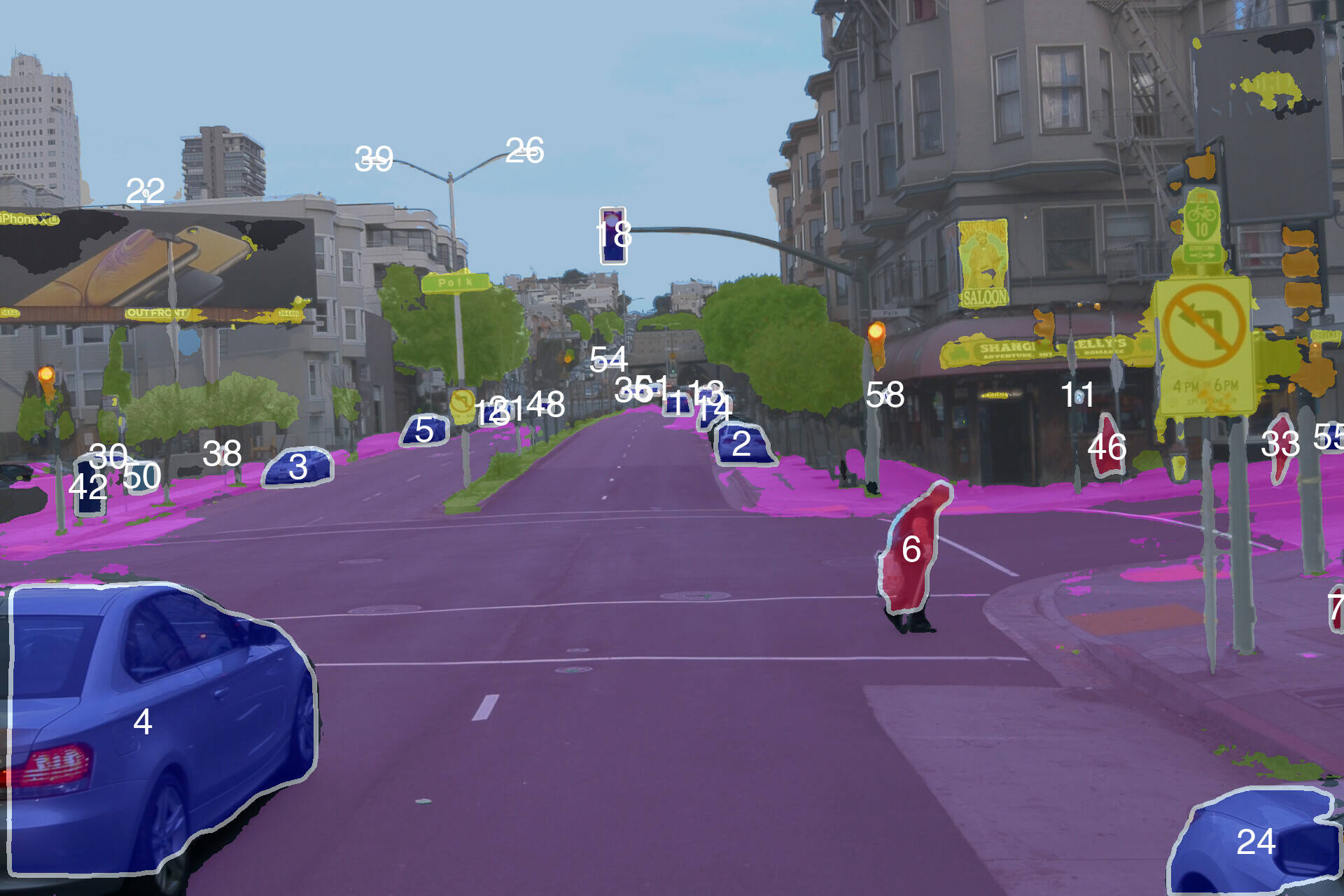}
& \includegraphics[width=\linewidth]{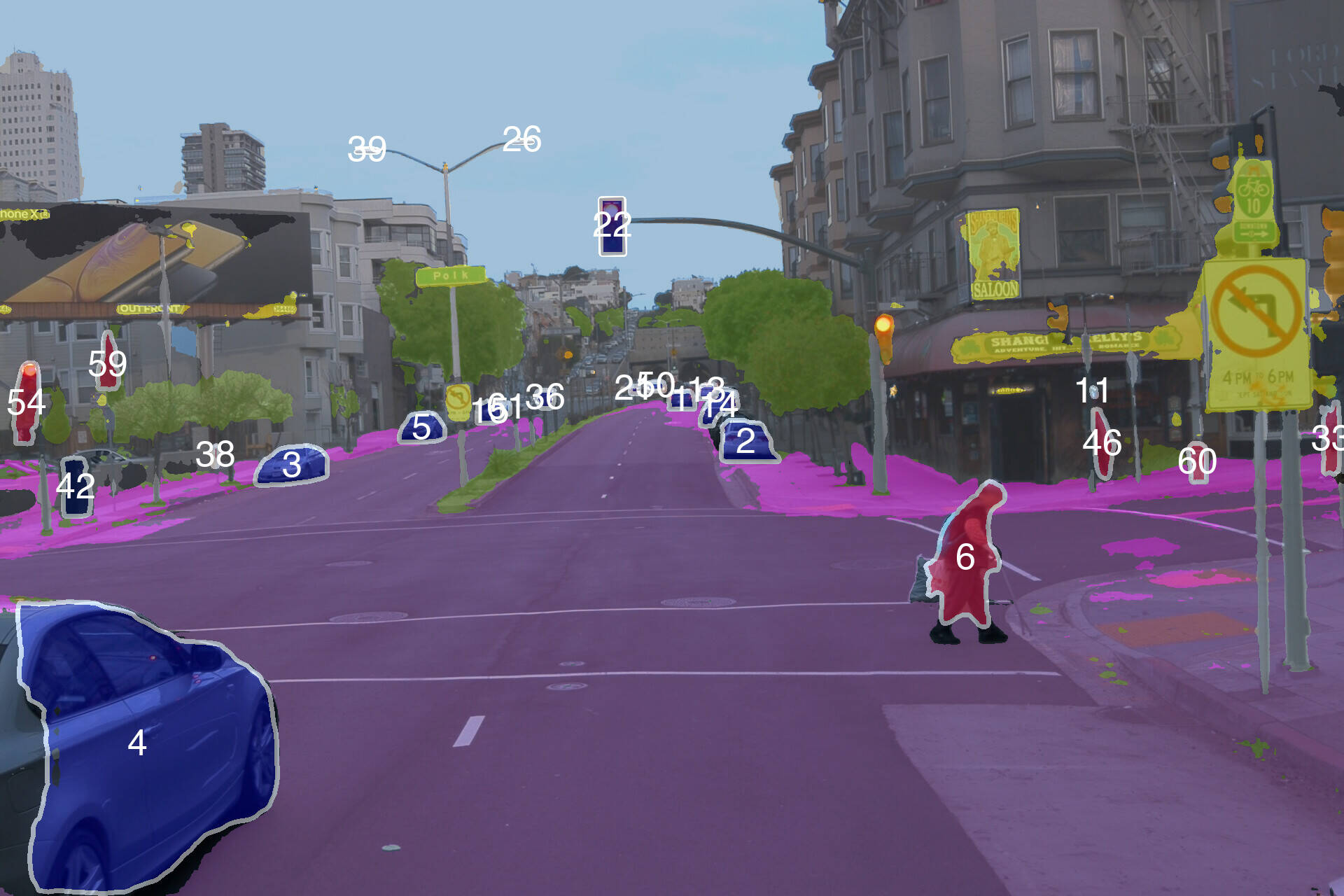}\\[-2.5pt]

&\tiny{$\phantom{.}t_0$}
&\tiny{$\phantom{.}t_1$} 
&\tiny{$\phantom{.}t_2$}
&\tiny{$\phantom{.}t_0$} 
&\tiny{$\phantom{.}t_1$} 
&\tiny{$\phantom{.}t_2$} \\

\end{tabular}

    \vspace{-1.2em}
    \caption{\textbf{Waymo---Qualitative unsupervised VPS examples.} We compare our proposed method \MethodName to the proposed baselines DepthG~\cite{Sick:2024:USS} +~VideoCutLER~\cite{Wang:2024:VCS}, U2Seg~\cite{Niu:2024:UUI}~+~SORT, CUPS~\cite{Hahn:2025:UPS}~+~SORT~\cite{Bewley:2016:SOR}, and CUPS\textsuperscript{\textdagger}~\cite{Hahn:2025:UPS}~+~SORT~\cite{Bewley:2016:SOR} on Waymo~\cite{Mei:2022:WAY, Sun:2020:SPA} val.\label{fig:qualitative_waymo}}
    \vspace{-0.5em}
    \vfill
\end{figure*}

\Cref{fig:qualitative_waymo} compares all methods on the Waymo dataset. DepthG~+~VideoCutLER captures only prominent foreground objects and frequently merges distant instances into single masks (\eg, car~1, left). U2Seg~+~SORT predicts good semantics but continues to merge multiple objects and exhibits noticeable artifacts in the instance predictions. CUPS~+~SORT achieves strong semantic segmentation results and recovers many instances with stable temporal identities. Under the pronounced domain shift and in cluttered scenes, both CUPS~+~SORT and \MethodName occasionally predict small false instance predictions (\eg, lamppost; right example). The monocular variant, CUPS\textsuperscript{\textdagger}~+~SORT, detects fewer objects and generates coarser instance masks (\eg, person~7, right). In contrast, \MethodName provides accurate semantics and numerous precise instance masks (\eg, person~6 and car~4, right) with consistent tracking across the entire sequence.

A qualitative assessment on the OOD dataset, MOTS, is provided in \cref{fig:qualitative_mots}. DepthG~+~VideoCutLER recovers many of the foreground pedestrians but frequently merges multiple individuals into a single instance mask. U2Seg~+~SORT predicts pedestrian instances reliably, yet suffers from artifacts (\eg, store signs in the right example). CUPS~+~SORT outputs precise instance masks but occasionally fails to maintain tracks (\eg, person~2; left example). The monocular variant, CUPS\textsuperscript{\textdagger}~+~SORT, yields coarser masks and more artifacts overall. Overall, \MethodName delivers the strongest qualitative results among all evaluated methods: it provides accurate instance masks with stable temporal associations, struggling only with very small distant objects.

We further assess the generalization ability of \MethodName. \Cref{fig:qualitative_davis} shows qualitative results on the DAVIS~\cite{davis} dataset, using the class assignments learned on Cityscapes-VPS to map pseudo-classes to ground-truth categories for visualization. \MethodName generalizes well to this unseen domain and correctly handles unseen semantic concepts, such as forest and mountains (top examples). In addition, \MethodName produces accurate instance masks with consistent tracking over time.

\begin{figure}[t!]%
    \centering
    \setlength{\tabcolsep}{1.27pt}
    \input{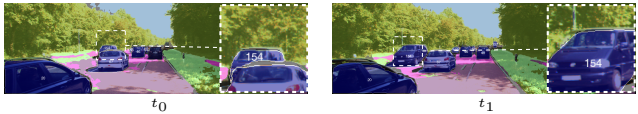}
    \vspace{-0.9em}
    \caption{\textbf{\MethodName partial occlusion example} on KITTI-STEP. While \MethodName struggles by design with full occlusion, \MethodName is still able to track objects through some partial occlusions (\cf object \num{154}). Zoom in for details.}
    \label{fig:occlusions}
    \vspace{-0.5em}
\end{figure}

\section{Limitations and Future Work}\label{sec:supp_limit_future}

\inparagraph{Moving objects assumption.} We show that unsupervised VPS is feasible by combining self-supervised visual representations with motion and depth cues. A current limitation is the requirement for independently moving objects to obtain initial video instance pseudo-labels. Although this assumption holds in many real-world scenarios, predominantly static objects, \eg, a painting mounted on a wall, remain challenging to segment in the unsupervised setting. MaskCut-based approaches such as U2Seg or VideoCutLER can, in principle, discover such objects, but they require object-centric imagery and exhibit poor results on scene-centric data. Integrating motion-based segmentation with MaskCut-based pseudo-labeling may enable the segmentation of predominantly static objects while still scaling to scene-centric videos.

\begin{figure*}[ht!]
    \centering
    \newcommand{\imgwidth}{0.159}
\newcommand{\dddviswidth}{0.182}

\tiny
\renewcommand{\arraystretch}{1.3}
\sffamily
\begin{tabularx}{0.997\textwidth}{*{2}{>{\centering\arraybackslash}X}}  
    \cellcolor{building}\textcolor{white}{Background}
    & \cellcolor{person}\textcolor{white}{Person}
\end{tabularx}

\vspace{1pt}

\scriptsize
\sffamily
\setlength{\tabcolsep}{1pt}
\renewcommand{\arraystretch}{1.0}

\begin{tabular}{
    >{\centering\arraybackslash}m{0.017\textwidth}
    >{\centering\arraybackslash}m{\imgwidth\textwidth}
    >{\centering\arraybackslash}m{\imgwidth\textwidth}
    >{\centering\arraybackslash}m{\imgwidth\textwidth}
    >{\centering\arraybackslash}m{\imgwidth\textwidth}
    >{\centering\arraybackslash}m{\imgwidth\textwidth}
    >{\centering\arraybackslash}m{\imgwidth\textwidth}
}

\rotatebox[origin=lB]{90}{\hspace{-0.5em}\tiny{Ground-\phantom{j}truth}}
& \includegraphics[width=\linewidth]{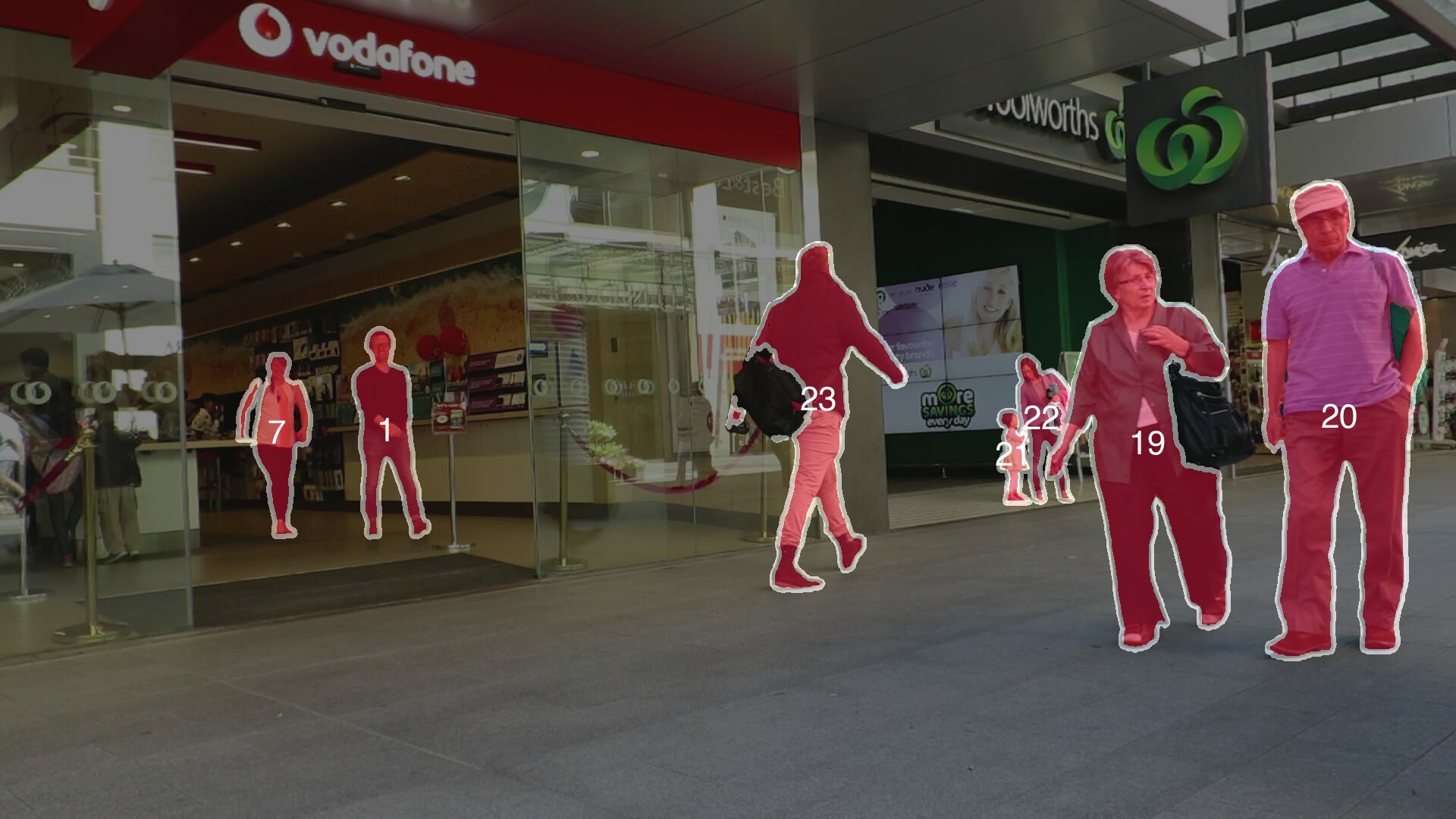} 
& \includegraphics[width=\linewidth]{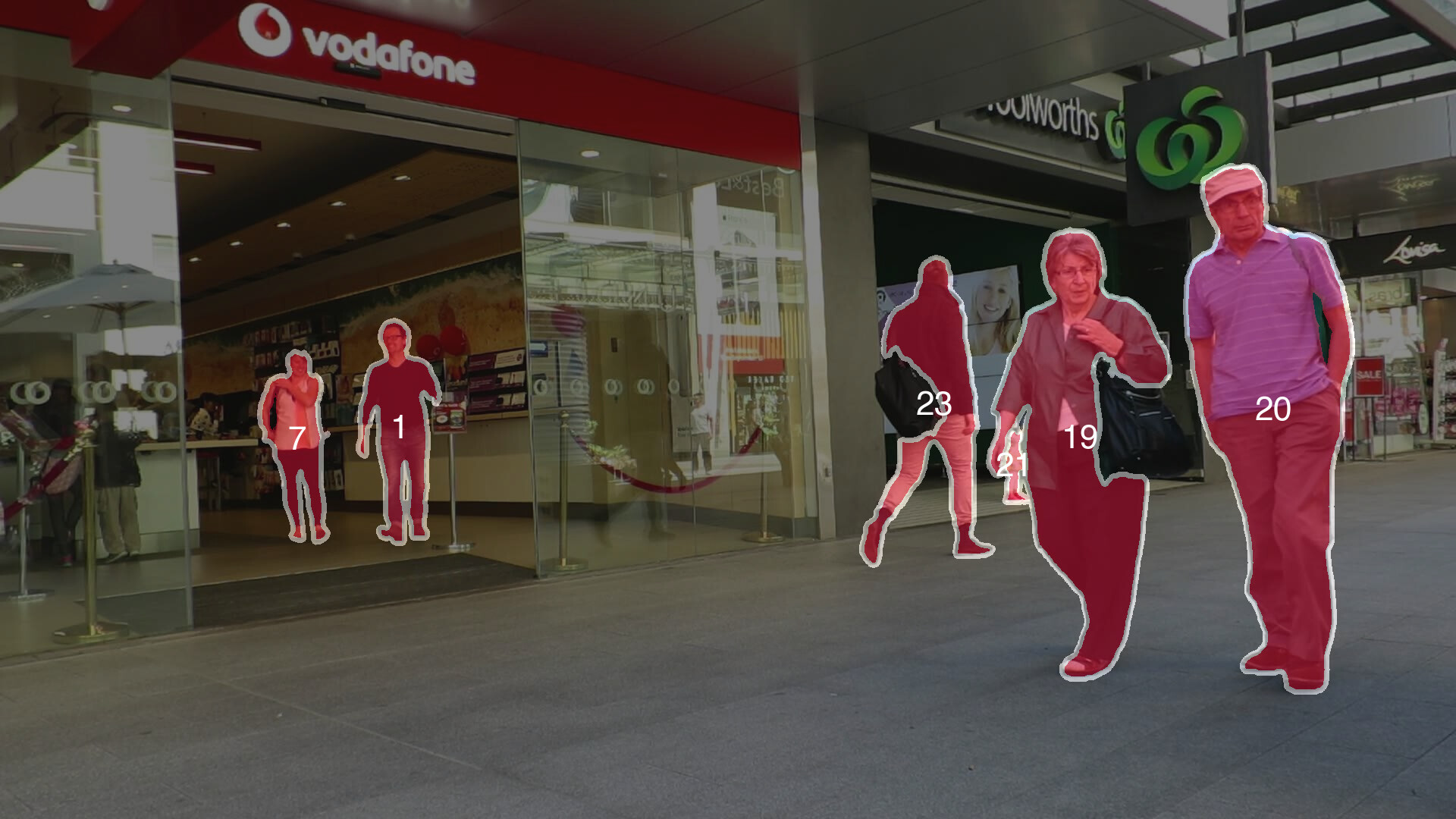} 
& \includegraphics[width=\linewidth]{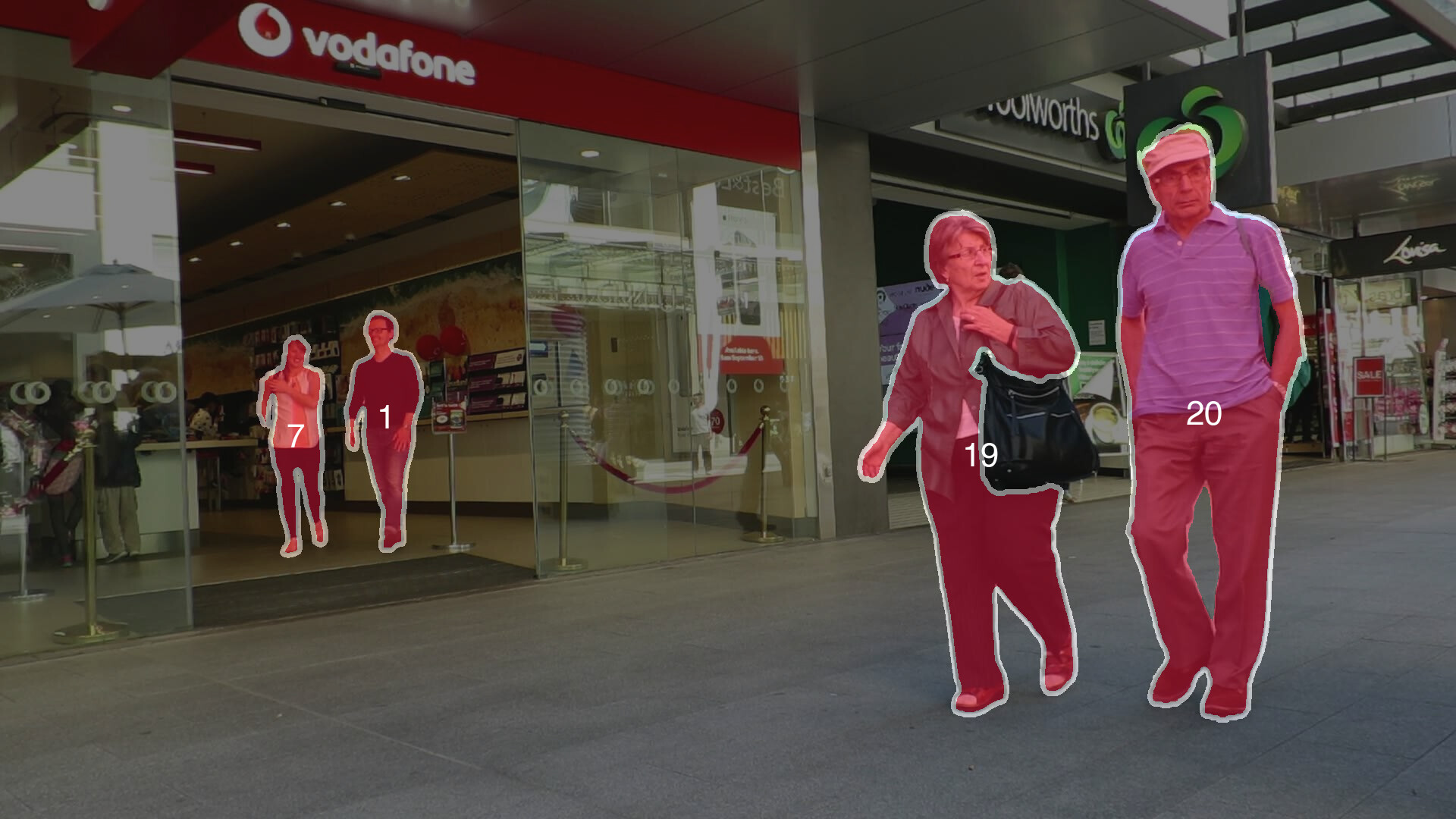} 
& \includegraphics[width=\linewidth]{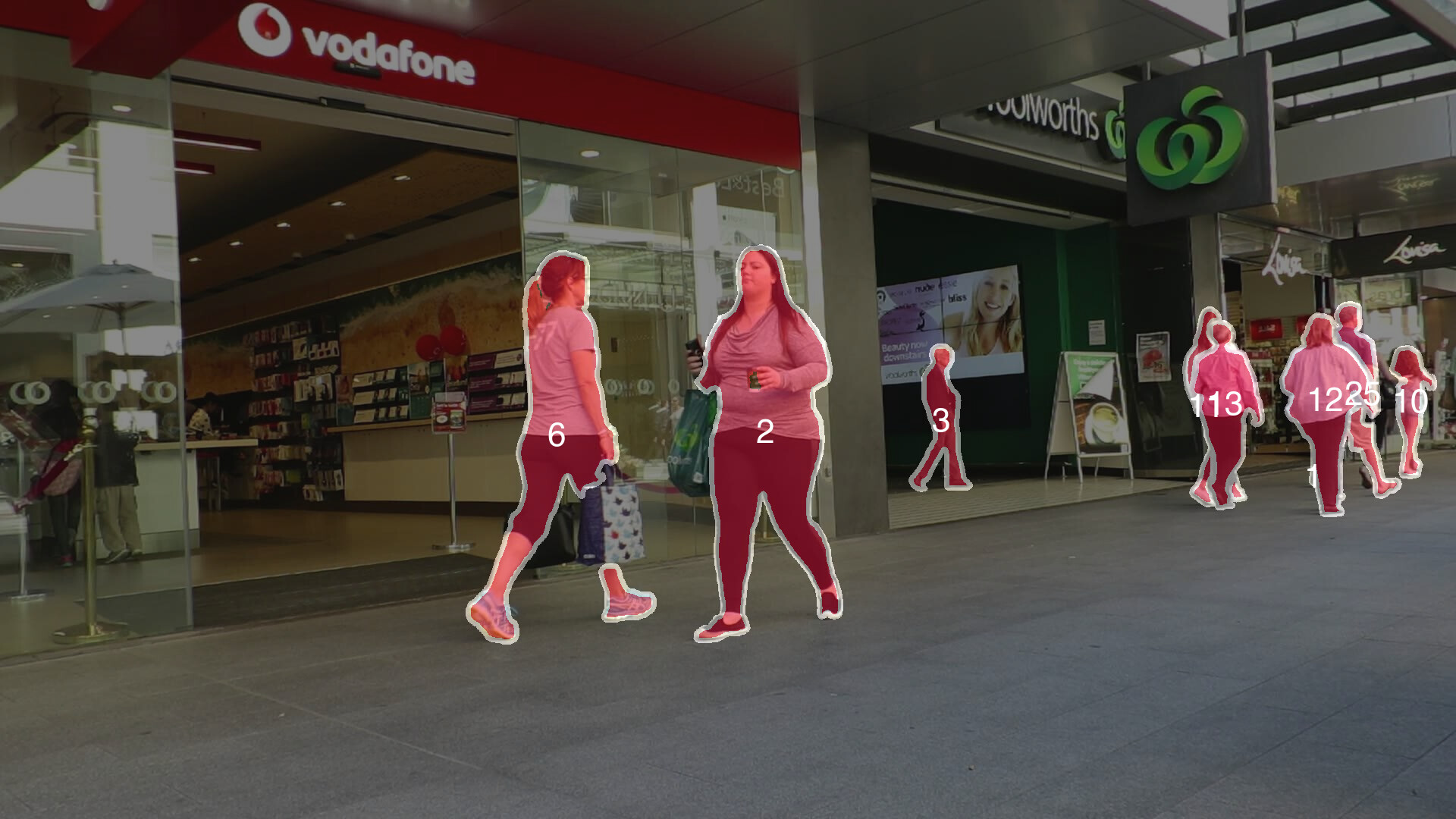}  
& \includegraphics[width=\linewidth]{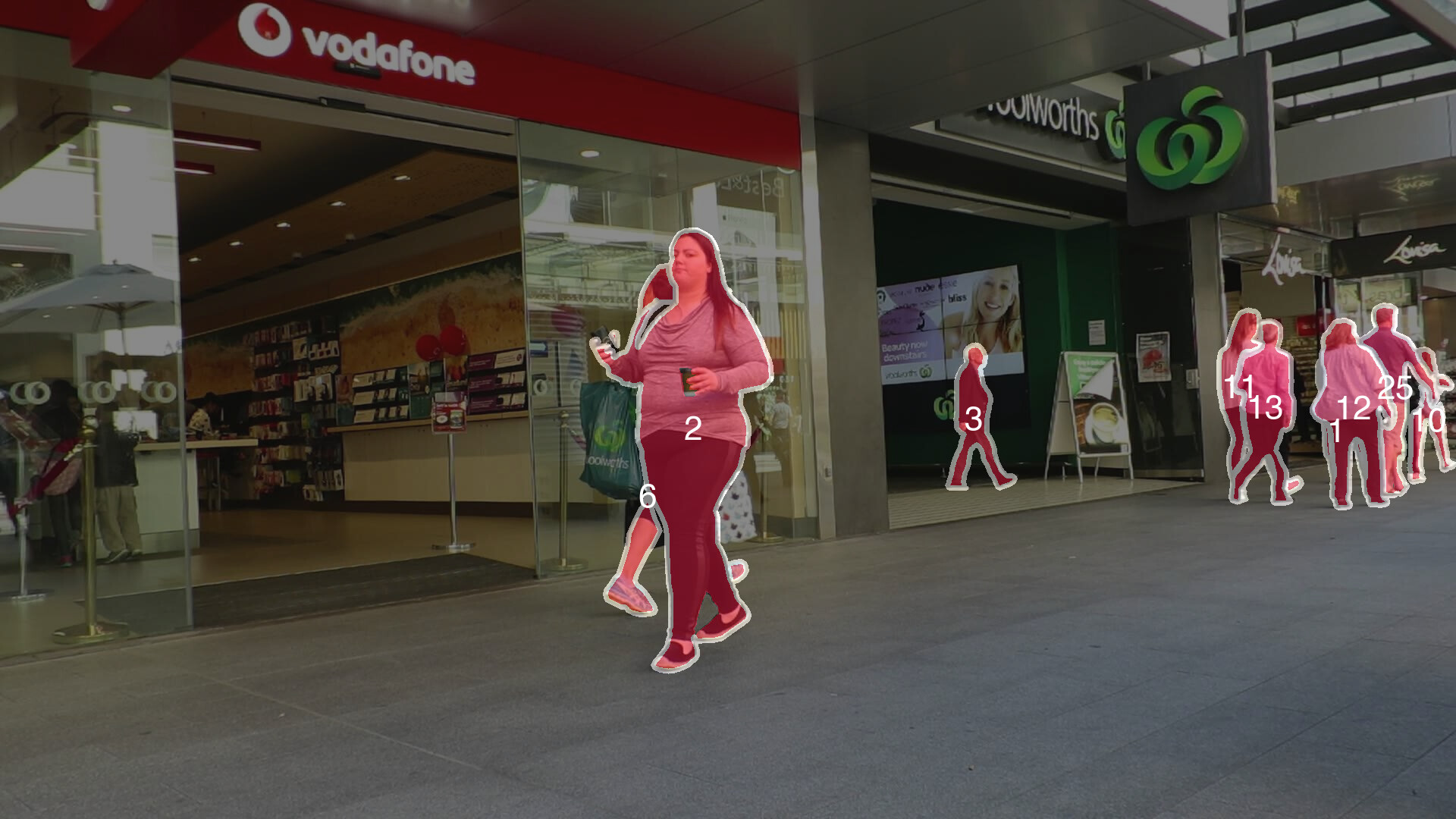} 
& \includegraphics[width=\linewidth]{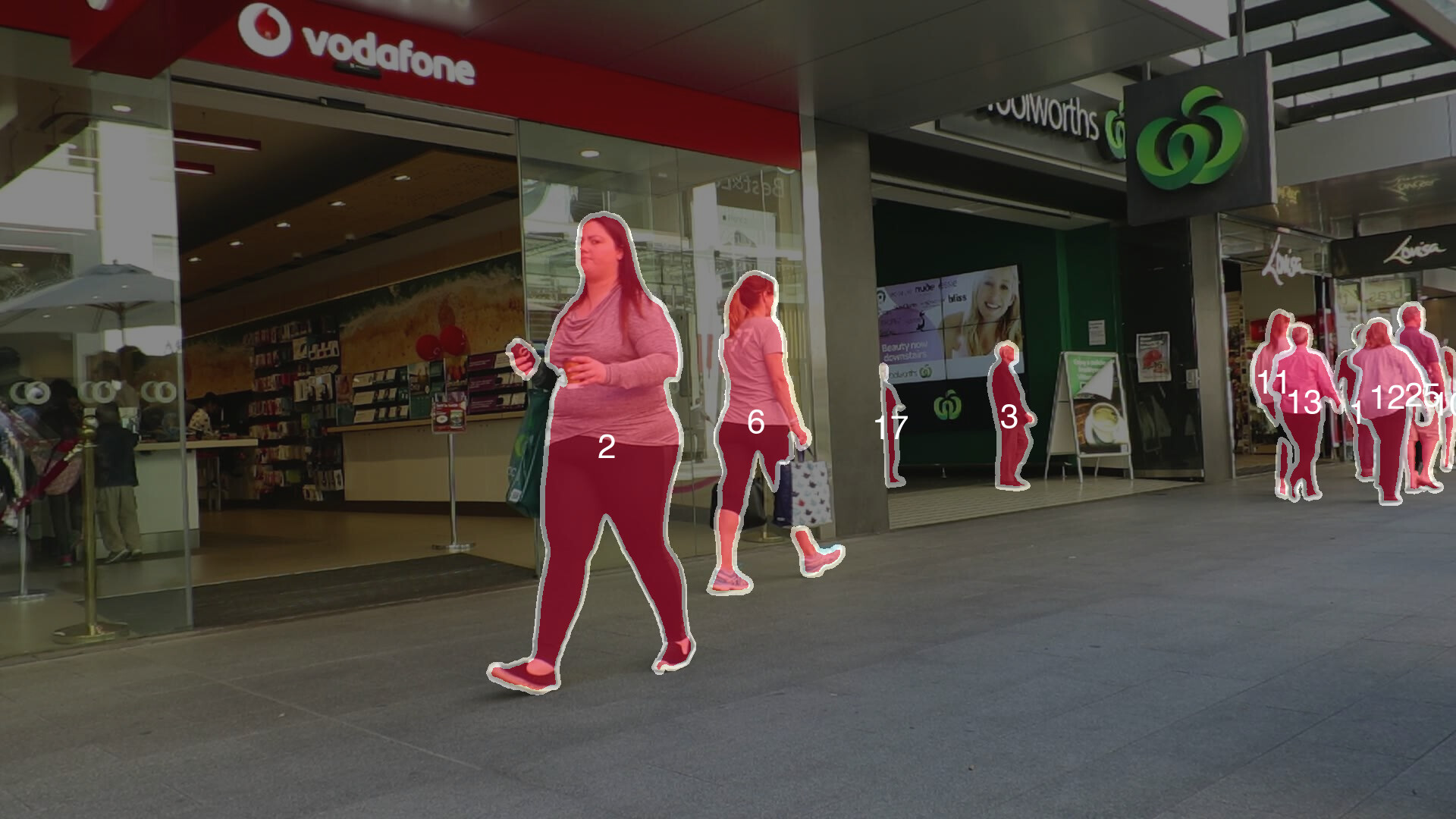}\\

{\rotatebox[origin=lB]{90}{\hspace{-0.3em}\tiny{\shortstack{DepthG~+\\[-3pt]VideoCutLER}}}}
& \includegraphics[width=\linewidth]{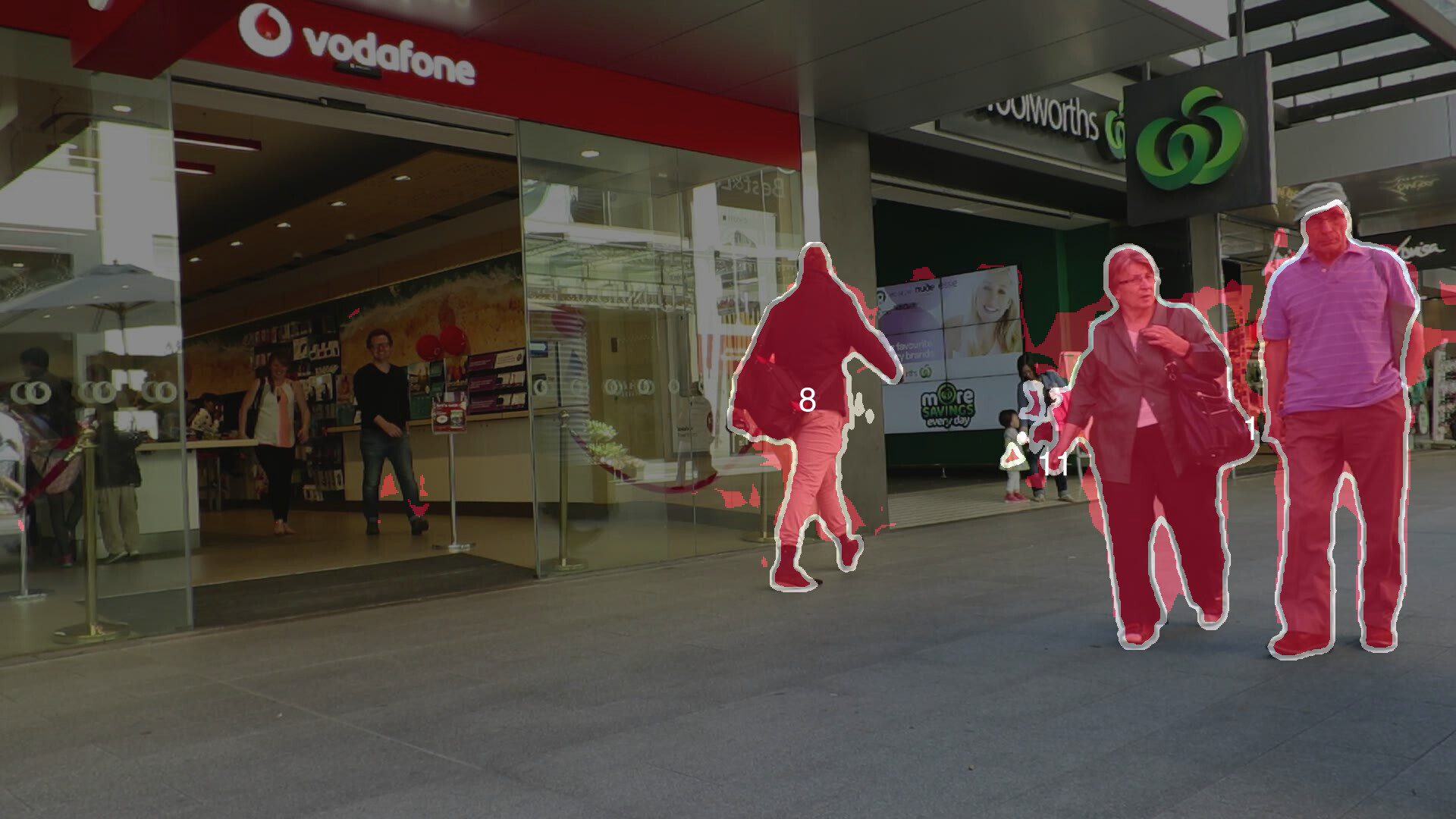} 
& \includegraphics[width=\linewidth]{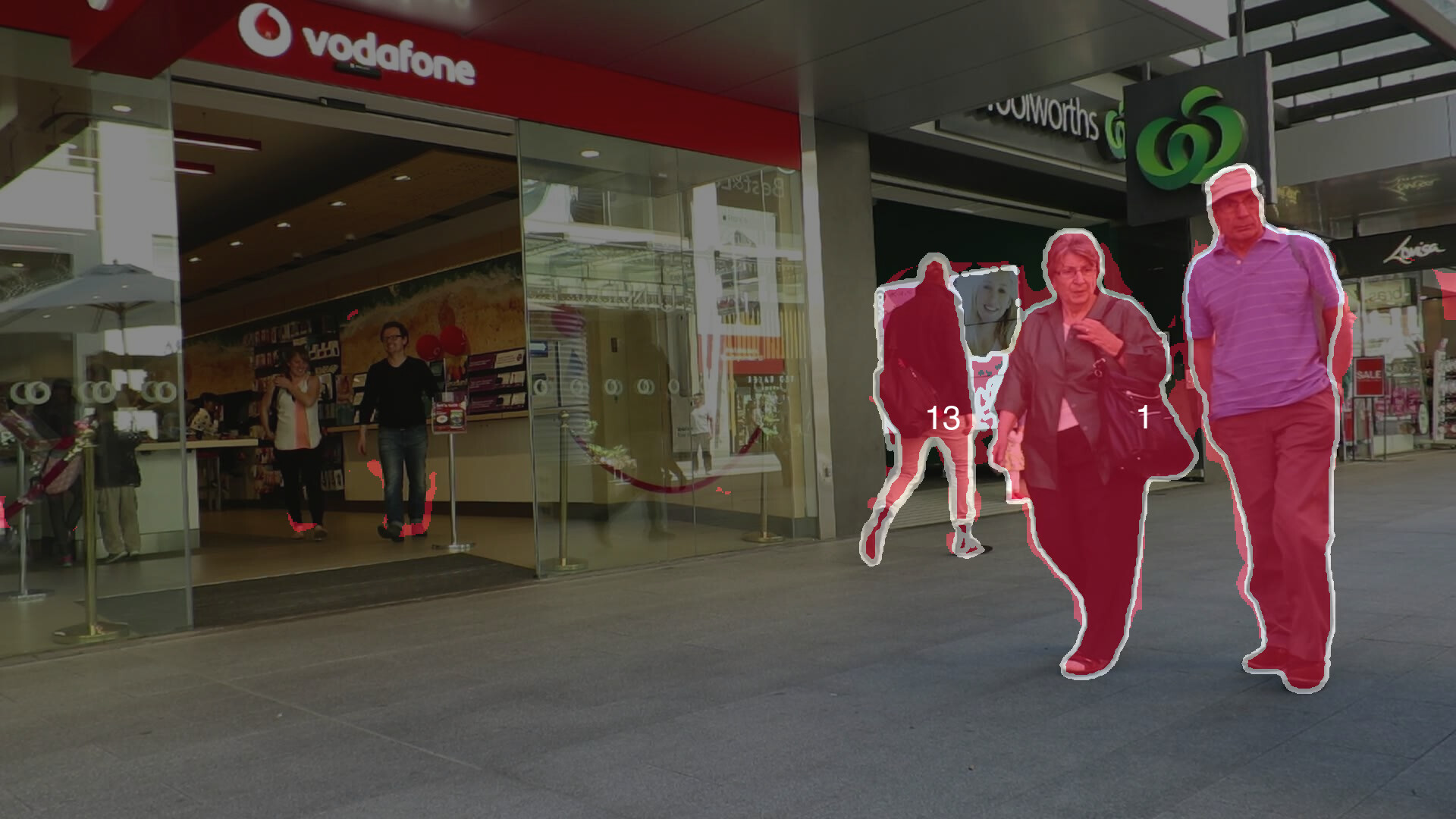} 
& \includegraphics[width=\linewidth]{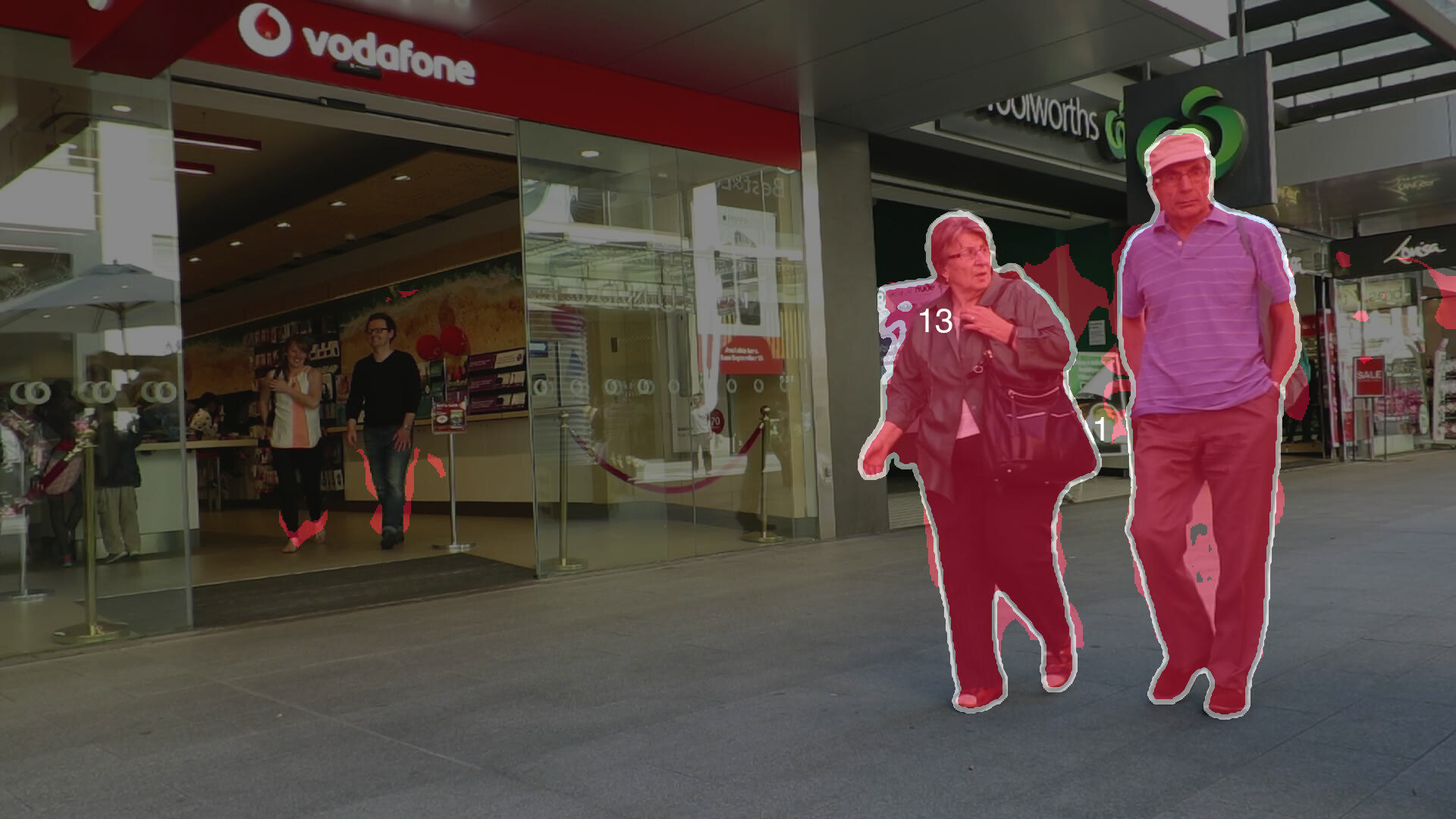} 
& \includegraphics[width=\linewidth]{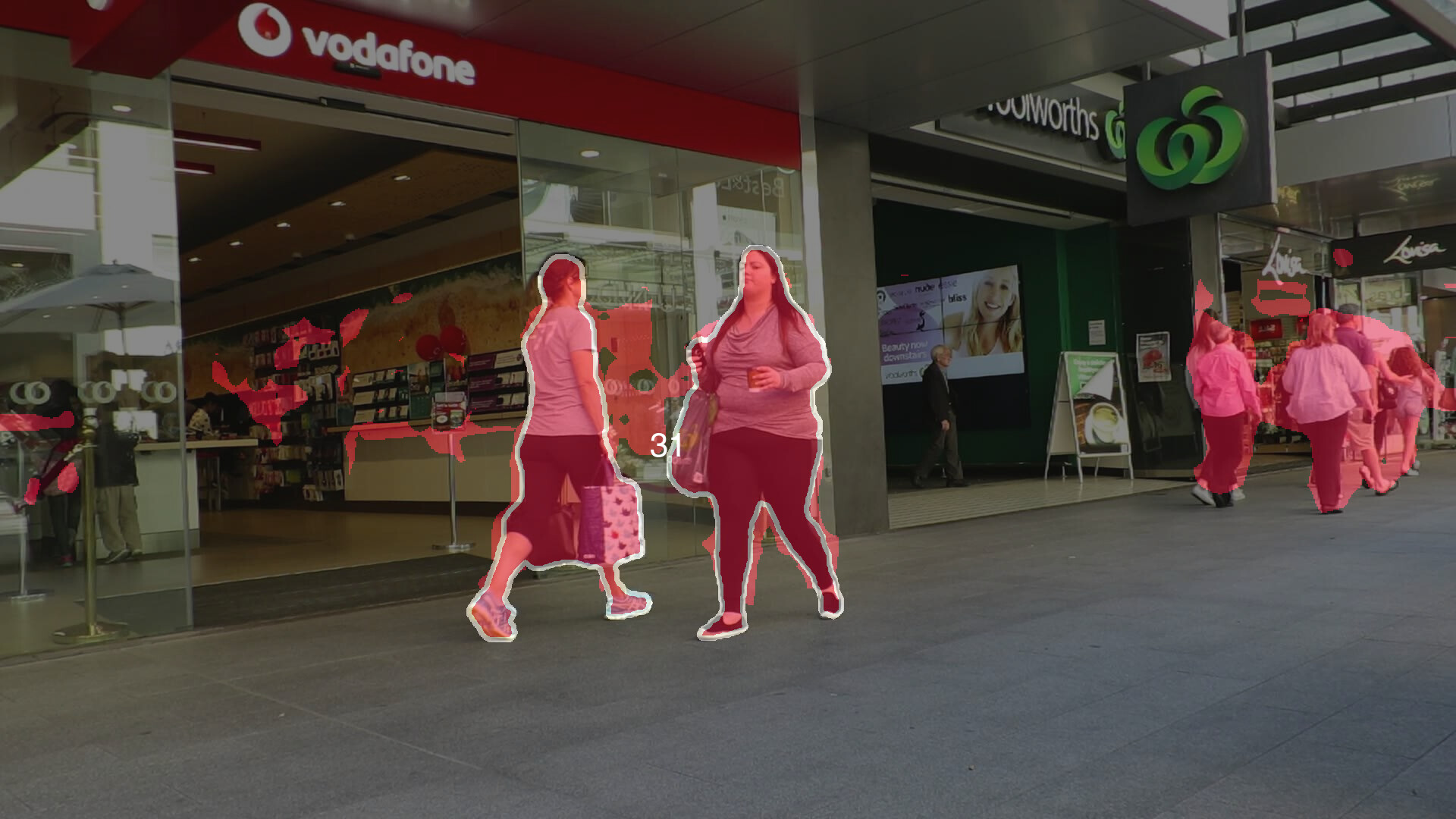}   
& \includegraphics[width=\linewidth]{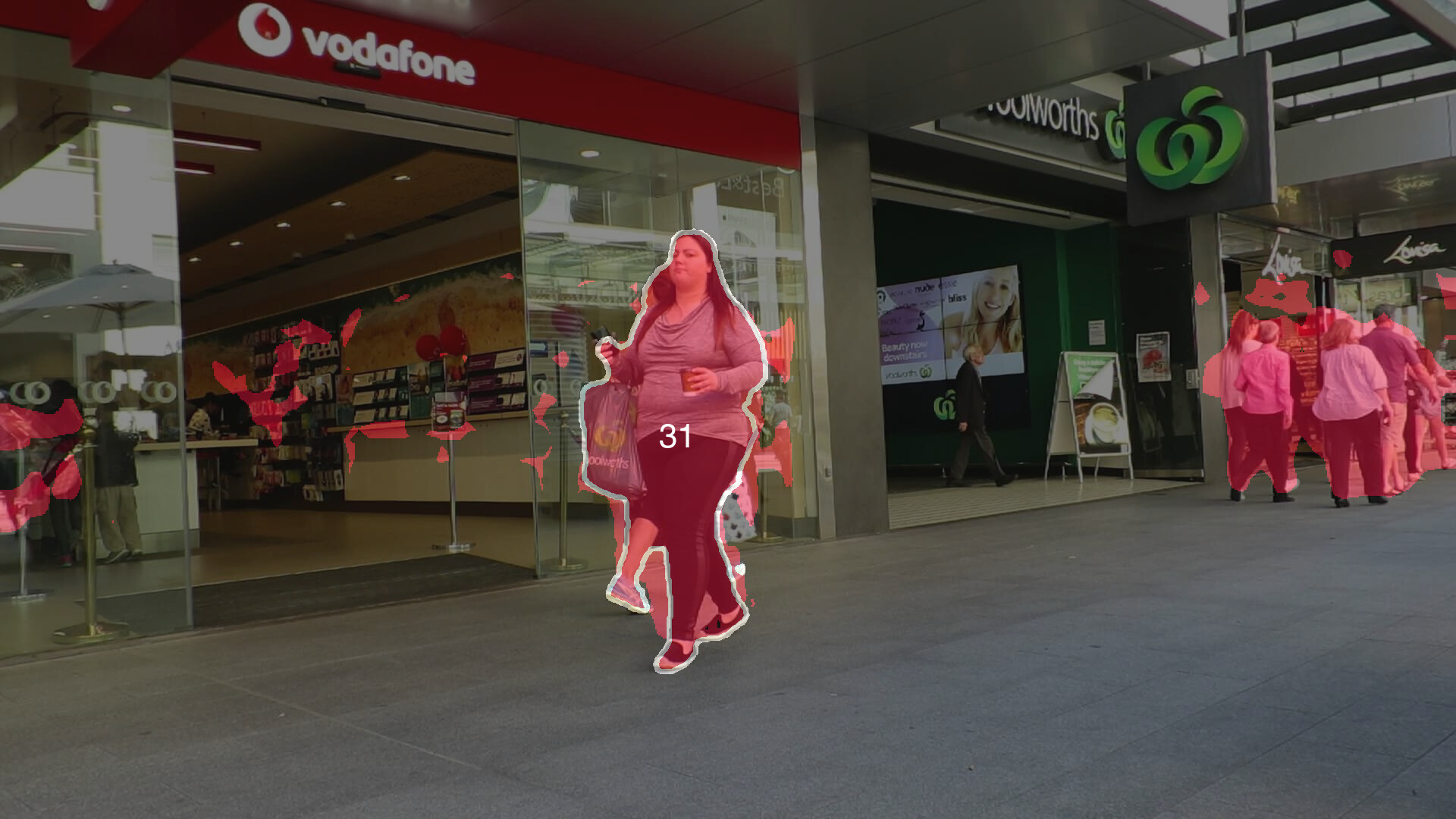}  
& \includegraphics[width=\linewidth]{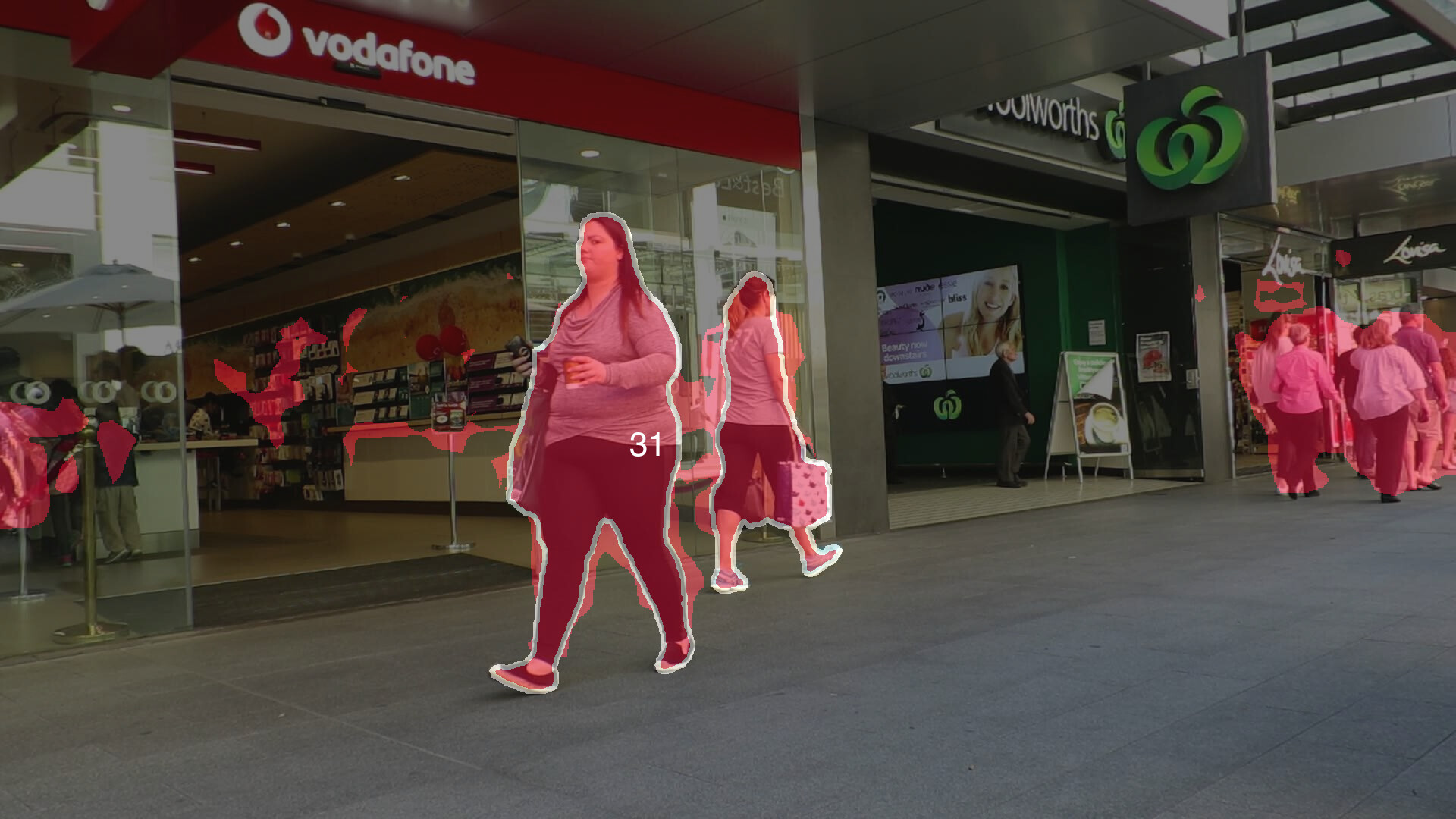} \\

\rotatebox[origin=lB]{90}{\hspace{-0.1em}\tiny{U2Seg~+~SORT}}
& \includegraphics[width=\linewidth]{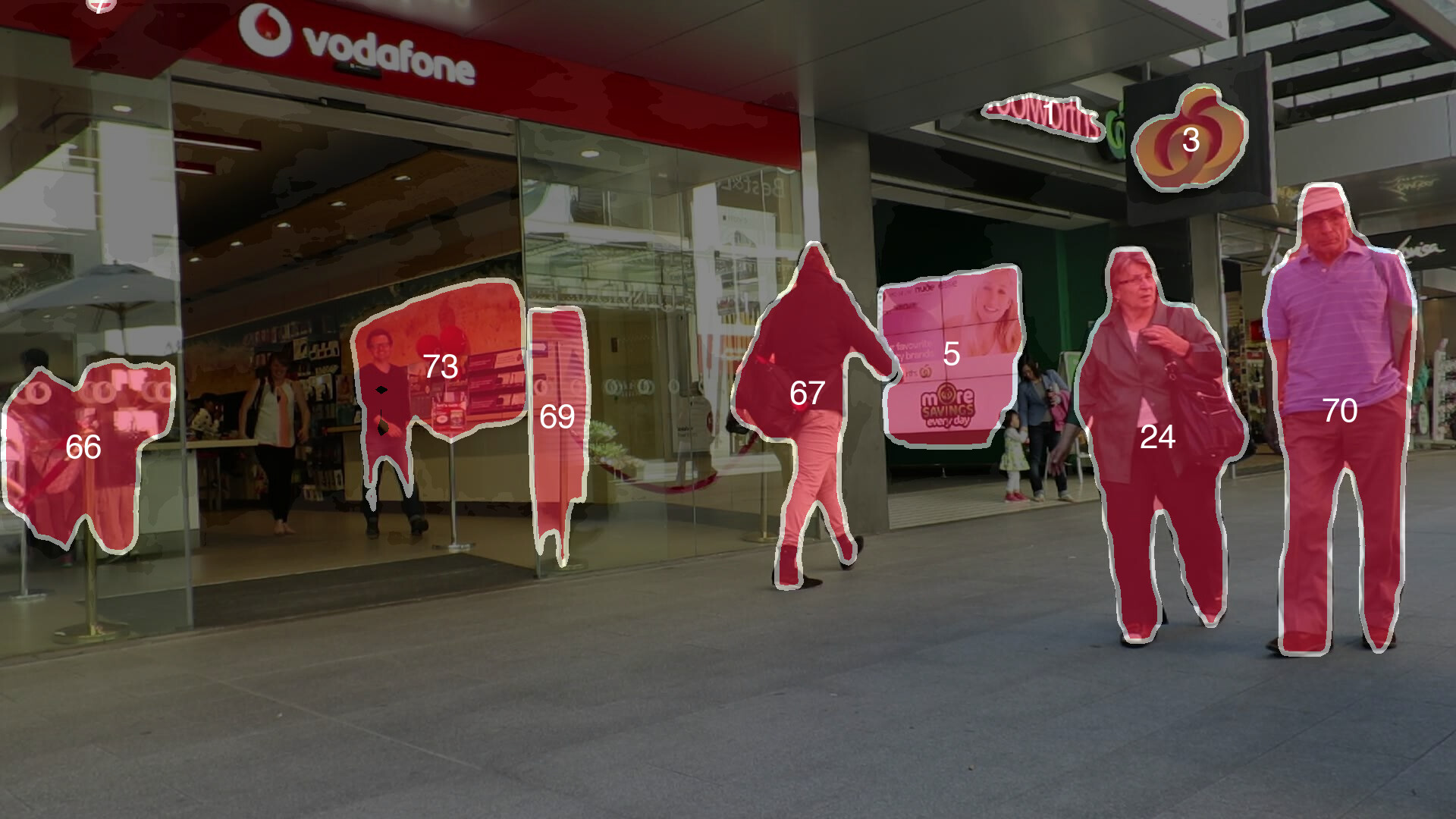} 
& \includegraphics[width=\linewidth]{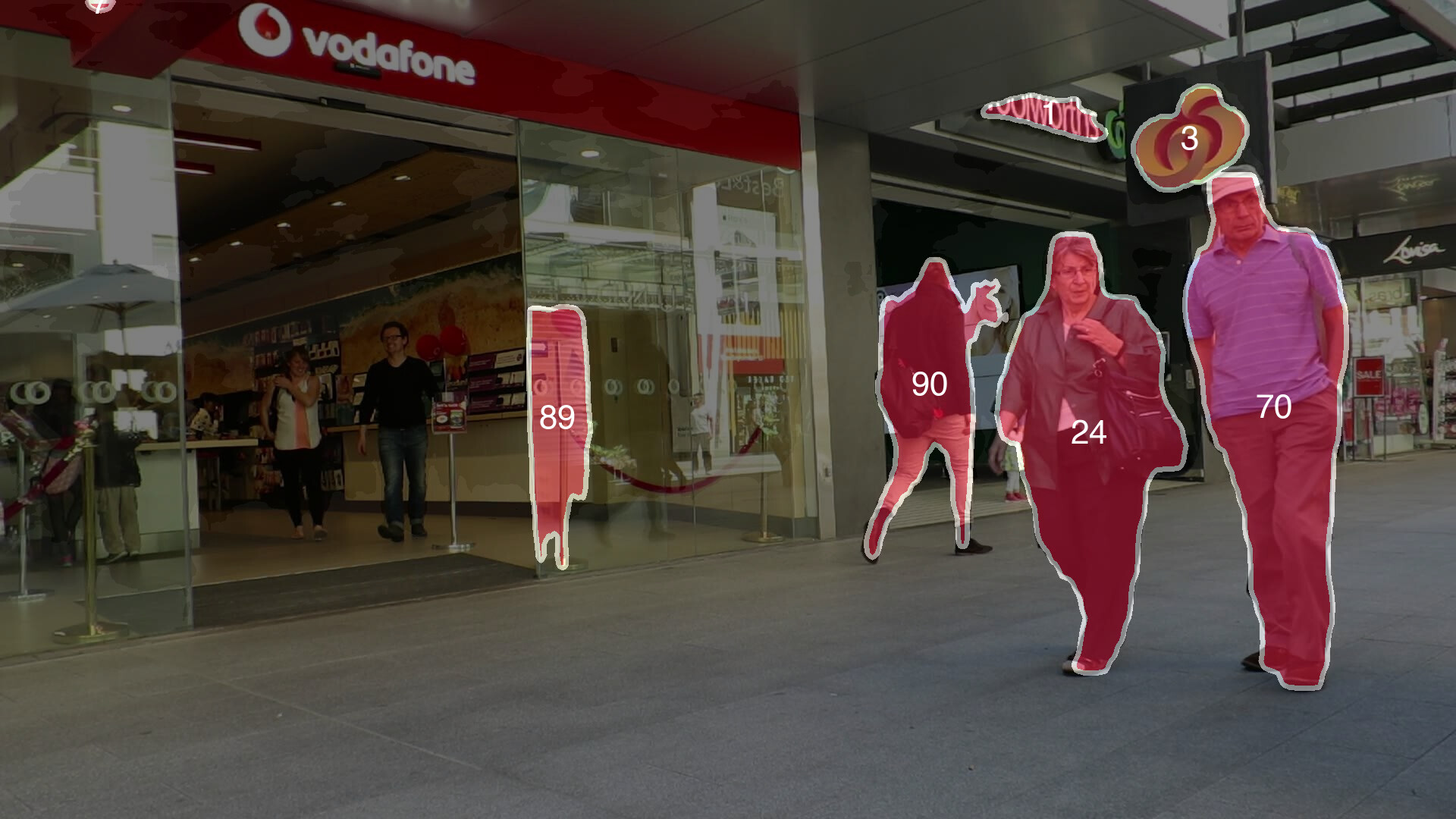} 
& \includegraphics[width=\linewidth]{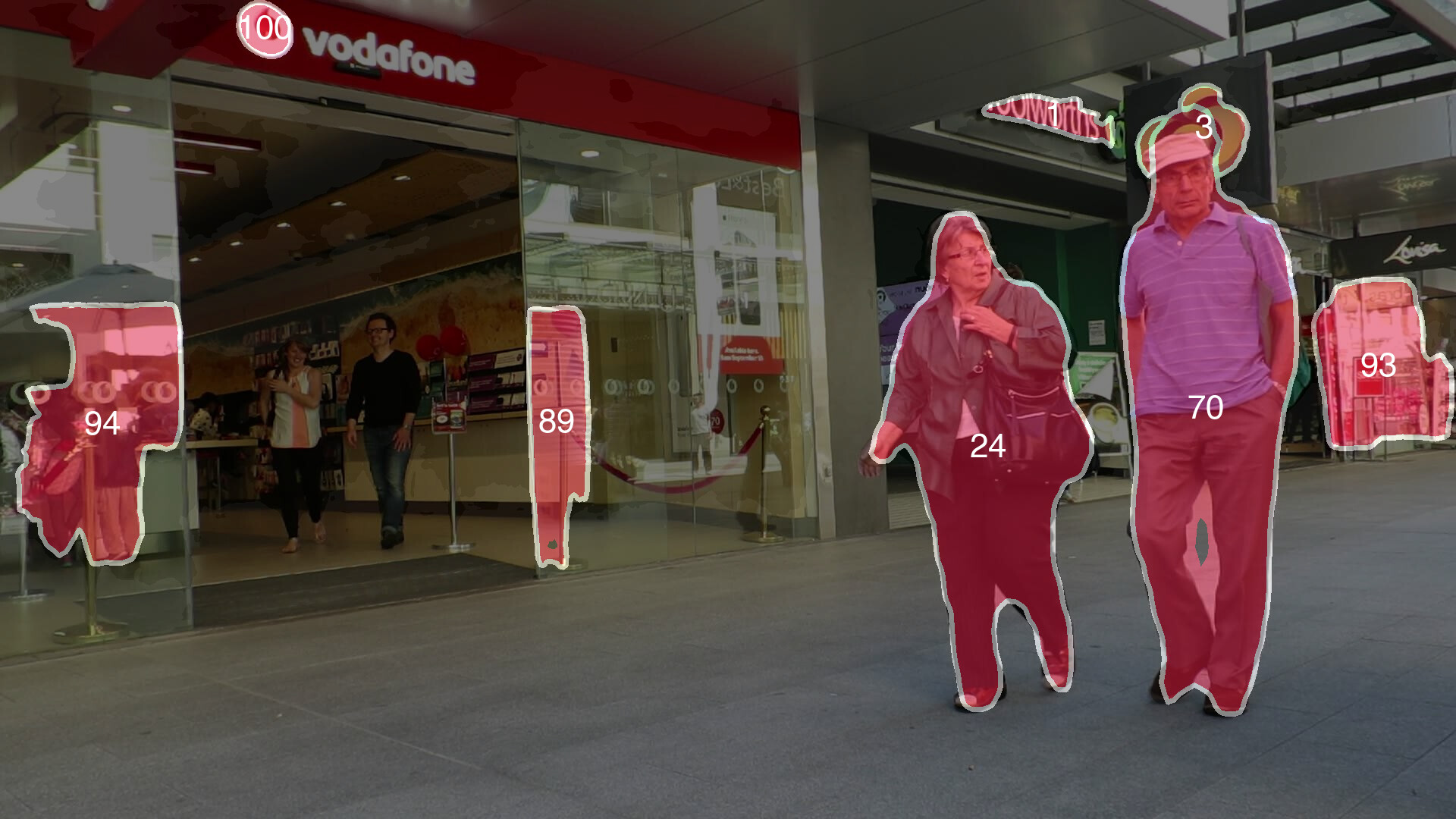} 
& \includegraphics[width=\linewidth]{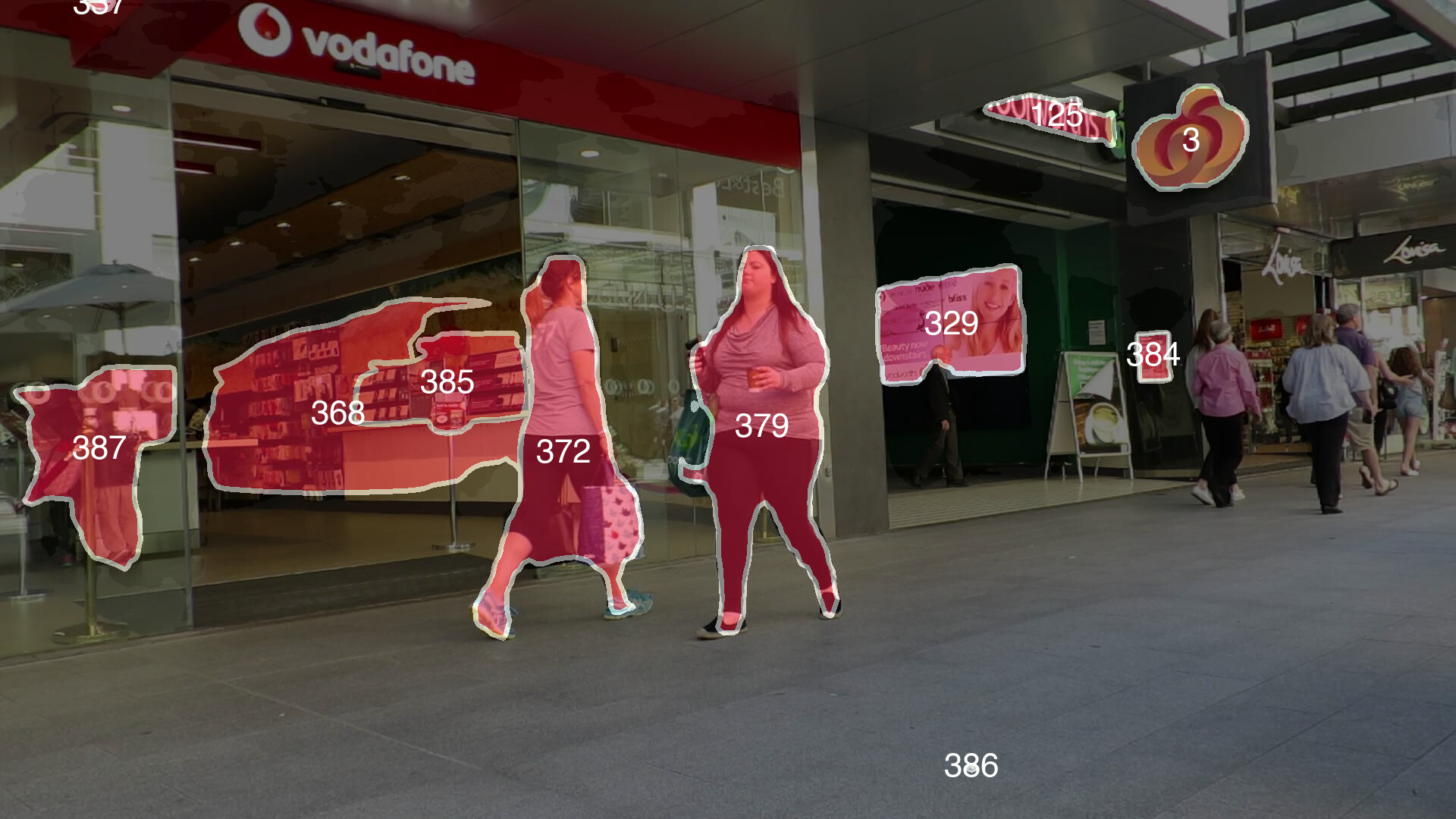}   
& \includegraphics[width=\linewidth]{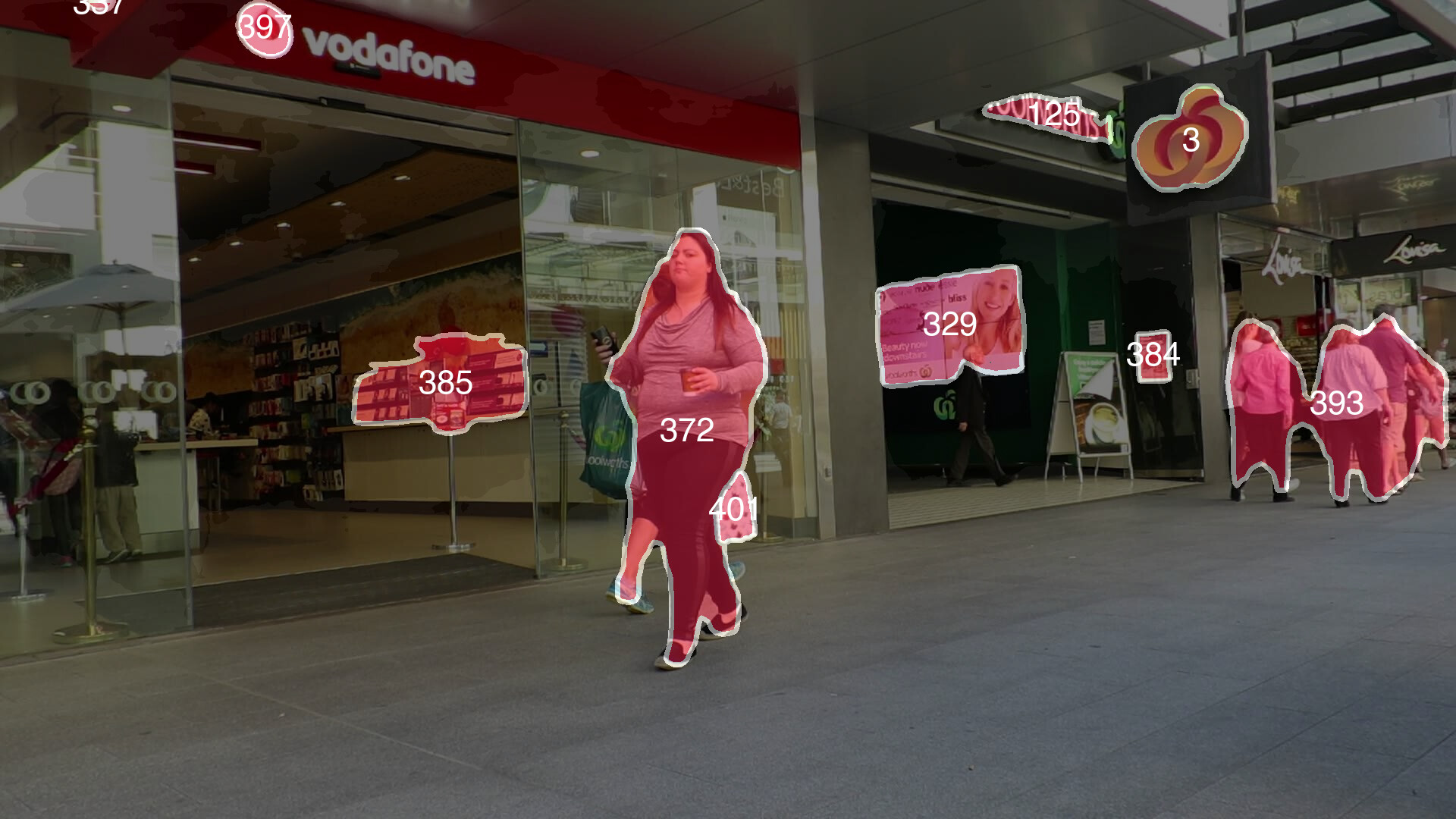}   
& \includegraphics[width=\linewidth]{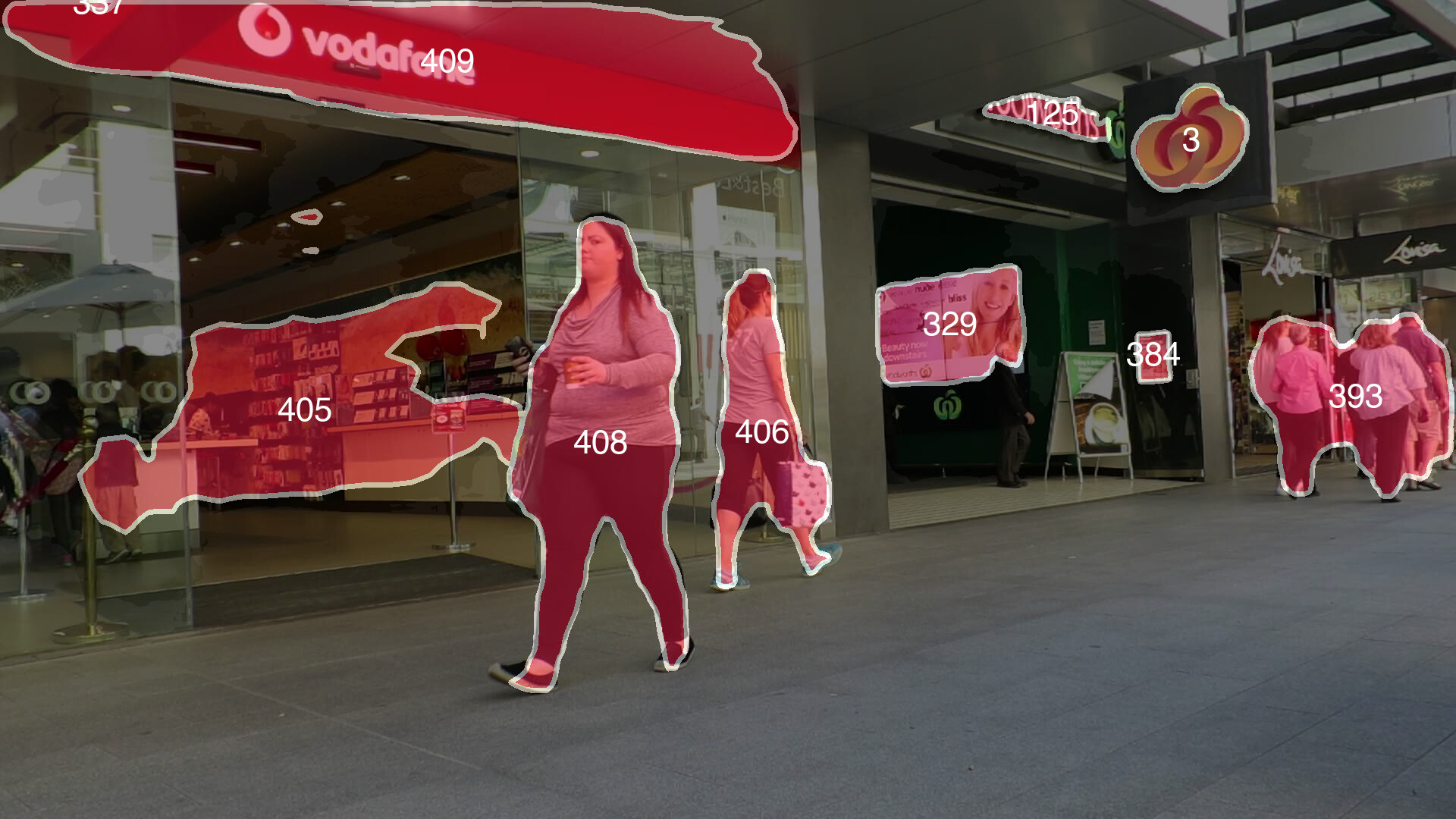}  \\

\rotatebox[origin=lB]{90}{\hspace{-0.1em}\tiny{CUPS~+~SORT}}
& \includegraphics[width=\linewidth]{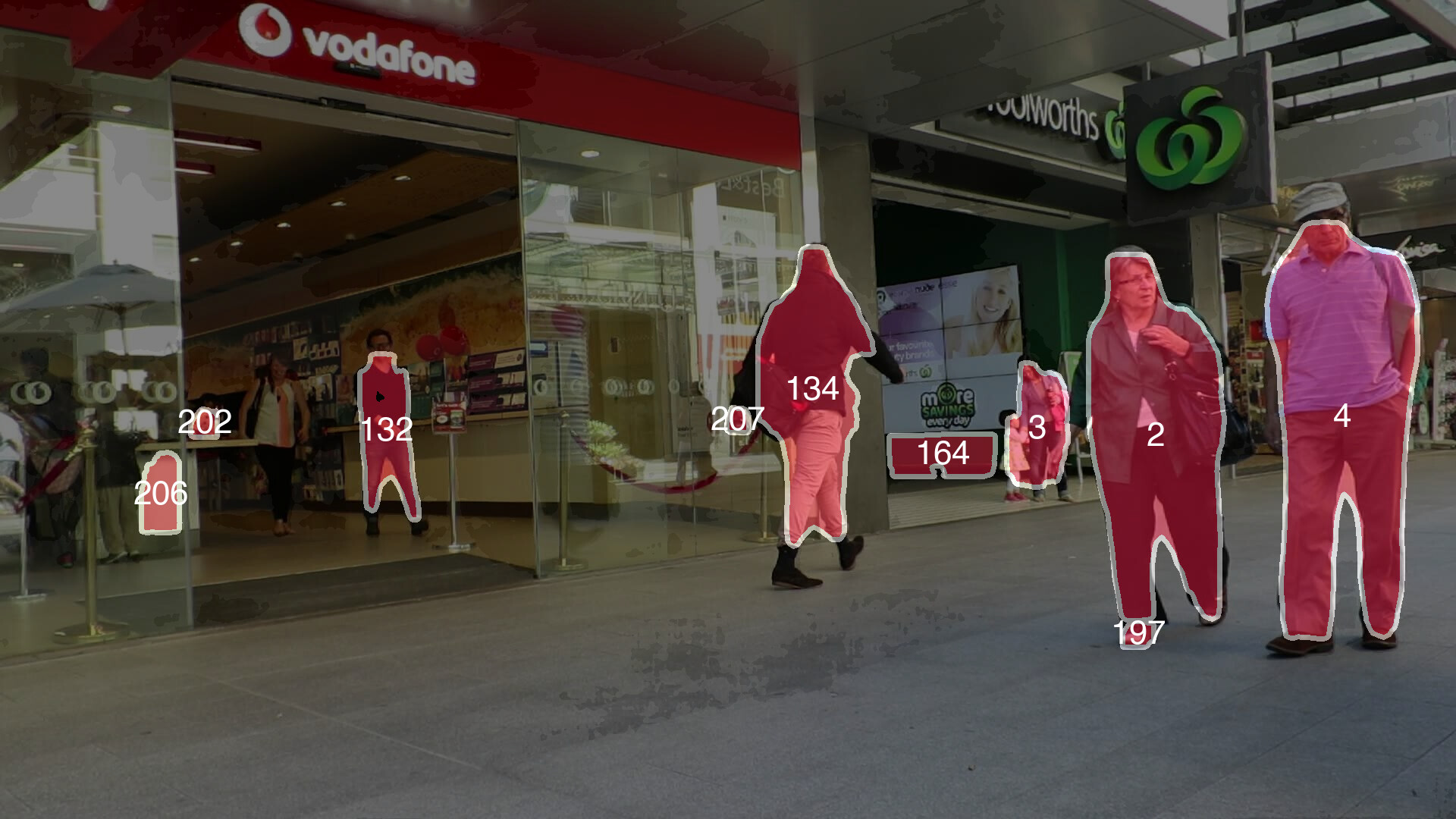} 
& \includegraphics[width=\linewidth]{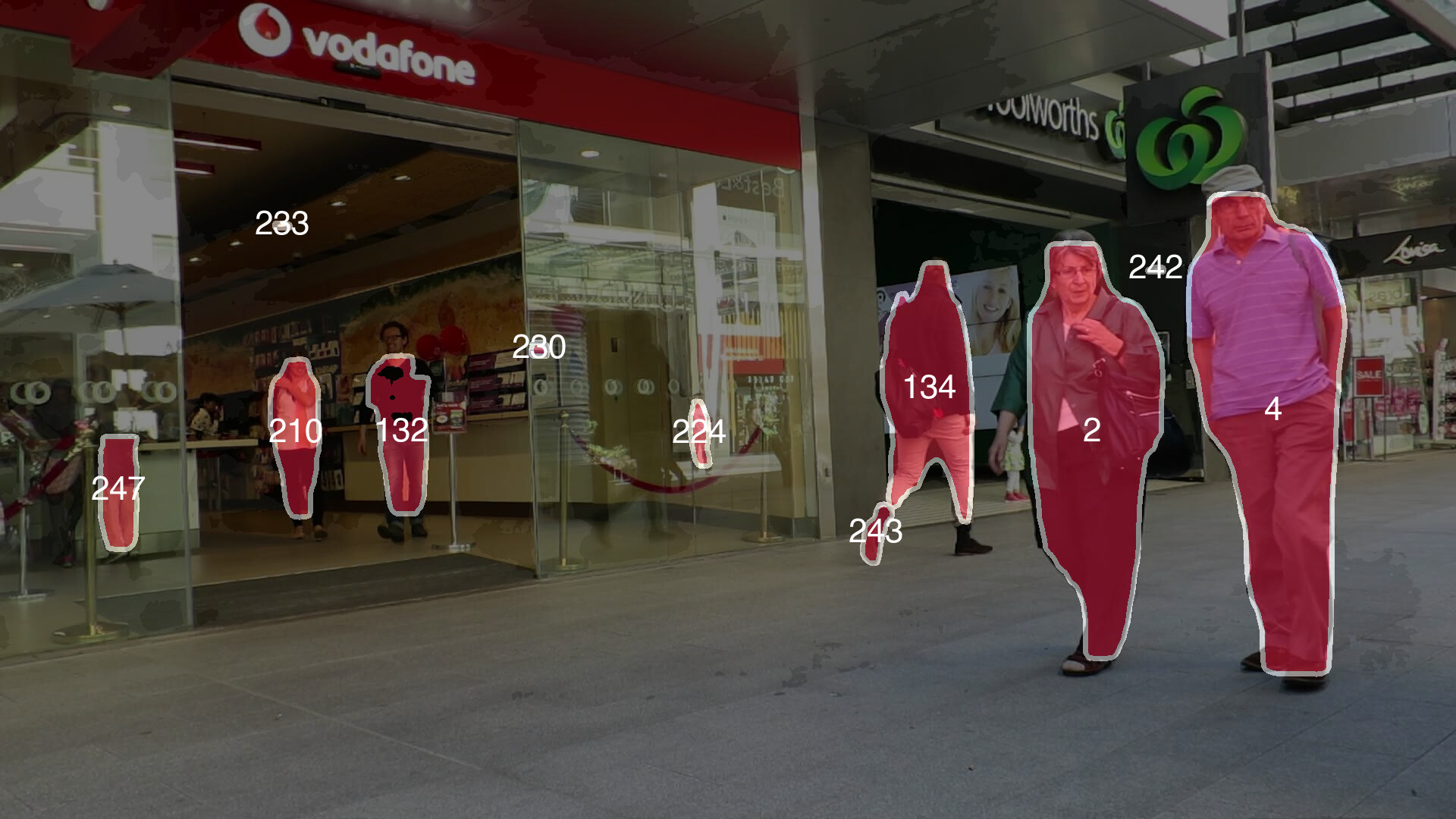}
& \includegraphics[width=\linewidth]{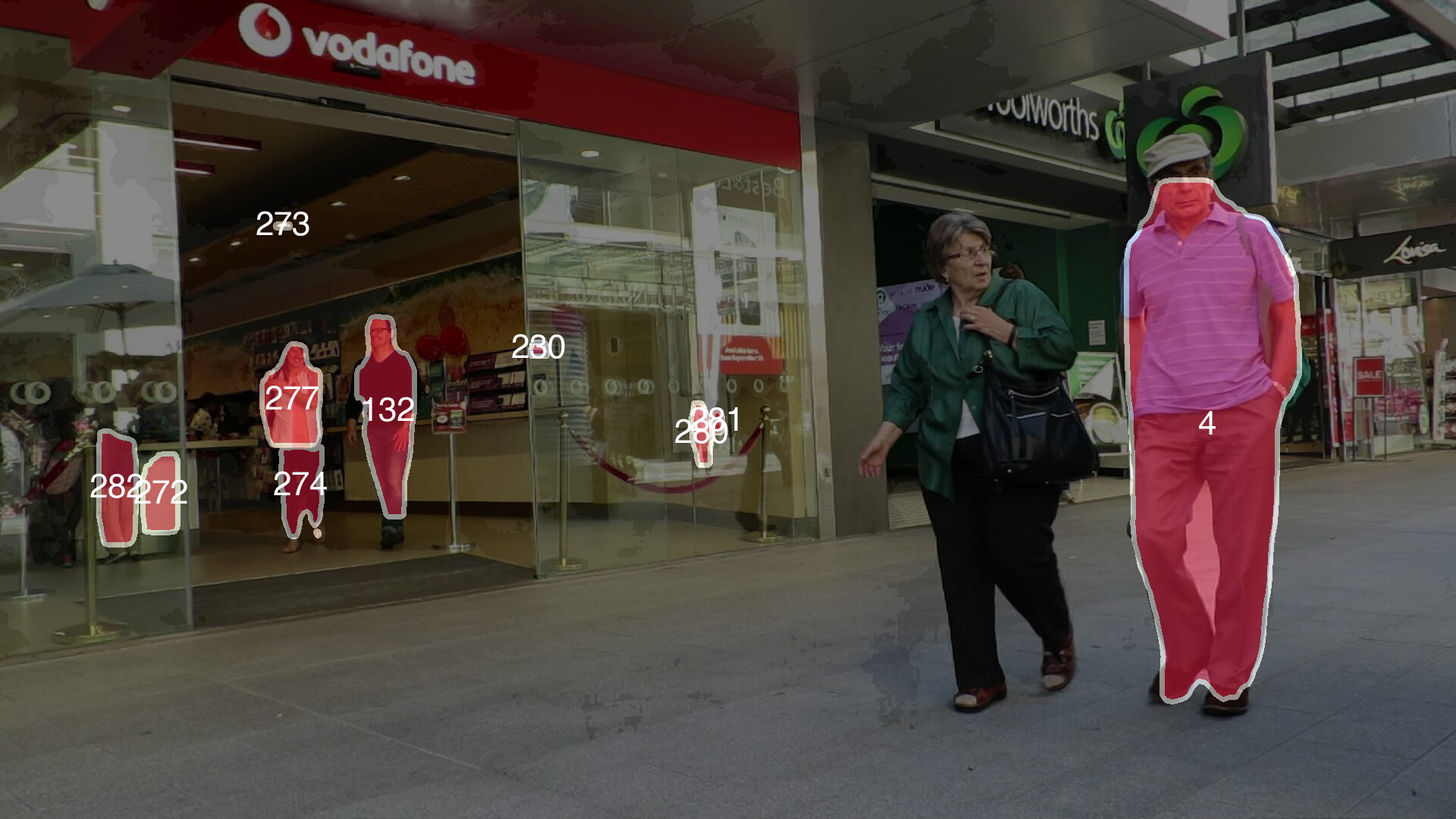} 
& \includegraphics[width=\linewidth]{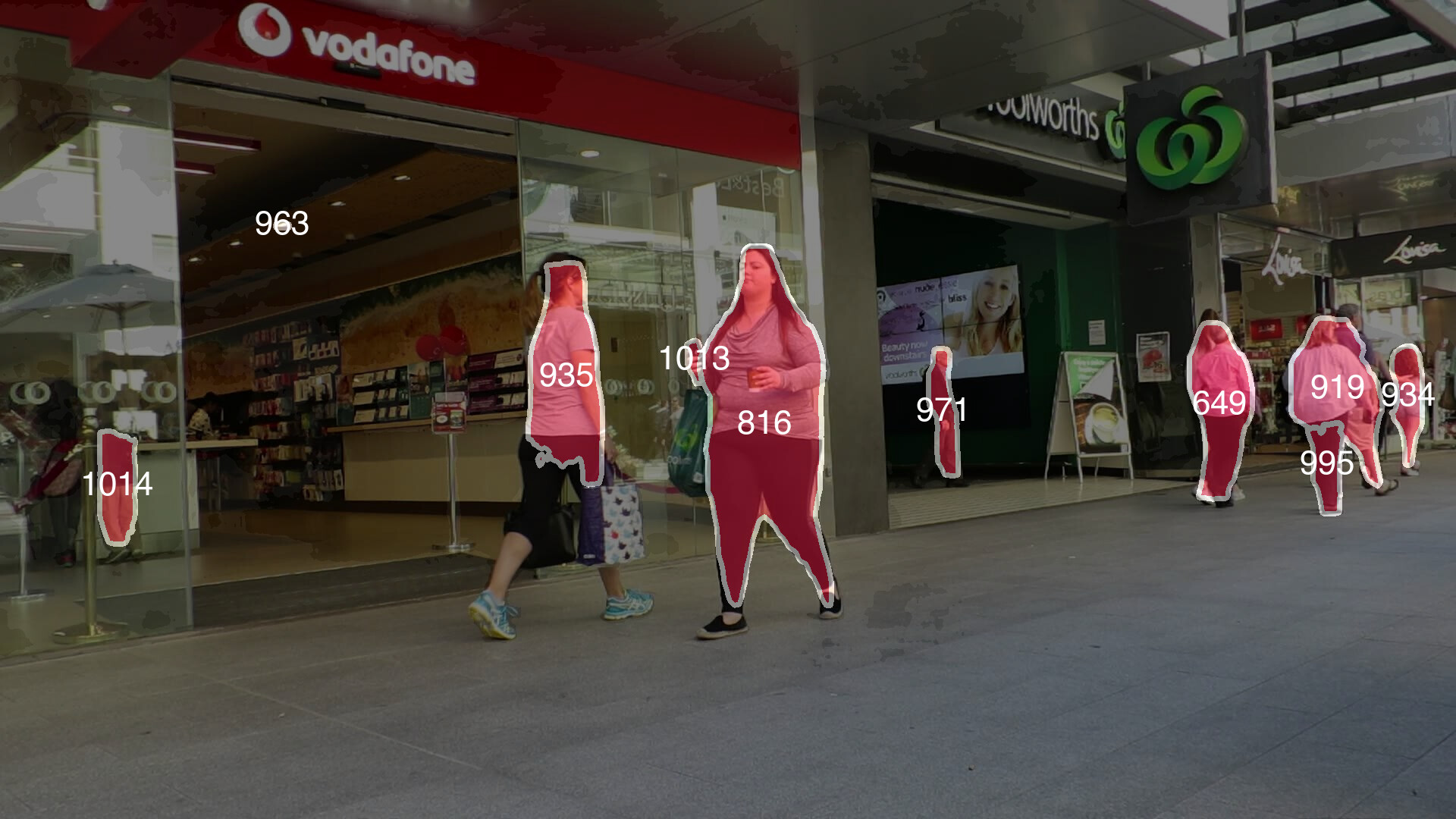}  
& \includegraphics[width=\linewidth]{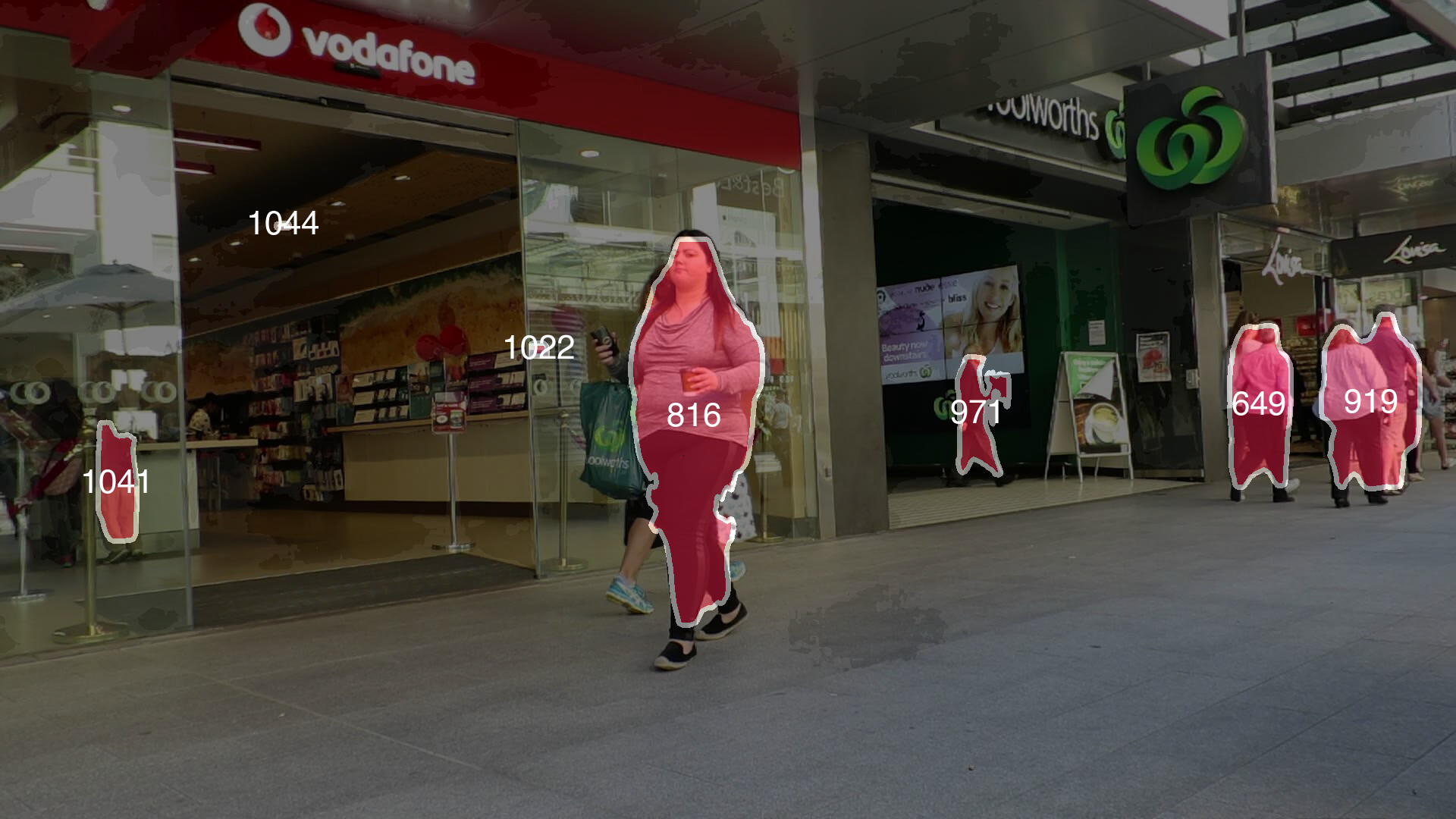}
& \includegraphics[width=\linewidth]{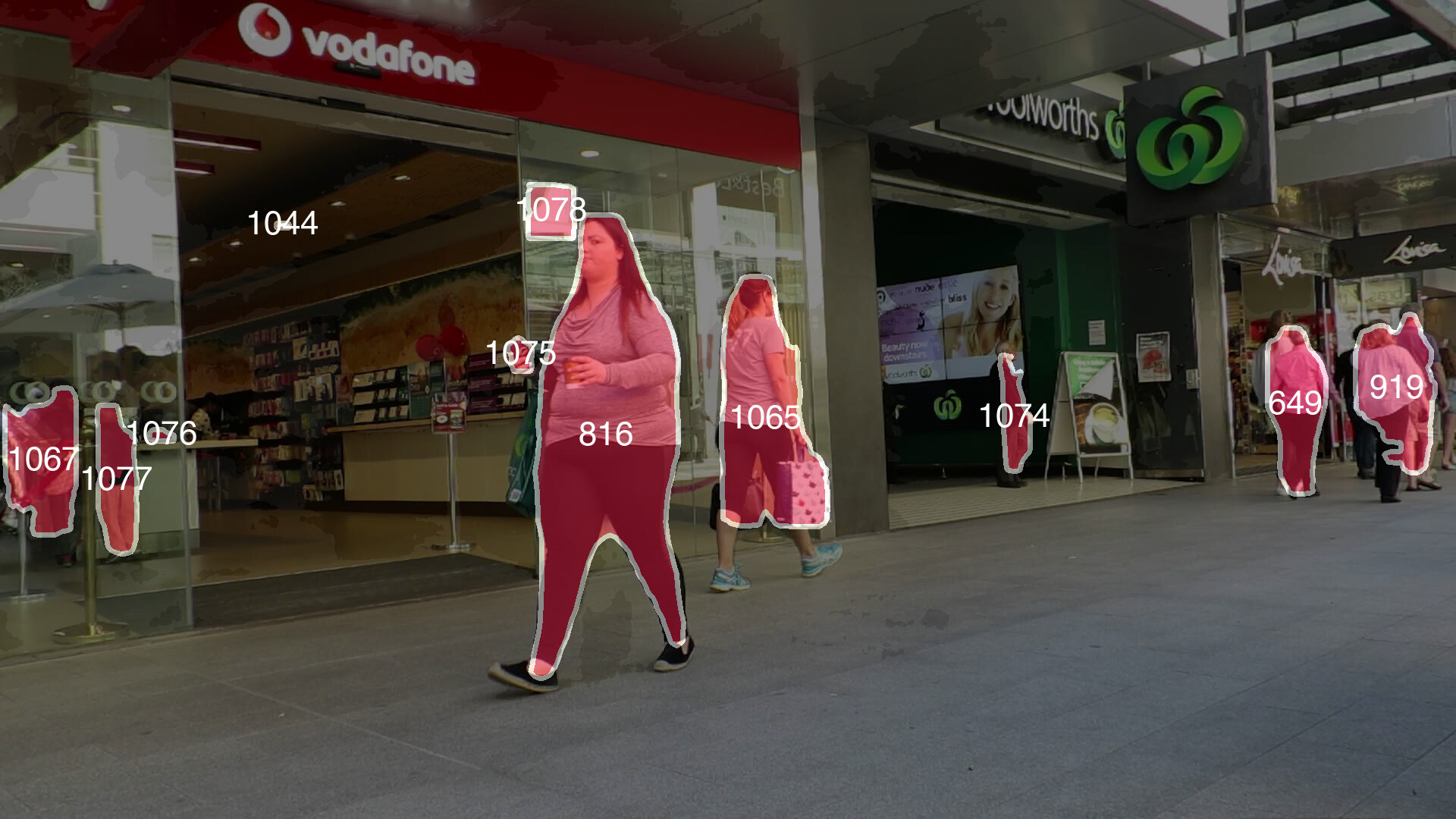}\\

\rotatebox[origin=lB]{90}{\hspace{-0.1em}\tiny{CUPS\textsuperscript{$\dagger$}\!+\phantom{j}SORT}}
& \includegraphics[width=\linewidth]{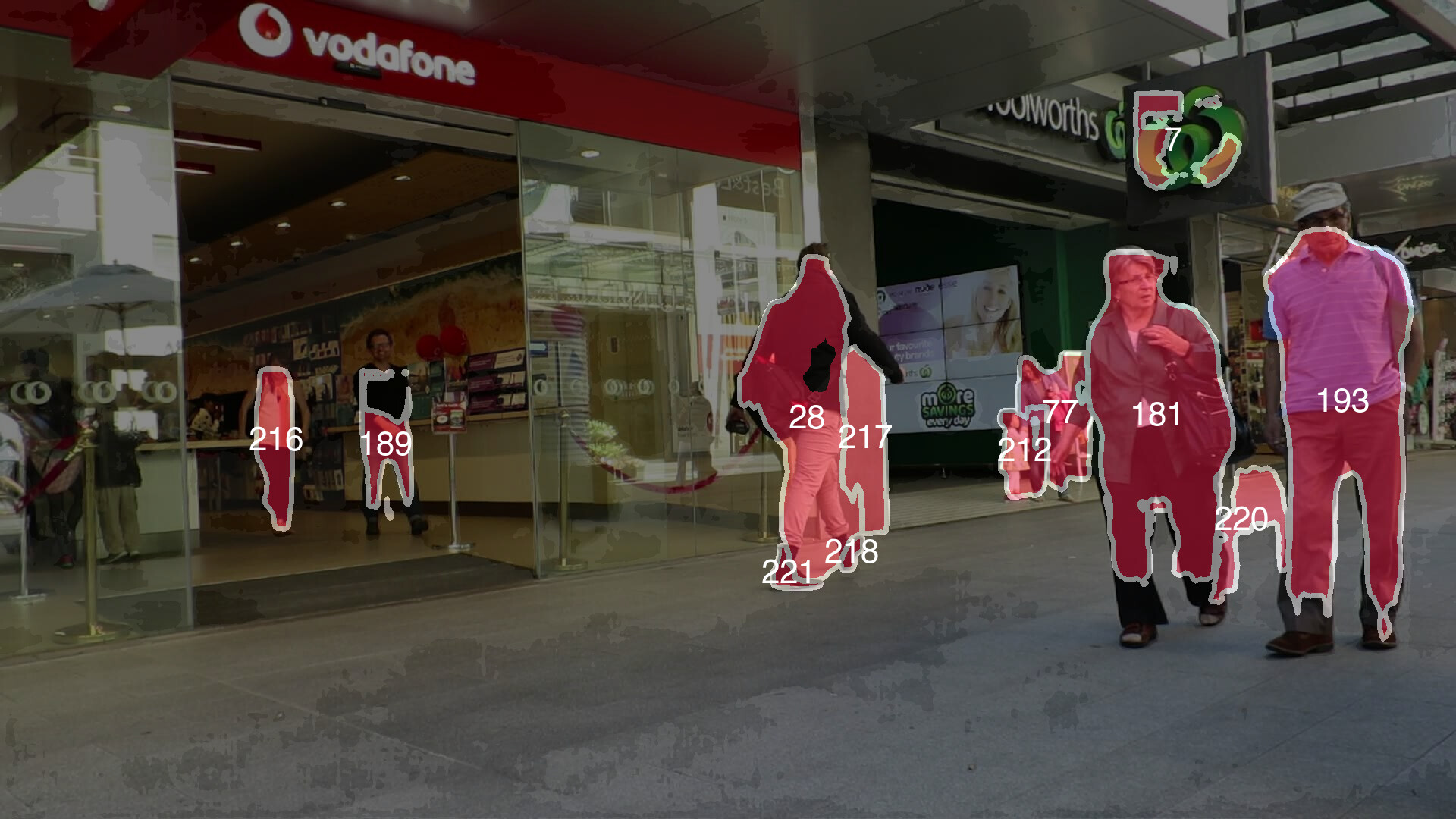} 
& \includegraphics[width=\linewidth]{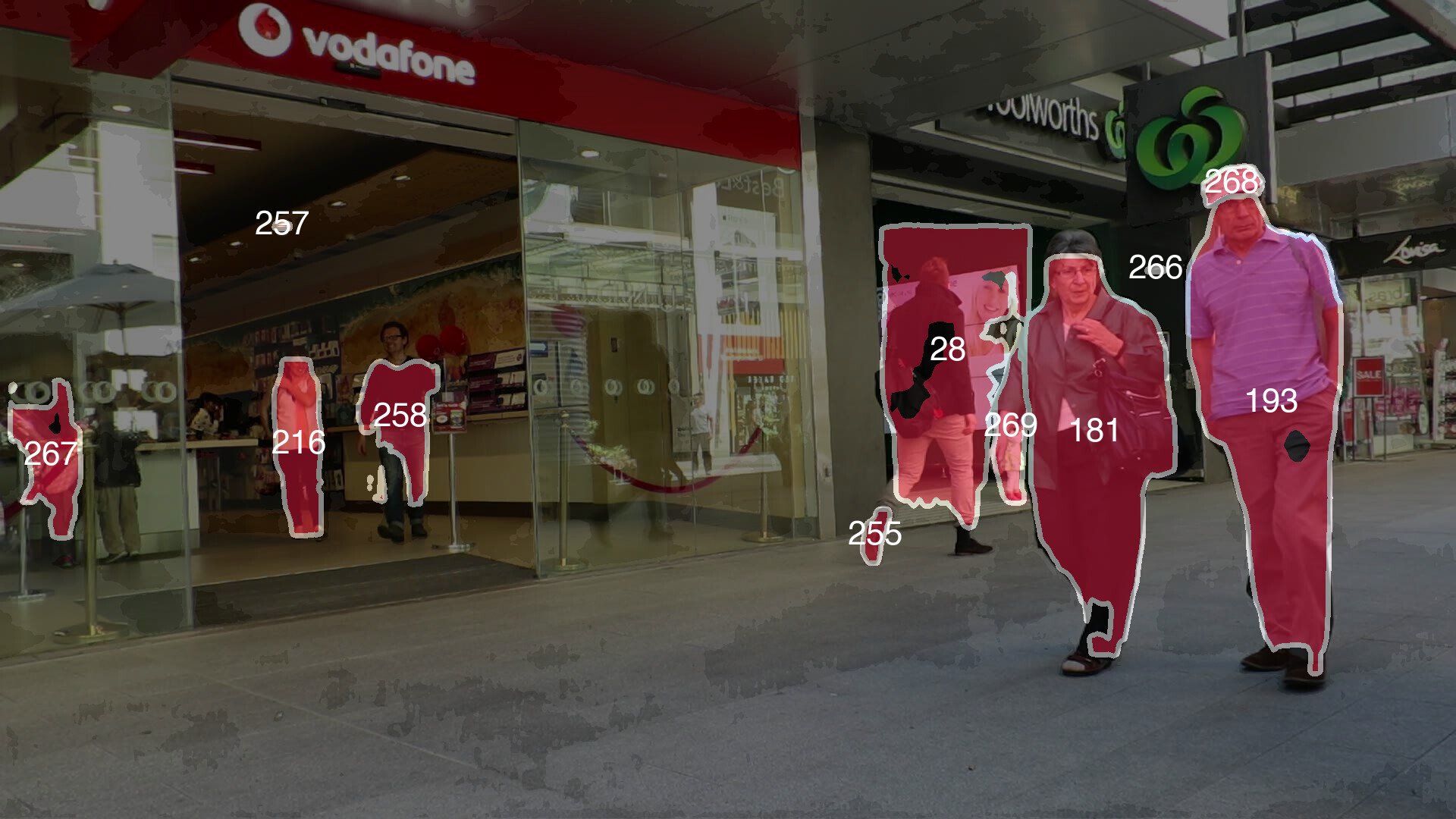} 
& \includegraphics[width=\linewidth]{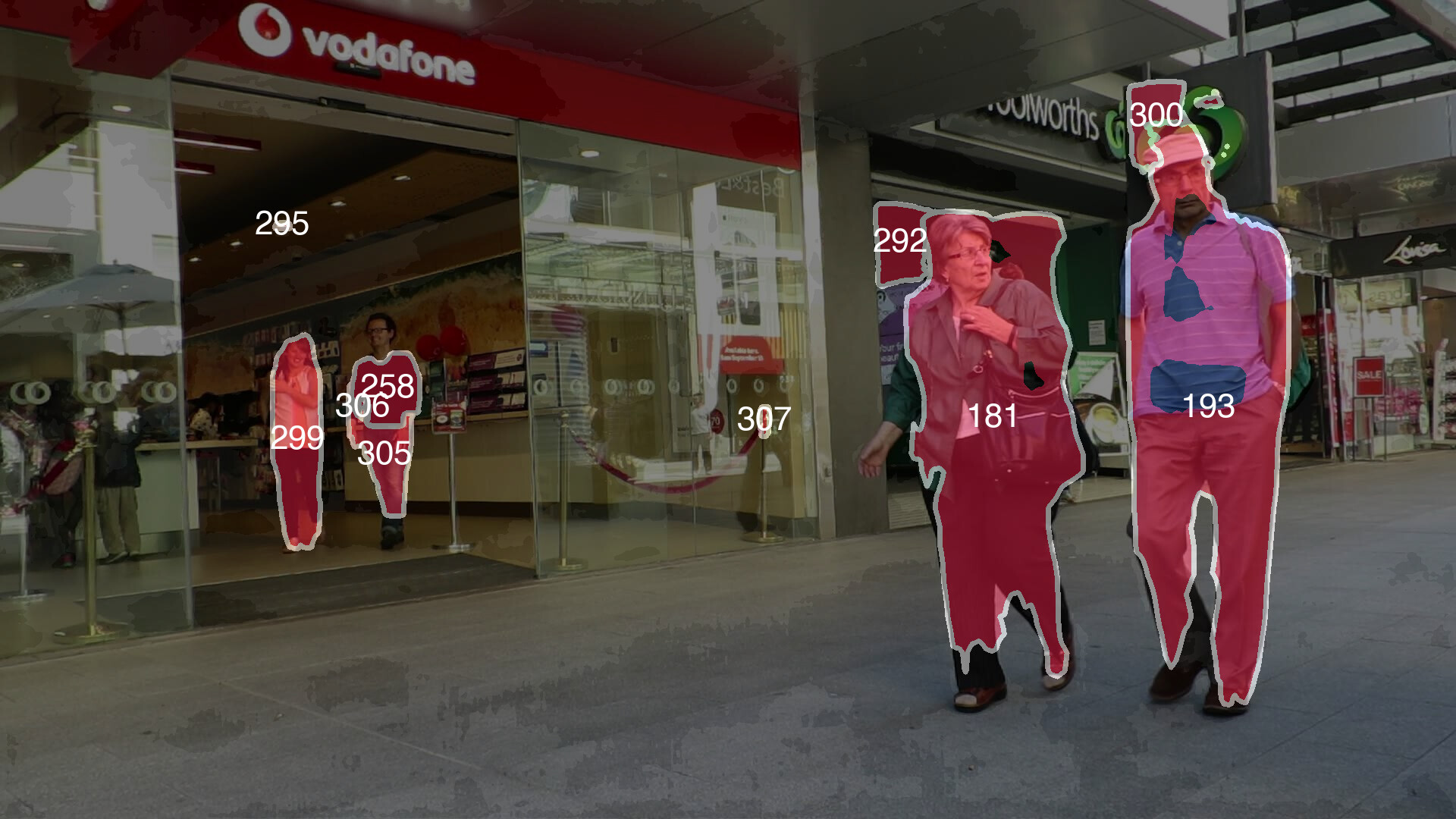}  
& \includegraphics[width=\linewidth]{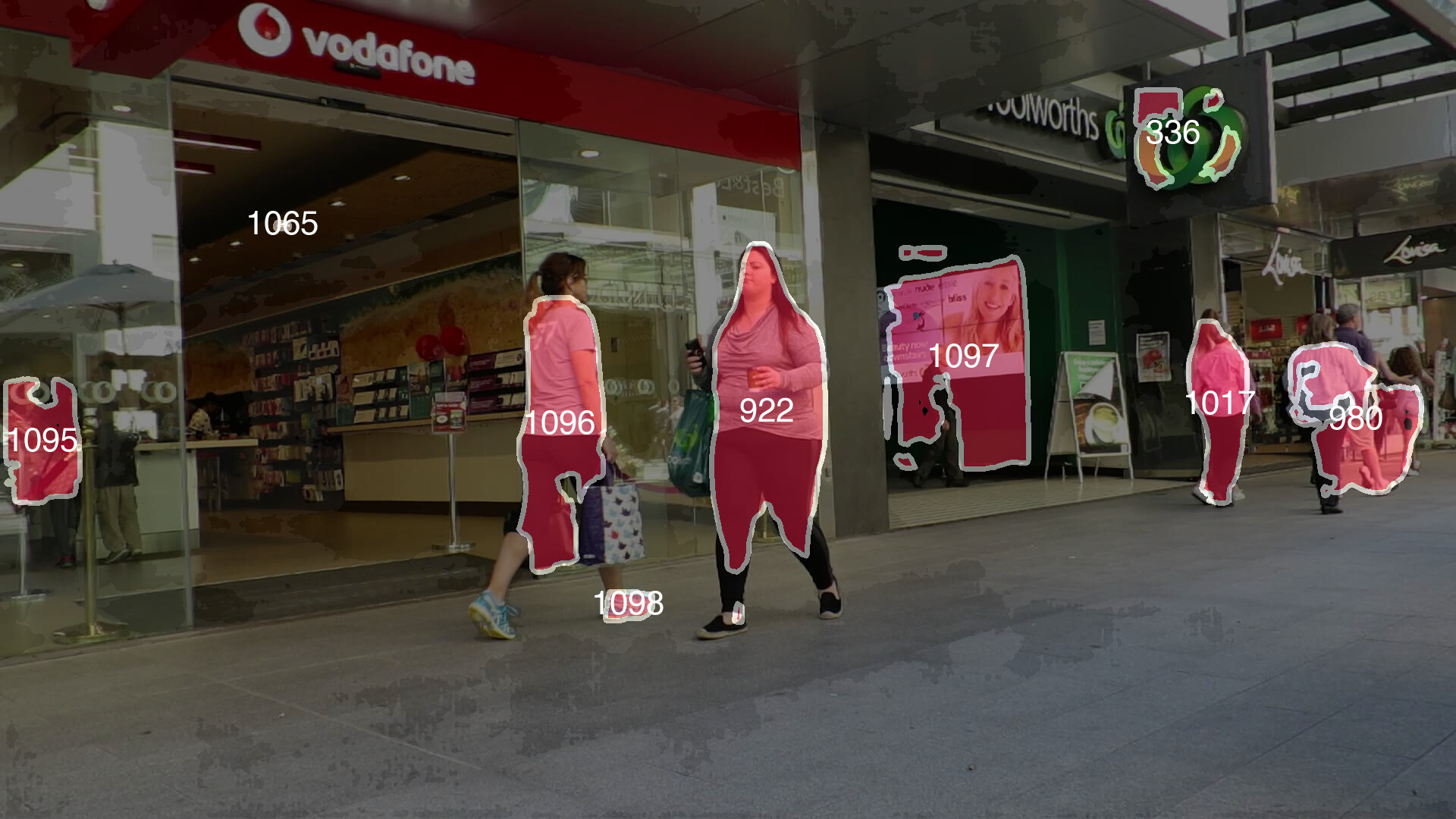}  
& \includegraphics[width=\linewidth]{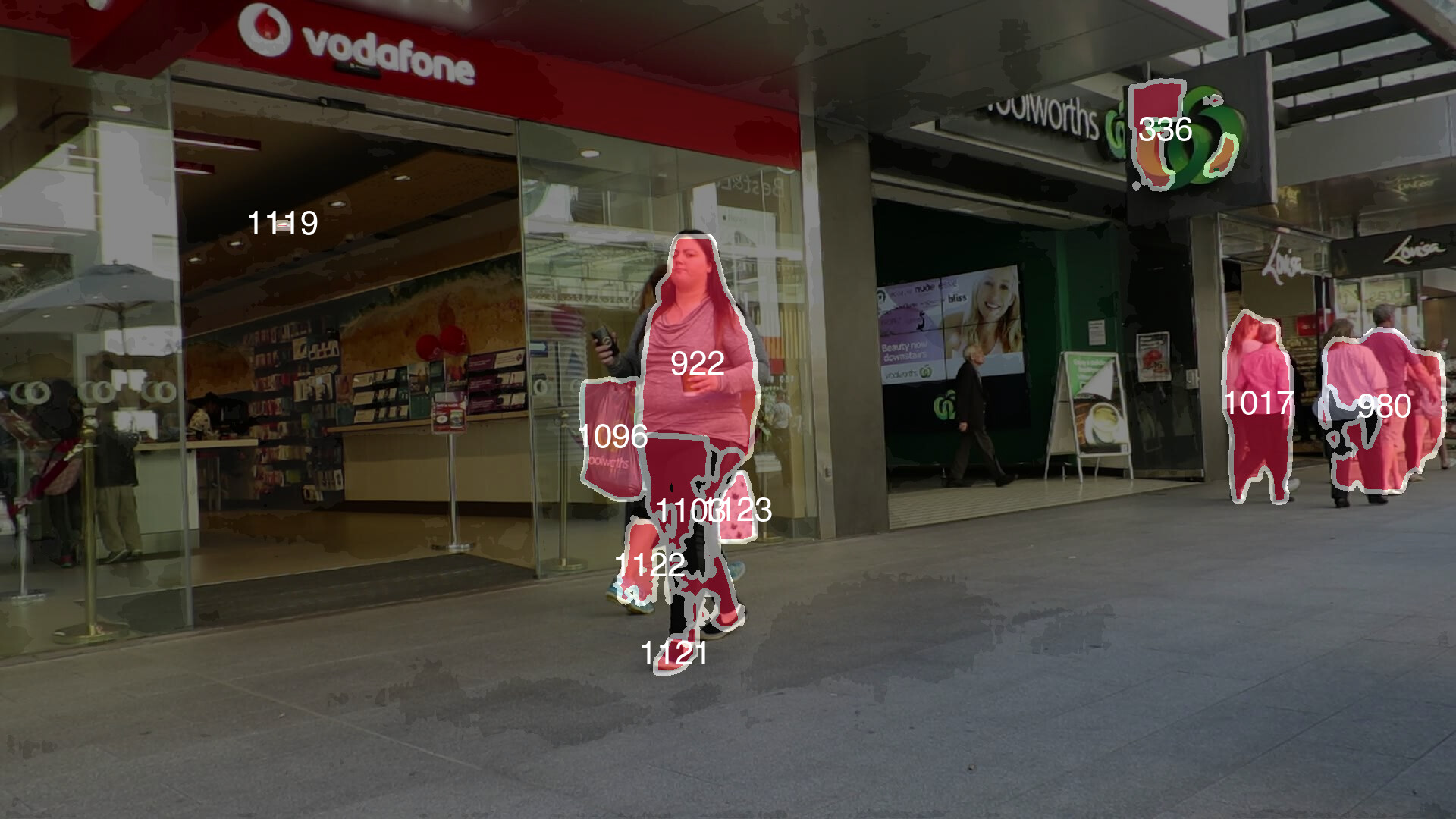}  
& \includegraphics[width=\linewidth]{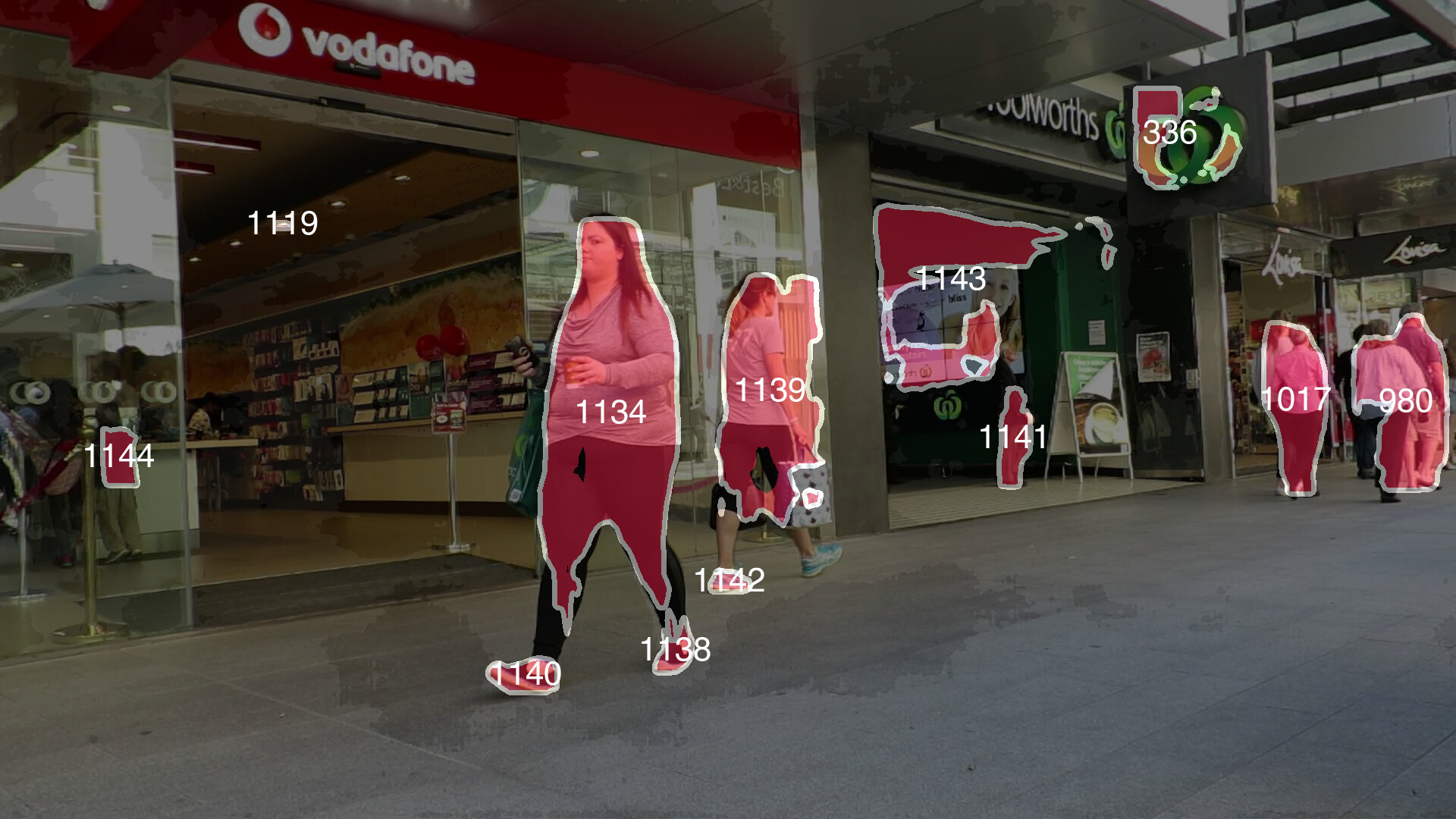} \\

\rotatebox[origin=lB]{90}{\hspace{-0.1em}\tiny{\MethodName~\textit{(Ours)}}}
& \includegraphics[width=\linewidth]{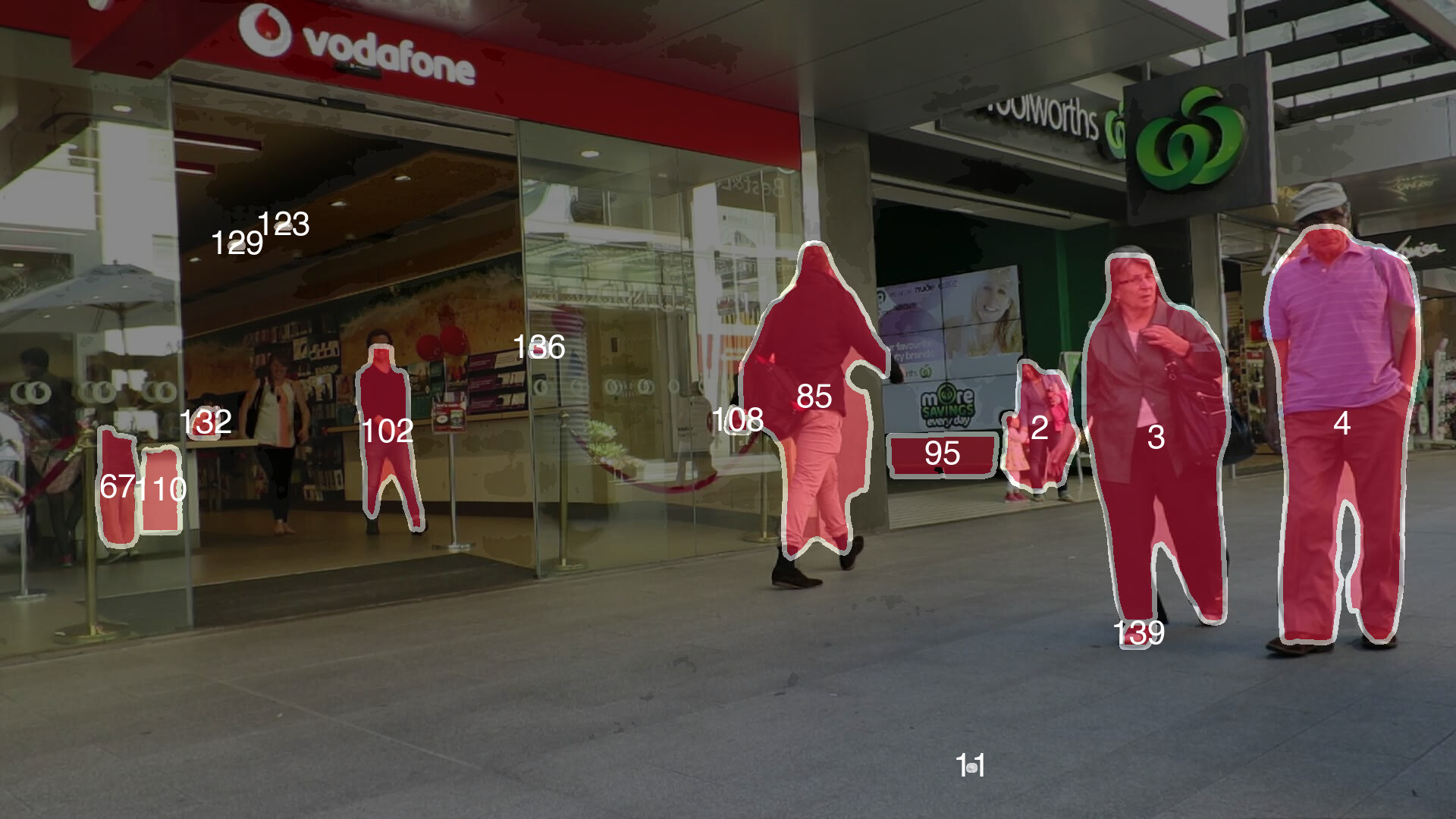} 
& \includegraphics[width=\linewidth]{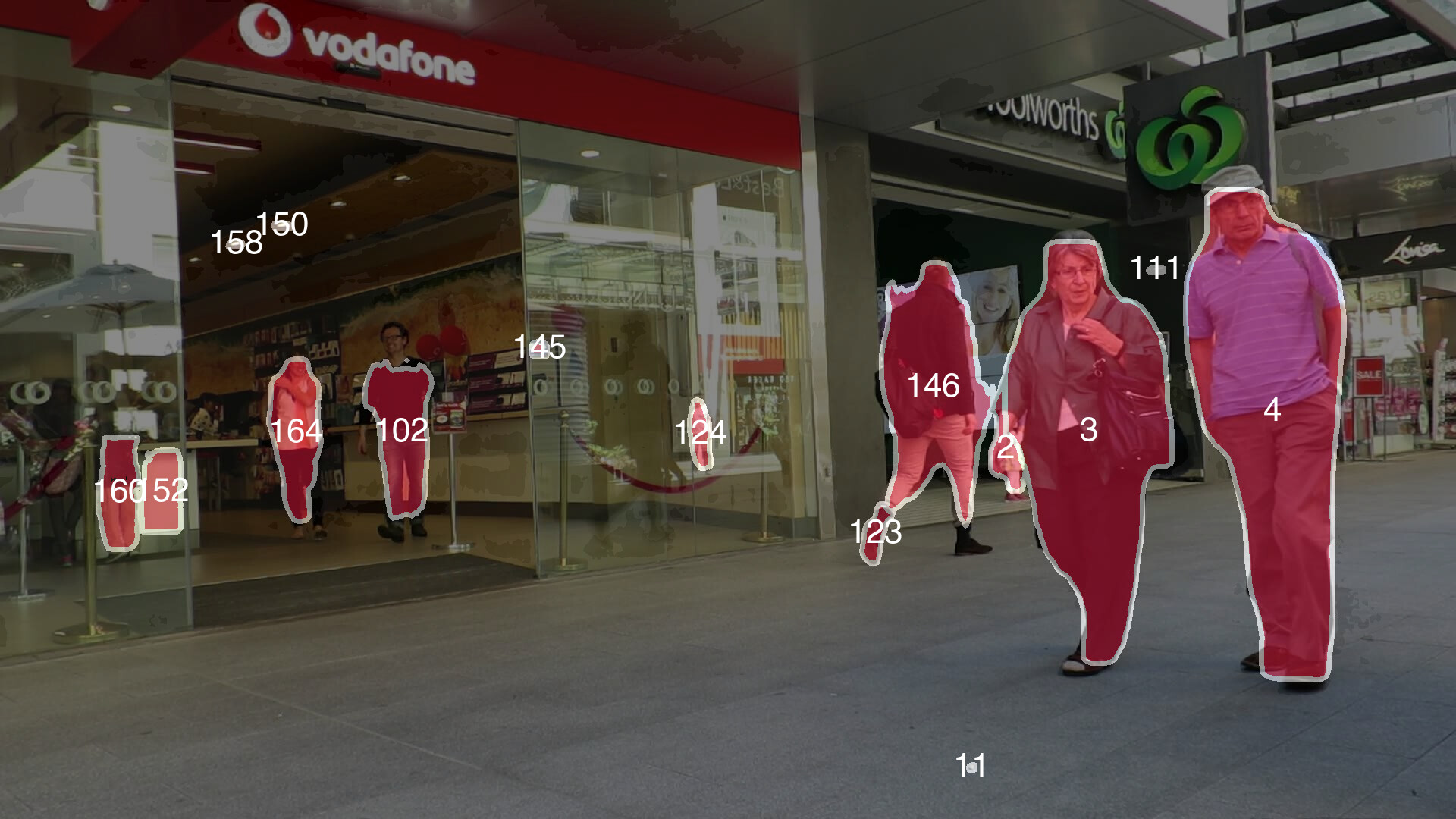}  
& \includegraphics[width=\linewidth]{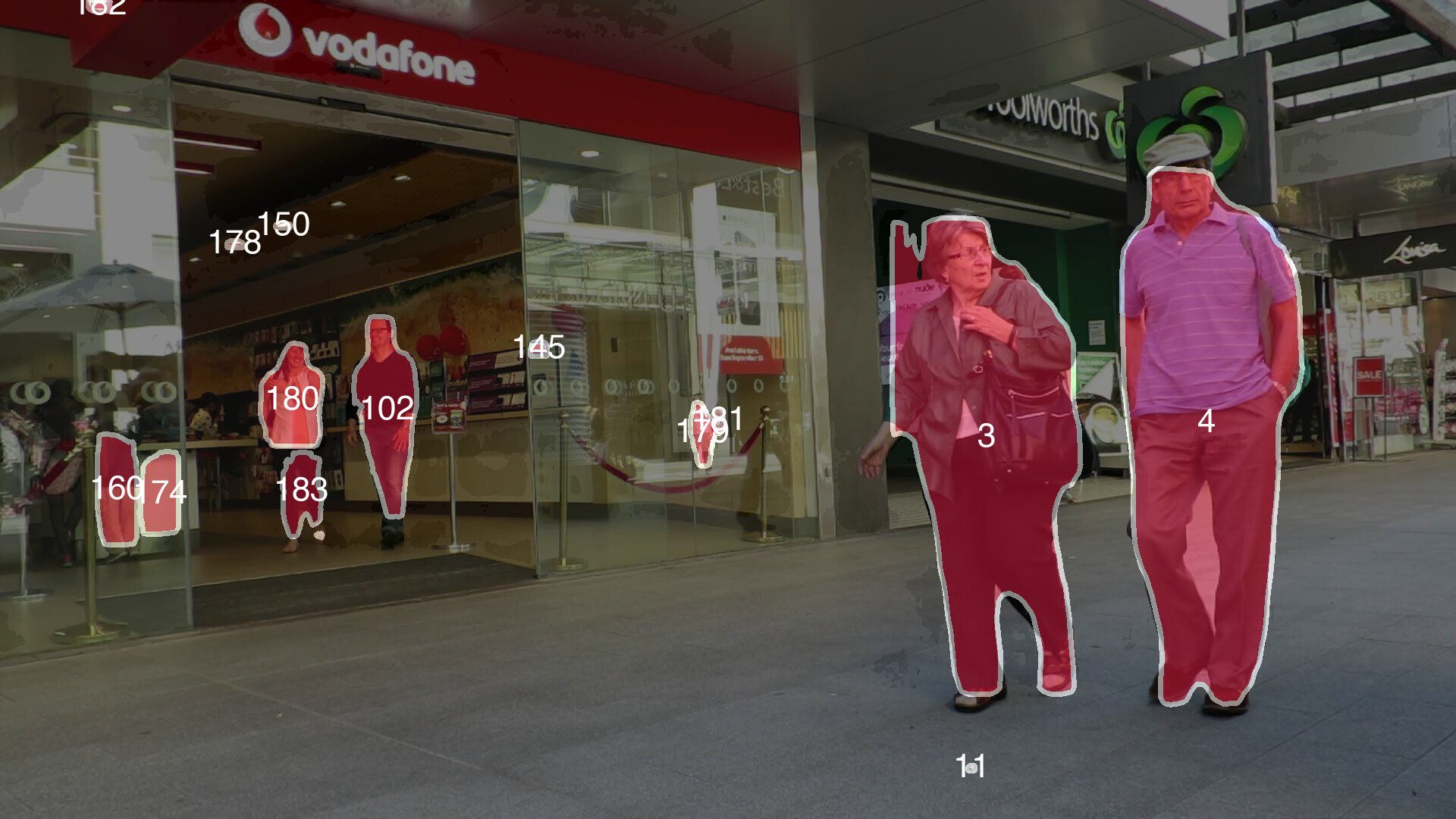} 
& \includegraphics[width=\linewidth]{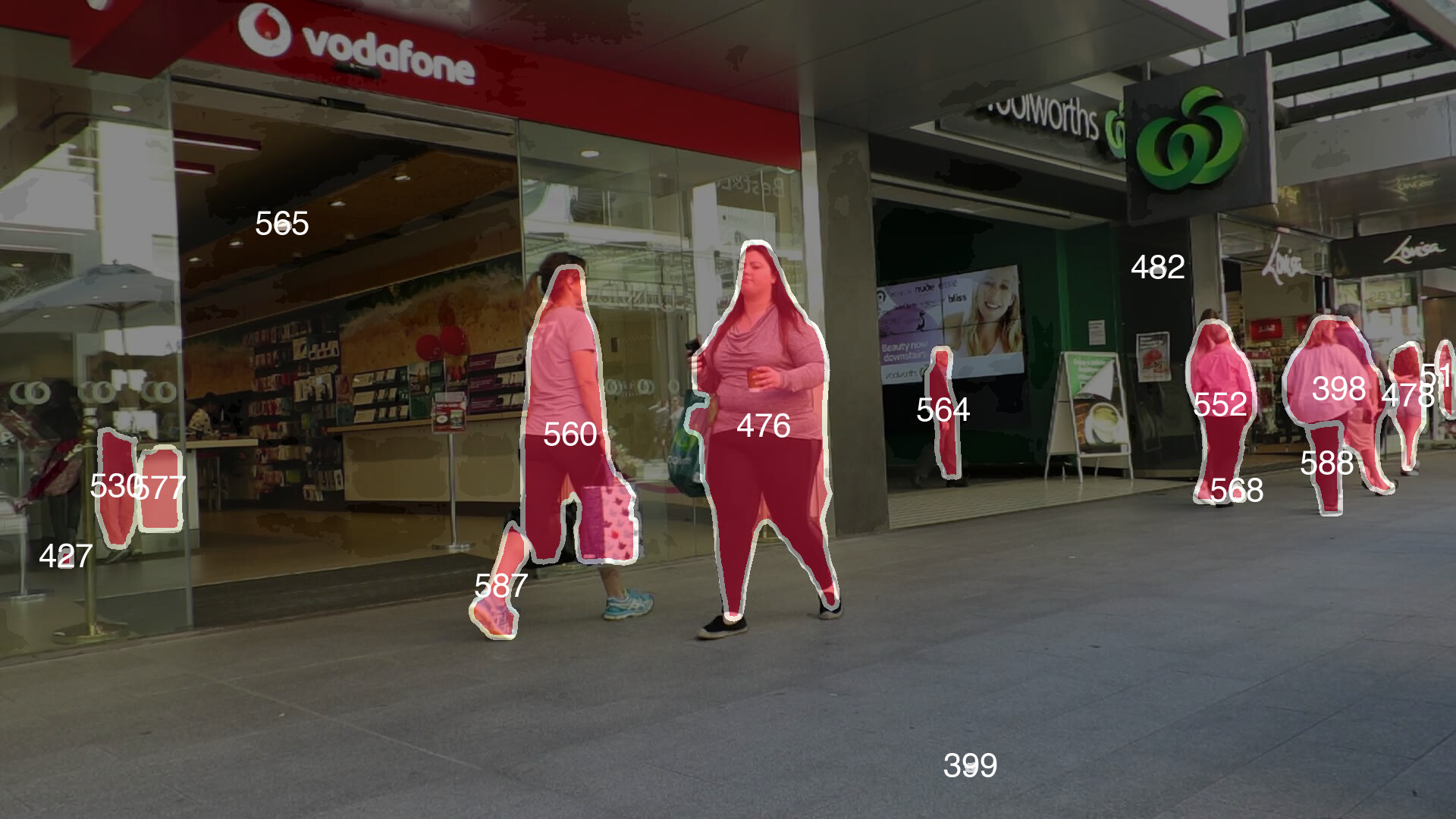}  
& \includegraphics[width=\linewidth]{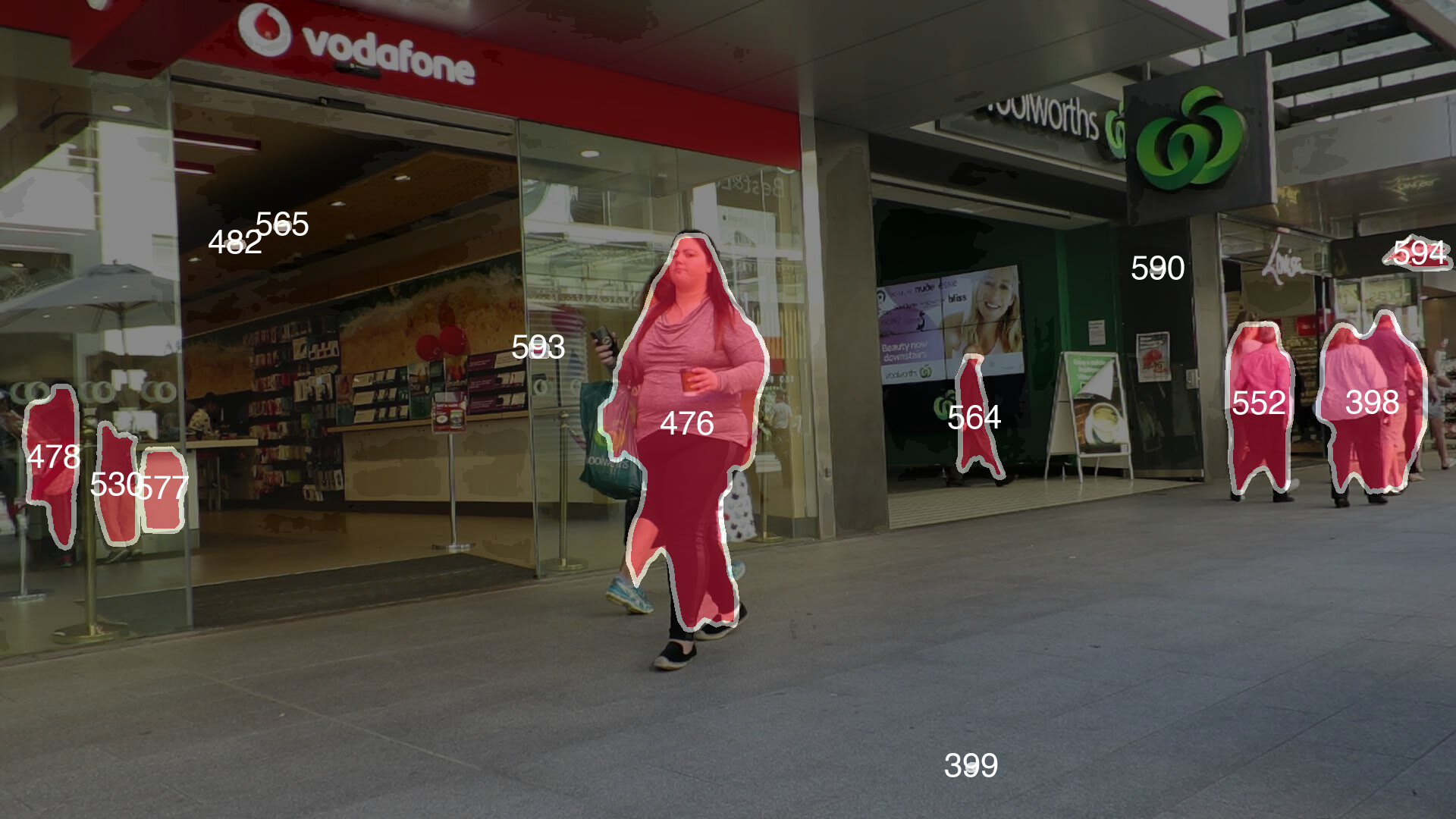}  
& \includegraphics[width=\linewidth]{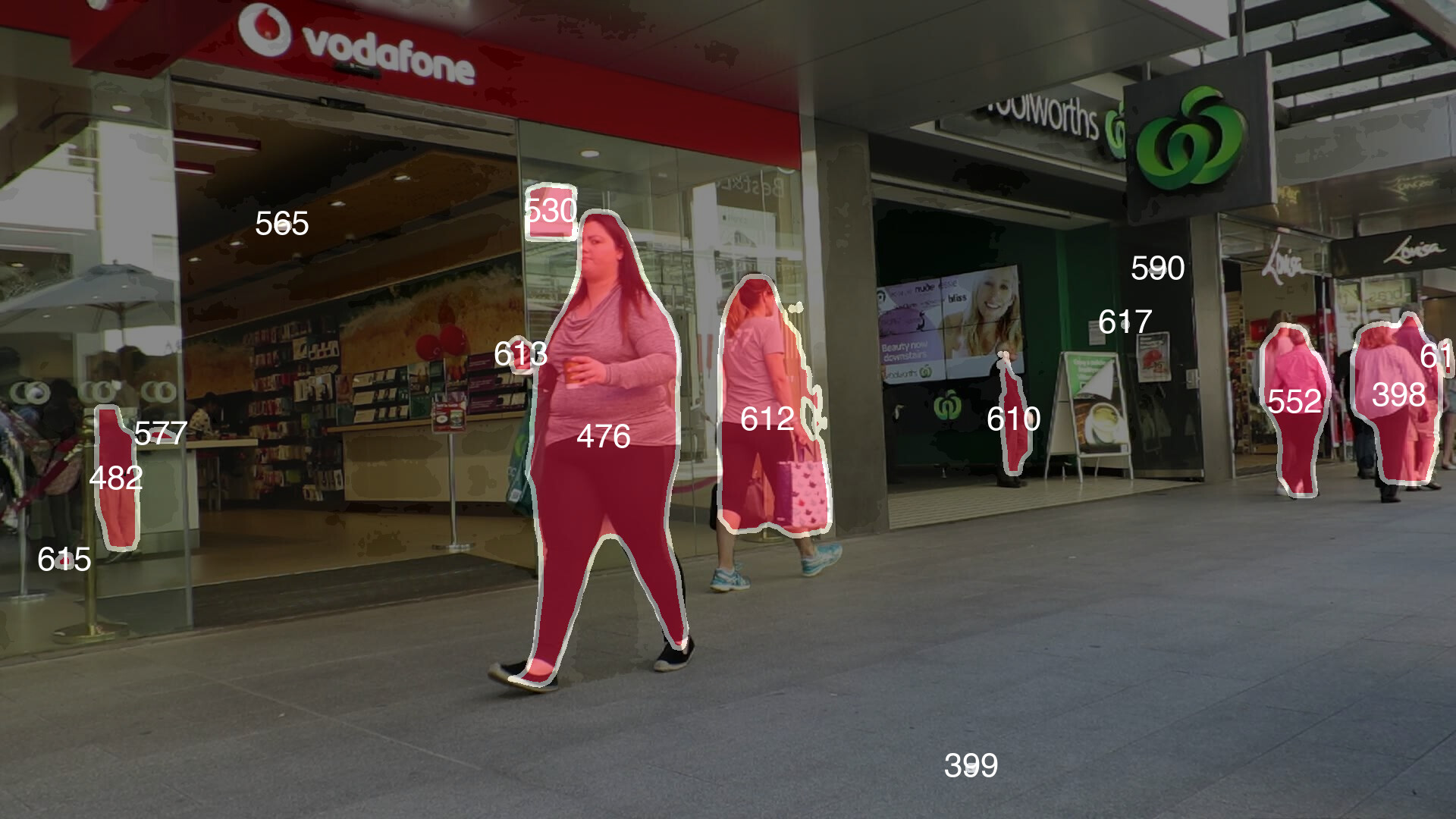}  \\[-2.5pt]

&\tiny{$\phantom{.}t_0$}
&\tiny{$\phantom{.}t_1$} 
&\tiny{$\phantom{.}t_2$}
&\tiny{$\phantom{.}t_0$} 
&\tiny{$\phantom{.}t_1$} 
&\tiny{$\phantom{.}t_2$} \\

\end{tabular}

    \vspace{-1.2em}
    \caption{\textbf{MOTS---Qualitative unsupervised VPS examples.} We compare our proposed method \MethodName to the proposed baselines DepthG~\cite{Sick:2024:USS} +~VideoCutLER~\cite{Wang:2024:VCS}, U2Seg~\cite{Niu:2024:UUI}~+~SORT, CUPS~\cite{Hahn:2025:UPS}~+~SORT~\cite{Bewley:2016:SOR}, and CUPS\textsuperscript{\textdagger}~\cite{Hahn:2025:UPS}~+~SORT~\cite{Bewley:2016:SOR} on MOTS~\cite{Voigtlaender:2019:MMO} val.\label{fig:qualitative_mots}}
    \vspace{-0.5em}
\end{figure*}

\begin{figure*}[t]
    \centering
    \newcommand{\imgwidth}{0.163}
\newcommand{\dddviswidth}{0.182}

\tiny
\renewcommand{\arraystretch}{1.3}
\sffamily
\begin{tabularx}{0.998\textwidth}{*{19}{>{\centering\arraybackslash}X}}
    \cellcolor{road}\textcolor{white}{Road} &
    \cellcolor{sidewalk}\!\!\textcolor{white}{Sidewalk} &
    \cellcolor{building}\!\textcolor{white}{Building} &
    \cellcolor{wall}\textcolor{white}{Wall} &
    \cellcolor{fence}\textcolor{white}{Fence} &
    \cellcolor{pole}\textcolor{white}{Pole} &
    \cellcolor{trafficlight}\textcolor{white}{T.~Light} &
    \cellcolor{trafficsign}\textcolor{white}{T.~Sign} &
    \cellcolor{vegetation}\textcolor{white}{Veget.} &
    \cellcolor{terrain}\textcolor{white}{Terrain} &
    \cellcolor{sky}\textcolor{white}{Sky} &
    \cellcolor{person}\textcolor{white}{Person} &
    \cellcolor{rider}\textcolor{white}{Rider} &
    \cellcolor{car}\textcolor{white}{Car} &
    \cellcolor{truck}\textcolor{white}{Truck} & 
    \cellcolor{bus}\textcolor{white}{Bus} & 
    \cellcolor{train}\textcolor{white}{Train} &
    \cellcolor{motorcycle}\textcolor{white}{M.cycle} &
    \cellcolor{bicycle}\textcolor{white}{Bicycle}
\end{tabularx}

\vspace{1pt}

\scriptsize
\sffamily
\setlength{\tabcolsep}{1pt}
\renewcommand{\arraystretch}{1.0}

\begin{tabular}{
    >{\centering\arraybackslash}m{\imgwidth\textwidth}
    >{\centering\arraybackslash}m{\imgwidth\textwidth}
    >{\centering\arraybackslash}m{\imgwidth\textwidth}
    >{\centering\arraybackslash}m{\imgwidth\textwidth}
    >{\centering\arraybackslash}m{\imgwidth\textwidth}
    >{\centering\arraybackslash}m{\imgwidth\textwidth}
}

\includegraphics[width=\linewidth]{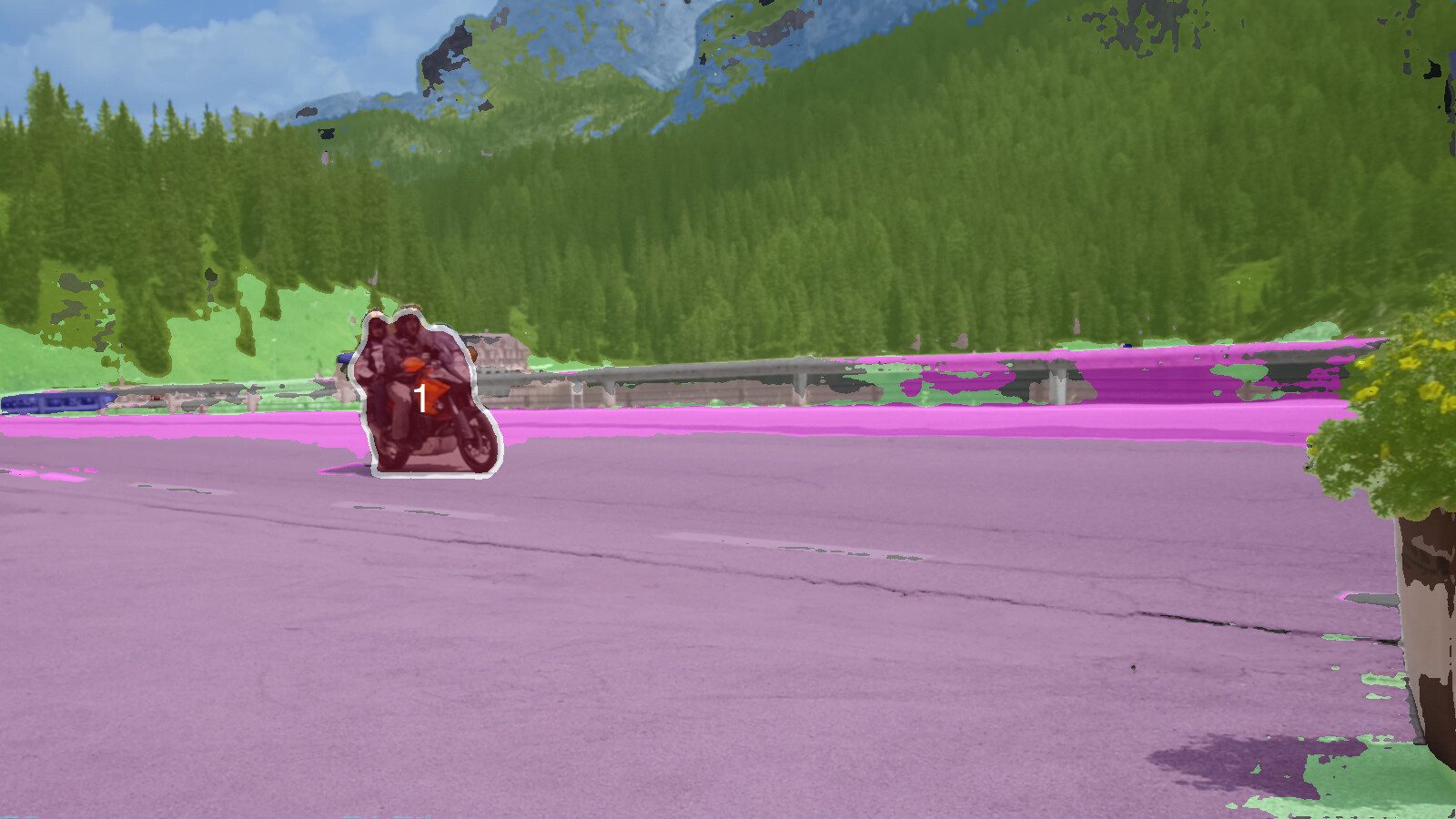} 
& \includegraphics[width=\linewidth]{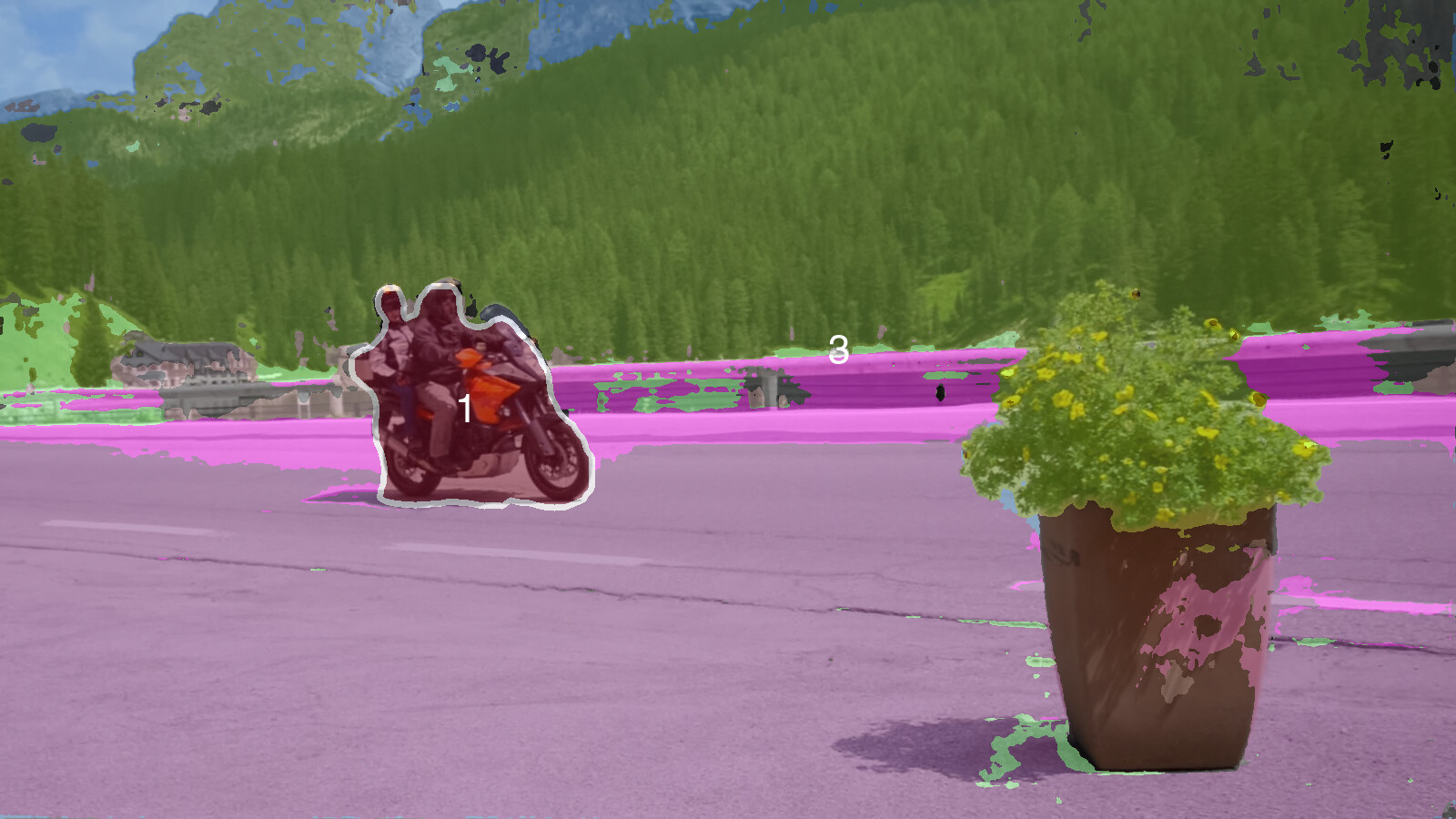}  
& \includegraphics[width=\linewidth]{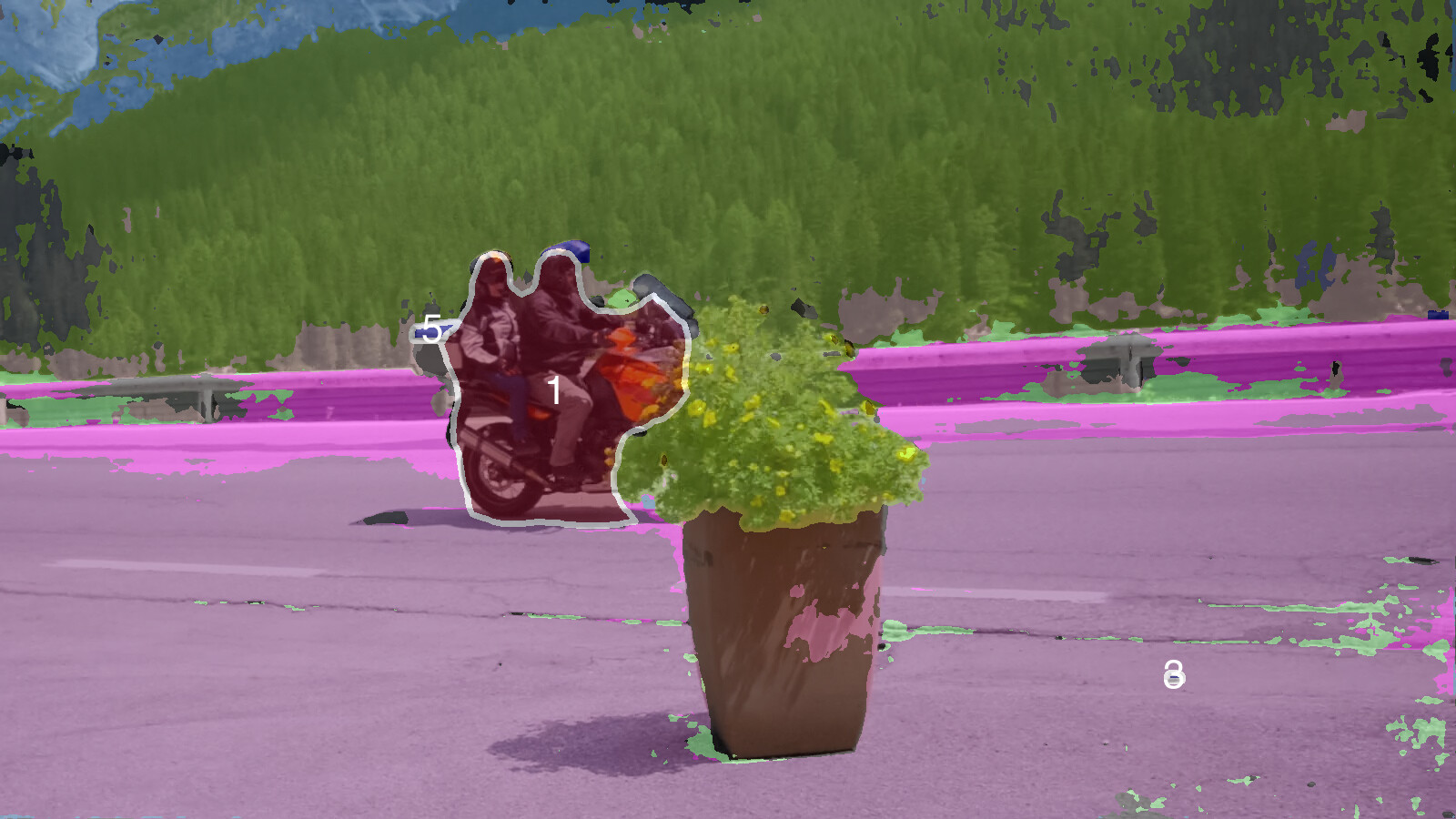}  
& \includegraphics[width=\linewidth]{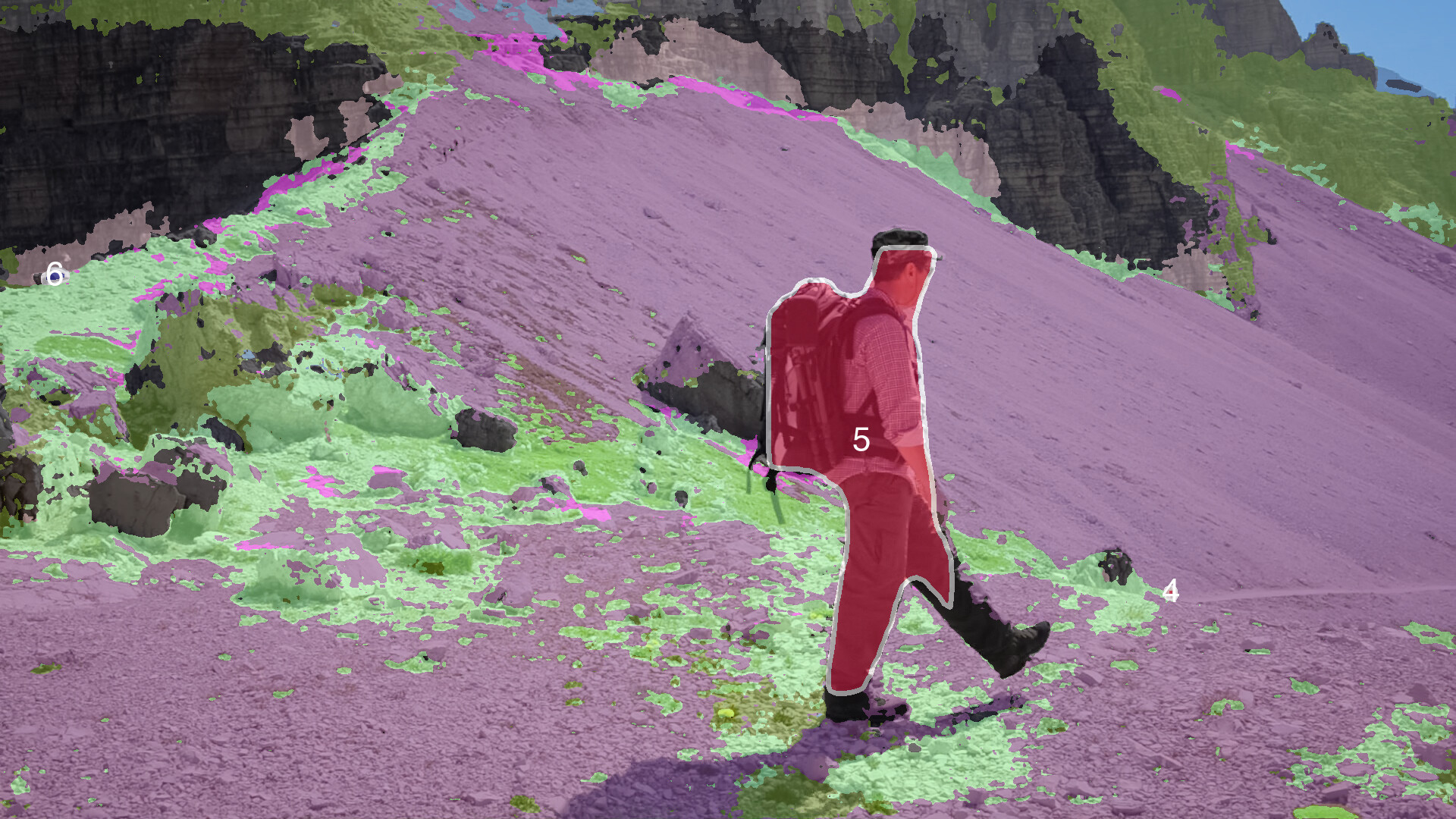}
& \includegraphics[width=\linewidth]{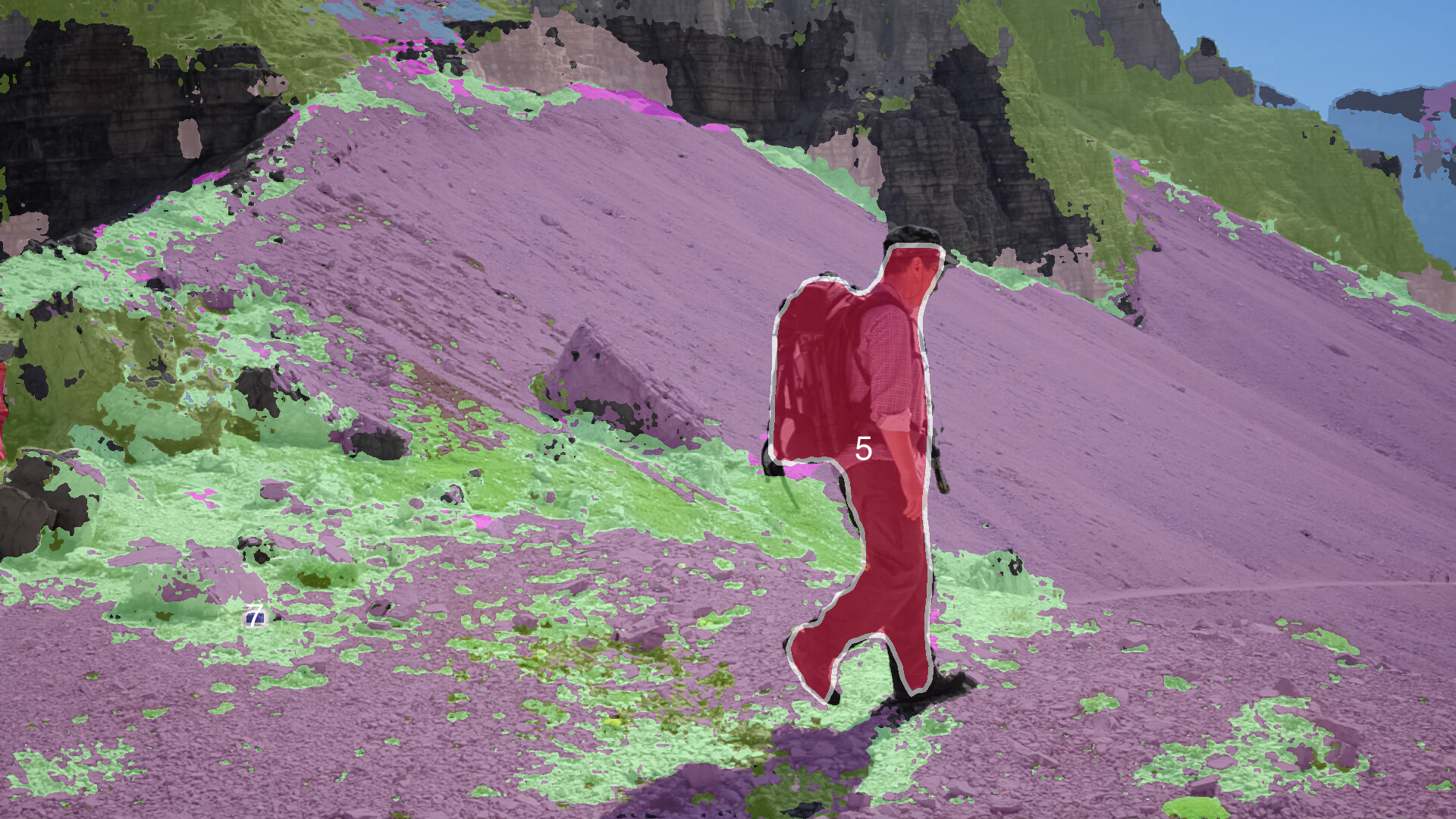}
& \includegraphics[width=\linewidth]{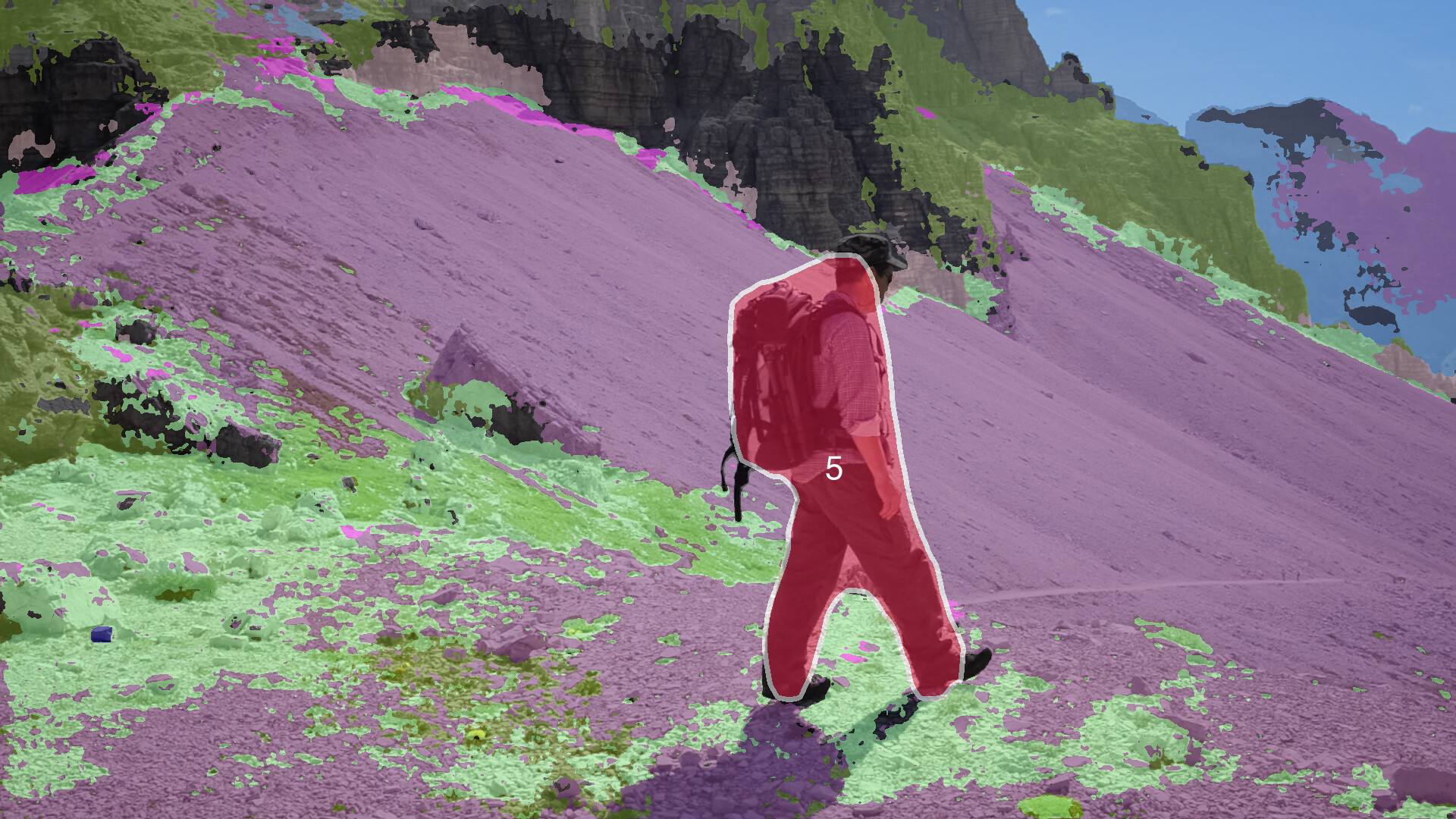}\\

\includegraphics[width=\linewidth]{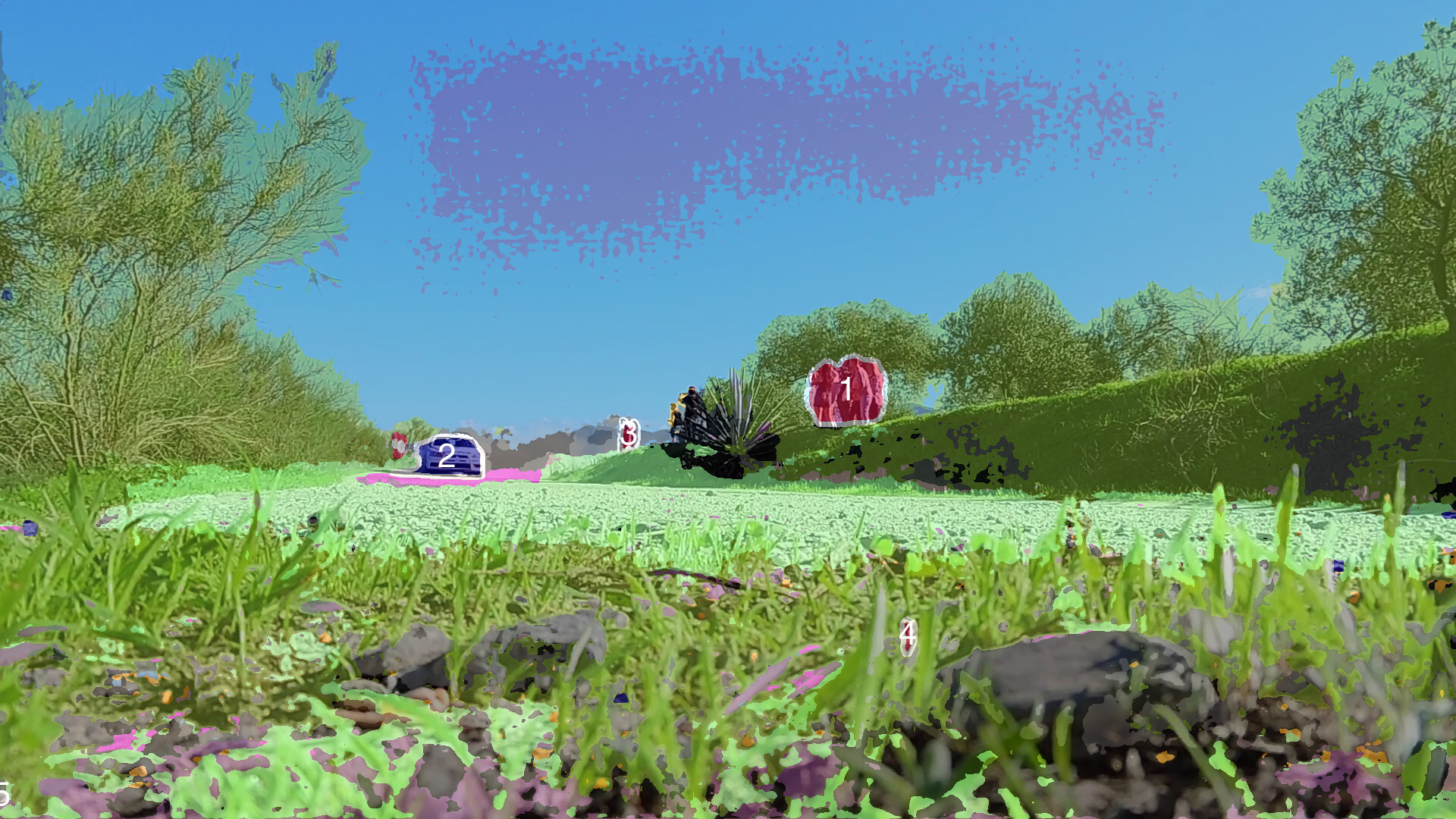} 
& \includegraphics[width=\linewidth]{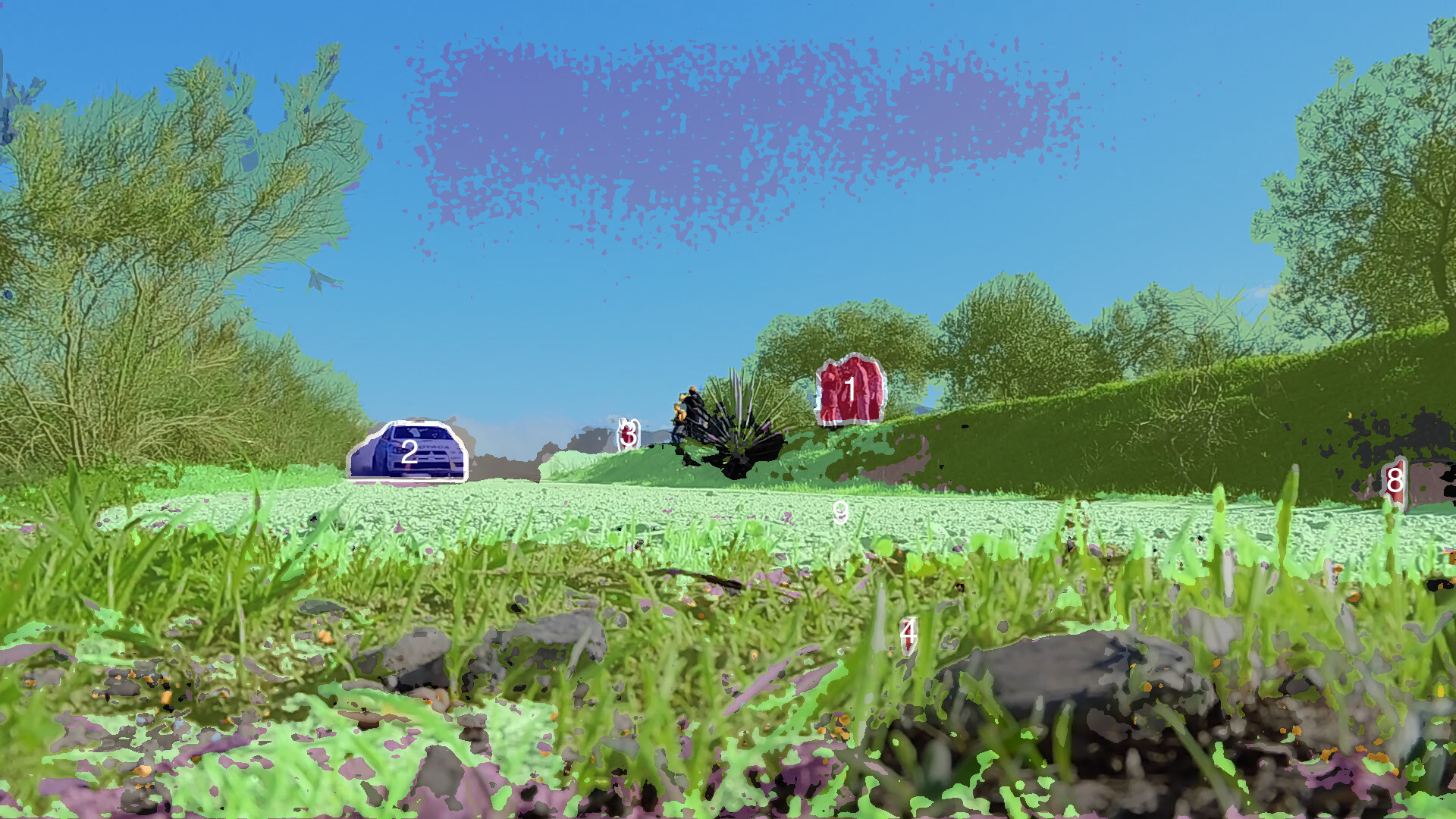} 
& \includegraphics[width=\linewidth]{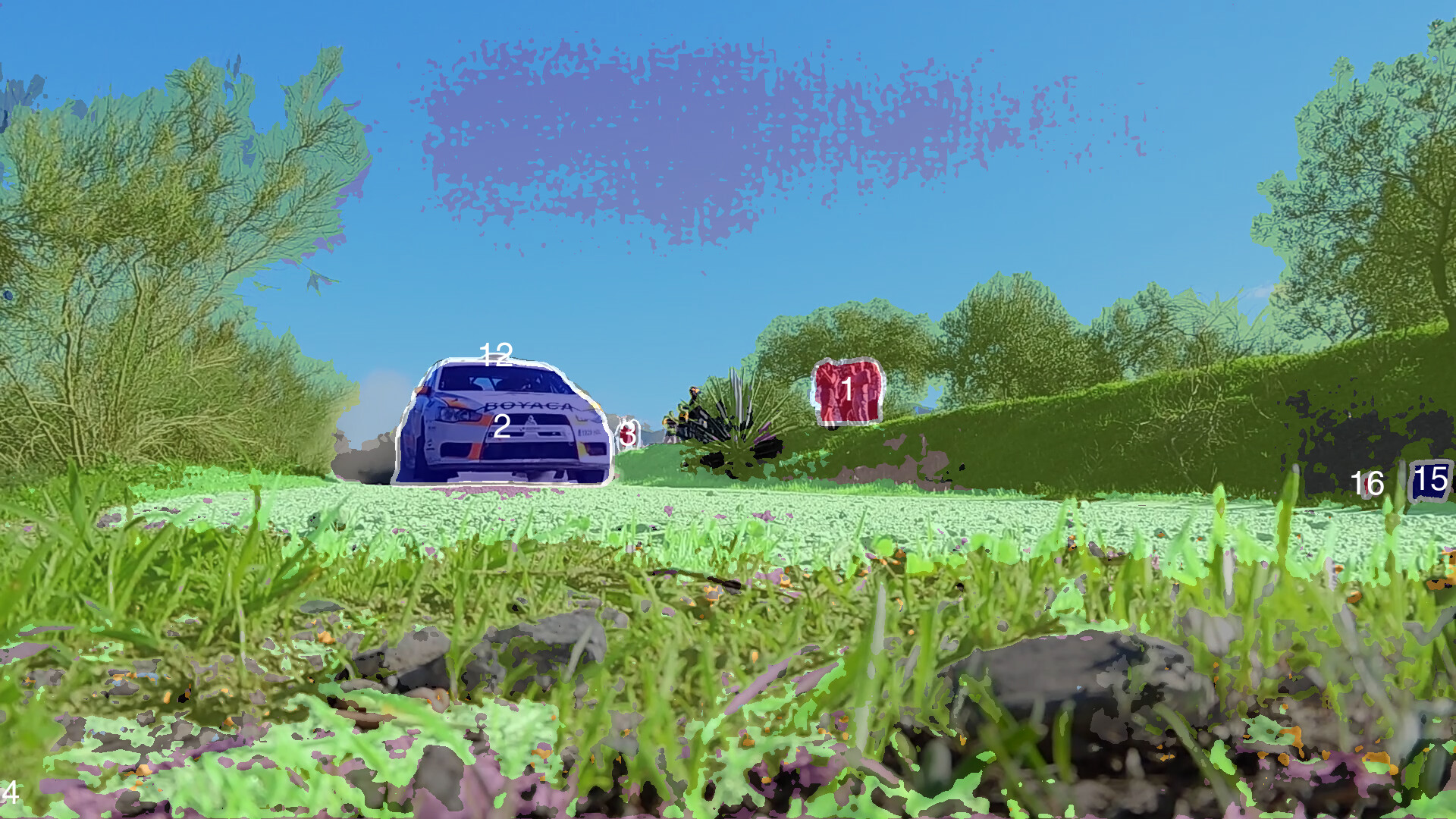} 
& \includegraphics[width=\linewidth]{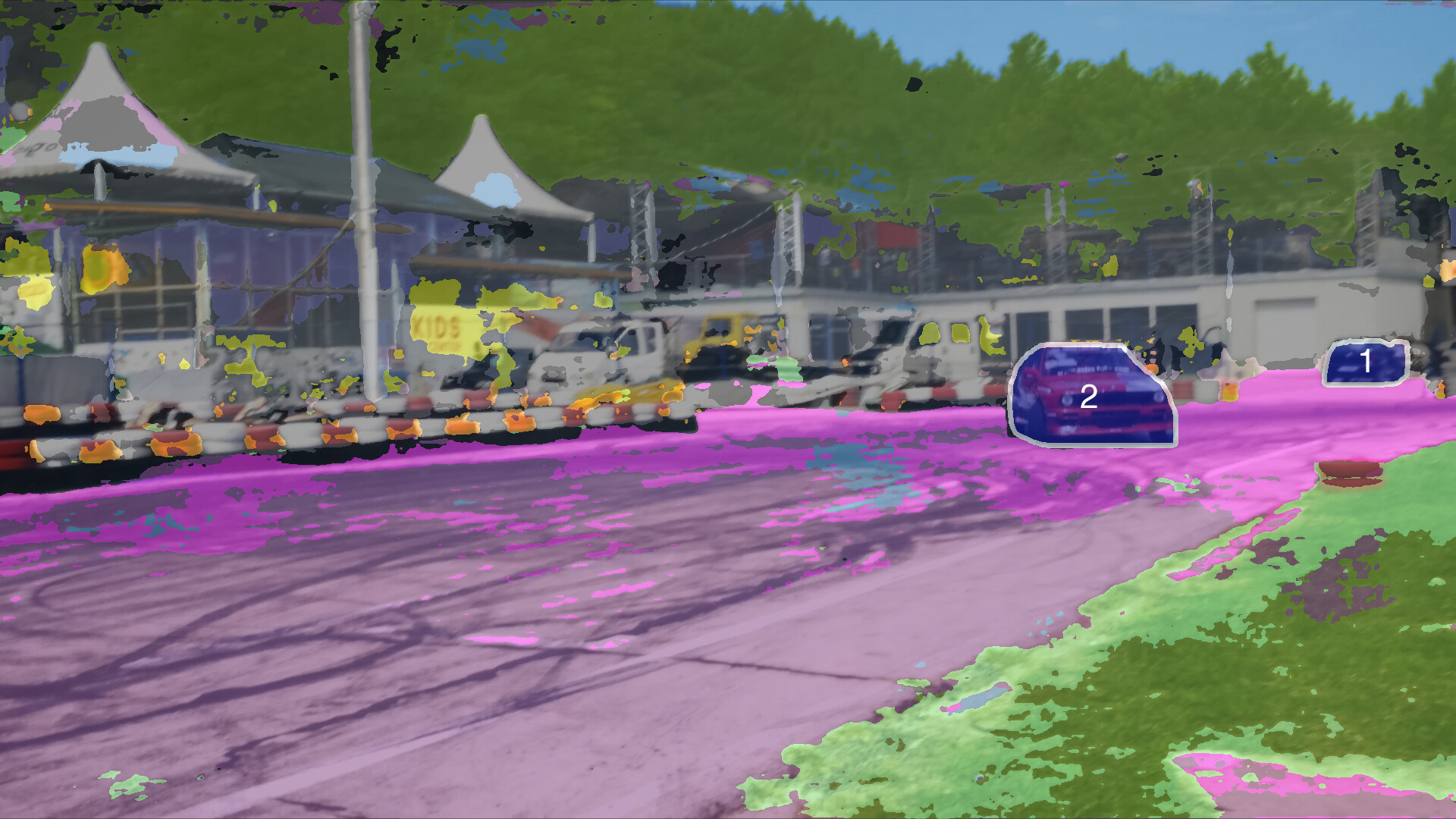}
& \includegraphics[width=\linewidth]{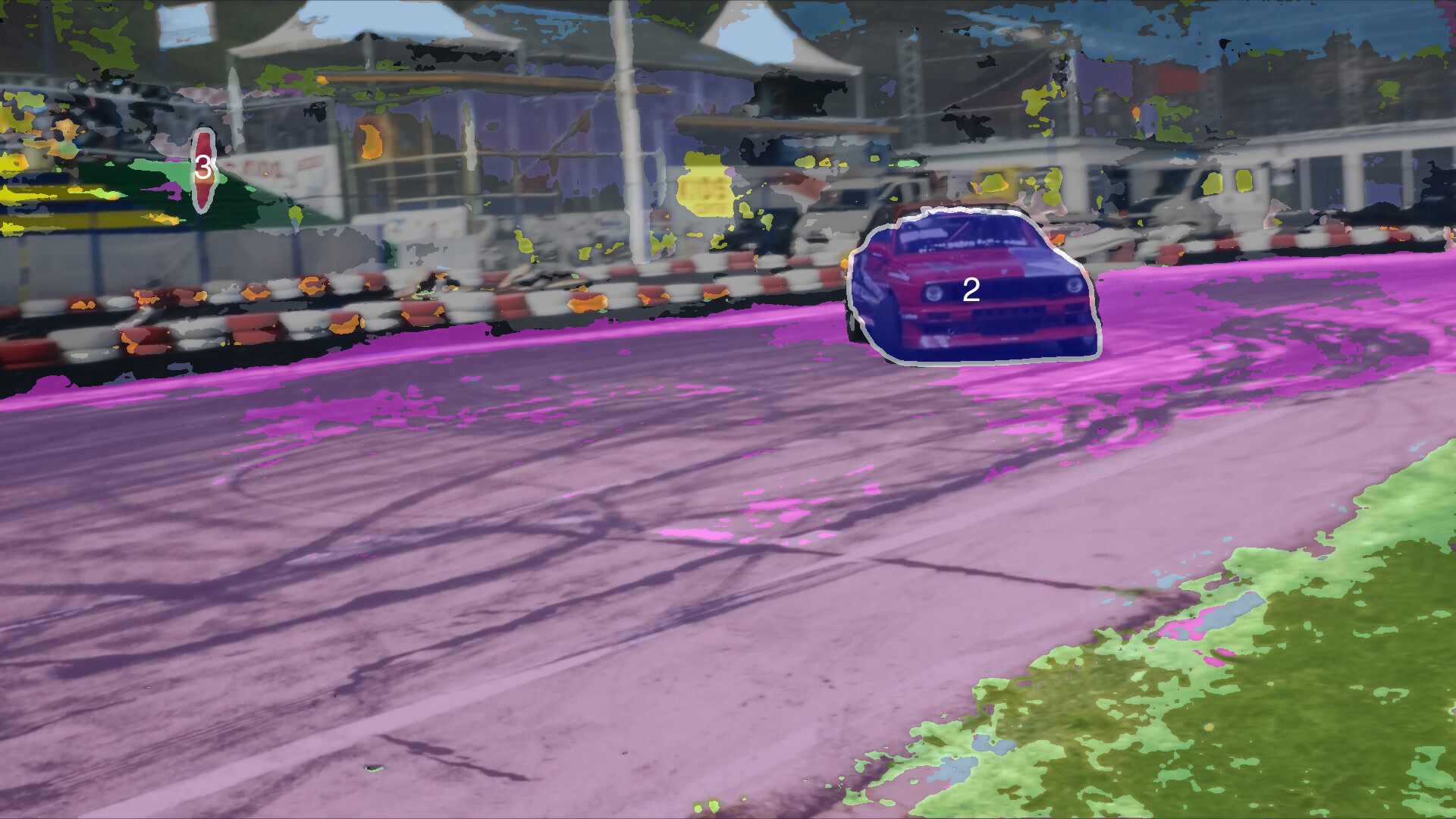}
& \includegraphics[width=\linewidth]{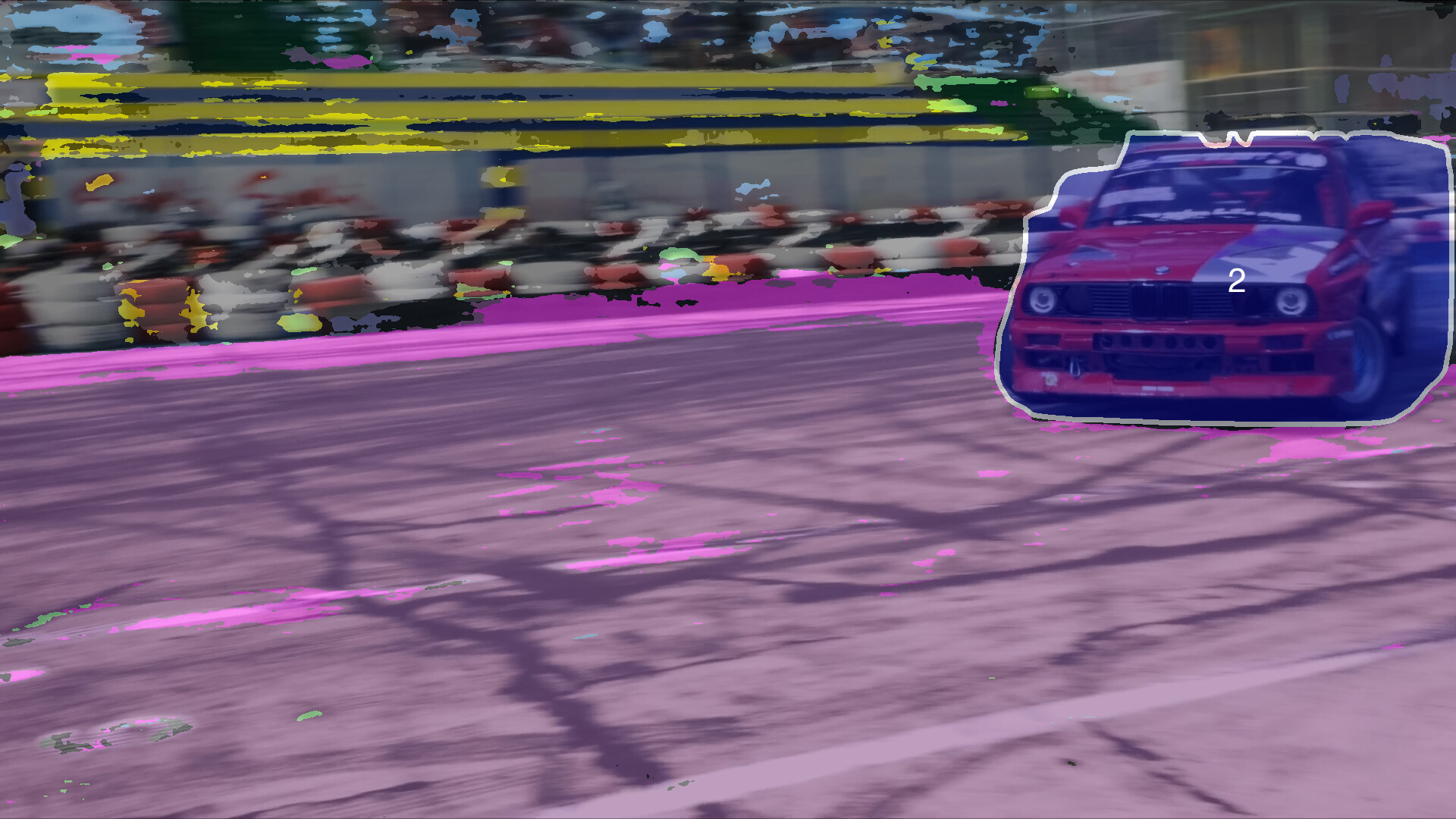}\\

\includegraphics[width=\linewidth]{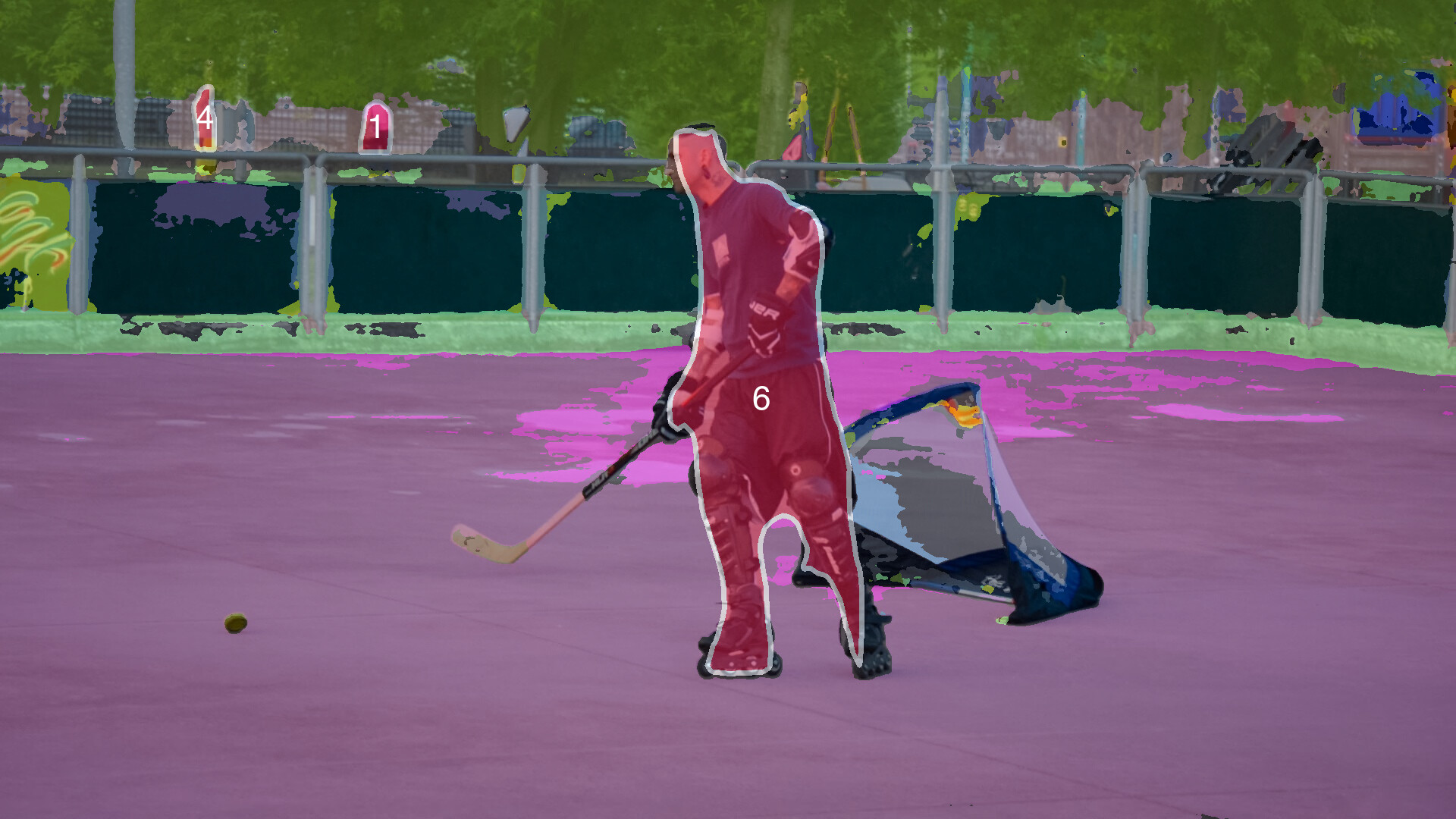} 
& \includegraphics[width=\linewidth]{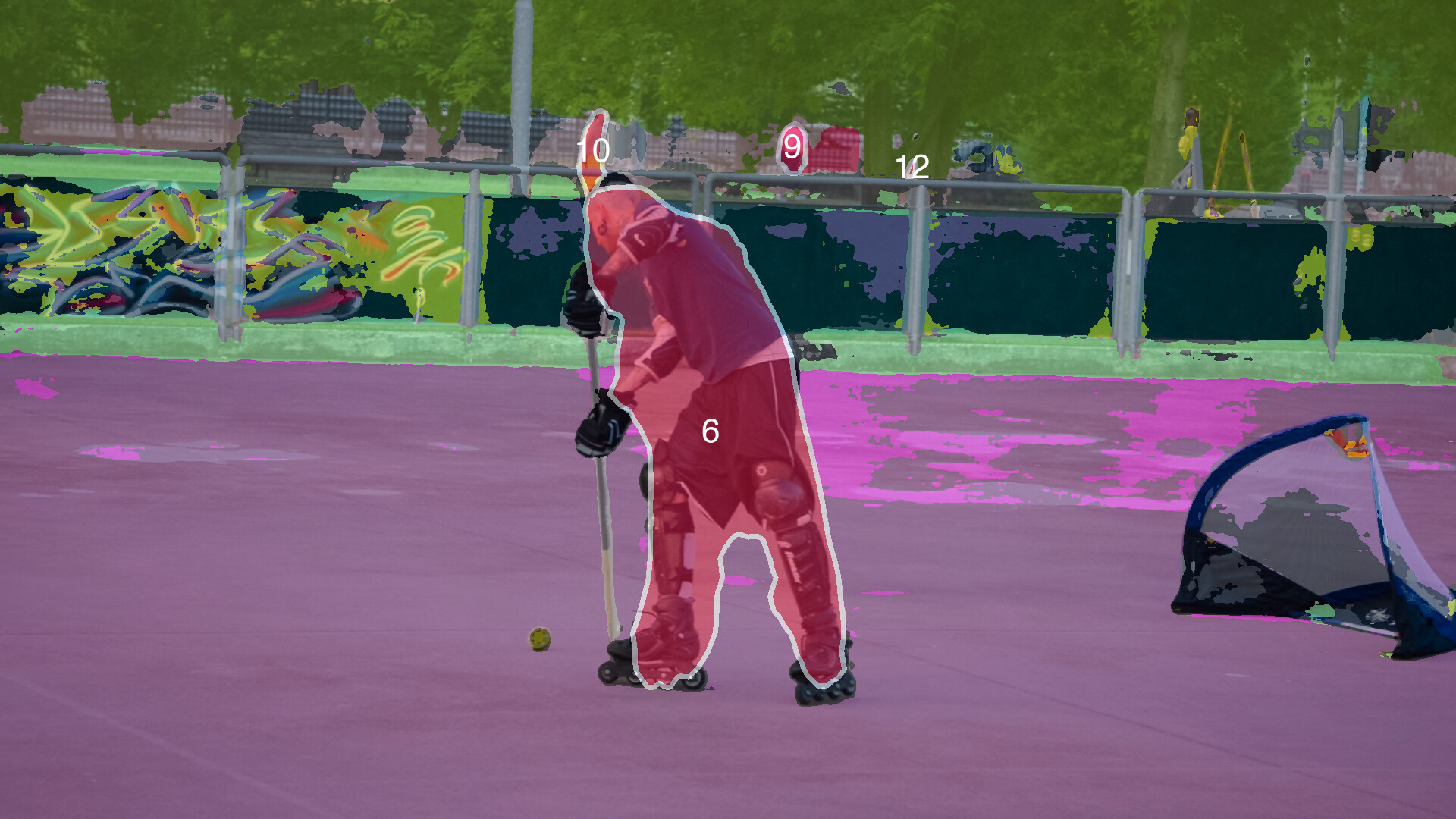}
& \includegraphics[width=\linewidth]{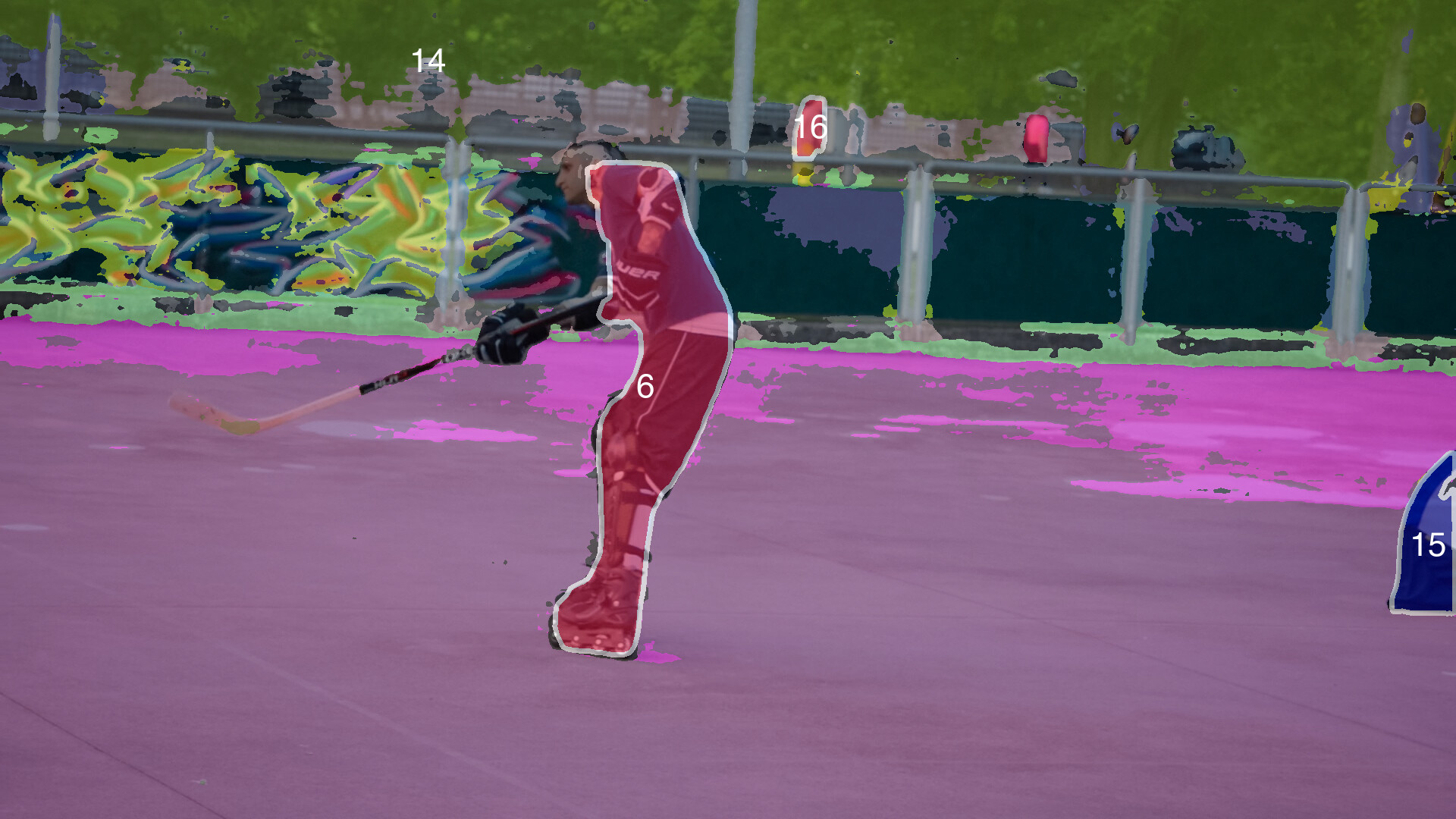}
& \includegraphics[width=\linewidth]{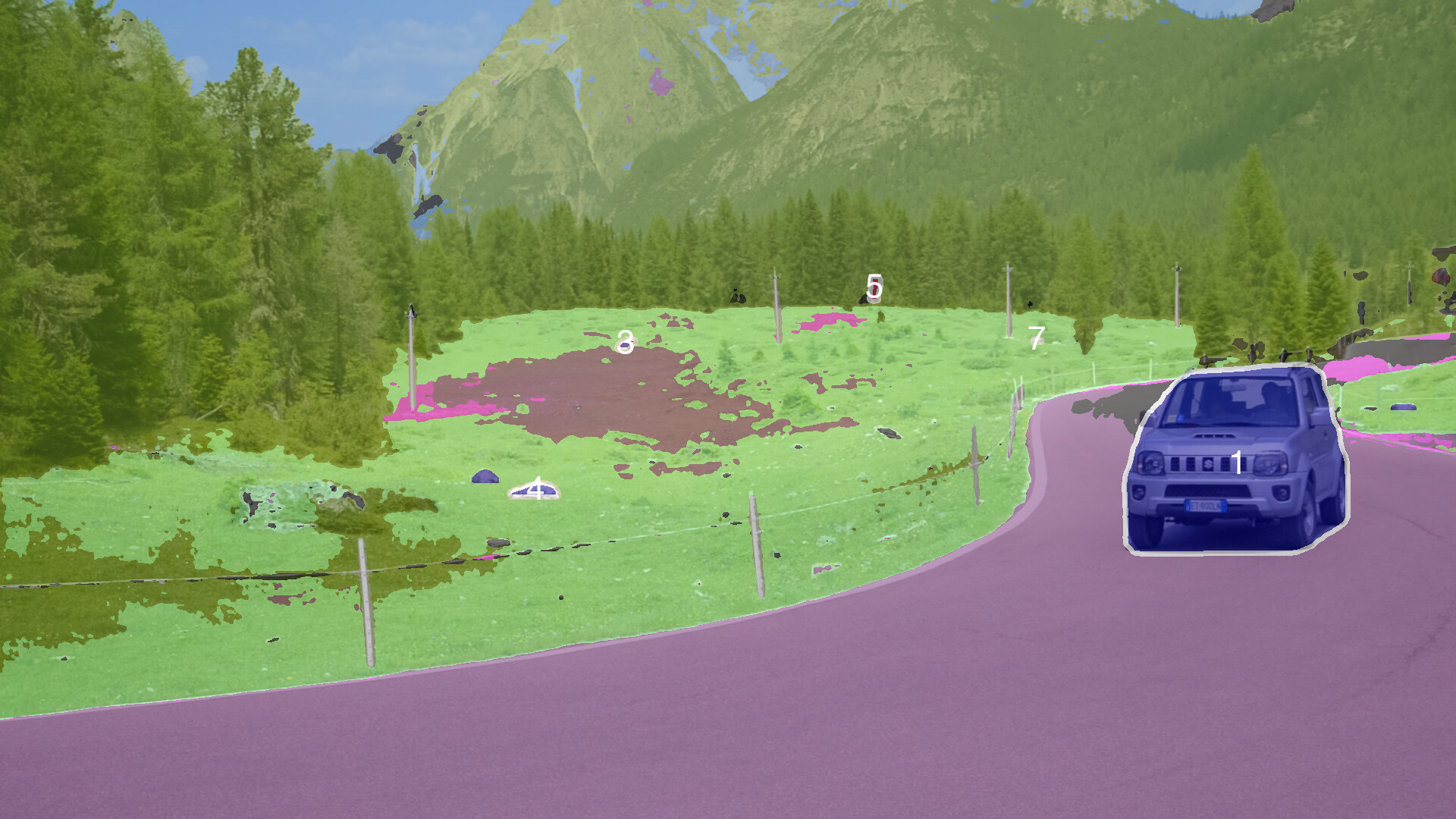} 
& \includegraphics[width=\linewidth]{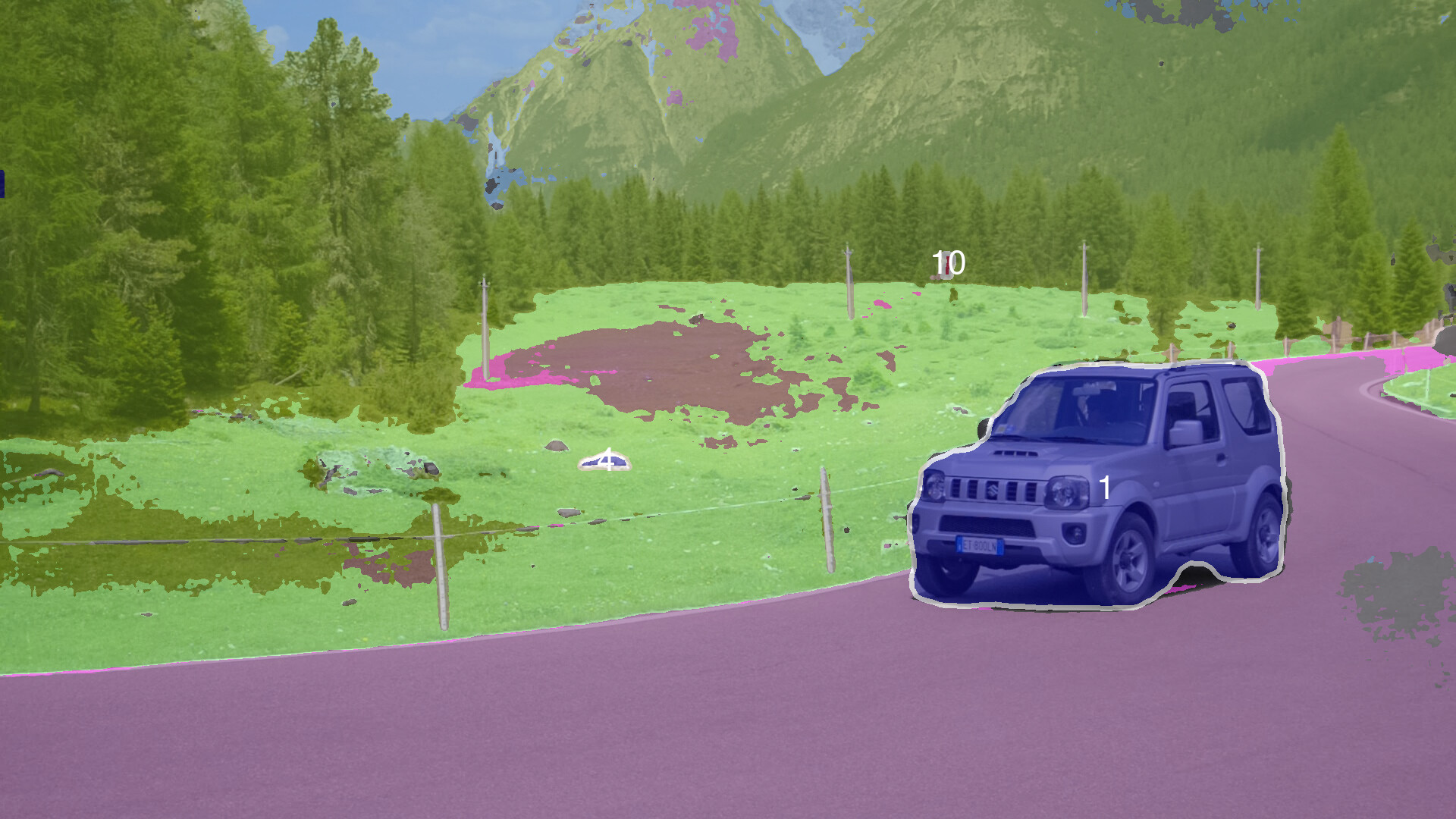} 
& \includegraphics[width=\linewidth]{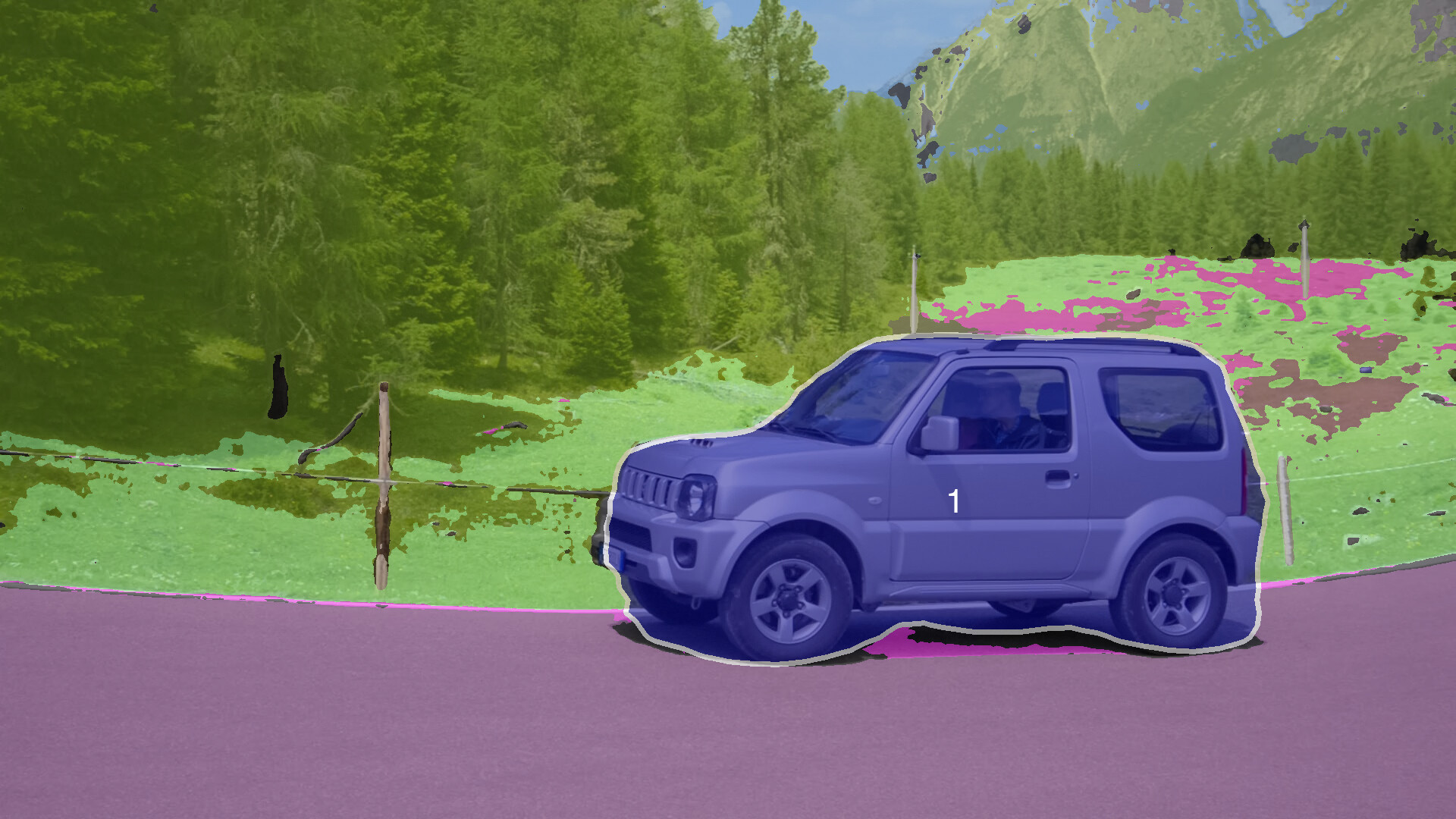}\\

\includegraphics[width=\linewidth]{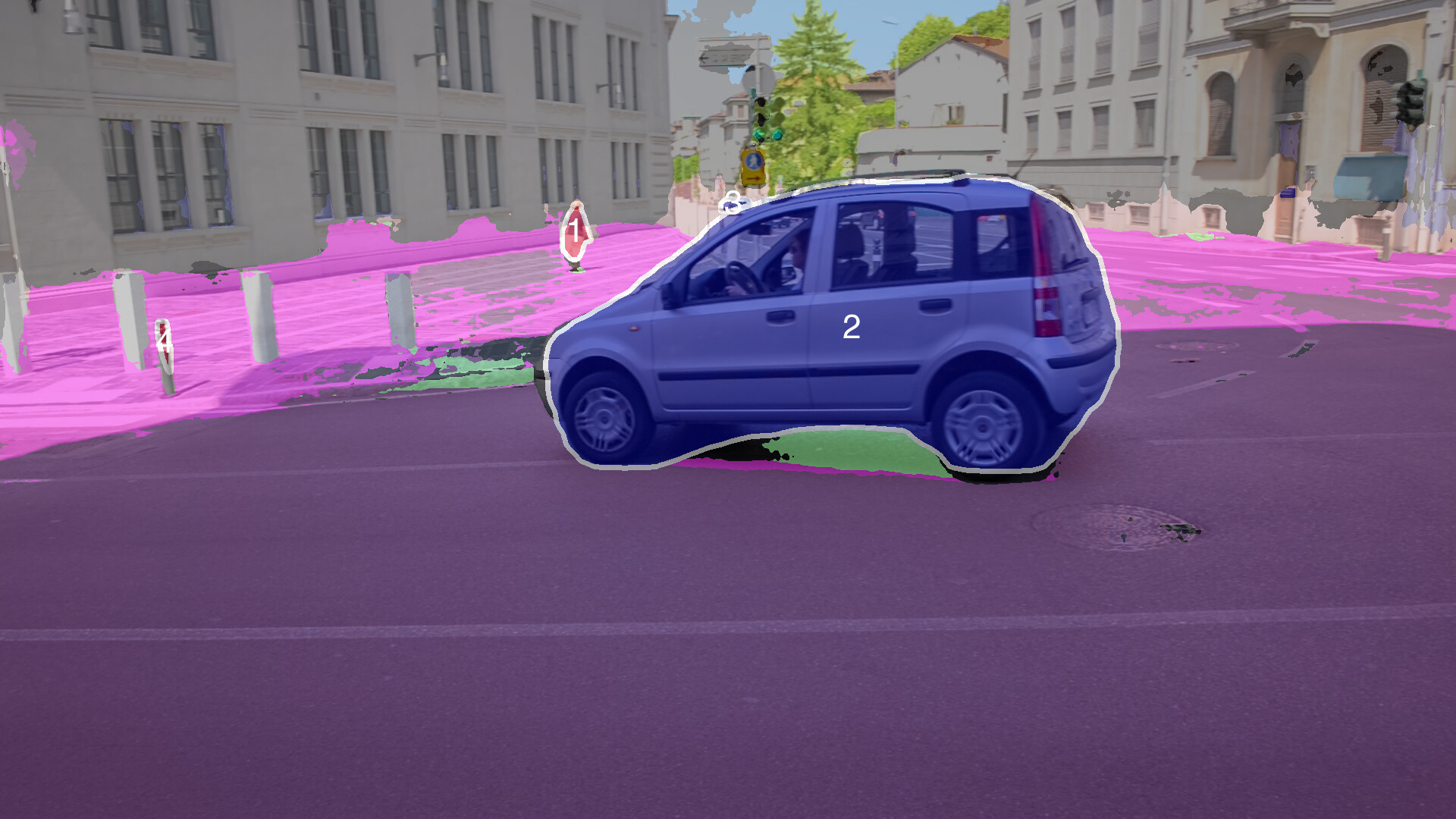} 
& \includegraphics[width=\linewidth]{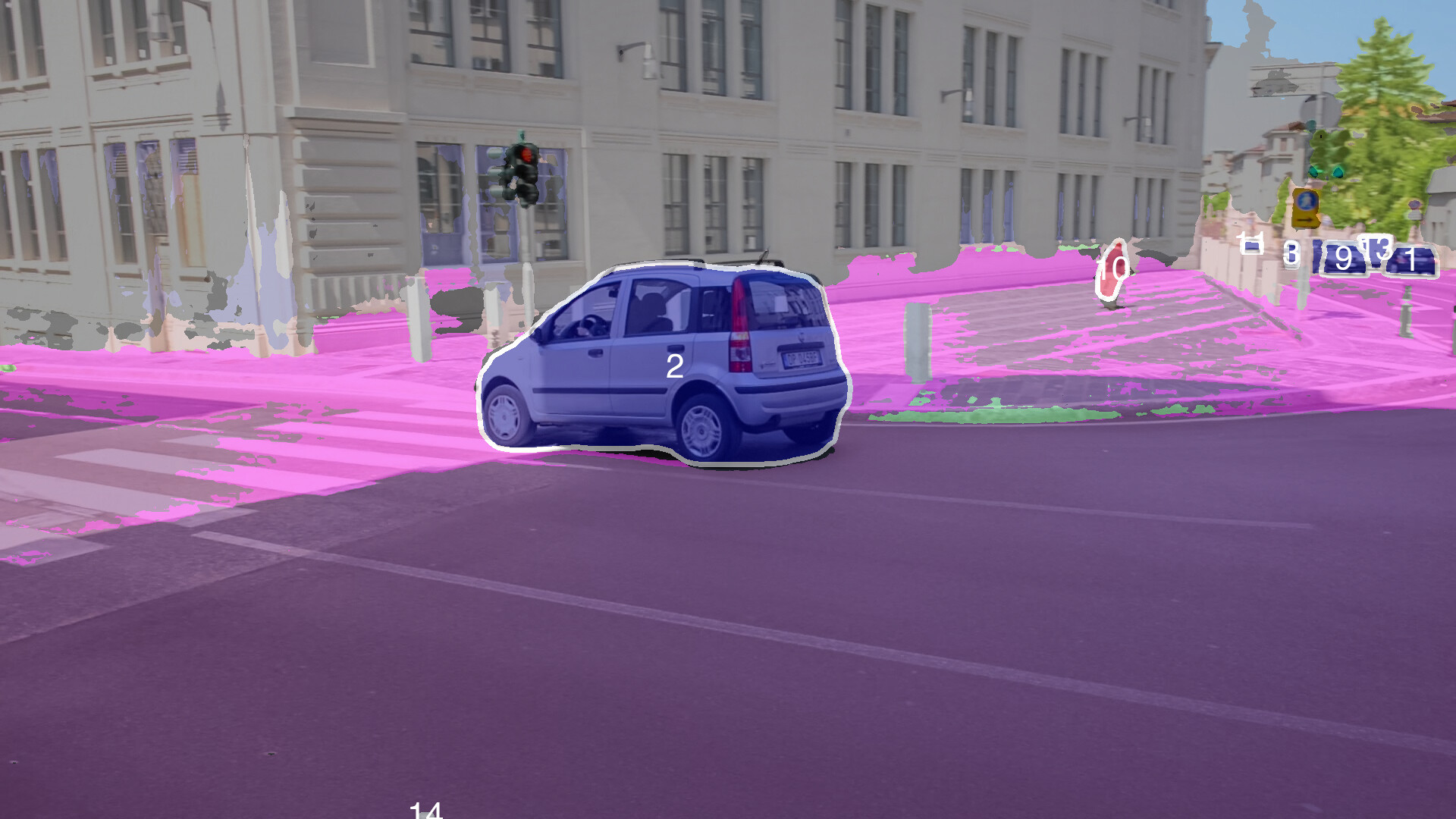} 
& \includegraphics[width=\linewidth]{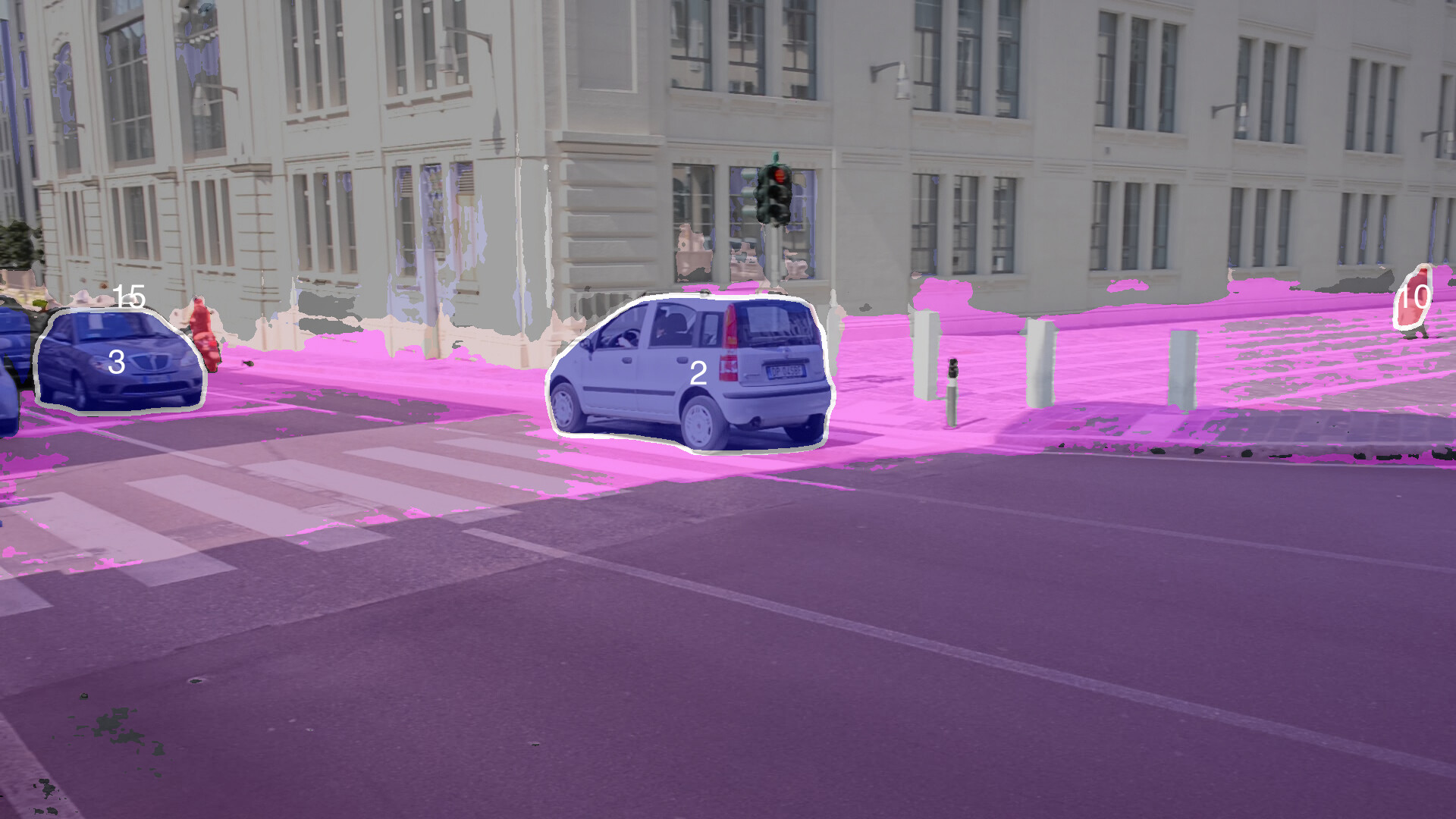}  
& \includegraphics[width=\linewidth]{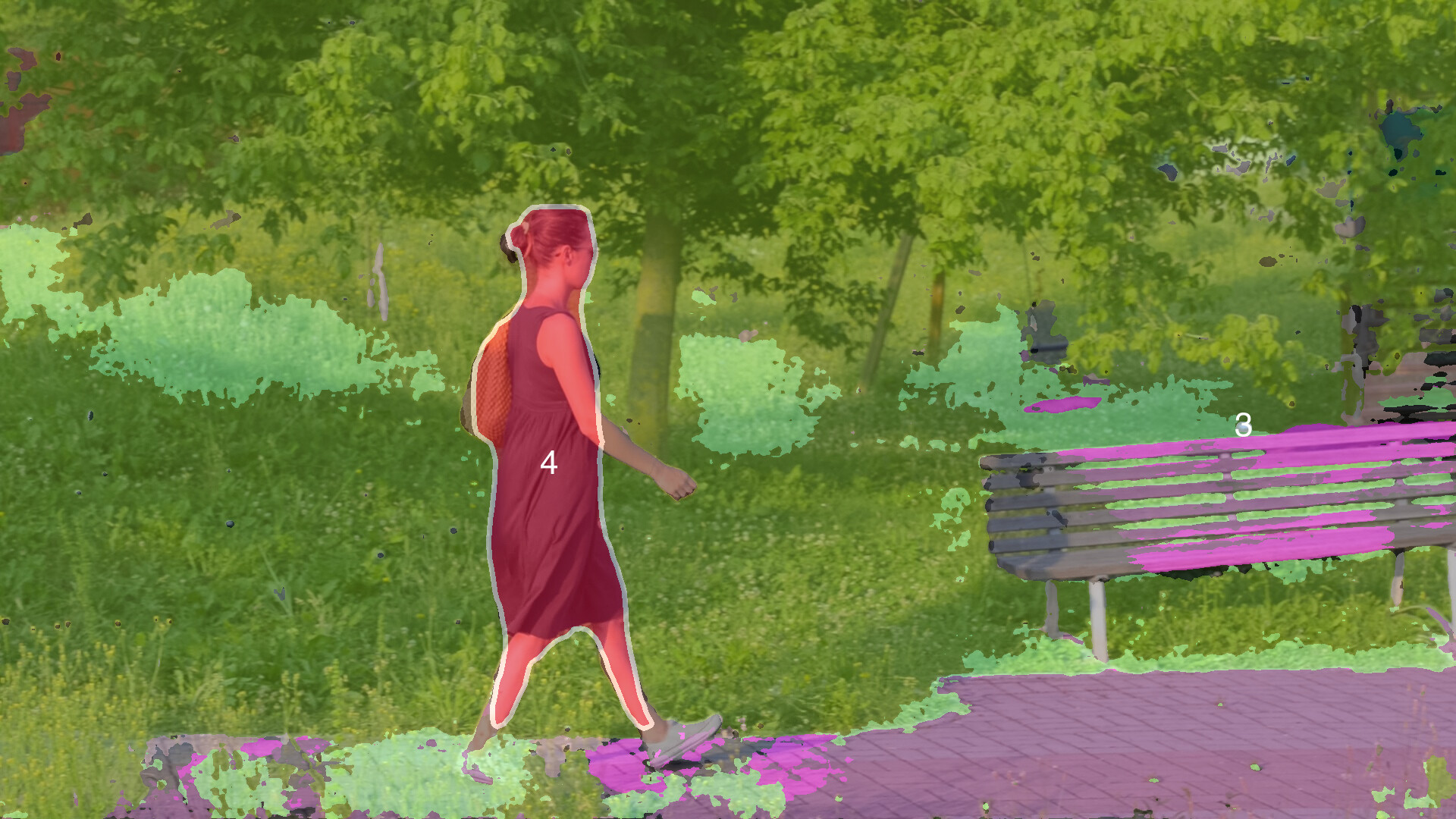}
& \includegraphics[width=\linewidth]{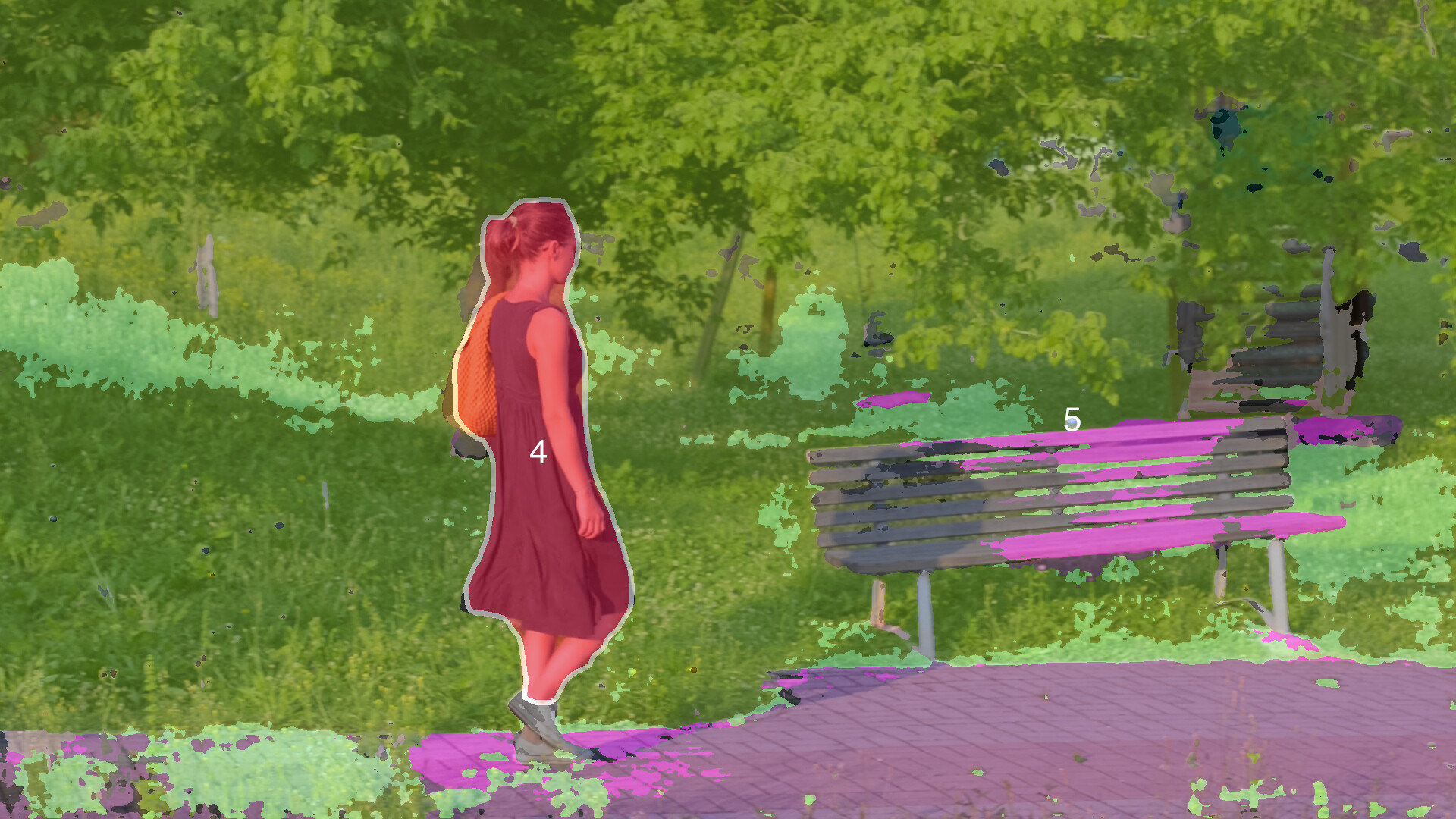}
& \includegraphics[width=\linewidth]{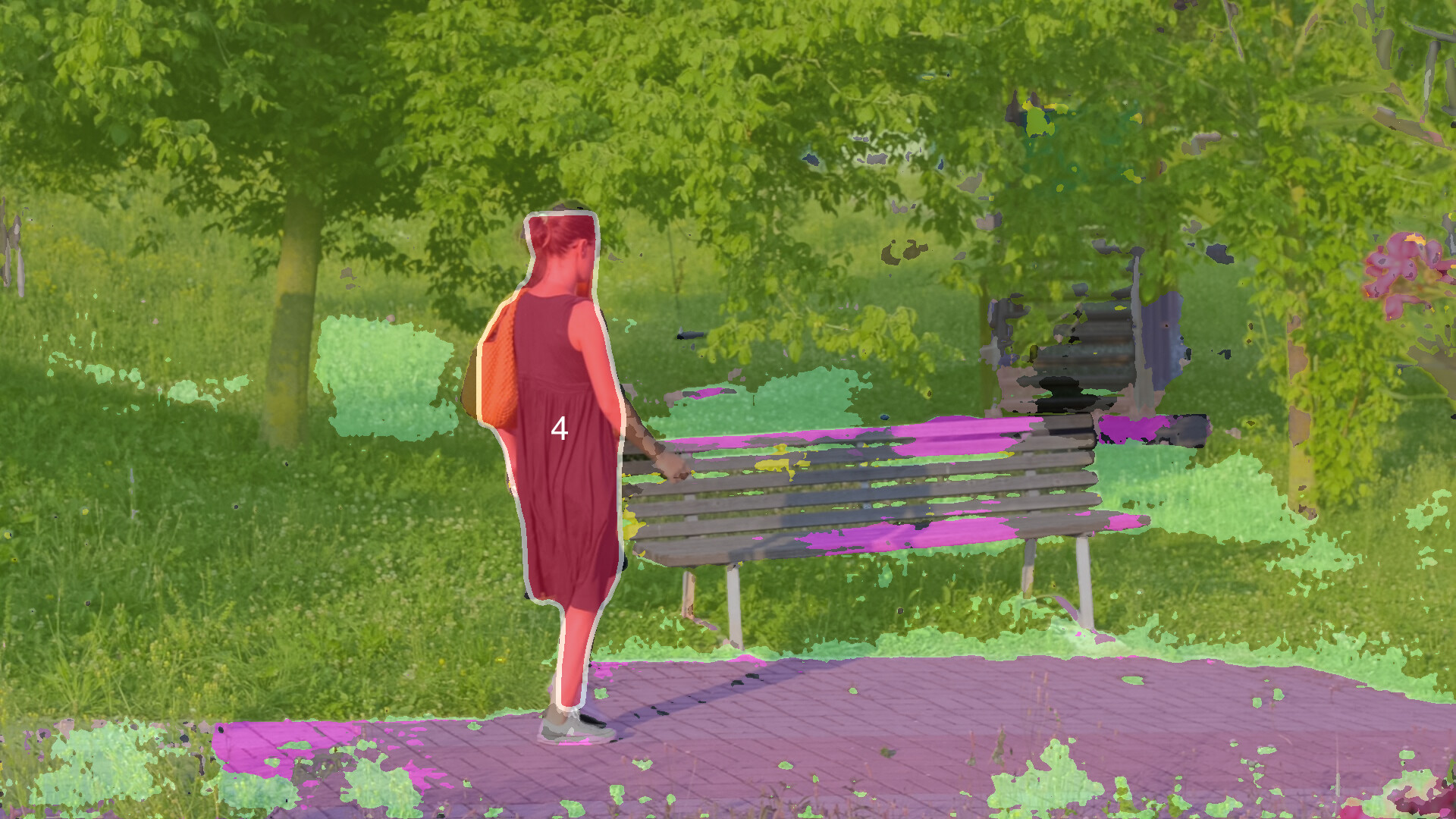}\\[-2.5pt]

\tiny{$\phantom{.}t_0$}
&\tiny{$\phantom{.}t_1$} 
&\tiny{$\phantom{.}t_2$}
&\tiny{$\phantom{.}t_0$} 
&\tiny{$\phantom{.}t_1$} 
&\tiny{$\phantom{.}t_2$} \\

\end{tabular}

    \vspace{-1.2em}
    \caption{\textbf{Qualitative unsupervised VPS examples on DAVIS} \cite{davis}. We provide qualitative samples for \MethodName inference on DAVIS videos using the Cityscapes-VPS class assignments for visualization purposes. \MethodName generalizes to the unseen dataset and even to unseen semantic concepts. \label{fig:qualitative_davis}}
    \vspace{-0.5em}
\end{figure*}

\inparagraph{Dependency on driving scenes.} While Most of our results are reported on driving scenes, \MethodName can be applied to non-driving-specific scenarios. Our approach only requires an agent moving through space, and target instances are movable, a common setting in robotics. Still, we require accurate unsupervised depth and motion, as well as VPS annotations for evaluation, which are mostly available for driving scenes (\eg, KITTI-STEP~\cite{Weber:2021:STE}). We show domain generalization of \MethodName beyond driving scenes on MOTS (\cf \cref{tab:uvps_gen}, \cref{fig:qualitative_mots}) and DAVIS (\cf \cref{fig:qualitative_davis}).

\inparagraph{Occlusions.} Motion segmentation, used for pseudo-labeling, can only detect non-occluded objects. Additionally, partial occlusions, \eg, a car behind a pole, can lead to two detections of the same object cut by the partial occlusion. Subsequently, \MethodName, trained using these pseudo-labels, struggles to detect partially occluded objects correctly and fails to track temporarily fully occluded objects. Still, though our self-enhanced video copy-paste augmentation \MethodName can handle some degree of partial occlusions (\cf \cref{fig:occlusions}). Enhancing the pasting strategy of video copy-paste augmentations by systematically introducing partial and full occlusions, as well as by extending training to longer clips, might offer potential avenues to mitigate this limitation.

\inparagraph{Unsupervised semantic taxonomy.} Unsupervised segmentation approaches, including \MethodName, learn a segmentation taxonomy from unsupervised cues and imposed hyperparameters. While we demonstrate that \MethodName learns a taxonomy that significantly correlates with human-defined taxonomies, ideally, unsupervised approaches would learn a flexible, hierarchical taxonomy capable of expressing and discovering novel semantic categories. Creating more flexible approaches and benchmarks that treat unsupervised VPS as an open-vocabulary task would provide a path to overcoming this limitation. Additionally, unsupervised taxonomies require matching to a ground-truth taxonomy for validation. While most likely not suitable for evaluation, exploring ground-truth-free alignment between taxonomies could provide a powerful way to adapt unsupervised segmentation models to new taxonomies and to analyze the structure of the learned taxonomy~\cite{platonic, blindmatch, hug}.

\inparagraph{Scaling to multiple datasets.} Existing scene-centric unsupervised panoptic methods, such as CUPS, rely on stereo video during training. In contrast, \MethodName's pseudo-labeling uses monocular videos. This provides an initial step toward scaling unsupervised panoptic video understanding to larger video datasets. Still, achieving true scalability to causal and monocular videos requires progress in two domains. \emph{First}, while stereo depth estimation is robust and generalizes well, unsupervised monocular depth estimation is still limited. Current models, including Dynamo-Depth~\cite{Sun:2023:DDF}, are typically trained on a single dataset and typically do not generalize well to different cameras and other datasets/domains. Robust and generalizable unsupervised monocular depth estimation, including a static and dynamic scene decomposition, would enable more high-quality pseudo-labels, enabling scaling \MethodName. \emph{Second}, unsupervised semantic segmentation approaches must produce consistent pseudo semantics across diverse datasets and a large set of pseudo-categories. Current DINO-based unsupervised semantic segmentation approaches, including DepthG~\cite{Sick:2024:USS}, typically train a segmentation head for a specific dataset. The resulting pseudo-categories do not necessarily align with pseudo-categories obtained when training on another dataset. Additionally, diffusion-based unsupervised semantic segmentation approaches, such as DiffCut~\cite{Couairon:2024:DCT}, use language supervision and often provide pseudo-categories only consistent within a single image, requiring per-image matching for validation. Obtaining an approach that can express a globally consistent, hierarchical, and large-scale taxonomy of pseudo-categories is needed for scaling unsupervised VPS and unsupervised scene understanding in general.

\end{document}